\documentclass[10pt,journal,compsoc]{IEEEtran}
\ifCLASSOPTIONcompsoc
  \usepackage[nocompress]{cite}
\else
  \usepackage{cite}
\fi
\ifCLASSINFOpdf
   \usepackage[pdftex]{graphicx}
\else
\fi
\usepackage{amssymb}
\usepackage{amsmath}
\usepackage[ruled,vlined]{algorithm2e}
\usepackage[table]{xcolor}
\usepackage{multirow}
\ifCLASSOPTIONcompsoc
  \usepackage[caption=false,font=footnotesize,labelfont=sf,textfont=sf]{subfig}
  \usepackage{adjustbox}
\else
  \usepackage[caption=false,font=footnotesize]{subfig}
\fi
\newcommand\MYhyperrefoptions{bookmarks=true,bookmarksnumbered=true,
pdfpagemode={UseOutlines},plainpages=false,pdfpagelabels=true,
colorlinks=true,linkcolor={black},citecolor={black},urlcolor={black},
pdftitle={Data Augmentation in HDLSS with VAE},%<!CHANGE!
pdfsubject={Data augmentation},%<!CHANGE!
pdfauthor={Clement Chadebec et. al},%<!CHANGE!
pdfkeywords={Variational autoencoders, data augmentation, latent space modeling}}%<^!CHANGE!
\usepackage[\MYhyperrefoptions,pdftex]{hyperref}
\begin{document}
%
% paper title
% Titles are generally capitalized except for words such as a, an, and, as,
% at, but, by, for, in, nor, of, on, or, the, to and up, which are usually
% not capitalized unless they are the first or last word of the title.
% Linebreaks \\ can be used within to get better formatting as desired.
% Do not put math or special symbols in the title.
\title{Data Augmentation in High Dimensional Low Sample Size Setting Using a Geometry-Based Variational Autoencoder}
%
%
% author names and IEEE memberships
% note positions of commas and nonbreaking spaces ( ~ ) LaTeX will not break
% a structure at a ~ so this keeps an author's name from being broken across
% two lines.
% use \thanks{} to gain access to the first footnote area
% a separate \thanks must be used for each paragraph as LaTeX2e's \thanks
% was not built to handle multiple paragraphs
%
%
%\IEEEcompsocitemizethanks is a special \thanks that produces the bulleted
% lists the Computer Society journals use for "first footnote" author
% affiliations. Use \IEEEcompsocthanksitem which works much like \item
% for each affiliation group. When not in compsoc mode,
% \IEEEcompsocitemizethanks becomes like \thanks and
% \IEEEcompsocthanksitem becomes a line break with idention. This
% facilitates dual compilation, although admittedly the differences in the
% desired content of \author between the different types of papers makes a
% one-size-fits-all approach a daunting prospect. For instance, compsoc 
% journal papers have the author affiliations above the "Manuscript
% received ..."  text while in non-compsoc journals this is reversed. Sigh.

\author{Clément Chadebec,
%~\IEEEmembership{Université de Paris, INRIA, Centre de Recherche des Cordeliers, INSERM, Sorbonne Université},
Elina Thibeau-Sutre,
%~\IEEEmembership{Paris Brain Institute, ICM, Inserm U 1127, CNRS UMR 7225, Sorbonne Université, Inria, Aramis project-team},
Ninon Burgos,
%~\IEEEmembership{Paris Brain Institute, ICM, Inserm U 1127, CNRS UMR 7225, Sorbonne Université, Inria, Aramis project-team}
and~Stéphanie~Allassonnière,
%~\IEEEmembership{Université de Paris, INRIA, Centre de Recherche des Cordeliers, INSERM, Sorbonne Université}
for the Alzheimer’s Disease Neuroimaging Initiative, and the Australian Imaging Biomarkers and Lifestyle flagship study of ageing

        % <-this % stops a space
%\IEEEcompsocitemizethanks{\IEEEcompsocthanksitem M. Shell was with the Department
%of Electrical and Computer Engineering, Georgia Institute of Technology, Atlanta,
%GA, 30332.\protect\\
% note need leading \protect in front of \\ to get a newline within \thanks as
% \\ is fragile and will error, could use \hfil\break instead.
%E-mail: see http://www.michaelshell.org/contact.html
%\IEEEcompsocthanksitem J. Doe and J. Doe are with Anonymous University.}% <-this % stops a space
%\thanks{Manuscript received April 19, 2005; revised August 26, 2015.}

\thanks{Data used in preparation of this article were obtained from the Alzheimer’s Disease Neuroimaging Initiative (ADNI) database (\url{http://adni.loni.usc.edu}). As such, the investigators within the ADNI contributed to  the design and implementation of ADNI and/or provided data but did not participate in analysis or writing of this report. A complete listing of ADNI investigators can be found at: \url{http://adni.loni.usc.edu/wp-content/uploads/how_to_apply/ADNI_Acknowledgement_List.pdf}}

\thanks{Data used in the preparation of this article was obtained from the Australian Imaging Biomarkers and Lifestyle flagship study of ageing (AIBL) funded by the Commonwealth Scientific and Industrial Research Organisation (CSIRO) which was made available at the ADNI database (\url{http://adni.loni.usc.edu}). The AIBL researchers contributed data but did not participate in analysis or writing of this report. AIBL researchers are listed at \url{www.aibl.csiro.au.}}

\IEEEcompsocitemizethanks{
    \IEEEcompsocthanksitem Clément~Chadebec and Stéphanie~Allassonnière are with the Université de Paris, Inria, Centre de Recherche des Cordeliers, Inserm, Sorbonne Université, Paris, France
    \IEEEcompsocthanksitem Elina~Thibeau-Sutre and Ninon~Burgos are with Sorbonne Université, Institut du Cerveau - Paris Brain Institute (ICM), Inserm U 1127, CNRS UMR 7225, AP-HP Hôpital de la Pitié Salpêtrière and Inria Aramis project-team, Paris, France
}

}

% note the % following the last \IEEEmembership and also \thanks - 
% these prevent an unwanted space from occurring between the last author name
% and the end of the author line. i.e., if you had this:
% 
% \author{....lastname \thanks{...} \thanks{...} }
%                     ^------------^------------^----Do not want these spaces!
%
% a space would be appended to the last name and could cause every name on that
% line to be shifted left slightly. This is one of those "LaTeX things". For
% instance, "\textbf{A} \textbf{B}" will typeset as "A B" not "AB". To get
% "AB" then you have to do: "\textbf{A}\textbf{B}"
% \thanks is no different in this regard, so shield the last } of each \thanks
% that ends a line with a % and do not let a space in before the next \thanks.
% Spaces after \IEEEmembership other than the last one are OK (and needed) as
% you are supposed to have spaces between the names. For what it is worth,
% this is a minor point as most people would not even notice if the said evil
% space somehow managed to creep in.

% The paper headers
% \markboth{Journal of \LaTeX\ Class Files,~Vol.~14, No.~8, August~2015}%
% {Shell \MakeLowercase{\textit{et al.}}: Bare Advanced Demo of IEEEtran.cls for IEEE Computer Society Journals}
\markboth{Data Augmentation in High Dimensional Low Sample Size Setting Using a Geometry-Based Variational Autoencoder}%
{Data Augmentation in High Dimensional Low Sample Size Setting Using a Geometry-Based Variational Autoencoder}
% The only time the second header will appear is for the odd numbered pages
% after the title page when using the twoside option.
% 
% *** Note that you probably will NOT want to include the author's ***
% *** name in the headers of peer review papers.                   ***
% You can use \ifCLASSOPTIONpeerreview for conditional compilation here if
% you desire.

% The publisher's ID mark at the bottom of the page is less important with
% Computer Society journal papers as those publications place the marks
% outside of the main text columns and, therefore, unlike regular IEEE
% journals, the available text space is not reduced by their presence.
% If you want to put a publisher's ID mark on the page you can do it like
% this:
%\IEEEpubid{0000--0000/00\$00.00~\copyright~2015 IEEE}
% or like this to get the Computer Society new two part style.
%\IEEEpubid{\makebox[\columnwidth]{\hfill 0000--0000/00/\$00.00~\copyright~2015 IEEE}%
%\hspace{\columnsep}\makebox[\columnwidth]{Published by the IEEE Computer Society\hfill}}
% Remember, if you use this you must call \IEEEpubidadjcol in the second
% column for its text to clear the IEEEpubid mark (Computer Society journal
% papers don't need this extra clearance.)

% use for special paper notices
%\IEEEspecialpapernotice{(Invited Paper)}

% for Computer Society papers, we must declare the abstract and index terms
% PRIOR to the title within the \IEEEtitleabstractindextext IEEEtran
% command as these need to go into the title area created by \maketitle.
% As a general rule, do not put math, special symbols or citations
% in the abstract or keywords.
\IEEEtitleabstractindextext{%
\begin{abstract}
In this paper, we propose a new
method to perform data augmentation in a reliable way in the High Dimensional Low Sample Size (HDLSS) setting using a geometry-based variational autoencoder (VAE). Our approach combines the proposal of 1) a new VAE model, the latent space of which is modeled as a Riemannian manifold and which combines both Riemannian metric learning and normalizing flows and 2) a new generation scheme which produces more meaningful samples especially in the context of small data sets. The method is tested through a wide experimental study where its robustness to data sets, classifiers and training samples size is stressed. It is also validated on a medical imaging classification task on the challenging ADNI database where a small number of 3D brain magnetic resonance images (MRIs) are considered and augmented using the proposed VAE framework. In each case, the proposed method allows for a significant and reliable gain in the classification metrics. For instance, balanced accuracy jumps from 66.3\% to 74.3\% for a \emph{state-of-the-art} convolutional neural network classifier trained with 50 MRIs of cognitively normal (CN) and 50 Alzheimer disease (AD) patients and from 77.7\% to 86.3\% when trained with 243 CN and 210 AD while improving greatly sensitivity and specificity metrics.

\end{abstract}

% Note that keywords are not normally used for peerreview papers.
}

% make the title area
\maketitle

% To allow for easy dual compilation without having to reenter the
% abstract/keywords data, the \IEEEtitleabstractindextext text will
% not be used in maketitle, but will appear (i.e., to be "transported")
% here as \IEEEdisplaynontitleabstractindextext when compsoc mode
% is not selected <OR> if conference mode is selected - because compsoc
% conference papers position the abstract like regular (non-compsoc)
% papers do!
\IEEEdisplaynontitleabstractindextext
% \IEEEdisplaynontitleabstractindextext has no effect when using
% compsoc under a non-conference mode.

% For peer review papers, you can put extra information on the cover
% page as needed:
% \ifCLASSOPTIONpeerreview
% \begin{center} \bfseries EDICS Category: 3-BBND \end{center}
% \fi
%
% For peerreview papers, this IEEEtran command inserts a page break and
% creates the second title. It will be ignored for other modes.
\IEEEpeerreviewmaketitle

\ifCLASSOPTIONcompsoc
\IEEEraisesectionheading{\section{Introduction}\label{sec:introduction}}
\else
\section{Introduction}
\fi

% Computer Society journal (but not conference!) papers do something unusual
% with the very first section heading (almost always called "Introduction").
% They place it ABOVE the main text! IEEEtran.cls does not automatically do
% this for you, but you can achieve this effect with the provided
% \IEEEraisesectionheading{} command. Note the need to keep any \label that
% is to refer to the section immediately after \section in the above as
% \IEEEraisesectionheading puts \section within a raised box.

% The very first letter is a 2 line initial drop letter followed
% by the rest of the first word in caps (small caps for compsoc).
% 
% form to use if the first word consists of a single letter:
% \IEEEPARstart{A}{demo} file is ....
% 
% form to use if you need the single drop letter followed by
% normal text (unknown if ever used by the IEEE):
% \IEEEPARstart{A}{}demo file is ....
% 
% Some journals put the first two words in caps:
% \IEEEPARstart{T}{his demo} file is ....
% 
% Here we have the typical use of a "T" for an initial drop letter
% and "HIS" in caps to complete the first word.
\IEEEPARstart{E}{ven} though always larger data sets are now available, the lack of labeled data remains a tremendous issue in many fields of application. Among others, a good example is healthcare where practitioners have to deal most of the time with (very) low sample sizes (think of small patient cohorts) along with very high dimensional data (think of neuroimaging data that are 3D volumes with millions of voxels). Unfortunately, this leads to a very poor representation of a given population and makes classical statistical analyses unreliable~\cite{button_power_2013,turner_small_2018}. Meanwhile, the remarkable performance of algorithms heavily relying on the deep learning framework~\cite{goodfellow_deep_2016} has made them extremely attractive and very popular. However, such results are strongly conditioned by the number of training samples since such models usually need to be trained on huge data sets to prevent over-fitting or to give statistically meaningful results~\cite{shorten_survey_2019}.

A way to address such issues is to perform data augmentation (DA)~\cite{tanner_calculation_1987}. In a nutshell, DA is the art of increasing the size of a given data set by creating synthetic labeled data. For instance, the easiest way to do this on images is to apply simple transformations such as the addition of Gaussian noise, cropping or padding, and assign the label of the initial image to the created ones. While such augmentation techniques have revealed very useful, they remain strongly data dependent and limited. Some transformations may indeed be uninformative or even induce bias. For instance, think of a digit representing a \emph{6} which gives a \emph{9} when rotated. While assessing the relevance of augmented data may be quite straightforward for simple data sets, it reveals very challenging for complex data and may require the intervention of an \emph{expert} assessing the degree of relevance of the proposed transformations. In addition to the lack of data, imbalanced data sets also severely limit generalizability since they tend to bias the algorithm toward the most represented classes. Oversampling is a method that aims at balancing the number of samples per class by up-sampling the minority classes. The Synthetic Minority Over-sampling TEchnique (SMOTE) was first introduced in~\cite{chawla_smote_2002} and consists in interpolating data points belonging to the minority classes in their feature space. This approach was further extended in other works where the authors proposed to over-sample close to the decision boundary using either the $k$-Nearest Neighbor ($k$-NN) algorithm~\cite{hutchison_borderline-smote_2005} or a support vector machine (SVM)~\cite{nguyen_borderline_2011} and so insist on samples that are potentially misclassified. Other over-sampling methods aiming at increasing the number of samples from the minority classes and taking into account their difficulty to be learned were also proposed~\cite{haibo_he_adasyn_2008,barua_mwmote--majority_2012}. However, these methods hardly scale to high-dimensional data~\cite{blagus_smote_2013, fernandez_smote_2018}.

The recent rise in performance of generative models such as generative adversarial networks (GAN)~\cite{goodfellow_generative_2014} or variational autoencoders (VAE)~\cite{kingma_auto-encoding_2014,rezende_stochastic_2014} has made them very attractive models to perform DA. GANs have already seen a wide use in many fields of application~\cite{zhu_emotion_2018, mariani_bagan_2018,antoniou_data_2018, lim_doping_2018,zhu_data_2018}, including medicine \cite{yi_generative_2019}. For instance, GANs were used on magnetic resonance images (MRIs)~\cite{shin_medical_2018, calimeri_biomedical_2017}, computed tomography (CT)~\cite{frid-adar_gan-based_2018,sandfort_data_2019}, X-ray~\cite{madani_chest_2018,salehinejad_generalization_2018, waheed_covidgan_2020 }, positron emission tomography (PET)~\cite{bi_synthesis_2017}, mass spectroscopy data~\cite{liu_wasserstein_2019}, dermoscopy~\cite{baur_generating_2018} or mammography~\cite{korkinof_high-resolution_2018,wu_conditional_2018} and demonstrated promising results. Nonetheless, most of these studies involved either a quite large training set (above 1000 training samples) or quite small dimensional data, whereas in everyday medical applications it remains very challenging to gather such large cohorts of labeled patients. As a consequence, as of today, the case of high dimensional data combined with a very low sample size remains poorly explored. When compared to GANs, VAEs have only seen a very marginal interest to perform DA and were mostly used for speech applications~\cite{hsu_unsupervised_2017,nishizaki_data_2017,wu_data_2019}. Some attempts to use such generative models on medical data either for classification~\cite{zhuang_fmri_2019,liu_data_2018} or segmentation tasks~\cite{painchaud_cardiac_2019,selvan_lung_2020,myronenko_3d_2018} can nonetheless be noted. The main limitation to a wider use of these models is that they most of the time produce blurry and fuzzy samples. This undesirable effect is even more emphasized when they are trained with a small number of samples which makes them very hard to use in practice to perform DA in the high dimensional (very) low sample size (HDLSS) setting.

In this paper, we argue that VAEs can actually be used for data augmentation in a reliable way even in the context of HDLSS data, provided that we bring some modeling of the latent space and amend the way we generate the data. Hence, in this paper we propose the following contributions:
\begin{itemize}
    \item We propose a new \emph{geometry-aware} VAE model, the latent space of which is seen as a Riemannian manifold and combining Riemannian metric learning and normalizing flows.
    \item We introduce a new \emph{non-prior} based generation procedure consisting in sampling from the inverse of the Riemannian metric volume element learned by the model. The choice of this framework is discussed, motivated and compared to other VAE models.\footnote{An implementation of the models may be found at \url{https://github.com/clementchadebec/benchmark_VAE} }
    \item We propose to use such a framework to perform data augmentation in the challenging context of HDLSS data. The robustness of the augmentation method to data sets and classifiers changes along with its reliance to the number of training samples and the complexity of the classifier is then tested through a series of experiments.\footnote{A software implementing the method was developed and is available at \url{https://github.com/clementchadebec/pyraug} }
    \item We validate the proposed method on several \emph{real-life} classification tasks on complex 3D MRI from ADNI and AIBL databases where the augmentation method allows for a significant gain in classification metrics even when only 50 samples per class are considered.
\end{itemize}

\section{Variational Autoencoder}

In this section, we quickly recall the idea behind VAEs along with some proposed improvements relevant to this paper.

\subsection{Model Setting}

Let $x \in \mathcal{X}$ be a set of data. A VAE aims at maximizing the likelihood of a given parametric model $\{\mathbb{P}_{\theta}, \theta \in \Theta\}$. It is assumed that there exist latent variables $z$ living in a lower dimensional space $\mathcal{Z}$, referred to as the \emph{latent space}, such that the marginal distribution of the data can be written as:
\begin{equation}\label{Eq:Objective}
    p_{\theta}(x) = \int \limits _{\mathcal{Z}} p_{\theta}(x|z)q(z) dz \,,
\end{equation}
where $q$ is a prior distribution over the latent variables acting as a regulation factor and $p_{\theta}(x|z)$ is most of the time taken as a simple parametrized distribution (\textit{e.g.} Gaussian, Bernoulli, etc.). Such a distribution is referred to as the \textit{decoder}, the parameters of which are usually given by neural networks. Since the integral of Eq.~\eqref{Eq:Objective} is most of the time intractable, so is the posterior distribution:
\[
p_{\theta}(z|x) = \frac{p_{\theta}(x|z) q(z)}{\int \limits_{\mathcal{Z}} p_{\theta}(x|z) q(z) dz}\,.
\]
This makes direct application of Bayesian inference impossible and so recourse to approximation techniques such as variational inference~\cite{jordan_introduction_1999} is needed. Hence, a variational distribution $q_{\phi}(z|x)$ is introduced and aims at approximating the true posterior distribution $p_{\theta}(z|x)$~\cite{kingma_auto-encoding_2014}. This variational distribution is often referred to as the \emph{encoder}. In the initial version of the VAE, $q_{\phi}$ is taken as a multivariate Gaussian whose parameters $\mu_{\phi}$ and $\Sigma_{\phi}$ are again given by neural networks. Importance sampling is then applied to get an unbiased estimate of  $p_{\theta}(x)$ we want to maximize in Eq.~\eqref{Eq:Objective}
\begin{equation}\label{eq: log-like estimate vae}
    \hat{p}_{\theta}(x) = \frac{p_{\theta}(x|z)q(z)}{q_{\phi}(z|x)} \hspace{2mm} \text{and} \hspace{2mm} \mathbb{E}_{z \sim q_{\phi}}\big[\hat{p}_{\theta}\big] = p_{\theta}(x)\,.
\end{equation}
Using Jensen's inequality allows finding a lower bound on the objective function of Eq.~\eqref{Eq:Objective}
\begin{equation}\label{eq: ELBO}
     \begin{aligned}
      \log p_{\theta}(x) &= \log \mathbb{E}_{z \sim q_{\phi}}\big[\hat{p}_{\theta}\big]\\
                         &\geq \mathbb{E}_{z \sim q_{\phi}}\big[\log \hat{p}_{\theta}\big]\\
                         & \geq \mathbb{E}_{z \sim q_{\phi}}\big[ \log p_{\theta}(x|z) \big] - D_{KL}(q_{\phi}(z|x)\lVert p(z))\,.
     \end{aligned}
\end{equation}
The Evidence Lower BOund (ELBO) is now tractable since  all distributions are known and so can be optimized with respect to the \textit{encoder} and \textit{decoder} parameters. 

\subsection{Improving the Model: Literature Review}

In recent years, many attempts to improve the VAE model have been made and we briefly discuss three main areas of improvement that are relevant to this paper in this section.
\subsubsection{Enhancing the Variational Approximate Distribution}\label{Subsec: Enhancing VAE}

When looking at Eq.~\eqref{eq: ELBO}, it can be noticed that we are nonetheless trying to optimize only a lower bound on the true objective function. Therefore, much efforts have been focused on making this lower bound tighter and tighter~\cite{burda_importance_2016, alemi_deep_2016,higgins_beta-vae_2017,cremer_inference_2018,zhang_advances_2018, ruiz_contrastive_2019}. One way to do this is to enhance the expressiveness of the approximate posterior distribution $q_{\phi}$. This is indeed due to the ELBO expression which can be also written as follows:
\[
    ELBO = \log p_{\theta}(x) - D_{KL}(q_{\phi}(z|x) || p_{\theta}(z|x))\,.
\]
This expression makes two terms appear. The first one is the function we want to maximize while the second one is the Kullback–Leibler (KL) divergence between the approximate posterior distribution $q_{\phi}(z|x)$ and the true posterior $p_{\theta}(z|x)$. This very term is always non-negative and equals 0 if and only if $q_{\phi} = p_{\theta}$ almost everywhere. Hence, trying to tweak the approximate posterior distribution so that it becomes \emph{closer} to the true posterior should make the ELBO tighter and enhance the model. To do so, a method proposed in~\cite{salimans_markov_2015} consisted in adding $K$ 
Markov chain Monte Carlo (MCMC) sampling steps on the top of the approximate posterior distribution and targeting the true posterior. More precisely, the idea was to start from $z_0 \sim q_{\phi}(z|x)$ and use parametrized \textit{forward} (resp. \textit{reverse}) kernels $r(z_{k+1}|z_k, x)$ (resp. $r(z_{k}|z_{k+1}, x)$) to create a new estimate of the true marginal distribution $p_{\theta}(x)$. With the same objective, parametrized invertible mappings $f_x$ called \textit{normalizing flows} were instead proposed in~\cite{rezende_variational_2015} to \emph{sample} $z$. A starting random variable $z_0$ is drawn from an initial distribution $q_{\phi}(z|x)$ and then $K$ normalizing flows are applied to $z_0$ resulting in a random variable $z_K = f_x^K \circ \cdots \circ f_x^1(z_0) $ whose density writes:
\[
q_{\phi}(z_K|x) = q_{\phi}(z_0|x) \prod \limits _{k=1}^K |\det \mathbf{J}_{f_x^k}|^{-1}\,,
\]
where $\mathbf{J}_{f_x^k}$ is the Jacobian of the $k^{\mathrm{th}}$ normalizing flow. Ideally, we would like to have access to normalizing flows targeting the true posterior and allowing enriching the above distribution and so improve the lower bound. In that particular respect, a model inspired by the Hamiltonian Monte Carlo sampler~\cite{neal_mcmc_2011} and relying on Hamiltonian dynamics was proposed in~\cite{salimans_markov_2015} and~\cite{caterini_hamiltonian_2018}. The strength of such a model relies in the choice of the normalizing flows which are guided by the gradient of the true posterior distribution. 
%Indeed, the main idea is to define the target density $\pi$ from which we want to sample to be the true posterior distribution $p_{\theta}(z|%x) \propto p_{\theta}(x, z) = \pi(z)$. The \textit{normalizing flows} corresponds to the leapfrog integrator used within the HMC framework to %"sample" efficiently $z$. 
 
\subsubsection{Improving the Prior Distribution}
While enhancing the approximate posterior distribution resulted in major improvements of the model, it was also argued that the prior distribution over the latent variables plays a crucial role as well~\cite{hoffman_elbo_2016}. Since the vanilla VAE uses a standard Gaussian distribution as prior, a natural improvement consisted in using a mixture of Gaussian instead~\cite{nalisnick_approximate_2016,dilokthanakul_deep_2017} which was further enhanced with the proposal of the variational mixture of posterior (VAMP)~\cite{tomczak_vae_2018}. In addition, other models trying to amend the prior and relying on hierarchical latent variables have been proposed~\cite{sonderby_ladder_2016,burda_importance_2016,klushyn_learning_2019}. Prior learning is also a promising idea that has emerged (\emph{e.g.}~\cite{chen_variational_2016}) or more recently~\cite{razavi_generating_2019,pang_learning_2020,aneja_ncp-vae_2020} and allows accessing complex prior distributions. In the same vein, \emph{ex-post} density estimation was also proposed and consists in fitting a simple distribution such as a mixture of Gaussian in the latent space post training \cite{ghosh_variational_2020}. This approach aimed at alleviating the poor expressiveness of the prior. Another approach relying on accept/reject sampling to improve the prior distribution~\cite{bauer_resampled_2019} can also be cited. While these proposals improved the model, the choice of the prior distribution remains tricky and strongly conditioned by the training data and the tractability of the ELBO. 

\subsubsection{Adding Geometrical Consideration to the Model}\label{Subsec: geometry to model}
In the mean time, several papers have been arguing that geometrical aspects should also be taken into account. For instance, on the ground that the vanilla VAE fails to apprehend data having a latent space with a specific geometry, several latent space modelings were proposed as a hypershere~\cite{davidson_hyperspherical_2018} where Von-Mises distributions are considered instead of Gaussian or as a Poincare disk~\cite{mathieu_continuous_2019,ovinnikov_poincare_2020}. Other works trying to introduce Riemannian geometry within the VAE framework proposed to model either the input data space~\cite{falorsi_explorations_2018, miolane_learning_2020} or the latent space (or both)~\cite{arvanitidis_locally_2016,chen_metrics_2018,shao_riemannian_2018, kalatzis_variational_2020} as Riemannian manifolds. 

\section{The Proposed Method}

In this section, we first present a new \emph{geometry-aware} VAE model bridging the gap between Sec.~\ref{Subsec: Enhancing VAE} and Sec.~\ref{Subsec: geometry to model}. It combines MCMC sampling and Riemannian metric learning to improve the expressiveness of the posterior distribution and learn meaningful latent representations of the data. Secondly, we propose a new \emph{non-prior} based generation scheme taking into account the learned geometry of the data. We indeed argue that while the vast majority of works dealing with VAE generate new data using the prior distribution, which is standard procedure, this is often sub-optimal, in particular in the context of small data sets. We believe that the choice of the prior distribution is strongly data set dependent and is also constrained to be simple so that the ELBO in Eq.~\eqref{eq: ELBO} remains tractable. Hence, the view adopted here is to consider the VAE only as a dimensionality reduction tool which is able to extract the latent structure of the data, \emph{i.e.} the latent space modeled as the Riemannian manifold $(\mathbb{R}^d, g)$ where $d$ is the dimension of the manifold and $g$ is the associated Riemannian metric. Before going further we first recall some elements on Riemannian geometry.

%The same idea of Riemannian metric learning was also re-used in \cite{arvanitidis_prior-based_2021} where it is combined with prior learning. While trying to learn %the best prior is of interest, in this paper we only see it as a regularization factor since it is not used in the generation procedure. Hence, we decided to keep %it simple ($\mathcal{N}(0, I_d)$) and so do not add any further parameters to the model. 

\subsection{Some Elements on Riemannian Geometry}

In the framework of differential geometry, one may define a (connected) Riemannian manifold $\mathcal{M}$ as a smooth manifold endowed with a Riemannian metric $g$ that is a smooth inner product $g: p \to \langle \cdot | \cdot \rangle_p$ on the tangent space $T_p\mathcal{M}$ defined at each point of the manifold $p \in \mathcal{M}$. We call a chart (or coordinate chart) $(U, \varphi)$ a homeomorphism mapping an open set $U$ of the manifold to an open set $V$ of an Euclidean space.  The manifold is called a $d-$dimension manifold if for each chart of an atlas we further have $V \subset \mathbb{R}^d$. That is there exists a neighborhood $U$ of each point $p$ of the manifold such that $U$ is homeomorphic to $\mathbb{R}^d$. Given $p \in U$, the chart $\varphi: (x^1, \dots, x^d)$ induces a basis $\Big (\frac{\partial}{\partial x^1}, \dots, \frac{\partial}{\partial x^d} \Big)_p $ on the tangent space $T_p\mathcal{M}$. Hence, a local representation of the metric of a Riemannian manifold in the chart $(U,\varphi)$ can be written as a positive definite matrix $\mathbf{G}(p) = ( g_{i, j})_{p, 0 \leq i, j \leq d} = (\langle \frac{\partial}{\partial x^i} | \frac{\partial}{\partial x^j} \rangle_p)_{0 \leq i, j \leq d}$ at each point $p \in U$. That is for $v, w \in T_p\mathcal{M}$ and $p \in U$, we have $\langle u | w \rangle_p = u^{\top} \mathbf{G}(p) w$. Since we propose to work in the ambient-like manifold ($\mathbb{R}^d$, $g$), there exists a global chart given by $\varphi=id$. Hence, for the following, we assume that we work in this coordinate system and so $\mathbf{G}$  will refer to the metric's matrix representation in this chart. The length of a curve $\gamma: [0, 1] \to \mathcal{M}$ travelling from $z_1 \in \mathcal{M}$ to $z_2 \in \mathcal{M}$ such that $\gamma(0) = z_1$ and $\gamma(1) = z_2$ is then given by 
\[
    \mathcal{L}(\gamma) = \int \limits _0 ^1 \lVert \dot{\gamma}(t) \rVert_{\gamma(t)} \mathrm{d}t = \int \limits _0 ^1 \sqrt{\langle \dot{\gamma}(t) | \dot{\gamma}(t) \rangle_{\gamma(t)}} \mathrm{d}t\,.
\]  
Curves minimizing $\mathcal{L}$ are called \textit{geodesics} and a distance $\mathrm{dist}$ between any $z_1, z_2 \in \mathcal{M}$ can be introduced as follows:
\begin{equation}\label{Eq: geodesic distance}
    \mathrm{dist}(z_1, z_2) = \inf_{\gamma} \mathcal{L}(\gamma) \hspace{5mm} \mathrm{s.t.} \hspace{5mm} \gamma(0) = z_1, \gamma(1) = z_2
\end{equation}
The manifold $\mathcal{M}$ is said to be \textit{geodesically complete} if all geodesic curves can be extended to $\mathbb{R}$. 
%In other words, at each point $p \in \mathcal{M}$ one may draw a \emph{straight} line indefinitely and in any direction. 

\subsection{A Geometry-Aware VAE}

    We now assume that the latent space is the Riemannian manifold $\mathcal{M}=(\mathbb{R}^d$, $\mathbf{G}$) with $\mathbf{G}$ being the Riemanian metric. Building upon the Hamiltonian VAE (HVAE) \cite{caterini_hamiltonian_2018}, we propose to exploit the assumed Riemannian structure of the latent space by using Riemannian Hamiltonian dynamics~\cite{girolami_riemann_2011} instead. The main goal remains the same and consists in using the Riemannian Hamiltonian Monte Carlo (RHMC) sampler to be able to enrich the variational posterior $q_{\phi}(z|x)$ such that it targets the true (unknown) posterior $p_{\theta}(z|x)$ while exploiting the properties of Riemannian manifolds. 
    
    \subsubsection{Riemannian Hamiltonian Monte Carlo Sampler}
    In a nutshell, given the Riemannian manifold $\mathcal{M}=(\mathbb{R}^d, \mathbf{G})$ and a target density $p_{\mathrm{target}}(z)$ we want to sample from with $z \in \mathcal{M}$, the idea of the RHMC sampler is to introduce a random variable $v \sim \mathcal{N}(0, \mathbf{G}(z))$ and rely on Riemannian Hamiltonian dynamics to sample from complex distributions. Likewise physical systems, $z$ is seen as the \emph{position} and $v$ as the \emph{velocity} of a particle traveling in $\mathcal{M}$ and whose potential energy $U(z)$ and kinetic energy $K(z, v)$ write 
    \begin{equation*}
    \begin{aligned}
         U(z) &= - \log p_{\mathrm{target}}(z)\\
         K(v, z)&= \frac{1}{2}\Big[\log\big( (2\pi)^d|\mathbf{G}(z)|\big )+ v^{\top} \textbf{G}^{-1}(z)v \Big ]\,.
    \end{aligned}
    \end{equation*}
     The sum of these energies give together the Hamiltonian $H(z, v)$ \cite{duane_hybrid_1987,leimkuhler_simulating_2004}. The RHMC simulates the evolution in time of such a particle by solving Hamilton's equations which can be integrated using a discretization scheme known as the generalized \emph{leapfrog} integrator.
     \begin{equation}\label{Eq: Riemann Stormer Verlet}
        \begin{aligned}
            v(t + \varepsilon/2) = v(t) &- \frac{\varepsilon}{2} \nabla_z H\Bigl(z(t), v(t+\varepsilon/2)\Bigr)\,,\\
            z(t + \varepsilon)      = z(t) &+ \frac{\varepsilon}{2} \Bigl[\nabla_{v} H\Bigl(z(t), v(t + \varepsilon/2)\Bigr) \\&+ \nabla_{v} H\Bigl(z(t+\varepsilon),  v(t + \varepsilon/2)\Bigr)\Bigr]\,,\\
            v(t + \varepsilon)   = v(t + \varepsilon/2) &- \frac{\varepsilon}{2} \nabla_z H \Bigl(z(t+\varepsilon), v(t+ \varepsilon/2) \Bigr)\,,
        \end{aligned}
    \end{equation}
    where $\varepsilon$ is the leapfrog stepsize. This integrator ensures that the target distribution is preserved by Hamiltonian dynamics and it was shown that it is also volume preserving and time reversible \cite{hairer_geometric_2006, leimkuhler_simulating_2004}. The RHMC then creates a Markov chain $(z^n)$ using this integrator. More precisely, given $z_0^n$, the current state of the chain, an initial \emph{velocity} is sampled $v_0 \sim \mathcal{N}(0, \mathbf{G}(z_0^n))$ and Eq.~\eqref{Eq: Riemann Stormer Verlet} are run $K$ times to move from ($z_0^n$, $v_0$) to ($z_K^n$, $v_K$). The proposal $z_K^n$ is then accepted with probability $\alpha = \min\Big(1, \frac{\exp (-H(z_K^n, v_K))}{\exp (-H(z_0^n, v_0))}\Big)$ and we iterate. It was shown that the chain $(z^n)$ converges to its stationary distribution $p_{\mathrm{target}}$\cite{duane_hybrid_1987, liu_monte_2008, neal_mcmc_2011}. We provide additional details in Appendix~\ref{appendix B}.

    \subsubsection{RHMC within the VAE}\label{Subsec: RHMC within VAE}
    Likewise the HVAE, we set $p_{\mathrm{target}}$ to the joint distribution $p_{\theta}(x, z) = p_{\theta}(x|z)p(z)$ since given an input data point $x\in \mathcal{X}$ we have $p_{\theta}(x, z) \propto p_{\theta}(z|x)$ the true posterior and so the RHMC sampler is guided by the gradient of the true posterior distribution through the leapfrog steps in Eq.~\eqref{Eq: Riemann Stormer Verlet}. Note that the target distribution is now tractable since both the prior and the conditional distribution are known. As in \cite{caterini_hamiltonian_2018}, we also use a tempering scheme  consisting in starting from an initial temperature $\beta_0$ (which can be learned) and decreasing the \textit{velocity} $v$ by a factor $\alpha_k = \sqrt{\beta_{k-1} / \beta_k}$ after each leapfrog step $k$ ($\beta_K=1$). The temperature is then updated:
    \[
        \sqrt{\beta_k} = \Biggl(\Biggl(1 - \frac{1}{\sqrt{\beta_0}} \Bigg) \frac{k^2}{K^2} + \frac{1}{\sqrt{\beta_0}} \Bigg)^{-1} \,.
    \] As discussed in \cite{salimans_markov_2015},  the acceptation/rejection step is omitted throughout training so that the flow is differentiable with respect to the encoder's parameters allowing optimization. Hence, the RHMC steps can be seen as a specific kind of normalizing flow informed both by the target distribution through Eq.~\eqref{Eq: Riemann Stormer Verlet} and by the latent space geometry thanks to the metric $\mathbf{G}$. Our intuition is that using the underlying geometry of the manifold in which the latent variables live would better guide the approximate posterior distribution leading to better variational posterior estimates. It must be nonetheless noted that the generalized \textit{leapfrog} integrator in Eq.~\eqref{Eq: Riemann Stormer Verlet} is no longer explicit and so requires the use of fixed point iterations to be solved. Fortunately, only few iterations are needed to stabilize the scheme (we use 3 iterations). To compute the gradient involved in the integrator we rely on automatic differentiation \cite{paszke_automatic_2017}. Finally, the volume preservation property of the flow leads to a closed form derivation of the extended approximate posterior:
\begin{equation*}
    \begin{aligned}
        q_{\phi}(z_K, v_K|x) &= q_{\phi}(z_0|x) p(v_0|z_0) \prod_{k=1}^K |\det \mathbf{J}_{g^k}{}|\\
    &= q_{\phi}(z_0|x) p(v_0|z_0) \prod_{k=1}^K \Big(\frac{\beta_{k-1}}{\beta_k}\Big)^{d/2} \,,
    \end{aligned}
\end{equation*}
where $\mathbf{J}_{g^k}$ is the Jacobian of $k^{\text{th}}$ leapfrog step. Now, an unbiased estimate of the marginal $p_{\theta}(x)$ is given by:
\begin{equation}\label{eq: log-like estimate rhvae}
    \hat{p}_{\theta}(x) = \frac{p_{\theta}(x, z_K, v_K)}{q_{\phi}(z_K, v_K|x)} = \frac{p_{\theta}(x|z_K)p(v_K|z_K)q(z_K)}{q_{\phi}(z_0|x)p(v_0|z_0)\beta_0^{-d/2}}\,.
\end{equation}
    Note that the expression of the variational posterior remains computable so that the ELBO remains tractable. 
\begin{equation}\label{eq: RHVAE ELBO}
    \mathrm{ELBO}_{\mathrm{Riemannian}} = \mathbb{E}_{(z_0, v_0) \sim q_{\phi}(\cdot, \cdot)}[\log \widehat{p}_{\theta}(x)]
\end{equation}
    We provide the training algorithm in Appendix~\ref{appendix B}.
    
    \subsubsection{The Metric} 
    Since the latent space is now seen as the Riemannian manifold $(\mathbb{R}^d, \mathbf{G})$, it is in particular characterised by the Riemannian metric $\mathbf{G}$ whose choice is crucial. While several attempts have been made to try to put a Riemannian structure over the latent space of VAEs~\cite{arvanitidis_latent_2018,chen_metrics_2018,shao_riemannian_2018,frenzel_latent_2019, kalatzis_variational_2020, arvanitidis_geometrically_2020}, the proposed metrics involved the Jacobian of the generator function which is hard to use in practice and is constrained by the generator network architecture. As a consequence, we instead decide to rely on the idea of Riemannian metric learning~\cite{lebanon_metric_2006}. Hence, we propose to use a parametric metric inspired from~\cite{louis_computational_2019} as follows:
    \begin{equation}\label{eq: metric}
        \mathbf{G}^{-1}(z) = \sum_{i=1}^N L_{\psi_i} L_{\psi_i}^{\top} \exp \Big(-\frac{\lVert z - c_i \rVert_2^2}{T^2} \Big) + \lambda I_d \,,
    \end{equation}
    where $N$ is the number of observations, $L_{\psi_i}$ are lower triangular matrices with positive diagonal coefficients learned from the data and parametrized with neural networks, $c_i$ are referred to as the \emph{centroids} and correspond to the mean $\mu_{\phi}(x_i)$ of the encoded distributions of the latent variables $z_i$ $(z_i \sim q_{\phi}(z_i|x_i) = \mathcal{N}(\mu_{\phi}(x_i), \Sigma_{\phi}(x_i))$, $T$ is a temperature scaling the metric close to the \emph{centroids} and $\lambda$ is a regularization factor that also scales the metric tensor far from the latent codes. The shape of this metric is very powerful since we have access to a closed-form expression of the inverse metric tensor which is usually useful to compute shortest paths (through the exponential map). Moreover, this metric is very smooth, differentiable everywhere and allows scaling the Riemannian volume element $\sqrt{\det \mathbf{G}(z)}$ far from the data very easily through the regularization factor $\lambda$.

    \subsubsection{Training Process}
\begin{figure}[!t]
      \centering
      \subfloat{\includegraphics[width=3.5in]{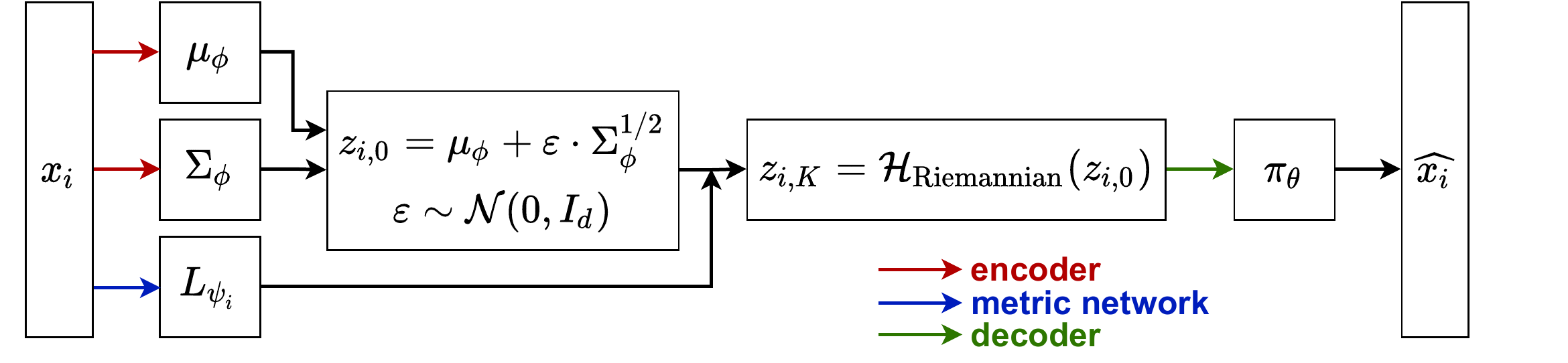}}
      \caption{Geometry-aware VAE framework. Neural networks are highlighted with the colored arrows and $\mathcal{H}_{\mathrm{Riemannian}}$ are the normalizing flows using Riemannian Hamiltonian equations.}
      \label{Fig: RHVAE framework}
    \end{figure}

    The model's architecture is displayed in Fig.~\ref{Fig: RHVAE framework}. The idea is to encode the input data points $x_i$ and so get the means $\mu_{\phi}(x_i)$ of the posterior distributions associated with the encoded latent variables $z_{i, 0} \sim \mathcal{N}(\mu_{\phi}(x_i), \Sigma_{\phi}(x_i))$. These means are then used to update the metric centroids $c_i$. In the mean time, the input data points $x_i$ are fed to another neural network which outputs the matrices $L_{\psi_i}$ used to update the metric. The updated metric is then used to \emph{sample} $z_{i,K} $ from $z_{i,0}$ using Eq.~\eqref{Eq: Riemann Stormer Verlet} as explained in Sec.~\ref{Subsec: RHMC within VAE}. The $z_{i,K}$ are then fed to the decoder network which outputs the parameters $\pi_{\theta}$ of the conditional distribution $p_{\theta}(x|z)$. The reparametrization trick is used to sample $z_{i,0}$ as is common and since the Riemannian Hamiltonian equations are \emph{deterministic} with respect to $z$, back-propagation can be performed. A scheme of the \emph{geometry-aware} VAE model framework can be found in Fig.~\ref{Fig: RHVAE framework}. In the following, we will refer to the proposed model either as \emph{geometry-aware VAE} or \emph{RHVAE} for short. An implementation using PyTorch~\cite{paszke_automatic_2017} is available in the supplementary materials.

    \subsubsection{Discussion on the Posterior Expressiveness}
    
    Theoretically, using \emph{geometry-aware} Hamiltonian normalizing flows should conduct to a better estimate of the true posterior $p_{\theta}(z|x)$ and so a better ELBO leading to a potentially higher likelihood $p_{\theta}(x)$. To validate this empirically, we report the estimated log-likelihood computed using Importance Sampling with the approximate posterior and Eq.~\eqref{eq: log-like estimate vae} and Eq.~\eqref{eq: RHVAE ELBO}. We use 100 importance samples and compute it three times on MNIST test set. Hamiltonian based models use 3 leapfrog steps. We also report the value of the ELBO and compute $D_{KL}(q_{\phi}(z|x) \lVert p_{\theta}(z|x))$. As shown in Table~\ref{table:NLL comparison}, using \emph{geometry-aware} normalizing flows leads to a higher estimated $p_{\theta}$ and a smaller gap between the estimated true posterior $ p_{\theta}(z|x)$ and the variational approximation $q_{\phi}(z|x)$ measured by the KL divergence between both distributions. Note that all models are trained with the same architectures and training settings.
    
    \begin{table}[ht]
    \caption{Effect of geometrical considerations on the estimated log-likelihood and ELBO on MNIST test set.}
    \label{table:NLL comparison}
    \begin{center}
    \scriptsize
    \begin{tabular}{c | c | c | c }
    \hline
    Model & $\log p_{\theta}(x)$ $\uparrow$ & ELBO & $D_{KL}(q_{\phi}(z|x) \lVert p_{\theta}(z|x))$ $\downarrow$ \\
    \hline
    VAE   & -92.94 (0.01) & -100.06 (0.09)& 7.12 (0.09) \\
    HVAE  & -85.33 (0.01) & -88.93 (0.02) & 3.61 (0.02) \\
    RHVAE & -82.64 (0.01) & -86.21 (0.04) & 3.57 (0.03)\\
    \hline
    \end{tabular}
    \end{center}
    \end{table}

    \subsubsection{Sampling from the Latent Space}\label{Sec: Sampling proposal}
    In this paper, we propose to amend the standard sampling procedure of classic VAEs after training to better exploit the Riemannian structure of the latent space. The \emph{geometry-aware} VAE is indeed here seen as a tool able to capture the intrinsic latent structure of the data and so we propose to exploit this property directly within the generation procedure. This differs greatly from the standard fully probabilistic view where the prior distribution is used to generate new data. We believe that such an approach remains far from being optimal when one considers small data sets since, depending on its choice, the prior may either poorly prospect the latent space or sample in locations without any usable information. In that respect, our approach can be seen as part of the recently proposed prior learning based methods or methods relying on \emph{ex-post} density estimation discussed earlier. Some of these methods were indeed proposed on the ground that there may exist a mismatch between the chosen prior distribution $p(z)$ and the optimal one given by the the aggregated posterior distribution $q(z) = \mathbb{E}_{x \sim p_{\mathrm{data}}(x)} [q_{\phi}(z|x)] $ \cite{hoffman_elbo_2016, dai_diagnosing_2018, bauer_resampled_2019, ghosh_variational_2020}, where $p_{\mathrm{data}}(x)$ is the empirical distribution of the data \cite{tomczak_vae_2018}. Moreover, since our method is mainly about increasing the expressiveness of the variational posterior $q_{\phi}$ there exists no apparent reason that the latent codes are distributed according to the prior either. However, instead of \emph{learning} a prior, we propose to directly use the metric that provides information on the geometry of the latent space as discussed and illustrated in Sec.~\ref{Sec: Sampling discussion} and Sec.~\ref{Sec: Generation comparison}. We indeed propose to sample from the following distribution:
    \begin{equation}\label{Eq: Target distribution}
    p(z) = \frac{\mathbf{1}_S(z) \sqrt{\det \mathbf{G}^{-1}(z)}}{\int \limits _{\mathbb{R}^d} \mathbf{1}_S(z) \sqrt{\det \mathbf{G}^{-1}(z) dz}}\,,
    \end{equation}
    where $S$ is a compact set\footnote{Take for instance $\{z \in \mathcal{Z}, \lVert z \rVert \leq 2 \cdot \max_i \lVert c_i \rVert \}$} so that the integral is well defined. Fortunately, since we use a parametrized metric given by Eq.~\eqref{eq: metric} and whose inverse has a closed form, it is pretty straightforward to evaluate the numerator of Eq.~\eqref{Eq: Target distribution}. Then, classic MCMC sampling methods can be employed to sample from $p$ on $\mathbb{R}^d$. In this paper, we propose to use the Hamiltonian Monte Carlo (HMC) sampler~\cite{neal_hamiltonian_2005} since the gradient of the log-density is computable. We provide some additional details in Appendix~\ref{appendix C}. 

  \begin{figure}[!t]
    \centering
    \subfloat{\includegraphics[width=1.5in]{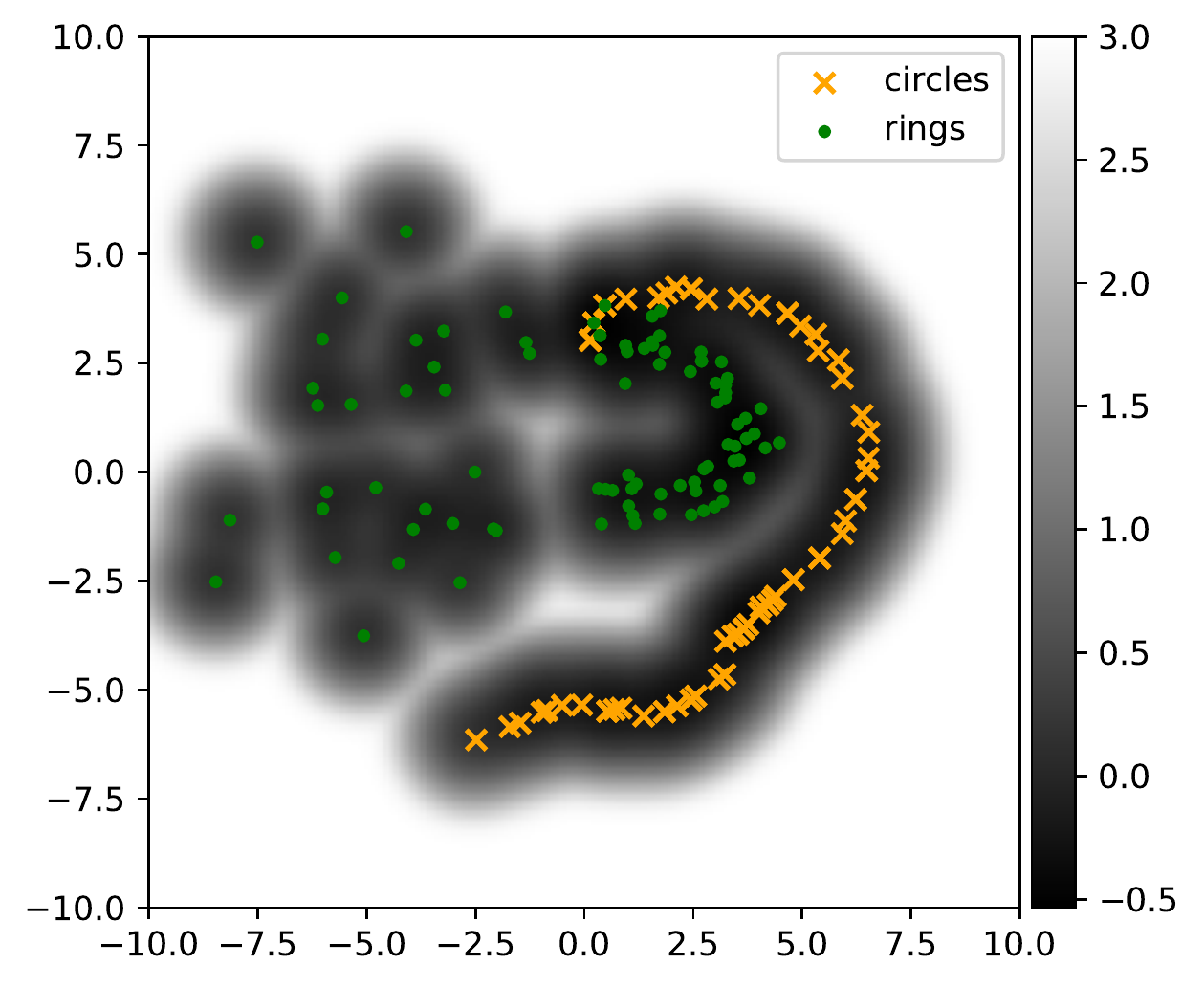}}
    \hfil
    \subfloat{\includegraphics[width=1.44in]{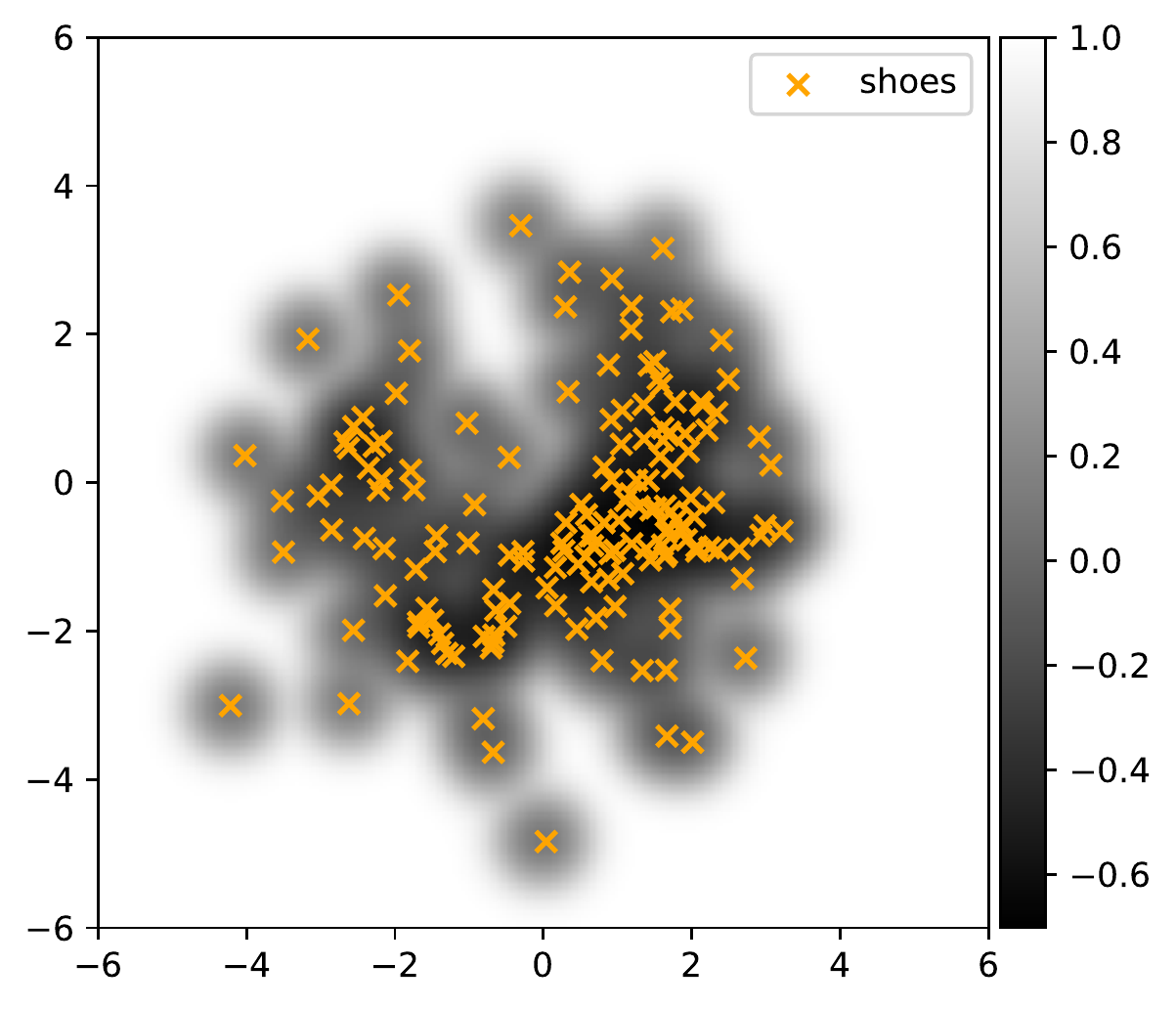}}
    \centering
    \vfil
    \vspace{-1.2em}
    \subfloat{\includegraphics[width=1.48in]{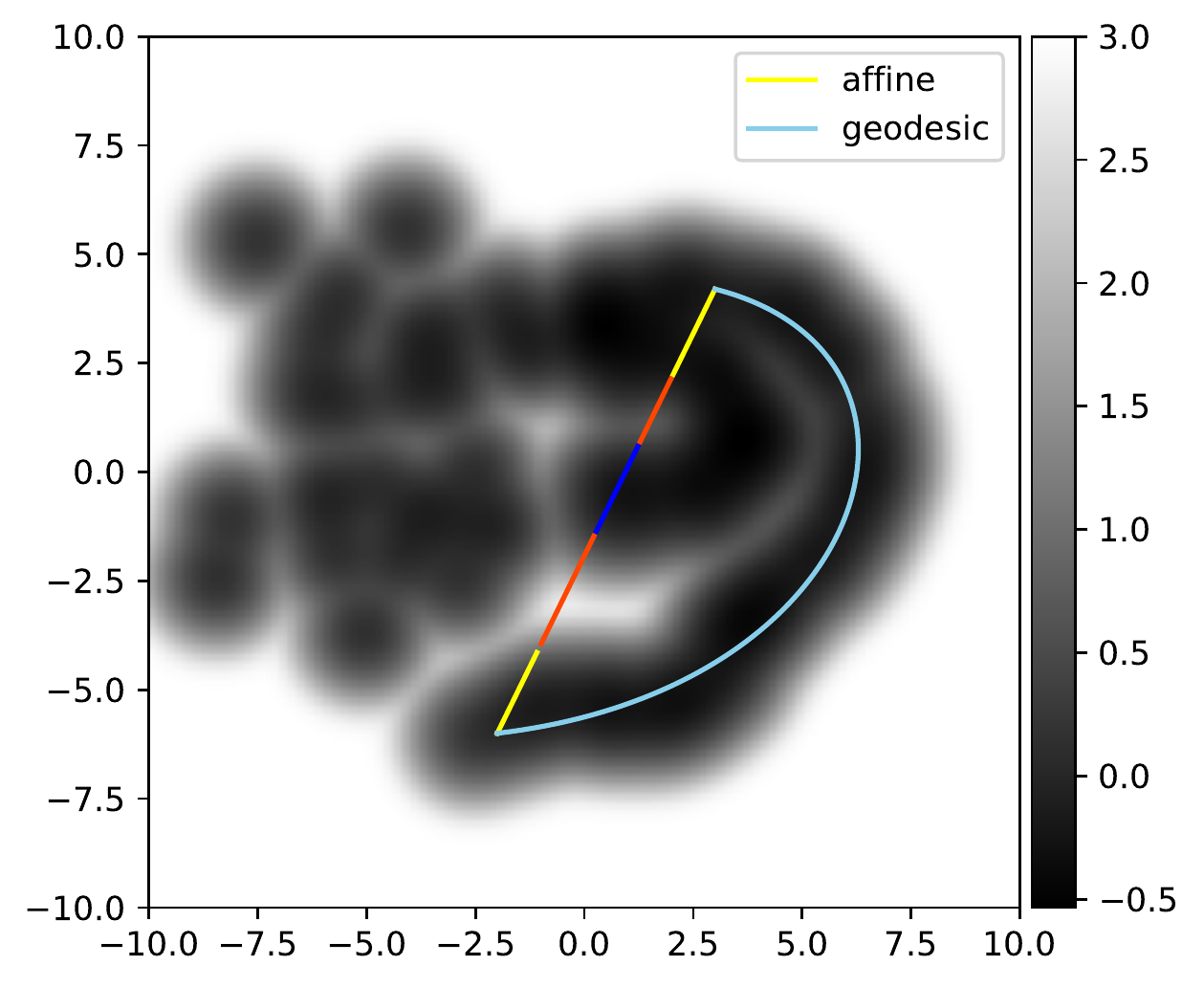}}
    \hfil
    \subfloat{\includegraphics[width=1.44in]{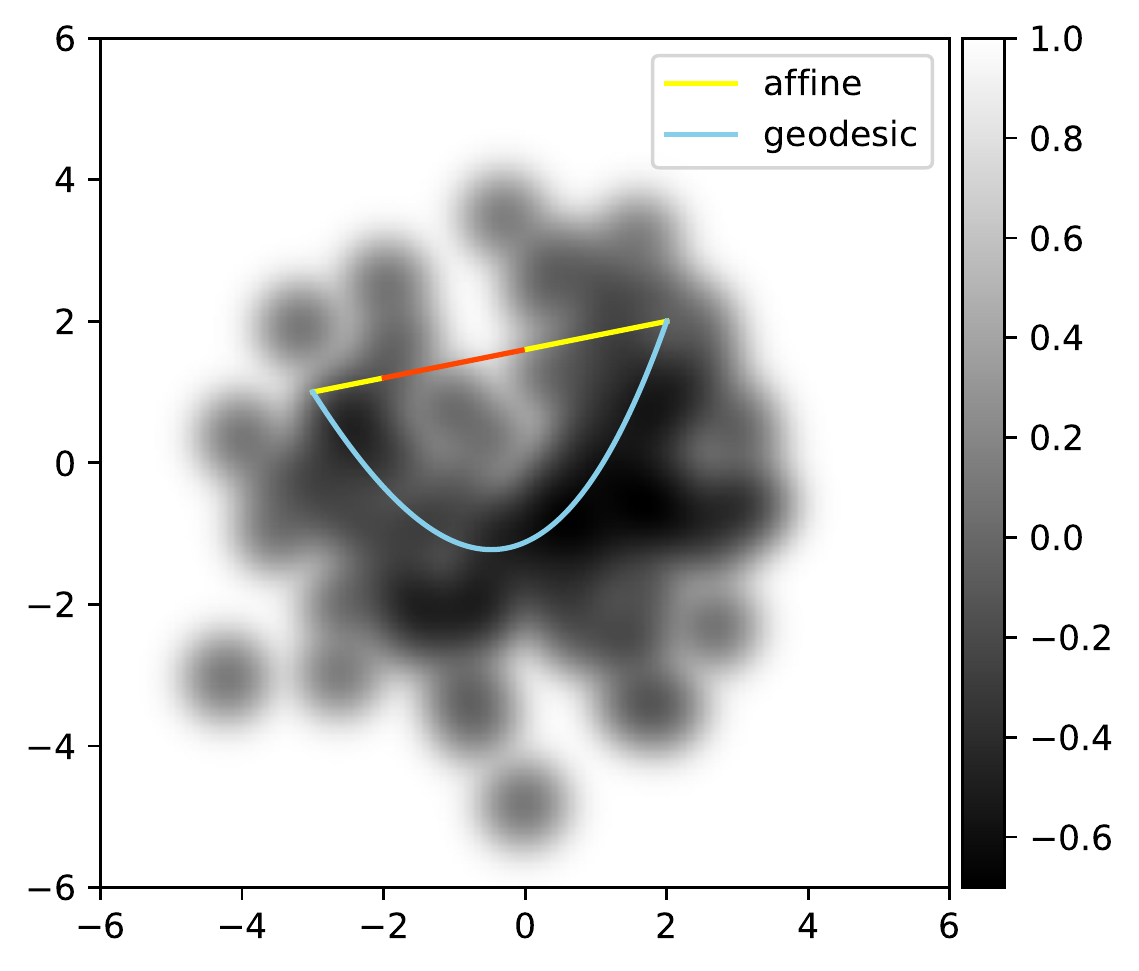}}
    \vfil
    \vspace{-1em}
    \subfloat{\includegraphics[width=3.1in]{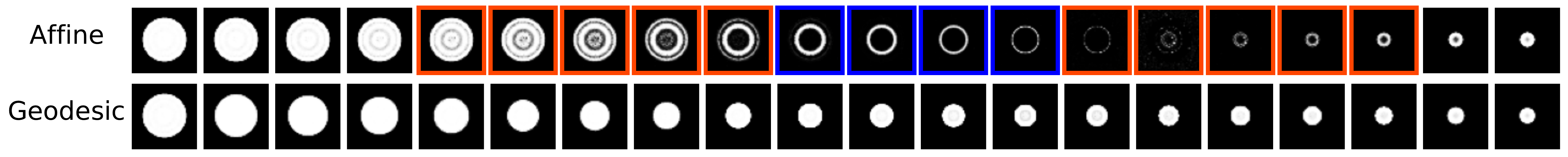}}
    \vfil
    \centering
    \vspace{-1em}
    \subfloat{\includegraphics[width=3.1in]{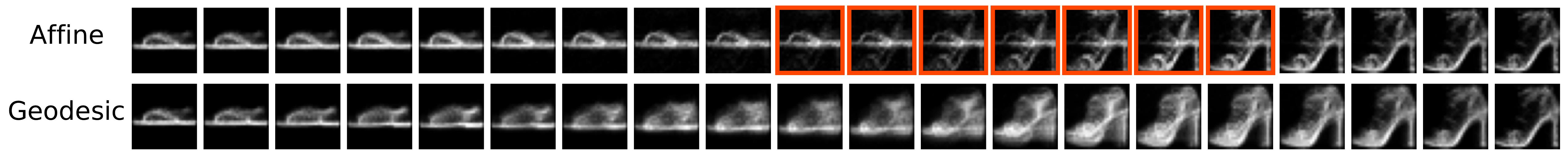}}
    \caption{Geodesic interpolations under the learned metric in two different latent spaces. Top: Latent spaces with the log metric volume element presented in gray scale. Second row: The resulting interpolations under the Euclidean metric or the Riemannian metric. Third row: The learned manifolds and corresponding decoded samples. Bottom: Decoded samples all along the interpolation curves.}
    \label{Fig: Geodesic computation}
    \end{figure}

    \begin{figure*}
      \centering
      \captionsetup[subfigure]{position=above, labelformat = empty}
      \subfloat[Vanilla VAE]{\includegraphics[width=1.5in]{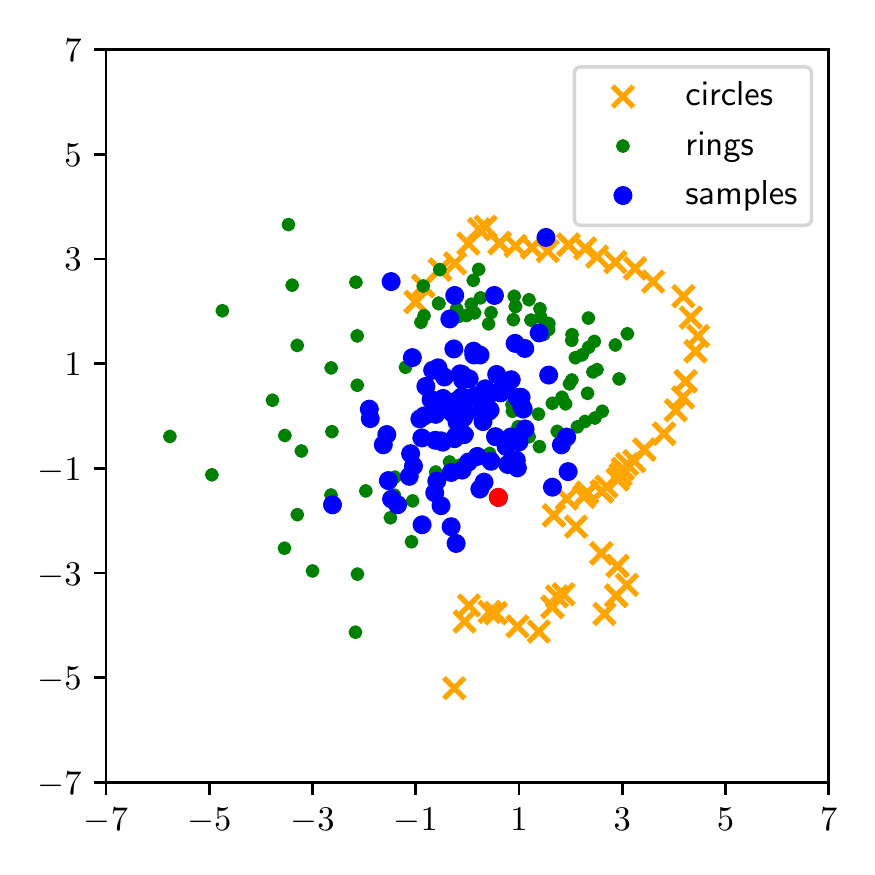}}
      \hfil
      \hspace{.05in}
      \subfloat[VAE - VAMP prior]{\includegraphics[width=1.5in]{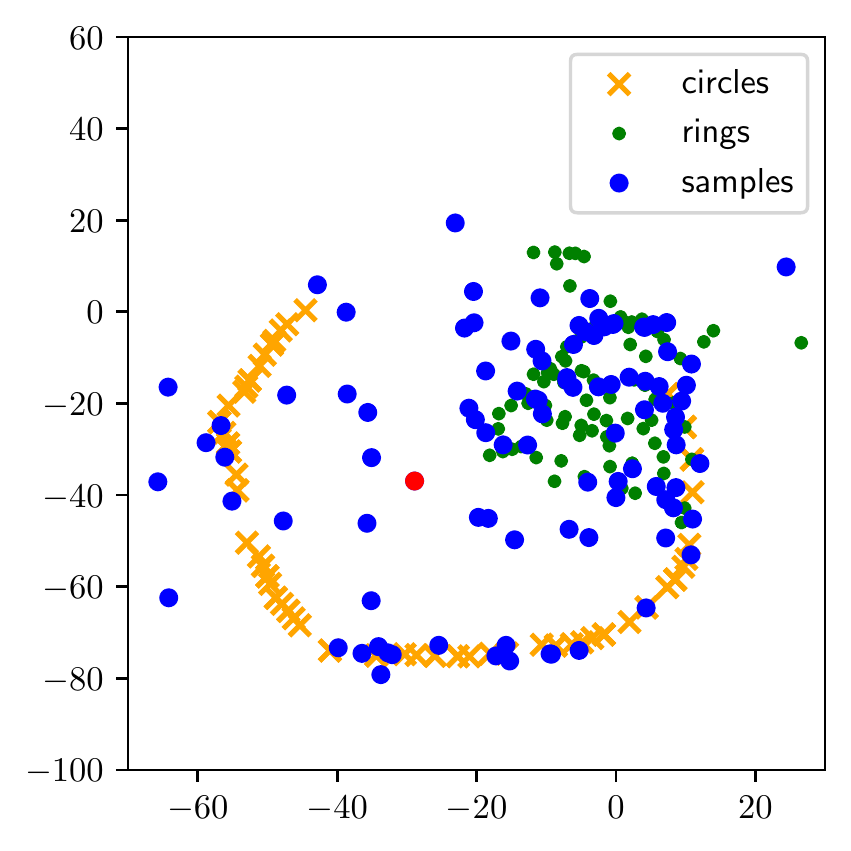}}
      \hfil
      \hspace{0.05in}
      \subfloat[Geometry-aware VAE - $\mathcal{N}(0, I_d)$]{\includegraphics[width=1.8in]{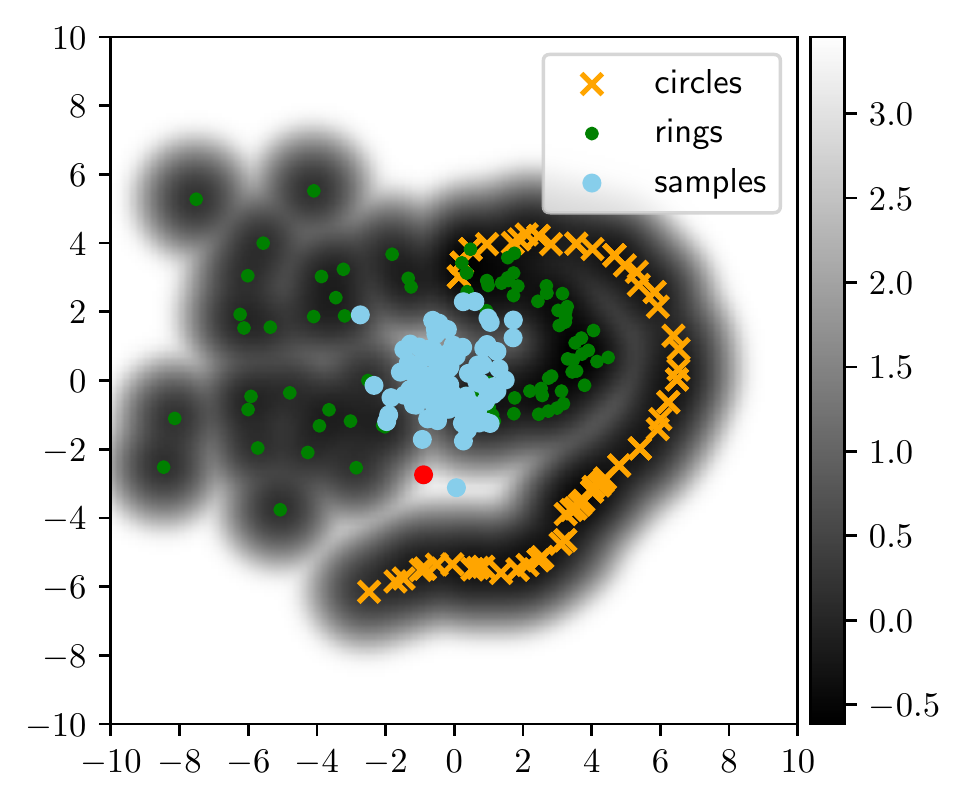}}
      \hspace{-0.1in}
      \subfloat[Geometry-aware VAE - Ours]{\includegraphics[width=1.87in]{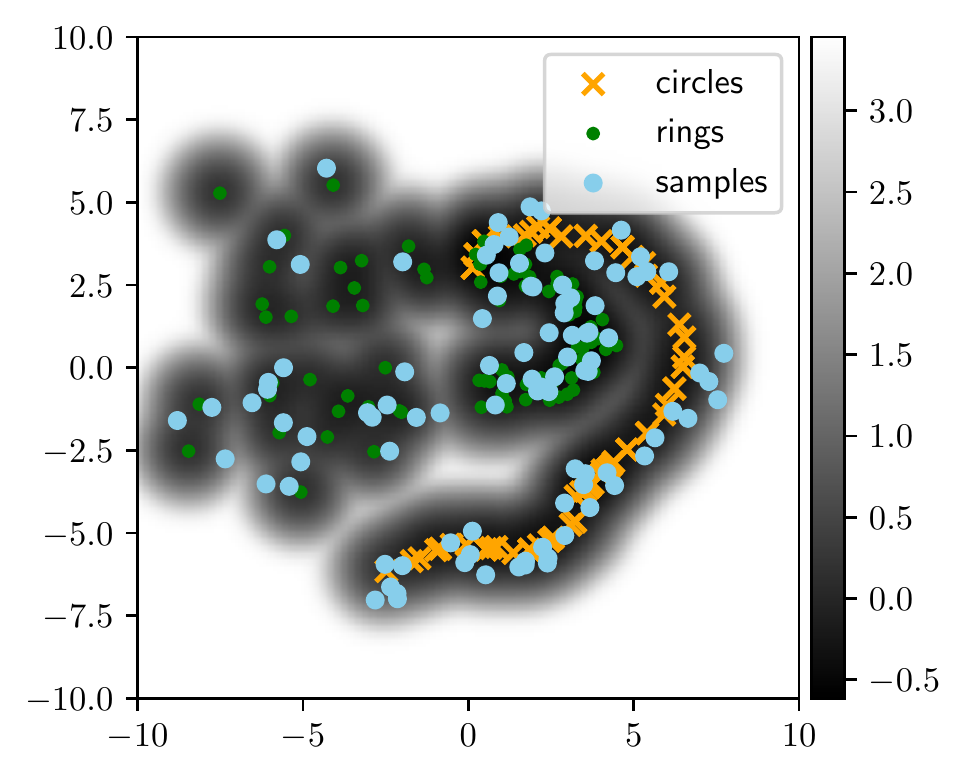}}
      %\hspace{0.3in}
      \vfil
      \vspace{-1.4em}
      \hspace{-0.17in}
      \subfloat{\includegraphics[width=1.3in]{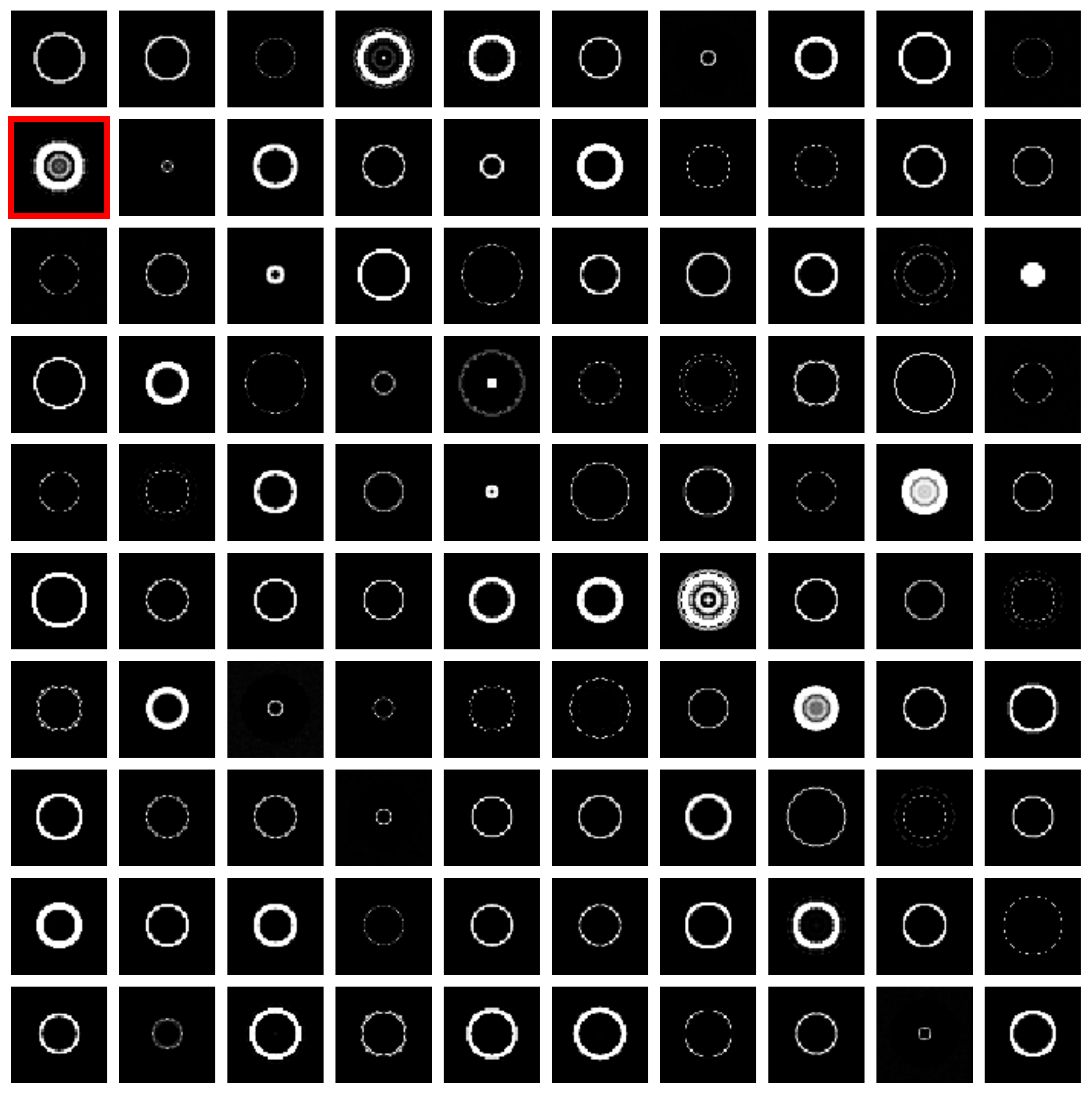}}
      \hfil
      \hspace{-0.05in}
      \subfloat{\includegraphics[width=1.3in]{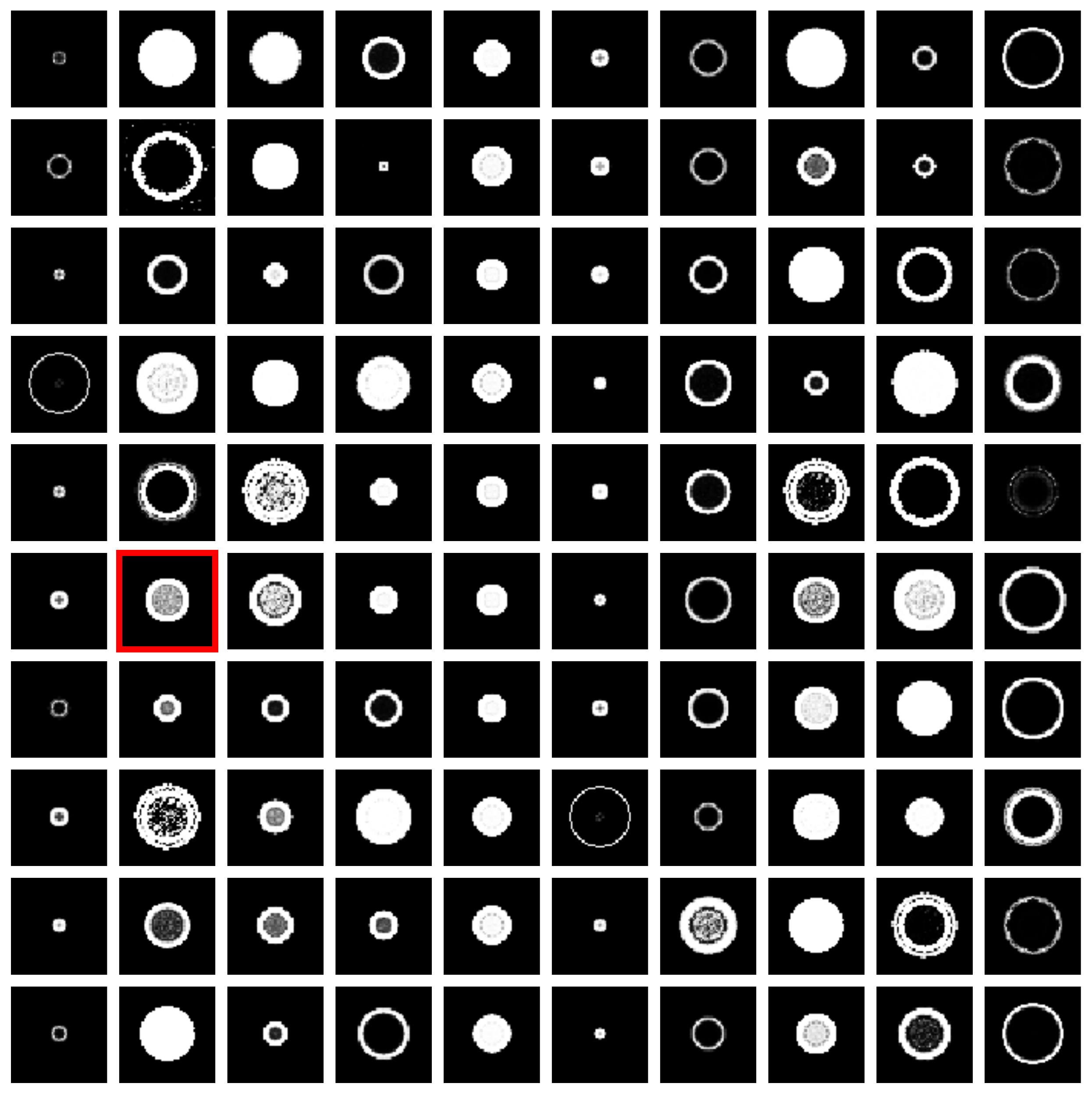}}
      \hfil
      \hspace{-0.00in}
      \subfloat{\includegraphics[width=1.3in]{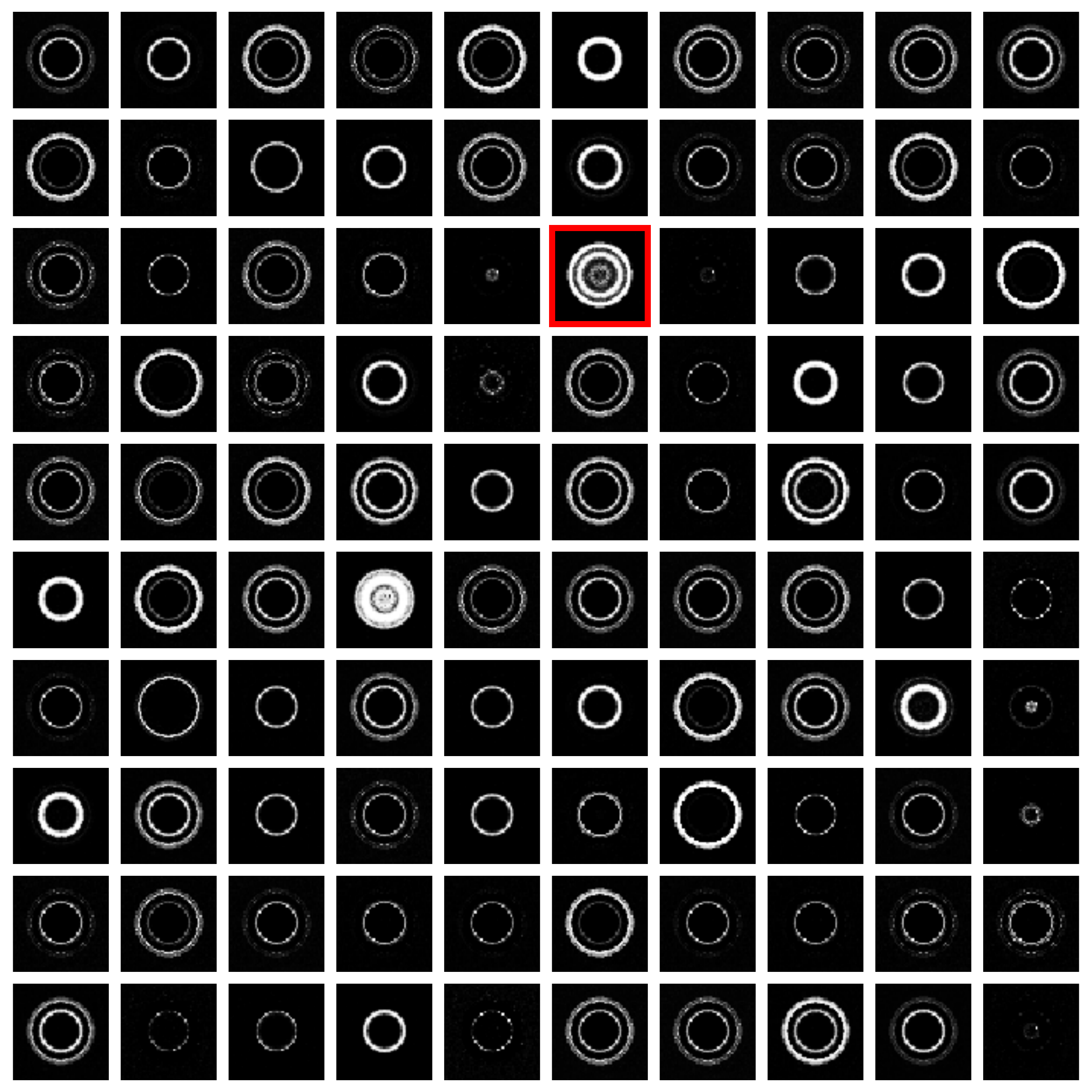}}
      \hfil
      \hspace{0.05in}
      \subfloat{\includegraphics[width=1.3in]{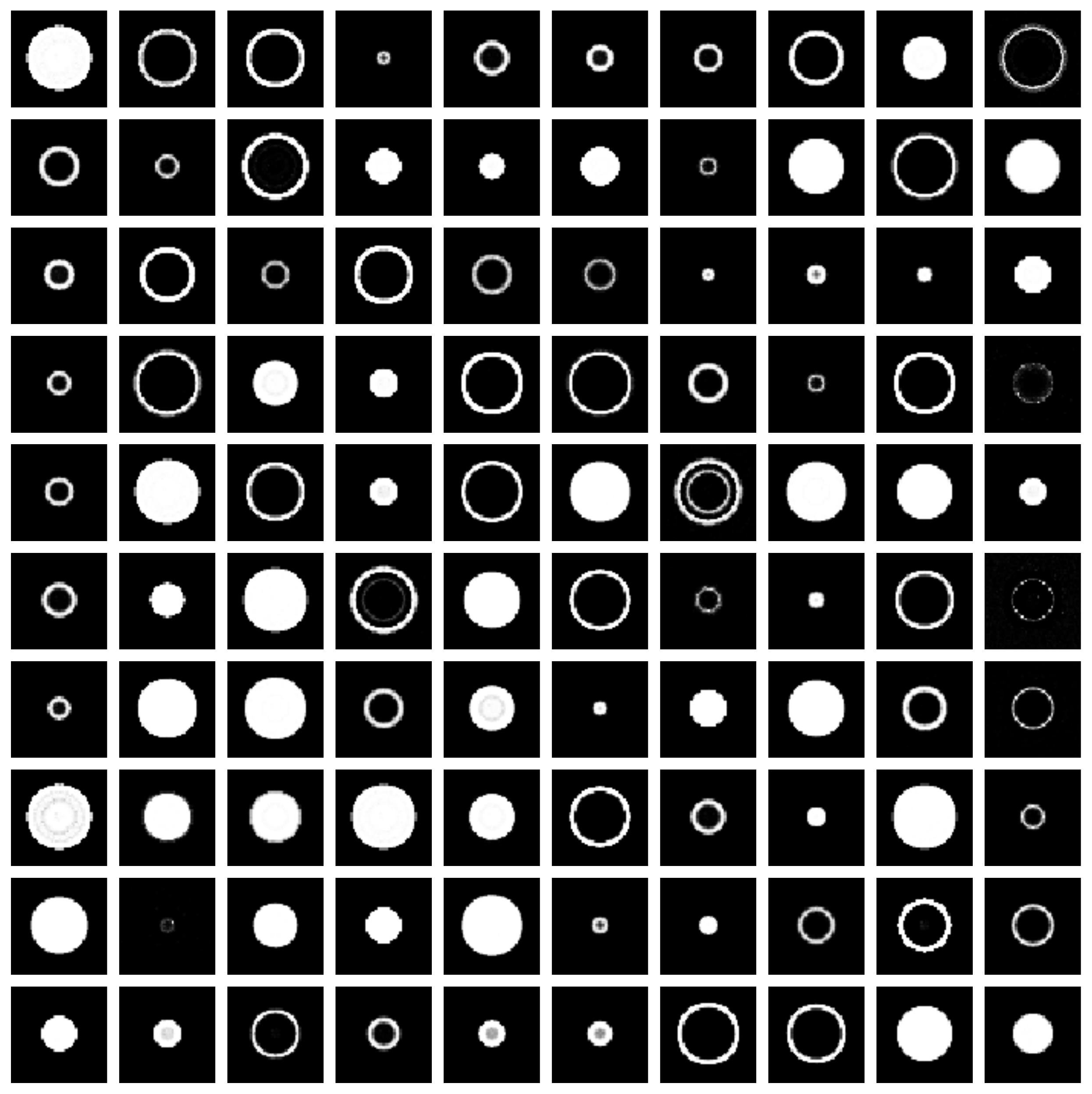}}
      \caption{VAE sampling comparison. Top: The learned latent space along with the means $\mu_{\phi}(x_i)$ of the latent code distributions (colored dots and crosses) and 100 latent space samples (blue dots) using either the prior distribution or the proposed scheme. For the \emph{geometry-aware} VAEs, the log metric volume element is presented in gray scale in the background. Bottom: The 100 corresponding decoded samples in the data space.}
      \label{Fig: Sampling comparison}
    \end{figure*}

    \subsubsection{Discussion on the Sampling Distribution}\label{Sec: Sampling discussion}
    One may wonder what is the rationale behind the use of the distribution $p$ formerly defined in Eq.~\eqref{Eq: Target distribution}. By design, the metric is such that the metric volume element $\sqrt{\det \mathbf{G}(z)}$ is scaled by the factor $\lambda$ far from the encoded data points. Hence, choosing a relatively small $\lambda$ imposes that shortest paths travel through the most populated area of the latent space, \emph{i.e.} next to the latent codes. As such, the metric volume element can be seen as a way to quantify the amount of information contained at a specific location of the latent space. The smaller the volume element the more information we have access to. Fig.~\ref{Fig: Geodesic computation} illustrates well these aspects. On the first row are presented two learned latent spaces along with the log of the metric volume element displayed in gray scale for two different data sets. The first one is composed of 180 binary disks and rings of different diameters and thicknesses while the second one is composed of 160 samples extracted from the FashionMNIST data set~\cite{xiao_fashion-mnist_2017}. The means $\mu_{\phi}(x_i)$ of the distributions associated with the latent variables are presented with the crosses and dots for each class. As expected, the metric volume element is smaller close to the latent variables since small $\lambda$'s were considered ($10^{-3}$ resp. $10^{-1}$). A common way to study the learned Riemannian manifold consists in finding geodesic curves, \emph{i.e.} the shortest paths with respect to the learned Riemannian metric. Hence, on the second row of Fig.~\ref{Fig: Geodesic computation}, we compare two types of interpolation in each latent space. For each experiment, we pick two points in the latent space and perform either a linear or a geodesic interpolation (\emph{i.e.} using the Riemannian metric). The bottom row illustrates the decoded samples all along each interpolation curve. The first outcome of such an experiment is that, as expected, geodesic curves travel next to the codes and so do not explore areas of the latent space with no information whereas linear interpolations do. Therefore, decoding along geodesic curves produces far better and more meaningful interpolations in the input data space since in both cases we clearly see the starting sample being progressively distorted until the path reaches the ending point. This allows for instance interpolating between two shoes and keep the intrinsic topology of the data all along the path since each decoded sample on the interpolation curve looks like a shoe. This is made impossible under the Euclidean metric where shortest paths are straight lines and so may travel through areas of least interest. For instance, the affine interpolation travels through areas with no latent data and so produces decoded samples that are mainly a superposition of samples (see the red lines and corresponding decoded samples framed in red) or crosses areas with codes belonging to the other class (see the blue line and the corresponding blue frames). This study demonstrates that most of the information in the latent space is contained next to the codes and so, if we want to generate new samples that look-like the input data, we need to sample around them and that is why we elected the distribution in Eq.~\eqref{Eq: Target distribution}. 
    Noteworthy is the fact that  likewise \cite{ghosh_variational_2020}, the prior $\mathcal{N}(0, I_d)$ is now only reduced to a latent code regularizer during training ensuring that the covariances do not collapse to $0_d$ and the codes remains close to the origin and is never used to generate samples.

    \subsection{Generation Comparison}\label{Sec: Generation comparison}
    
        In this section, we propose to compare the new generation procedure with other generation methods in the context of low sample size data sets. 

        \subsubsection{Qualitative Comparison} \label{Subsec: qualitative comp}
        First, we validate the proposed generation method on a hand-made synthetic data set composed of 180 binary disks and rings of different diameters and thicknesses (see Appendix~\ref{appendix E}). We then train 1) a vanilla VAE, 2) a VAE with VAMP prior~\cite{tomczak_vae_2018}, 3) a \emph{geometry-aware} VAE but using the prior to generate and 4) a \emph{geometry-aware} VAE with the proposed generation scheme, and compare the generated samples. Each model is trained until the ELBO does not improve for 20 epochs and any relevant parameter setting is made available in Appendix~\ref{appendix D}. In Fig.~\ref{Fig: Sampling comparison}, we compare the sampling obtained with each model. The first row shows the learned latent spaces along with the means of the encoded training data points for each class (crosses and dots) and 100 samples issued by the generation methods (blue dots). For the RHVAE models, the log metric volume element $ \sqrt{\det \mathbf{G}}$ is also displayed in gray scale in the background. The bottom row shows the resulting 100 decoded samples in the data space. 
        
        The first outcome of this experiment is that sampling from the prior distribution leads to a quite poor latent space prospecting. This drawback is very well illustrated when a standard Gaussian distribution is used to sample from the latent space (see 1$^{\mathrm{st}}$ and 3$^{\mathrm{rd}}$ column of the 1$^{\mathrm{st}}$ row). The prior distribution having a higher mass close to zero will insist on latent samples close to the origin. Unfortunately, in such a case, latent codes close to the origin only belong to a single class (rings). Therefore, even though the number of training samples was roughly the same for disks and rings, we end up with a model over-generating samples belonging to a certain class (rings) and even to a specific type of data within this very class. This undesirable effect seems even ten-folded when considering the \emph{geometry-based} VAE model since adding MCMC steps in the training process, as explained in Fig.~\ref{Fig: RHVAE framework}, tends to stretch the latent space. It can be nonetheless noted that using a multi-modal prior such as the VAMP prior mitigates this and allows for a better prospecting.
        However, such a model remains hard to fit when trained with small data sets as it may overfit (resp. underfit) the training samples if the number of pseudo-inputs is too high (resp. low). Another limitation of prior-based generation methods lies in their inability to assess a given sample quality. They may indeed sample in areas of the latent space containing very few information and so conduct to generated samples that are meaningless. This appears even more striking when small data sets are considered. An interesting observation that was noted among others in~\cite{arvanitidis_latent_2018} is that neural networks tend to interpolate very poorly in \textit{unseen} locations (i.e. far from the training data points). When looking at the \textit{decoded} latent samples (bottom row of Fig.~\ref{Fig: Sampling comparison}) we eventually end up with the same conclusion. Actually, it appears that the networks interpolate quite linearly between the training data points in our case. This may be illustrated for instance by the red dots in the latent spaces in Fig.~\ref{Fig: Sampling comparison} whose corresponding decoded sample is framed in red. The sample is located \emph{between} two classes and when decoded it produces an image mainly corresponding to a superposition of samples belonging to different classes. This aspect is also supported by the observations made when discussing the relevance of geodesic interpolations on Fig.~\ref{Fig: Geodesic computation} of Sec.~\ref{Sec: Sampling discussion}.
        Therefore, these drawbacks may conduct to a (very) poor representation of the actual data set diversity while presenting quite a few \textit{irrelevant} samples. Obviously the notion of \textit{irrelevance} is here disputable but if the objective is to represent a given set of data we expect the generated samples to be close to the training data while having some specificities to enrich it. Impressively, sampling against the inverse of the metric volume element as proposed in Sec.~\ref{Sec: Sampling proposal} allows for  a far more meaningful sample generation. Furthermore, the new sampling scheme avoids regions with no latent code, which thus contain poor information, and focuses on areas of interest so that almost every decoded sample is visually satisfying. Similar effects are observed on reduced versions of EMNIST~\cite{cohen_emnist_2017},  MNIST~\cite{lecun_mnist_1998} and FashionMNIST data sets and higher dimensional latent spaces (dimension 10) where samples are most of the time degraded when the classic generation is employed while the new one allows the generation of more diverse and sharper samples (see Appendix~\ref{appendix E}). Finally, the proposed method does not overfit the train data since the samples are not located on the centroids. The quantitative metrics of the next section also support this point.

    \begin{table*}[!ht]
      \caption{\emph{GAN-train} (the higher the better) and \emph{GAN-test} (the closer to the baseline the better) scores. A benchmark DenseNet model is trained with five independent runs on the generated data $\mathcal{S}_g$ (resp. the \emph{real} train set $\mathcal{S}_{\mathrm{train}}$) and tested on the \emph{real} test set $\mathcal{S}_{\mathrm{test}}$ (resp. $\mathcal{S}_g$) to compute the \emph{GAN-train} (resp. \emph{GAN-test}) score. 1000 synthetic samples per class are considered for $\mathcal{S}_g$ so that it matches the size of $\mathcal{S}_{\mathrm{test}}$.}
      \label{table: GAN - scores}
      \begin{center}
      \setlength{\tabcolsep}{3pt}
      \begin{tabular}{l |c c |c c |c c} 
        \cline{2-7}
        &\multicolumn{2}{c|}{\emph{reduced} MNIST} & \multicolumn{2}{c|}{\emph{reduced} MNIST} & \multicolumn{2}{c}{\multirow{2}{*}{\emph{reduced} EMNIST}} \\
        &\multicolumn{2}{c|}{(balanced)} & \multicolumn{2}{c|}{(unbalanced)} &  \\
      \hline
      Metric & GAN-train & GAN-test & GAN-train & GAN-test & GAN-train & GAN-test \\
      \hline
      \hline
      Baseline                        & $90.6 \pm 1.2 $ & - & $ 82.8 \pm 0.7 $  & - & $ 84.5  \pm 1.3 $ & - \\ 
      \hline
      VAE - $\mathcal{N}(0, I_d)$     & $83.4 \pm 2.4 $ & $ 67.1 \pm 4.9 $  & $ 74.7 \pm 3.2$  & $ 52.8 \pm 10.6 $  & $75.3 \pm 1.4$ & $ 54.5 \pm 6.5 $ \\
      VAMP                            & $72.8 \pm 6.7 $ & $ 77.6 \pm 4.8 $  & $ 68.2 \pm 6.6$  & $ 76.7 \pm 11.0  $  & $ 70.7 \pm 8.0 $ & $ 69.0 \pm 6.4 $ \\
      VAE - GMM & $82.9 \pm 2.4 $& $76.5 \pm 8.9$ & $ 74.4 \pm 3.8 $ & $ 68.4 \pm 12.3 $ & $ 74.0 \pm 2.6 $ & $ 57.6 \pm 4.6 $\\
      %RAE - L2 - GMM(2) & $89.3 \pm 1.1$ & $92.5 \pm 1.6$ & $84.0 \pm 3.3$ & $\underline{87.6} \pm 4.8 $ & $81.7 \pm 1.8$ & $86.0 \pm 3.0$ \\
      %RAE - L2 - GMM(10)& $89.6 \pm 2.8$ & $\underline{95.0} \pm 1.2$ & $84.2 \pm 3.6$ & $\underline{94.2} \pm 2.9 $ & $ 82.1 \pm 3.5 $ & $\underline{88.6} \pm 2.2$ \\
      %RAE - L2 - GMM(20)& $ 90.3 \pm 1.2 $ & $\underline{95.9} \pm 1.5$ & $ 83.0 \pm 3.1 $ & $ \underline{94.6} \pm 3.1 $ & $ 84.2 \pm 0.7 $ & $\underline{90.7} \pm 2.1$\\
      RAE - GMM(2) & $ 90.8 \pm 3.0 $ & $ 91.7 \pm 1.9 $ & $ 85.5 \pm 1.3 $ & $83.8 \pm 6.2$  & $ 80.3 \pm 1.5 $ & $ 69.8 \pm 7.2 $ \\
      RAE - GMM(10) & $ 90.3 \pm 2.3 $ & $ \underline{95.3} \pm 1.6 $ & $ 81.0 \pm 4.4 $ & $\underline{93.3} \pm 3.2$  & $ 80.6 \pm 1.6 $ & $ 83.4 \pm 4.8 $ \\
      RAE - GMM(20) & $ \boldsymbol{91.1 \pm 1.6}  $ & $ \underline{96.6} \pm 1.5  $ & $ 84.3 \pm 1.7 $ & $\underline{95.4} \pm 3.1 $  & $ 79.5 \pm 1.1 $ & $\boldsymbol{ 85.0 \pm 4.8} $ \\
      2-stage VAE & $84.8 \pm 2.3 $ & $71.4 \pm 8.3$ & $ 80.8 \pm 2.7 $ &  $ 60.2 \pm 9.2 $ & $ 79.6 \pm 2.3$ & $ 55.9 \pm 3.9 $ \\
      RHVAE - $\mathcal{N}(0, I_d)$   & $82.0 \pm 2.9 $ & $ 63.1 \pm 4.1 $  & $ 69.3 \pm 1.8$  & $ 46.9 \pm 8.4  $  & $73.6 \pm 4.1$ & $ 55.6 \pm 5.0$ \\
      \hline
      Ours                            & $90.1 \pm 1.4 $ & $ \boldsymbol{ 88.1 \pm 2.7} $  & $ \boldsymbol{86.2 \pm 1.8}$  & $ \mathbf{ 83.8 \pm 4.0 }  $  & $\boldsymbol{82.6 \pm 1.3}$ & $ 76.0 \pm 4.0$ \\
      \hline
      \end{tabular}
      \end{center}
      \end{table*}

        \subsubsection{Quantitative Comparison}\label{Subsec: quantitative comparison}
        In order to compare quantitatively the diversity and relevance of the samples generated by a generative model, several metrics were proposed~\cite{salimans_improved_2016,heusel_gans_2017,karras_progressive_2017, lucic_are_2018}. Since they suffer from some drawbacks~\cite{shmelkov_how_2018,borji_pros_2019}, we decide to use the \emph{GAN-train / GAN-test} measure discussed in~\cite{shmelkov_how_2018} as it appears to us well suited to measure the ability of a generative model to perform data augmentation. These two metrics consist in comparing the accuracy of a benchmark classifier trained on a set of generated data $\mathcal{S}_g$ and tested on a set of \emph{real} images $\mathcal{S}_{\mathrm{test}}$ (\emph{GAN-train}) or trained on the original train set $\mathcal{S}_{\mathrm{train}}$ (\emph{real} images used to train the generative model) and tested on $\mathcal{S}_g$ (\emph{GAN-test}). Those accuracies are then compared to the baseline accuracy given by the same classifier trained on $\mathcal{S}_{\mathrm{train}}$ and tested on $\mathcal{S}_{\mathrm{test}}$. These two metrics are quite interesting for our application since the first one (\emph{GAN-train}) measures the quality and diversity of the generated samples (the higher the better) while the second one (\emph{GAN-test}) accounts for the generative model's tendency to overfit (a score significantly higher than the baseline accuracy means overfitting). Ideally, the closer to the baseline the \emph{GAN-test} score the better. To stick to our low sample size setting, we compute these scores on three data sets created by down-sampling well-known databases. The first data set is created by extracting 500 samples from MNIST ensuring balanced classes (\emph{reduced} MNIST). For the second one, 500 samples of the MNIST database are again considered but a random split is applied such that some classes are under-represented (\emph{reduced} unbalanced MNIST). The last one consists in selecting 500 samples from 10 classes of the EMNIST data set having both lowercase and uppercase letters (\emph{reduced} EMNIST) so that we end up with a small database with strong variability within classes. The balance matches the one in the initial data set (\emph{by merge}). These three data sets are then divided into a baseline train set $\mathcal{S}_{\mathrm{train}}$ (80\%) and a validation set $\mathcal{S}_{\mathrm{val}}$ (20\%) used for the classifier training. Since the initial databases are huge, we use the original test set for $\mathcal{S}_{\mathrm{test}}$ so that it provides statistically meaningful results. For this comparison, we add a regularized autoencoder (RAE) \cite{ghosh_variational_2020}, a 2-stage VAE \cite{dai_diagnosing_2018} and a VAE where we use a 10-components mixture of Gaussian (GMM) instead of the prior to generate \cite{ghosh_variational_2020}, to the models presented in Sec.~\ref{Subsec: qualitative comp}. Each model is then trained on each class of $\mathcal{S}_{\mathrm{train}}$ to generate 1000 samples per class and $\mathcal{S}_g$ is created for each VAE by gathering all generated samples. A benchmark classifier chosen as a DenseNet\footnote{We used the PyTorch implementation provided in~\cite{amos_bamosdensenetpytorch_2020}.}~\cite{huang_densely_2017} is then 1) trained on $\mathcal{S}_{\mathrm{train}}$ and tested on $\mathcal{S}_{\mathrm{test}}$ (\emph{baseline}); 2) trained on $\mathcal{S}_g$ and tested on $\mathcal{S}_{\mathrm{test}}$ (\emph{GAN-train}) and 3) trained on  $\mathcal{S}_{\mathrm{train}}$ and tested on $\mathcal{S}_g$ (\emph{GAN-test}) until the loss does not improve for 50 epochs on $\mathcal{S}_{\mathrm{val}}$. For each experiment, the model is trained five times and we report the mean score and the associated standard deviation in Table~\ref{table: GAN - scores}. For the RAE we use a GMM and indicate the number of components between parentheses. As expected, the proposed method allows producing samples that are far more meaningful and relevant, in particular to perform DA. This is first illustrated by the \emph{GAN-train} scores that are either very close to the accuracy obtained with the \emph{baseline} or higher (see MNIST (unbalanced) in Table~\ref{table: GAN - scores}). The fact that we are able to enhance the classifier's accuracy even when trained only with synthetic data is very encouraging. Firstly, it proves that the created samples are close to the \emph{real} ones and so we were able to capture the true distribution of the data. Secondly, it shows that we do not overfit the initial training data since we are able to add some relevant information through the synthetic samples. This last observation is also supported by the \emph{GAN-test} scores for the proposed method which are quite close to the accuracies achieved on the \emph{baseline}. In case of overfitting, the \emph{GAN-test} score would be significantly higher than the \emph{baseline} since the classifier is tested on the generated samples while trained on the \emph{real} data that were also used to train the generative model. This is for instance the case for the RAE (underlined scores) where the number of components in the GMM impacts greatly the \emph{GAN-test} metric. Having a score close to the \emph{baseline} illustrates that the generative model is able to capture the distribution of the data and does not only \emph{memorize} it \cite{shmelkov_how_2018}. Finally, this study again shows the relevance of considering new ways to generate data from VAEs, such as fitting a mixture of Gaussian in the latent space, using a 2-stage VAE or using the proposed method, as they all improve in almost all cases the metrics when compared to prior-based methods (lines 2 and 9 of Table~\ref{table: GAN - scores}).

\section{Data Augmentation: Evaluation and Robustness}\label{Sec: Data Augmentation}

In this section we show the relevance of the proposed improvements to perform data augmentation in a HDLSS setting through a series of experiments.

\subsection{Setting}
\begin{figure}[!ht]
  \centering
  \subfloat{\includegraphics[width=3.2in]{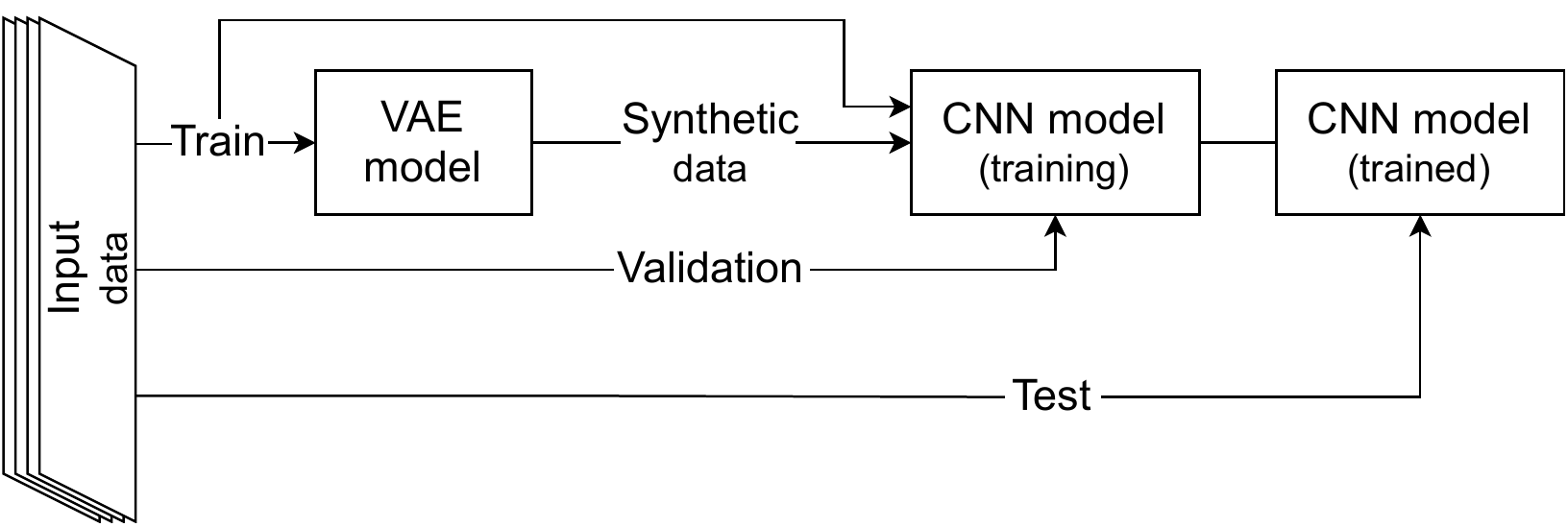}
  }
  \caption{Overview of the data augmentation procedure. The input data set is divided into a train set (the \emph{baseline}), a validation set and a test set. The train set is augmented using the VAE framework and generated data are then added to the \emph{baseline} to train a benchmark classifier.}
  \label{Fig: DA framework}
  \end{figure}

The setting we employ for DA consists in selecting a data set and splitting it into a train set (the \emph{baseline}), a validation set and a test set. The \emph{baseline} is then augmented using the proposed VAE framework and generation procedure. The generated samples are finally added to the original train set (\emph{i.e.} the \emph{baseline}) and fed to a classifier. The whole data augmentation procedure is illustrated in Fig.~\ref{Fig: DA framework}.

\subsection{Toy Data Sets}
The proposed VAE framework is here used to perform DA on several down-sampled well-known databases such that only tens of \emph{real} training samples per class are considered so that we stick to the low sample size setting. First, the robustness of the method across these data sets is tested with a standard benchmark classifier. Then, its reliability across other common classifiers is stressed. Finally, its scalability to larger data sets and more complex models is discussed.

\subsubsection{Materials}\label{Subsec: Toy materials}
In this section, we use the same three data sets described in Sec.~\ref{Subsec: quantitative comparison} and add one using the FashionMNIST data set and three classes we find hard to distinguish (\textit{i.e.} \emph{T-shirt}, \emph{dress} and \emph{shirt}). The data set is composed of 300 samples ensuring balanced classes (\textit{reduced} Fashion). Finally, we also select 150 samples from three balanced classes of CIFAR10 \cite{krizhevsky2009learning} hard to classify (\emph{cat}, \emph{dog} and \emph{horse}). In summary, we built five data sets having different class numbers, class splits and sample sizes. These data sets are again pre-processed such that 80\% is allocated for training (referred to as the \textit{Baseline}) and 20\% for validation. Since the original data sets are huge, we use the test set provided in the original databases (\emph{e.g.} $\approx$1000 samples per class for MNIST) so that it provides statistically meaningful results while allowing for a reliable assessment of the model's generalization power on unseen data.

\subsubsection{Robustness Across Data Sets}

\begin{table*}[!t]
  \caption{Data augmentation with a DenseNet model as benchmark. Mean accuracy and standard deviation across five independent runs are reported. The first three rows (Aug.) correspond to basic transformations (noise, crop, etc.). In gray are the cells where the accuracy is higher on synthetic data than on the \emph{baseline} (\textit{i.e.} the raw data). The test set is the one proposed in the entire original data set (\textit{e.g.} $\approx$1000 samples per class for MNIST) so that it provides statistically meaningful results and allows for a good assessment of the model's generalization power.}
  \label{Tab: Data Augmentation Toys}
 \centering
 %\small
 \scriptsize
   \begin{tabular}{c|ccccc|ccccc}

   \cline{2-11}
    & \multirow{2}{*}{MNIST}& MNIST &EMNIST &  \multirow{2}{*}{FASHION} & \multirow{2}{*}{CIFAR} & \multirow{2}{*}{MNIST}& MNIST &EMNIST &  \multirow{2}{*}{FASHION} & \multirow{2}{*}{CIFAR} \\
                  &   &(unbal.)& (unbal.)& & & &(unbal.)& (unbal.)& \\
    \hline
    &\multicolumn{5}{c|}{Baseline + Synthetic} & \multicolumn{5}{c}{Synthetic Only}\\
     \hline
     Baseline  & $89.9/0.6$ & $81.5/0.7$ & $82.6/1.4$ & $76.0/1.5$ & $42.6/7.6$ & - & - & - & - & - \\
     \hline
     Aug. (X5) & $92.8/0.4$ & $86.5 /0.9$ & $85.6 /1.3$ & $77.5 /2.0$ & $47.7 /2.3$ & - & - & - & - & -\\
     Aug. (X10)& $88.2 /2.2$ & $82.0 /2.4$ & $85.7 /0.3$ & $79.2 /0.6$ & $48.2 /1.7 $ & - & - & - & - & -\\
     Aug. (X15)& $92.8 /0.7$ & $85.8 /3.4$ & $86.6 /0.8$ & $80.0 /0.5$ & $48.0 /2.2 $ & - & - & - & - & -\\
     \hline
     VAE - 200 & $88.5 /0.9$ & $84.0 /2.0$ & $81.7 /3.0$ & $78.6 /0.4$ & 
     $ 46.9 /1.3$
     &$69.9 /1.5$ & $64.6 /1.8$ & $65.7 /2.6$ & $73.9 /3.0$ & $40.5 /4.1$ \\
     %VAE-500$^*$ & $90.4 /1.3$ & $87.3 /1.2$ & $83.4 /1.6$ & $78.7 /0.3$  \\
     VAE - 1k  & $91.2 /1.0$ & $86.0 /2.5$ & $84.3 /1.6$ & $77.6 /2.1$ & 
     $47.7 /1.4$
     &$83.4 /2.4$ & $74.7 /3.2$ & $75.3 /1.4$ & $71.4 /6.1$ & $41.3 /2.4$\\
     %VAE-2k$^*$  & $92.2 /1.6$ & $88.0 /2.2$ & $86.0 /0.2$ & $79.3 /1.1$  \\
     VAMP - 200         & $ 91.4 /1.9 $  & $ 81.1 /2.7 $ & $ 84.2 /0.8 $ & $ 79.8 /0.8 $& 
     $45.6 /6.9$
     & $ 61.3 /3.2 $ & $ 52.4 /3.0 $ & $ 67.4 /1.4 $ & $ 70.4 /3.2 $ &$40.6 /6.6 $ \\
     VAMP - 1k          & $ \boldsymbol{93.6 /0.9} $  & $ 88.0 /1.1 $ & $ 86.2 /1.1 $ & $ 79.6 /0.4 $ & 
     $45.2 /6.1$
     &$ 72.8 /6.7 $ & $ 68.2 /6.6 $ & $ 70.7 /8.0 $ & $ 69.2 /5.4 $ & $39.7 /7.7$\\
     RHVAE - 200$^*$ & $ 89.9 /0.5$ & $82.3 /0.9$ & $83.0 /1.3$ & $77.6/1.3$&
     $45.2 /1.9$
     & $76.0 /1.8$ & $61.5 /2.9$ & $59.8 /2.6$ & $72.8 /3.6$ & $42.4 /1.2$ \\
     %RHVAE-500$^*$ & $ 90.9 /1.1$ & $84.0 /3.2$ & $84.4 /1.2$ & $78.0 \pm1.3$  \\
     RHVAE - 1k$^*$  & $ 91.7 /0.8$ & $84.7 /1.8$ & $84.7 /2.4$ & $79.3/1.6$ & 
     $42.1 /2.9$
     &$82.0 /2.9$ & $69.3 /1.8$ & $73.6 /4.1$ & $76.0 /4.1$ & $ 40.7 /3.2$\\
     %RHVAE-2k$^*$  & $ 92.7 /1.4$ & $86.8 /1.0$ & $84.9 /2.1$ & $79.0 \pm1.4$ \\
    %Ours-2k & $\mathbf{94.3 /0.8}$& $89.1 /1.9$ & $\mathbf{87.6 /0.8}$ & $78.1 /1.8$   \\
     \hline
     VAE GMM - 200 & $ 90.5 /1.1 $  & $ 82.9 /2.2 $ & $ 84.8 /1.0 $ & $ 79.6 /0.7 $ & 
     $44.9 /1.9$
     &$76.5 /1.5$ & $ 64.0 /2.6 $ & $ 70.5 /1.5 $ & $ 71.9 /2.2 $ &$38.7 /4.2$ \\
     VAE GMM - 1k & $ 92.0 /1.8 $  & $ 86.7 /1.0 $ & $ 86.1 /1.1 $ & $ 79.5 /0.7 $  & 
     $38.9 /2.4$
     &$82.9 /2.4$ & $ 74.4 /3.8 $ & $ 74.0 /2.6 $ & $ 73.9 /2.5 $ & $41.6 /2.7$\\
     2-stage VAE - 200 & $ 91.2 /1.2 $  & $ 83.5 /1.5 $ & $ 85.3 /1.9 $ & $ \boldsymbol{80.5 /0.6}$ & 
     $44.4 /2.3$
     &$ 82.3 /1.1 $ & $ 74.9 /2.3 $ & $ 76.7 /1.3 $ & $ 76.2 /2.0 $ & $38.1 /2.6$\\
     2-stage VAE - 1k & $ 93.3 /0.7 $  & $ 87.7 /2.4 $ & $ 86.7 /1.1 $& $ 79.5 /0.9 $  & 
     $38.8 /3.0$
     &$ 84.8 /2.3 $ & $ 80.8 /2.7 $ & $ 79.6 /2.3 $ & $ 75.8 /1.8 $ & $37.9 /3.6$\\
     %RAE - L2 - 200     & $ 91.7 /0.9 $  & $ 82.1 /1.2 $ & $ 85.5 /0.9 $& $ 80.3 /1.1 $\\
     %RAE - L2 - 1k      & $  /$  & $ /$ &  $ \pm$ & $ / $\\
     RAE - 200     & $ 91.6 /1.1 $  & $ 81.3 /1.3 $ & $ 85.2 /0.9 $& $ 80.1 /0.8 $ & 
     $46.2 /2.9$
     &$ 83.6 /2.8 $  & $ 74.5 /1.6 $ & $ 76.9 /1.6 $& $ 66.5 /4.4 $ & $33.7 /1.7 $\\
     RAE - 1k      & $ 93.3 /0.8 $  & $ 88.3 /1.1 $ & $ 85.8 /0.9 $ & $ 79.8 /1.3 $ & 
     $44.1 /2.6$
     &$ \cellcolor{gray!30}\boldsymbol{90.3 /2.3} $  & $ 81.0 /4.4 $ &  $ 80.6 /1.6 $ & $ 62.0 /5.1 $ & $33.6 /0.4 $\\
     \hline
     Ours - 200  & $91.0 /1.0$         & $84.1 /2.0$                       & $85.1 /1.1$ & $ 77.0 /0.8$ & 
     $46.8 /2.2$
     &$87.2 /1.1$ & $79.5 /1.6$                   &  $77.0 /1.6$                   & $\cellcolor{gray!30}77.0 /0.8$ & $\cellcolor{gray!30} \boldsymbol{47.3 /1.7}$   \\
     %Ours-500  & $92.3 /1.1$         & $87.7 /0.9$                       & $85.1 /1.1$ & $ 78.5 /0.9$   \\
     Ours - 1k & $93.2/0.8$         & $\mathbf{89.7/0.8}$              & $\boldsymbol{87.0/1.0}$ & $80.2/0.8$ & 
     $ \boldsymbol{49.2/2.3}$
     &$\cellcolor{gray!30}90.1/1.4$ & $\cellcolor{gray!30}\boldsymbol{86.2/1.8}$ & $\cellcolor{gray!30}\boldsymbol{82.6/1.3}$ & $\cellcolor{gray!30}\boldsymbol{79.3/0.6}$ & $\cellcolor{gray!30}46.7/3.1$ \\
     \hline

   \end{tabular}%}
 \end{table*}

The first experiment we conduct consists in assessing the method's robustness across the five aforementioned data sets. For this study, we propose to consider a DenseNet model as benchmark classifier. On the one hand, the training data (the \emph{baseline}) is augmented by a factor 5, 10 and 15 using classic data augmentation methods (random noise, random crop, rotation, etc.) so that the proposed method can be compared with classic and simple augmentation techniques. On the other hand, the protocol described in Fig.~\ref{Fig: DA framework} is employed with the same VAEs as before. The generative models are trained individually on each class of the \emph{baseline} until the ELBO does not improve for 20 epochs. The VAEs are then used to produce 200 or 1000 new synthetic samples per class using the same generation protocols as described in Sec.~\ref{Subsec: quantitative comparison}. Finally, the benchmark DenseNet model is trained with five independent runs on either 1) the \emph{baseline}, 2) the augmented data using classic augmentation methods, 3) the augmented data using the VAEs or 4) only the synthetic data created by the generative models. For each experiment, the mean accuracy and the associated standard deviation across those five runs are reported in Table~\ref{Tab: Data Augmentation Toys}. An early stopping strategy is employed and CNN training is stopped if the loss does not improve on the validation set for 50 epochs. 

The first outcome of such a study is that, as expected, generating synthetic samples with the proposed method seems to enhance their relevance in particular for data augmentation tasks. This is for instance illustrated by the first column of Table~\ref{Tab: Data Augmentation Toys} where synthetic samples are added to the \emph{baseline}. While adding samples generated either by a VAE or RHVAE and using the prior distribution seems to improve the classifier accuracy when compared with the \emph{baseline}, the gain remains limited since it struggles to exceed the gain reached with classic augmentation methods. On the contrary, methods using either more complex priors (VAMP), a second VAE or a GMM allow improving classification results on MNIST and Fashion but still under-perform on EMNIST and CIFAR. Finally, the proposed generation method is able to produce very useful samples for the CNN model since in all cases it allows the classifier to either achieve the best result (highlighted in bold) or comparable performance than peers while keeping a relatively low standard deviation. Secondly, the relevance of the samples produced by the proposed scheme is even more supported by the second column of Table~\ref{Tab: Data Augmentation Toys} where the classifier is trained only using the synthetic samples generated by the VAEs. First, even with a quite small number of samples generated with our method (200 per class), the classifier is almost able to reach the accuracy achieved with the \emph{baseline}. For instance, when the CNN is trained on \emph{reduced} MNIST with 200 synthetic samples per class generated with our method, it is able to achieve an accuracy of 87.2\% vs. 89.9\% with the \emph{baseline}. In comparison, any other method fails to produce meaningful samples since a quite significant loss in accuracy is observed. The fact that the classifier almost performs as well on the synthetic data as on the \emph{baseline} is good news since it shows that the proposed framework is able to produce samples accounting for the original data set diversity even with a small number of generated samples. Even more interesting, as the number of synthetic data increases, the classifier is able to perform much better on the synthetic data than on the \emph{baseline} since a gain of 3 to 6 points in accuracy is observed. Again, this strengthens the observations made in Sec.~\ref{Subsec: qualitative comp} and \ref{Subsec: quantitative comparison} where we noted that \textbf{the proposed method is able to enrich the initial data set} with relevant and realistic samples.

Finally, it can be seen in this experiment why geometric data augmentation methods are still questionable and remain data set dependent. For example, augmenting the \textit{baseline} by a factor 10 (where we add flips and rotations on the original data) seems to have no significant effect on the \emph{reduced} MNIST data sets while it still improves results on \emph{reduced} EMNIST, Fashion and CIFAR. We see here how the \emph{expert} knowledge comes into play to assess the relevance of the transformations applied to the data. Fortunately, the method we propose does not require such knowledge and \textbf{appears to be quite robust to data set changes.}

\subsubsection{Robustness Across Classifiers}\label{Subsec: classifier robustness}

 \begin{figure*}[!ht] 
  \centering
  \subfloat[\emph{reduced} MNIST balanced \label{Subfig: Classifiers rob MNIST bal}]{\includegraphics[width=3.5in]{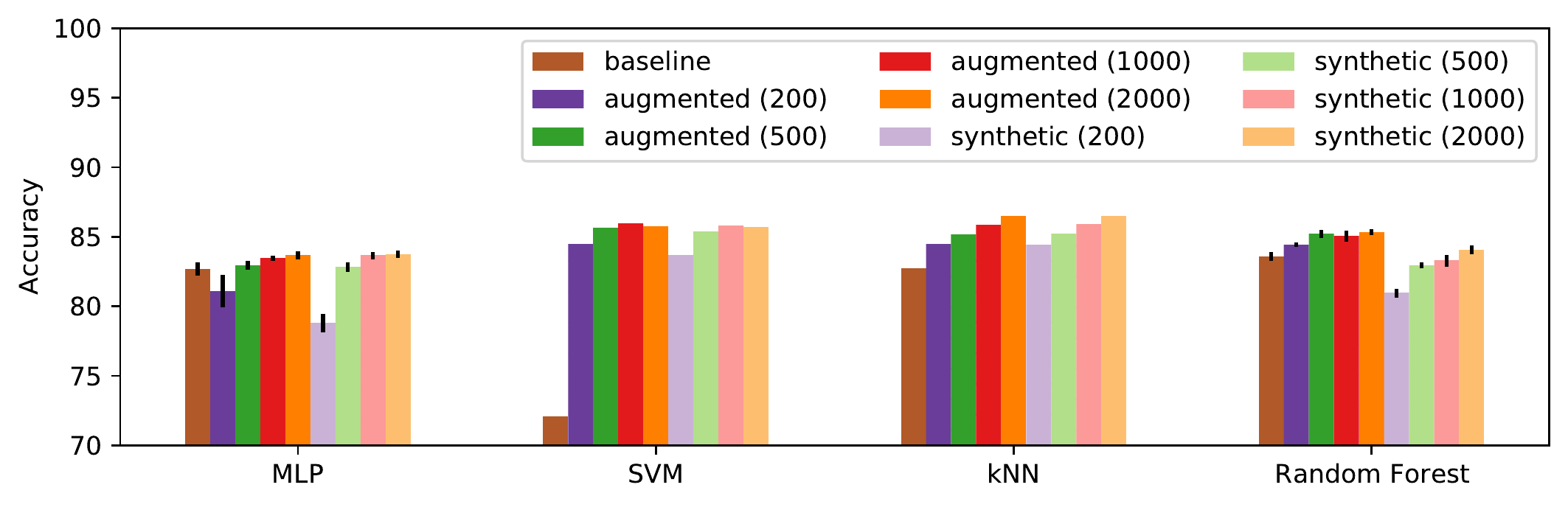}
  }
  \hfil
  \centering
  \subfloat[\emph{reduced} MNIST unbalanced \label{Subfig: Classifiers rob MNIST unbal}]{\includegraphics[width=3.5in]{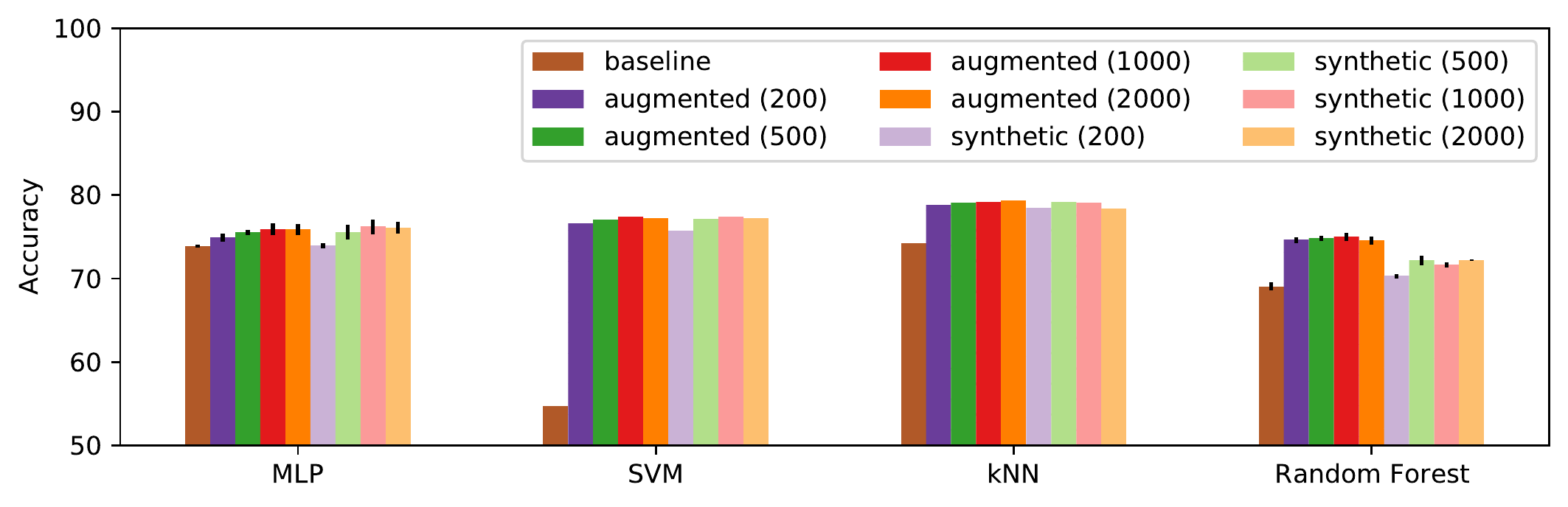}
  }
  
  \caption{Evolution of the accuracy of four benchmark classifiers on \emph{reduced} balanced MNIST (left) and \emph{reduced} unbalanced MNIST data sets (right). Stochastic classifiers are trained with five independent runs and we report the mean accuracy and standard deviation on the test set.}
  \label{Fig: Classifiers rob}
  \end{figure*}
  
In addition to assessing the robustness of the method to data sets changes, we also propose to evaluate its reliability across classifiers. To do so, we consider very different common supervised classifiers: a multi layer perceptron (MLP)~\cite{goodfellow_deep_2016}, a random forest~\cite{breiman_random_2001}, the $k$-NN algorithm and a SVM~\cite{kotsiantis_supervised_2007}. Each of the aforementioned classifiers is again trained either on 1) the original training data set (the \emph{baseline}); 2) the augmented data using the proposed method and 3) only the synthetic data generated by our method with five independent runs and using the same data sets as presented in Sec.~\ref{Subsec: Toy materials}. Finally, we report the mean accuracy and standard deviation across these runs for each classifier and data set. The results for the balanced (resp. unbalanced) \emph{reduced} MNIST data set can be found in Fig.~\ref{Subfig: Classifiers rob MNIST bal} (resp. Fig.~\ref{Subfig: Classifiers rob MNIST unbal}). Metrics obtained on \emph{reduced} EMNIST and Fashion are available in Appendix~\ref{appendix F} but reflect the same tendency. 

As illustrated in Fig.~\ref{Fig: Classifiers rob}, the method appears quite robust to classifier changes as well since it allows improving the model's accuracy significantly for almost all classifiers (the accuracy achieved on the \emph{baseline} is represented by the leftmost bar in Fig.~\ref{Fig: Classifiers rob} for each classifier).  The method's strength is even more striking when unbalanced data sets are considered since the method is able to produce meaningful samples even with a very small number of training data and so it is able to over-sample the minority classes in a reliable way. Moreover, as observed earlier, synthetic samples are again helpful to enhance classifiers' generalization power since they perform better when trained only on synthetic data than on the \emph{baseline} in almost all cases.

\subsubsection{A Note on the Method Scalability}

\begin{figure*}[!ht]
    \centering
    \subfloat{\includegraphics[width=2.1in]{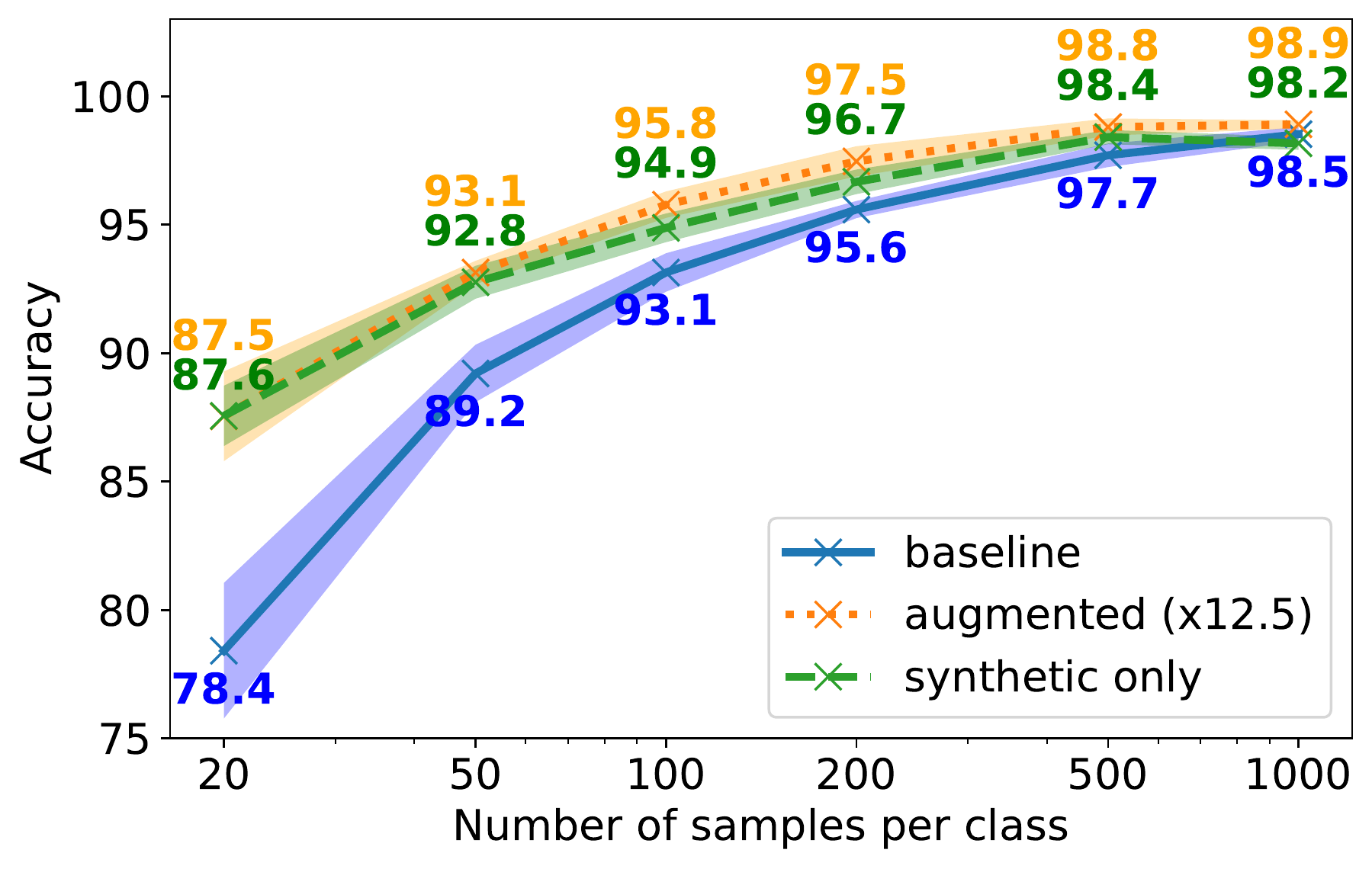}}
\hfil 
 \centering
    \subfloat{\includegraphics[width=2.1in]{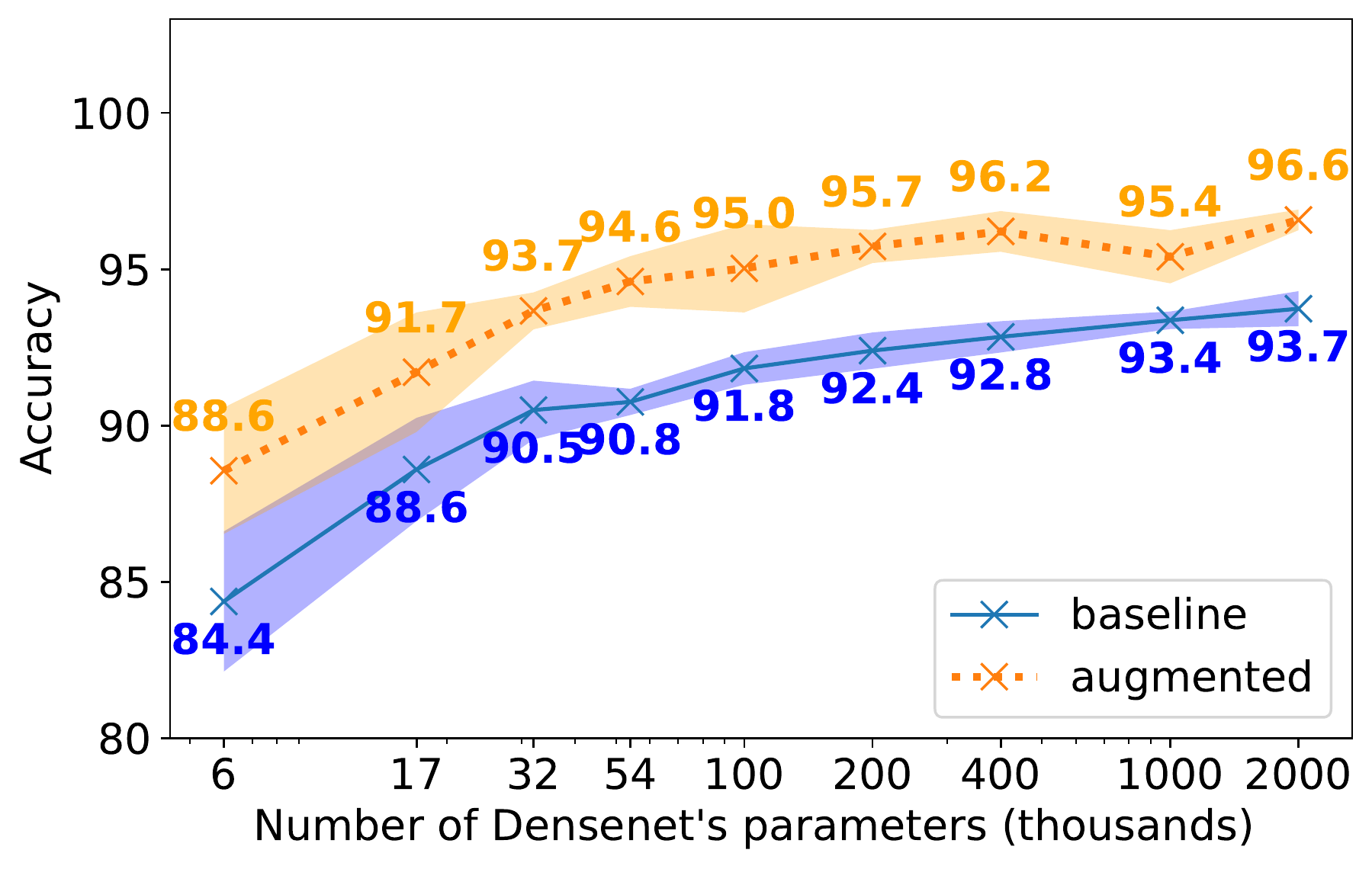}}
\hfil
 \centering
    \subfloat{\includegraphics[width=2.1in]{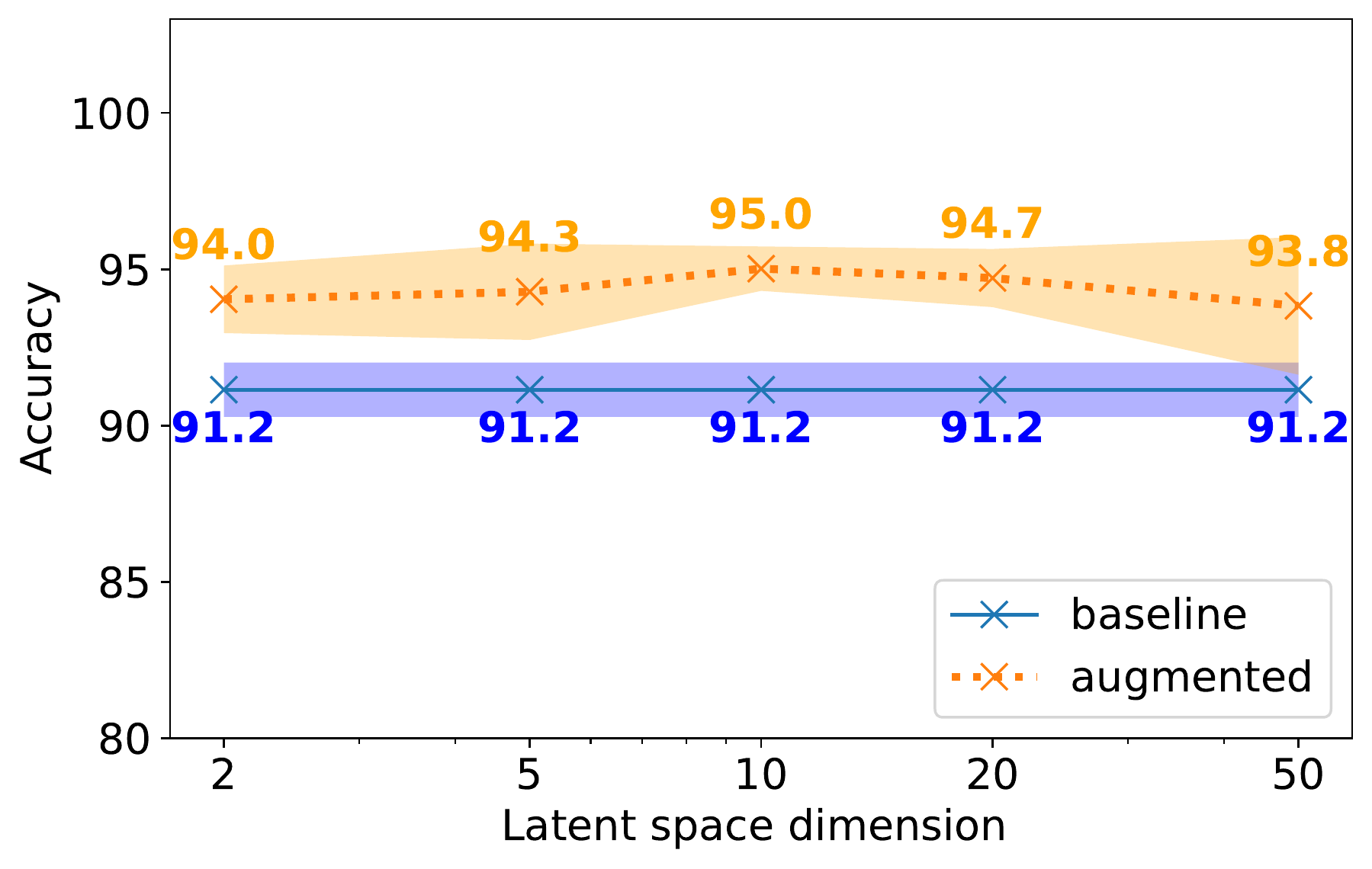}}
    
    \caption{Evolution of the accuracy of a benchmark DenseNet classifier according to the number of samples in the train set (\emph{i.e.} the \emph{baseline}) (\emph{left}), the number of parameters of the Densenet (\emph{middle}) and the latent space dimension of the VAE (\emph{right}) on MNIST. Curves show the mean accuracy and standard deviation across 5 runs on the original test set for the \emph{baseline} (blue), the augmented data (orange) and the synthetic ones (green).}
    \label{Fig: Toy scalability}
\end{figure*}

Finally, we also discuss the method scalability to larger data sets, bigger models and higher dimensional latent spaces. To do so, we consider the MNIST data set and a benchmark classifier taken as a DenseNet which performs well on such data. First, we down-sample the original MNIST database in order to progressively decrease the number of samples per class. We start by creating a data set having 1000 samples per class to finally reach 20 samples per class. For each created data set, we allocate 80\% for training (the \emph{baseline}) and reserve 20\% for the validation set. A \emph{geometry-aware} VAE is then trained on each class of the \emph{baseline} until the ELBO does not improve for 50 epochs and is used to generate synthetic samples (12.5$\times$ the \emph{baseline}). The benchmark CNN is trained with five independent runs on either 1) the \emph{baseline}, 2) the augmented data or 3) only the synthetic data generated with our model. The evolution of the mean accuracy on the original test set ($\approx$1000 samples per class) according to the number of samples per class is presented in Fig.~\ref{Fig: Toy scalability} (left). Second, we only consider 50 samples per class and train the VAE on each class to generate 1000 samples per class. The number of the classifier's parameters is  also progressively changed and we report the mean accuracy of the CNN according to the number of parameters in Fig.~\ref{Fig: Toy scalability} (middle). Finally, we consider several latent space dimensions for the VAE ranging from 2 to 50 and plot the evolution of the CNN accuracy according the latent space dimension (right).

First, this experiment shows that the fewer samples in the training set, the more useful the method appears. Using the proposed augmentation framework indeed allows for a gain of more than 9.0 points in the CNN accuracy when only 20 samples per class are considered. In other words, as the number of samples increases, the marginal gain seems to decrease. Nevertheless, this reduction must be put into perspective since it is commonly acknowledged that, as the results on the \emph{baseline} increase (and thus get closer to the perfect score), it is even more challenging to improve the score with the augmented data. In this experiment, we are nonetheless still able to improve the model accuracy even when it already achieves a very high score. For instance, with 500 samples per class, the augmentation method still allows increasing the model accuracy from 97.7\% to 98.8\%. Finally, for data sets with fewer than 500 samples per class, the classifier is able to outperform the \emph{baseline} even when trained only with the synthetic data. This shows again the strong generalization power of the proposed method which allows creating new relevant data for the classifier. Another interesting take from these experiments is that the augmentation method seems to benefit both simple and more complex models since the gain in the model accuracy remains quite steady ($\approx$ 3 pts) regardless of the number of parameters in the classifier (Fig.~\ref{Fig: Toy scalability} (middle)). Finally, the impact of the dimension of the latent space remains limited for such a framework as the classification accuracy remains stable. Nonetheless, this may be due to the simplicity of the database and more complex data might need higher dimensional latent spaces.

\section{Validation on Medical Imaging}\label{Sec: Medical images}
With this last series of experiments, we assess the validity of our data augmentation framework on a binary classification task consisting in differentiating Alzheimer's disease (AD) patients from cognitively normal (CN) subjects based on T1-weighted (T1w) MR images of human brains. Such a task is performed using a CNN trained, as before, either on 1) \emph{real} images, 2) synthetic samples or 3) both. In this section, label definition, preprocessing, quality check, data split and CNN training and evaluation is done using \href{https://github.com/aramis-lab/clinica}{Clinica}\footnote{https://github.com/aramis-lab/clinica} \cite{routierClinicaOpenSourceSoftware2021} and \href{https://github.com/aramis-lab/clinicadl}{ClinicaDL}\footnote{https://github.com/aramis-lab/clinicadl} \cite{thibeau-sutreClinicaDLOpensourceDeep2021}, two open-source software packages for neuroimaging processing.

% 

%To assess the validity of the data augmentation procedure on medical images, it was performed on T1-weighted (T1w) MR images of human brains. The synthetic data produced were used to learn a binary classification task: differentiate Alzheimer's disease patients from cognitively normal participants. This classification task was performed by a convolutional neural network (CNN) pretrained on real images, synthetic images or both. 
 
\subsection{Data Augmentation Literature for AD vs CN Task}

Even though many studies use CNNs to differentiate AD from CN subjects with anatomical MRI~\cite{wen_convolutional_2020}, we did not find any meta-analysis on the use of data augmentation for this task. Some results involving DA can nonetheless be cited and are presented in Table~\ref{table:background}. However, assessing the real impact of data augmentation on the performance of the models remains challenging. For instance, this is illustrated by the works of~\cite{aderghal_classification_2017} and~\cite{aderghal_classification_2018}, which are two examples in which DA was used and led to two significantly different results, although a similar framework was used in both studies. Interestingly, as shown in Table~\ref{table:background}, studies using DA for this task only relied on simple affine and pixel-level transformations, which may reveal data dependent. Note that complex DA was actually performed for AD~vs~CN classification tasks on PET images, but PET is less frequent than MRI in neuroimaging data sets~\cite{islam_gan-based_2020}. As noted in the previous sections, our method would apply pretty straightforwardly to this modality as well. For MRI, other techniques such as transfer learning~\cite{oh_classification_2019} and weak supervision~\cite{liu_weakly_2020} were preferred to handle the small amount of samples in data sets and may be coupled with DA to further improve the classifier performance.

\begin{table}[ht]
\caption{Accuracy obtained by studies performing AD vs CN classification with CNNs applied on T1w MRI and using data augmentation}
\label{table:background}
\begin{center}
\setlength{\tabcolsep}{3pt}
\begin{tabular}{c | c | c | c | c | c} 
\cline{5-6}
\multicolumn{4}{c}{} & \multicolumn{2}{|c}{Accuracy} \\
\hline
Study & Methods & Subj. & Images & Baseline & Augmented \\
\hline
\cite{valliani_deep_2017} & rotate, flip, shift & 417 & 417 & 78.8 & 81.3 \\
\cite{backstrom_efficient_2018} & flip & 340 & 1198 & -- & 90.1 \\
\cite{cheng_cnns_2017} & shift, sample, rotate & 193 & 193 &  -- & 85.5 \\
\cite{aderghal_classification_2017} & shift, blur, flip & 720 & 720 & 82.8 & 83.7 \\ 
\cite{aderghal_classification_2018} & shift, blur & 720 & 720 & -- & 90.0 \\
\hline
\end{tabular}
\end{center}
\end{table}
%\begin{table*}[!ht]
%\caption{Summary of participant demographics, mini-mental state examination (MMSE) and global clinical dementia rating (CDR) scores at baseline.}
%\label{table:population}
%\begin{center}
%\setlength{\tabcolsep}{3pt}

%\begin{tabular}{l l l l l l l}
%\hline
%Data set & Label & Images & Age & Gender & MMSE & CDR \\
%\hline
%\multirow{2}{*}{ADNI} & CN & 403 & $73.3\pm6.0$ & 185 M / 218 F & $29.1\pm1.1$ & 0: 403 \\ 
%& AD & 362 & $74.9\pm7.9$ & 202 M / 160 F & $23.1\pm2.1$ & 0.5: 169, 1: 192, 2: 1 \\ 
%\hline
%\multirow{2}{*}{AIBL} & CN & 429 & $73.0\pm6.2$ & 183 M / 246 F & $28.8\pm1.2$ & 0: 406, 0.5: 22, 1: 1 \\ 
%& AD & 76 & $74.4\pm8.0$ & 33 M / 43 F & $20.6\pm5.5$ & 0.5: 31, 1: 36, 2: 7, 3: 2 \\ 
%\hline
%\end{tabular}

%\end{center}
%\end{table*}

\subsection{Materials}

Data used in this section were obtained from the Alzheimer’s Disease
Neuroimaging Initiative (ADNI) database  (\href{http://adni.loni.usc.edu/}{adni.loni.usc.edu}) and the Australian Imaging, Biomarkers and Lifestyle (AIBL) study (\href{https://aibl.csiro.au/}{aibl.csiro.au}).

The ADNI was launched in 2003 as a public-private partnership, led by Principal Investigator Michael W. Weiner, MD. The primary goal of ADNI has been to test whether serial MRI, PET, other biological markers, and clinical and neuropsychological assessment can be combined to measure the progression of mild cognitive impairment and early AD. For up-to-date information, see \href{www.adni-info.org}{www.adni-info.org}. The ADNI data set is composed of four cohorts: ADNI-1, ADNI-GO, ADNI-2 and ADNI-3. The data collection of ADNI-3 has not ended yet, hence our data set contains all images and metadata that were already available on May 6, 2019.
Similarly to ADNI, the AIBL data set seeks to discover which biomarkers, cognitive characteristics, and health and lifestyle factors determine the development of AD. This cohort is also longitudinal and the diagnosis is given according to a series of clinical tests~\cite{ellis_australian_2009}. Data collection for this cohort is over. 

Two diagnoses are considered for the classification task:
\begin{itemize}
    \item CN: baseline session of participants who were diagnosed as cognitively normal at baseline and stayed stable during the follow-up;
    \item AD: baseline session of participants who were diagnosed as demented at baseline and stayed stable during the follow-up.
\end{itemize}
Table~\ref{table:population} summarizes the demographics, the mini-mental state examination (MMSE) and global clinical dementia rating (CDR) scores at baseline of the participants included in our data set. The MMSE and the CDR scores are classical clinical scores used to assess dementia. The MMSE score has a maximal value of 30 for cognitively normal persons and decreases if symptoms are detected. The CDR score has a minimal value of 0 for cognitively normal persons and increases if symptoms are detected. 

\begin{table}[!ht]
\caption{Summary of participant demographics, mini-mental state examination (MMSE) and global clinical dementia rating (CDR) scores at baseline.}
\label{table:population}
\begin{center}
\setlength{\tabcolsep}{3pt}
\scriptsize
\begin{tabular}{l l l l l l l}
\hline
Data set & Label & Subj. & Age & Sex M/F & MMSE & CDR \\
\hline
\multirow{3}{*}{ADNI} & CN & 403 & $73.3\pm6.0$ & 185/218 & $29.1\pm1.1$ & 0: 403 \\ 
& \multirow{2}{*}{AD} & \multirow{2}{*}{362} & \multirow{2}{*}{$74.9\pm7.9$} & \multirow{2}{*}{202/160} & \multirow{2}{*}{$23.1\pm2.1$} & 0.5: 169, 1: 192\\
&&&&&& 2: 1 \\ 
\hline
\multirow{4}{*}{AIBL} & \multirow{2}{*}{CN} & \multirow{2}{*}{429} & \multirow{2}{*}{$73.0\pm6.2$} & \multirow{2}{*}{183/246} & \multirow{2}{*}{$28.8\pm1.2$} & 0: 406, 0.5: 22\\
&&&&&& 1: 1 \\ 
& \multirow{2}{*}{AD} & \multirow{2}{*}{76} & \multirow{2}{*}{$74.4\pm8.0$} & \multirow{2}{*}{33/43} & \multirow{2}{*}{$20.6\pm5.5$} & 0.5: 31, 1: 36\\
&&&&&& 2: 7, 3: 2 \\ 
\hline
\end{tabular}
\end{center}
\end{table}

\subsection{Preprocessing of T1-Weighted MRI}

The steps performed in this section correspond to the procedure followed in \cite{wen_convolutional_2020} and are listed below:

\begin{enumerate}
    \item Raw data are converted to the BIDS standard~\cite{gorgolewski_brain_2016},
    \item Bias field correction is applied using N4ITK~\cite{tustison_n4itk_2010},
    \item T1w images are linearly registered to the MNI standard space~\cite{fonov_unbiased_2009, fonov_unbiased_2011} with ANTS~\cite{avants_insight_2014} and cropped. This produced images of size 169$\times$208$\times$179 with 1~mm$^3$ isotropic voxels.
    \item An automatic quality check is performed using an open-source pretrained network~\cite{fonov_deep_2018}. All images passed the quality check.
    \item NIfTI files are converted to tensor format.
    \item (Optional) Images are down-sampled with a trilinear interpolation leading to a size of 84$\times$104$\times$89.
    \item Intensity rescaling between the minimum and maximum values of each image is performed. 
\end{enumerate}
These steps lead to 1) down-sampled images (84$\times$104$\times$89) or 2) high-resolution images (169$\times$208$\times$179).
%\begin{itemize}
%    \item downsampled images: image of size 84x104x89,
%    \item high-resolution images: image of size 169x208x179.
%\end{itemize}

\subsection{Evaluation Procedure}

The ADNI data set is split into three sets: training, validation and test.
First, the test set is created using 100 randomly chosen participants for each diagnostic label (i.e. 100 CN, 100 AD). The rest of the data set is split between the training (80\%) and the validation (20\%) sets. We ensure that age, sex and site distributions between the three sets are not significantly different.

A smaller training set (denoted as \textit{train-50}) is extracted from the obtained training set (denoted as \textit{train-full}). This set comprises only 50 images per diagnostic label, instead of 243 CN and 210 AD for \textit{train-full}. We ensure that age and sex distributions between \textit{train-50} and \textit{train-full} are not significantly different. This is not done for the site distribution as there are more than 50 sites in the ADNI data set (so they could not all be represented in this smaller training set). AIBL data are never used for training or hyperparameter tuning and are only used as an independent test set.

% Finally, a 5-fold cross-validation split was performed on \textit{train-full}. This cross-validation split was used for the architecture search with random search.

\subsection{CNN Classifiers}

A CNN takes as input an image and outputs a vector of size $C$ corresponding to the number of labels existing in the data set. Then, a CNN predicts the label of a given image by selecting the highest probability in the output vector.

\subsubsection{Hyperparameter Choices}

As for the VAE, the architecture of the CNN depends on the size of the input. Then, there is one architecture per input size: down-sampled images and high-resolution images (see Appendix~\ref{appendix D.4}). Moreover, two different paradigms are used to choose the architecture. 
First, we reuse the same architecture as in~\cite{wen_convolutional_2020}. This architecture was obtained by optimizing manually the networks on the ADNI data set for the same task (AD vs CN). A slight adaption is done for the down-sampled images, which consists in resizing the number of nodes in the fully-connected layers to keep the same ratio between the input and output feature maps in all layers. We denote these architectures as \textbf{baseline}. Secondly, we launch a random search~\cite{bergstra_random_2012} that allows exploring different hyperperameter values. The hyperparameters explored for the architecture are the number of convolutional blocks, of filters in the first layer and of convolutional layers in a block, the number of fully-connected layers and the dropout rate. Other hyperparameters such as the learning rate and the weight decay are also part of the search. 100 different random architectures are trained on the 5-fold cross-validation done on \textit{train-full}. For each input, we choose the architecture that obtained the best mean balanced accuracy across the validation sets of the cross-validation. We denote these architectures as \textbf{optimized}.

\subsubsection{Network Training}

The weights of the convolutional and fully-connected layers are initialized as described in~\cite{he_delving_2015}, which corresponds to the default initialization method in PyTorch. Networks are trained for 100 epochs for \textbf{baseline} and 50 epochs for \textbf{optimized}. The training and validation losses are computed with the cross-entropy loss. For each experiment, the final model is the one that obtained the highest validation balanced accuracy during training. The balanced accuracy of the model is evaluated at the end of each epoch.

\subsection{Experimental Protocol}

\begin{figure}[ht] 
      \centering
      \subfloat{\includegraphics[width=.8in]{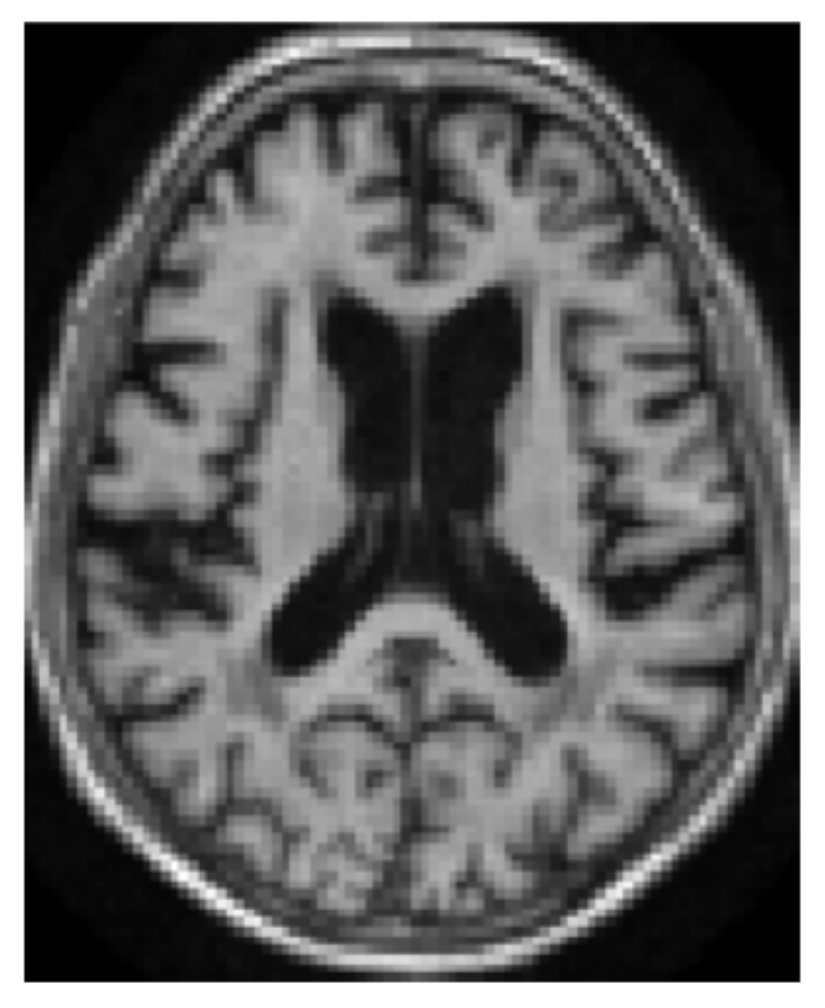}
      }
      \centering
      \subfloat{\includegraphics[width=.8in]{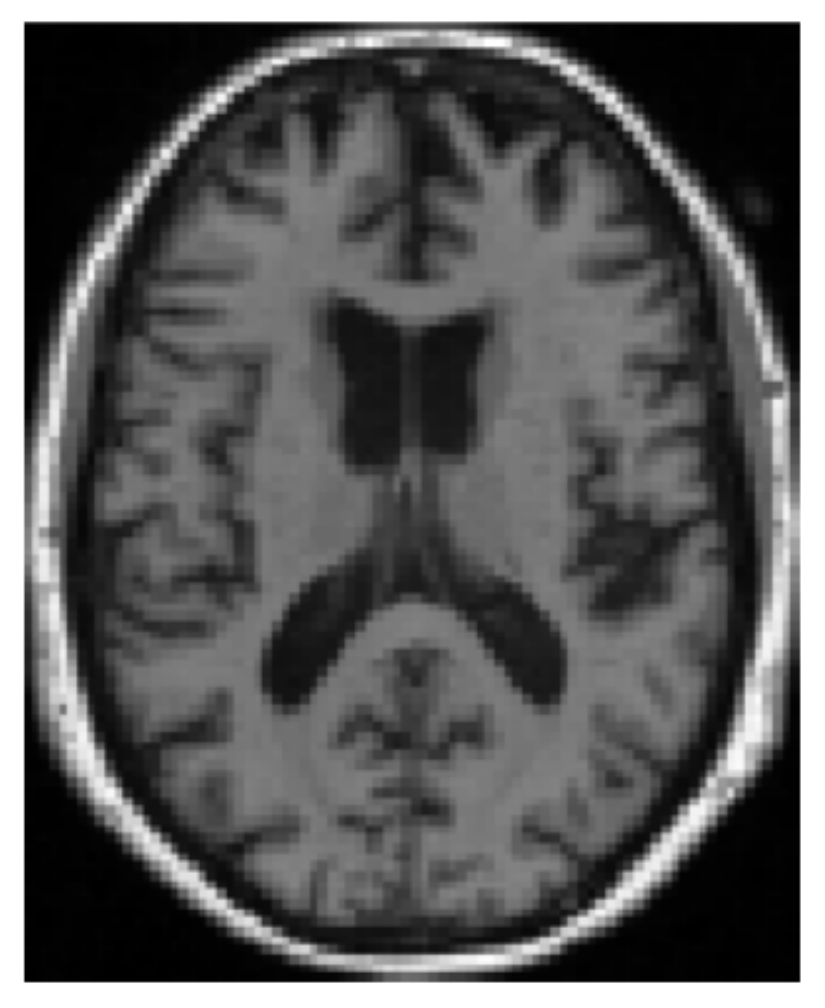}
      }
      \centering
      \subfloat{\includegraphics[width=.8in]{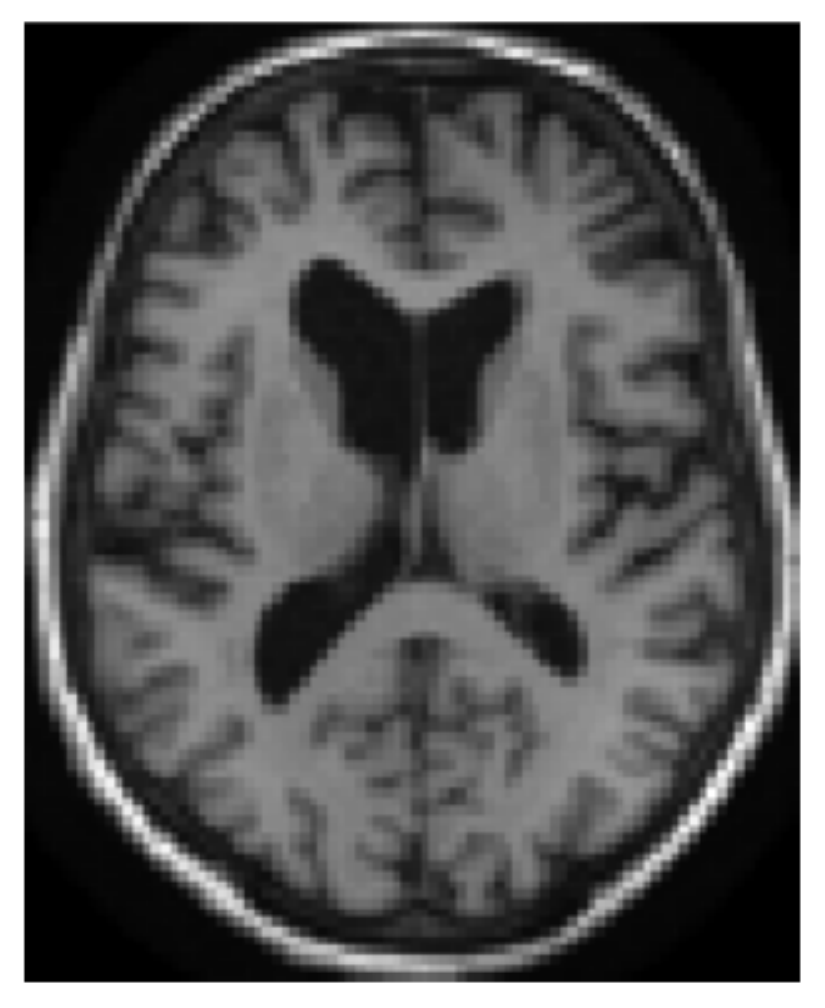}
      }
      \centering
      \subfloat{\includegraphics[width=.8in]{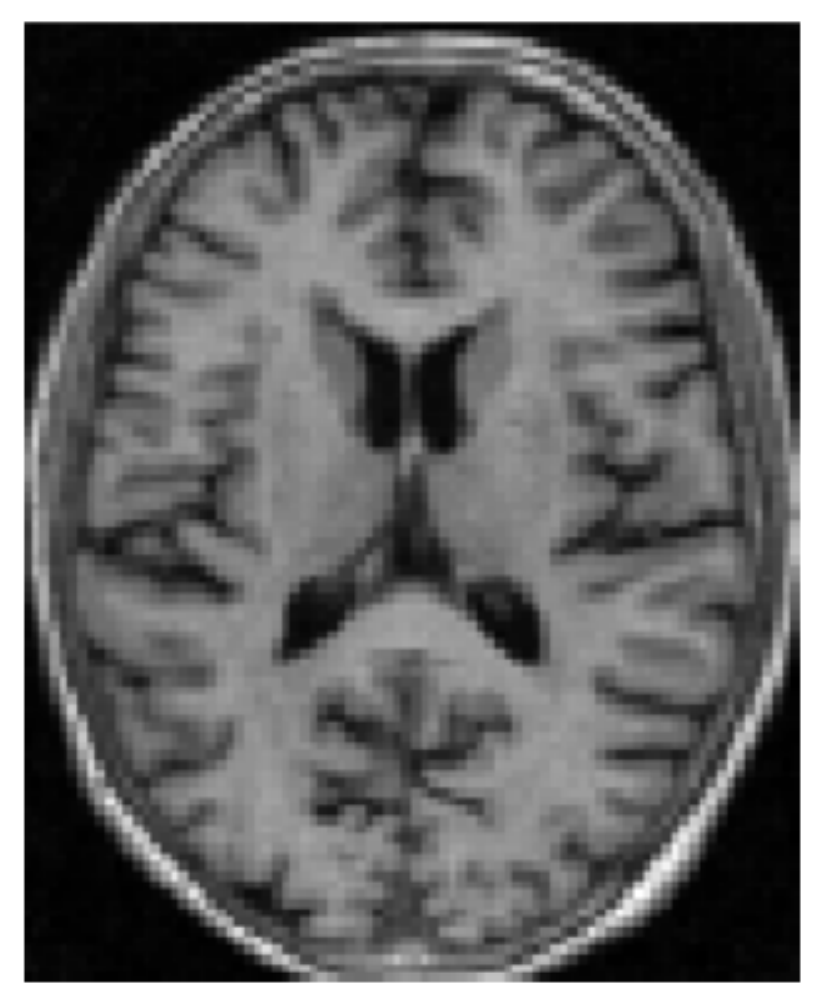}
      }
      %\vfil
      \vspace{-1.2em}
      \centering
      \subfloat{\includegraphics[width=.8in]{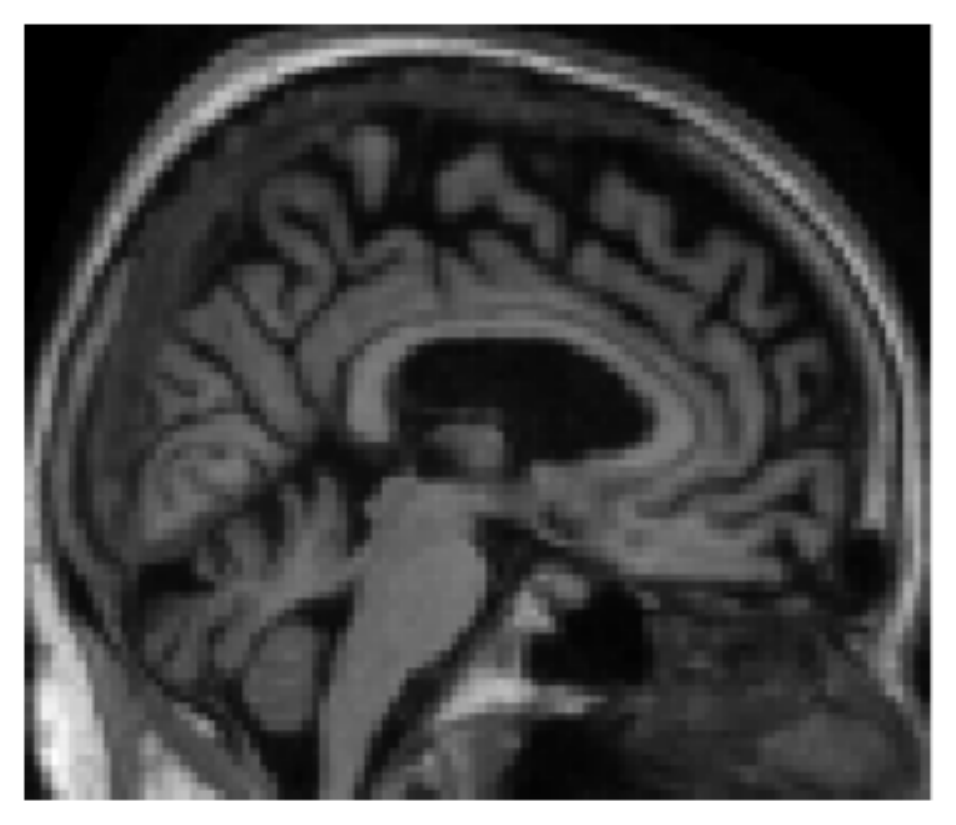}
      }
      \centering
      \subfloat{\includegraphics[width=.8in]{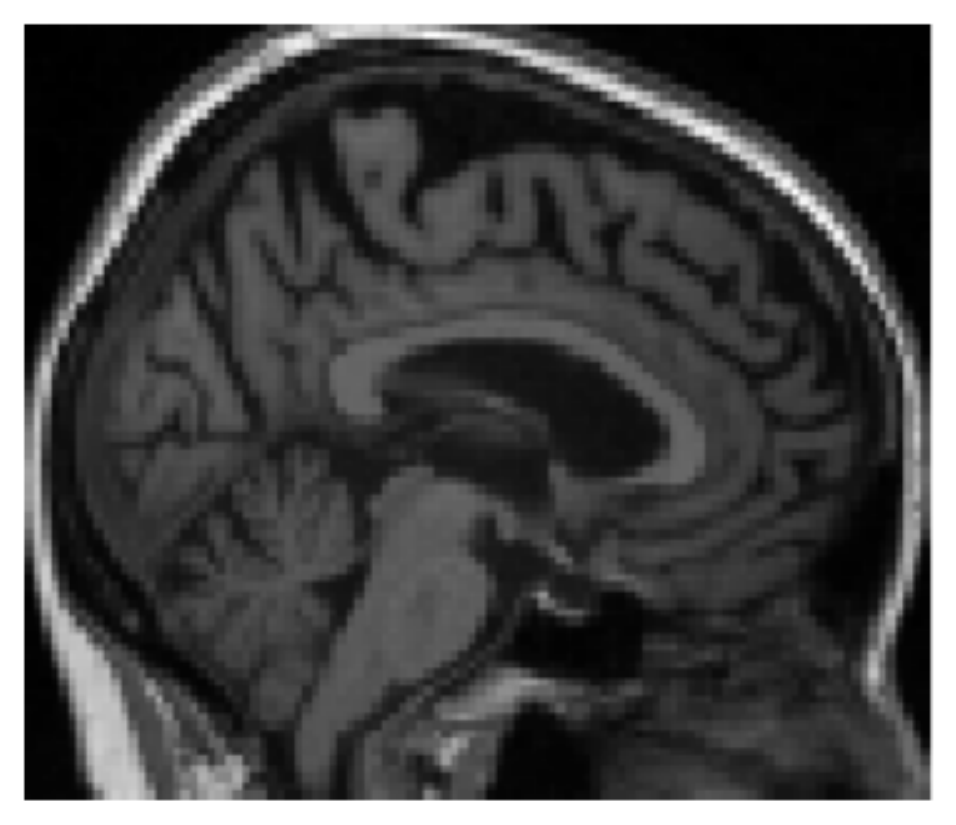}
      }
      \centering
      \subfloat{\includegraphics[width=.8in]{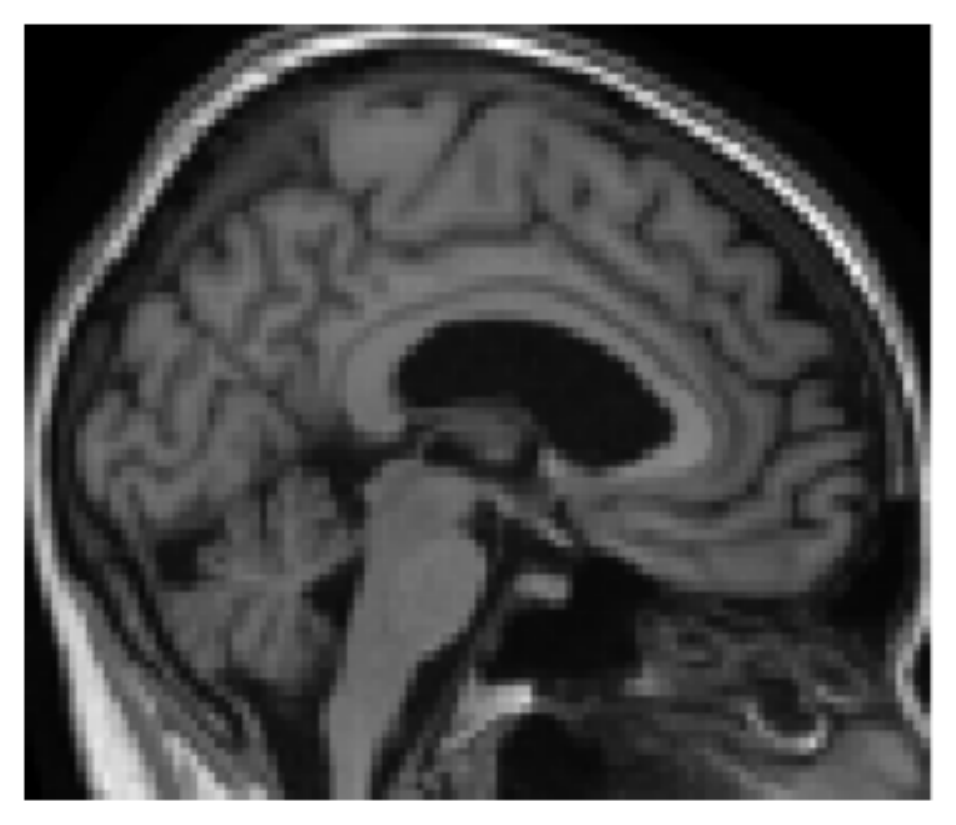}
      }
      \centering
      \subfloat{\includegraphics[width=.8in]{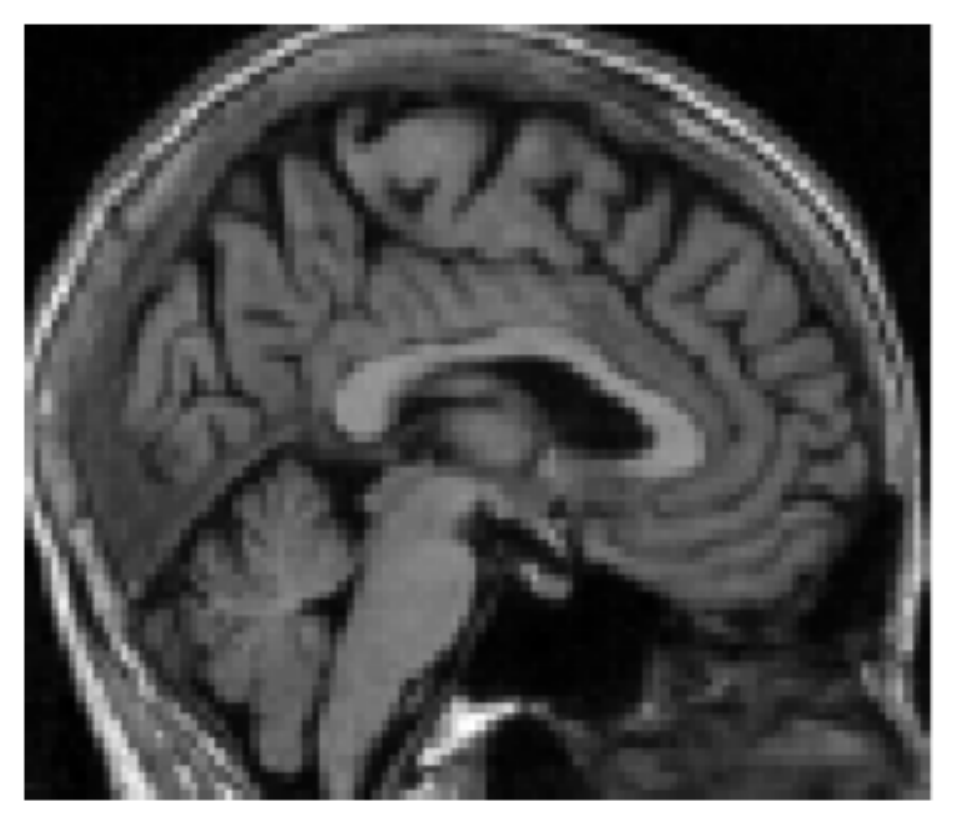}
      }   
      %\vfil
      \vspace{-1.2em}
      \centering
      \subfloat{\includegraphics[width=.8in]{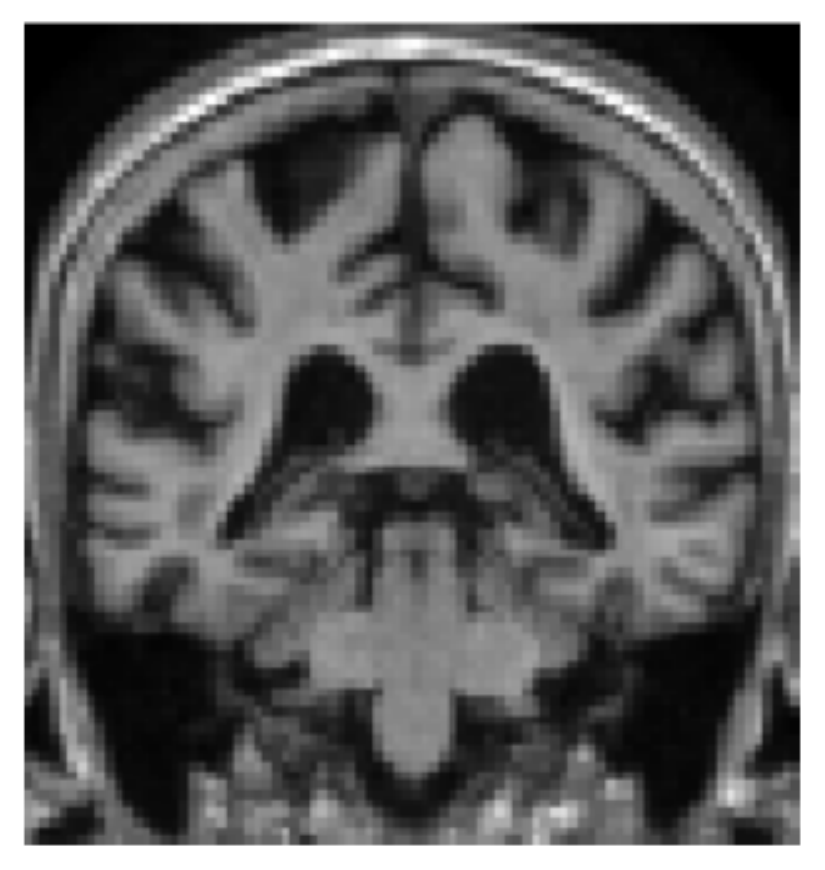}
      }
      \centering
      \subfloat{\includegraphics[width=.8in]{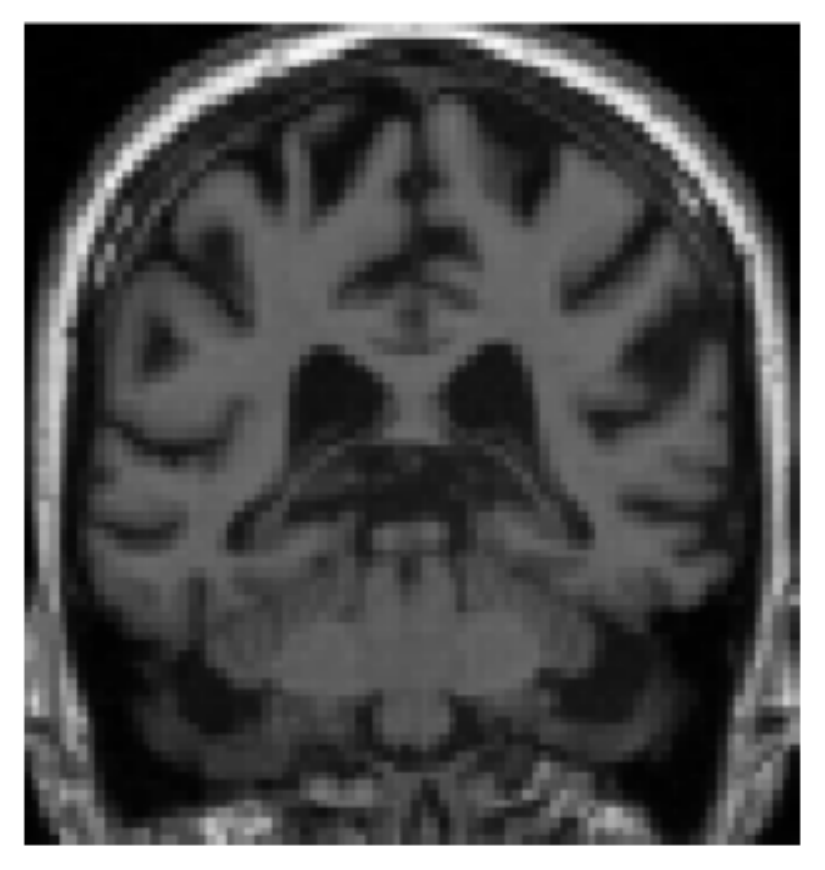}
      }
      \centering
      \subfloat{\includegraphics[width=.8in]{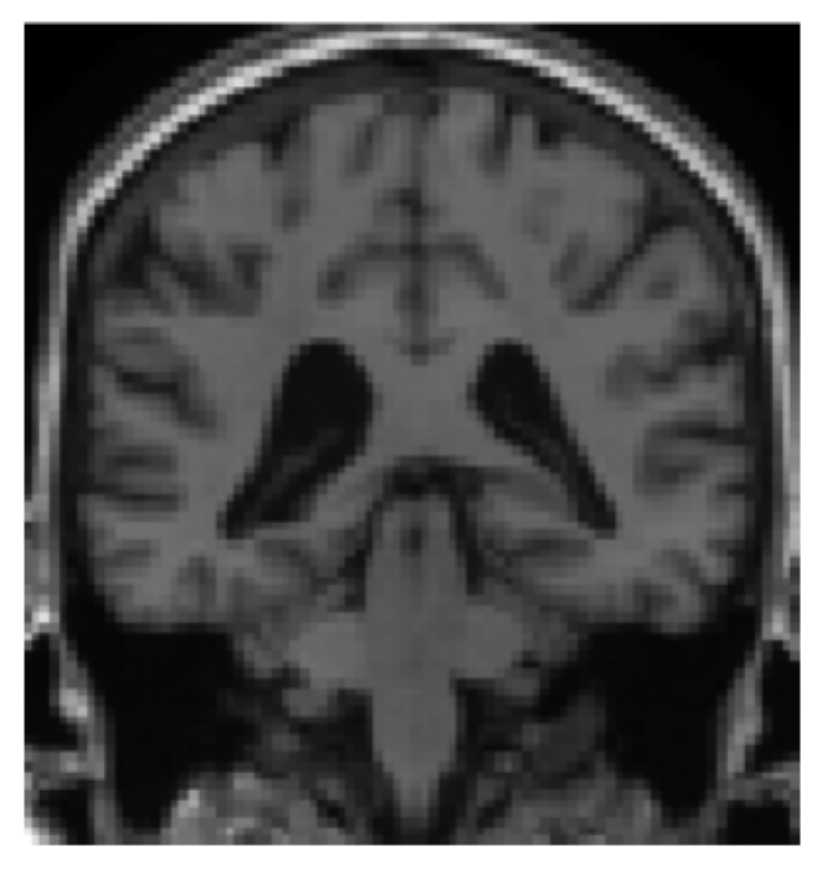}
      }
      \centering
      \subfloat{\includegraphics[width=.8in]{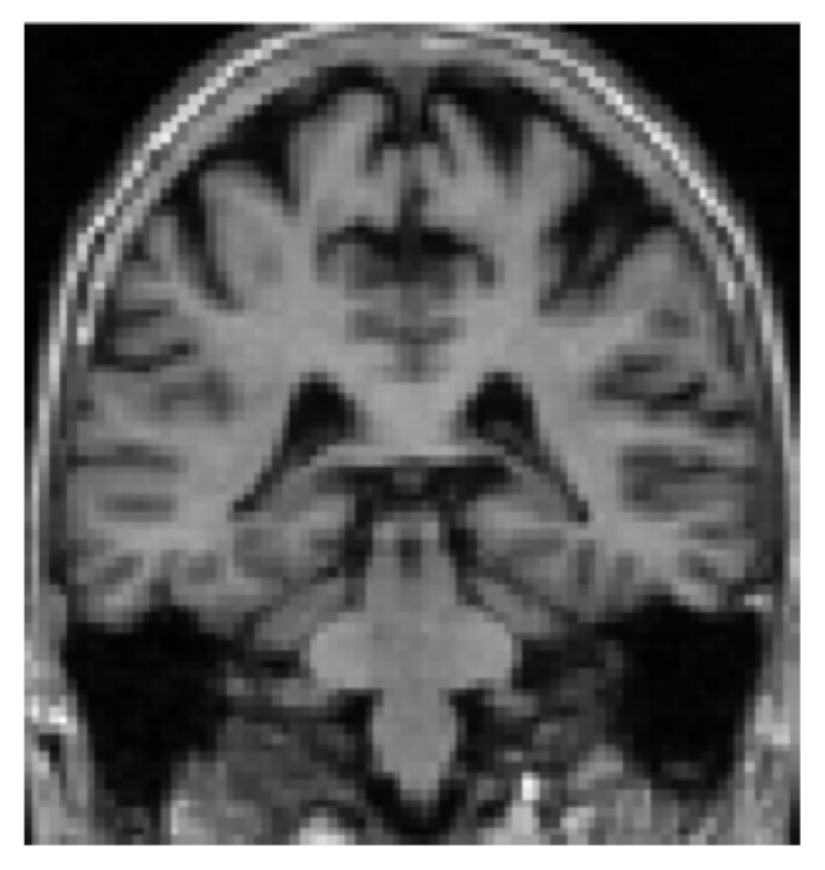}
      }        
      \caption{Example of two \emph{true} patients compared to two generated by our method. Can you find the intruders ? Answers in Appendix~\ref{appendix H}.}
      \label{Fig: ADNI generation}
      \end{figure}

As done in the previous sections, we perform three types of experiments and train the model on 1) only the \emph{real} images, 2) only on synthetic data and 3) on synthetic and real images. Due to the current implementation, augmentation on high-resolution images is not possible due to computational time and so these images are only used to assess the baseline performance of the CNN with the maximum information available. Each series of experiments is done once for each training set (\textit{train-50} and \textit{train-full}). The CNN and the VAE share the same training set, and the VAE does not use the validation set during its training. For each training set, two VAEs are trained, one on the AD label only and the other on the CN label only. Examples of real and generated AD images are shown in Fig.~\ref{Fig: ADNI generation}. For each experiment 20 runs of the CNN training are launched. The use of a smaller training set \textit{train-50} allows mimicking the behavior of the framework on smaller data sets, which are frequent in the medical domain.
%Three types of experiments were done using downsampled images:
%\begin{enumerate}
%    \item training on real images only,
%    \item training on synthetic data only,
%    \item training on synthetic data and real images.
%\end{enumerate}

%Augmentation on high-resolution images is not possible with the current implementation because of computational issues and so these images are only used to assess the baseline performance of the CNN with the maximum information available. Each series of experiments is done once for each training set (\textit{train-50} and \textit{train-full}). The CNN and the VAE share the same training set, and the VAE does not use the validation set during its training. For each training set, two VAEs are trained, one on the AD label only and the other on the CN label only. Examples of real and generated AD images are shown in Fig.~\ref{Fig: ADNI generation}. For each experiment 20 runs of the CNN training are launched. The use of a smaller training set \textit{train-50} allows mimicking the behavior of the framework on smaller data sets, which are frequent in the medical domain.

\subsection{Results}

Results presented in Table~\ref{table: perf_baseline} (resp. Table~\ref{table: perf_optimized}) are obtained with \textbf{baseline} (resp. \textbf{optimized}) hyperparameters and using either the \textit{train-full} or \textit{train-50} data set. Scores on synthetic images only are given in Appendix~\ref{appendix I}. Experiments are done on down-sampled images unless \emph{high-resolution} is specified.

\begin{table*}
\caption{Mean test performance of each series of 20 runs trained with the \textbf{baseline} hyperparameters}
\label{table: perf_baseline}
%\resizebox{\textwidth}{!}{%
\centering
\scriptsize
  \begin{tabular}{|c|c||c|c|c||c|c|c|}
    \cline{3-8}
    \multicolumn{2}{c|}{} & \multicolumn{3}{c||}{ADNI} & \multicolumn{3}{c|}{AIBL} \\
    \hline
    training & \multirow{2}{*}{data set} & \multirow{2}{*}{sensitivity} & \multirow{2}{*}{specificity} & balanced & \multirow{2}{*}{sensitivity} & \multirow{2}{*}{specificity} & balanced \\
    set & & & & accuracy & & & accuracy \\
    \hline
    \hline
    \multirow{6}{*}{\textit{train-50}} & real & $70.3\pm12.2$ & $62.4\pm11.5$ & $66.3\pm2.4$ & $60.7\pm13.7$ & $73.8\pm7.2$ & $67.2\pm4.1$ \\
    & real (high-resolution) & $78.5\pm9.4$ & $57.4\pm8.8$ & $67.9\pm2.3$ & $57.2\pm11.2$ & $75.8\pm7.0$ & $66.5\pm3.0$ \\
    \cline{2-8}
    & 500 synthetic + real & $71.9\pm5.3$ & $67.0\pm4.5$ & $69.4\pm1.6$ & $55.9\pm6.8$ & $81.1\pm3.1$ & $68.5\pm2.5$ \\
    & 2000 synthetic + real & $72.2\pm4.4$ & $70.3\pm4.3$ & $71.2\pm1.6$ & $66.6\pm7.1$ & $79.0\pm4.1$ & $72.8\pm2.2$ \\
    & 5000 synthetic + real & $\boldsymbol{74.7\pm5.3}$ & $\boldsymbol{73.5\pm4.8}$ & $\boldsymbol{74.1\pm2.2}$ & $\boldsymbol{71.7\pm10.0}$ & $80.5\pm4.4$ & $\boldsymbol{76.1\pm3.6}$ \\
    & 10000 synthetic + real & $74.7\pm7.0$ & $73.4\pm6.1$ & $74.0\pm2.7$ & $69.1\pm9.9$ & $\boldsymbol{80.7\pm5.1}$ & $74.9\pm3.2$ \\
    \hline
    \hline
    \multirow{6}{*}{\textit{train-full}} & real & $79.1\pm6.2$ & $76.3\pm4.2$ & $77.7\pm2.5$ & $70.6\pm6.7$ & $86.3\pm3.6$ & $78.4\pm2.4$ \\
    & real (high-resolution) & $84.5\pm3.8$ & $76.7\pm4.0$ & $80.6\pm1.1$ & $71.6\pm6.4$ & $89.2\pm2.7$ & $80.4\pm2.6$ \\
    \cline{2-8}
    & 500 synthetic + real & $82.5\pm3.4$ & $81.9\pm5.4$ & $82.2\pm2.4$ & $76.0\pm6.3$ & $89.7\pm3.3$ & $82.9\pm2.5$ \\
    & 2000 synthetic + real & $\boldsymbol{85.4\pm4.0}$ & $86.4\pm5.9$ & $85.9\pm1.6$ & $77.2\pm6.9$ & $90.4\pm3.8$ & $83.8\pm2.2$ \\
    & 5000 synthetic + real & $84.6\pm4.2$ & $86.9\pm3.6$ & $85.7\pm2.1$ & $76.9\pm5.2$ & $\boldsymbol{91.4\pm3.0}$ & $84.2\pm2.2$ \\
    & 10000 synthetic + real & $84.2\pm2.8$ & $\boldsymbol{88.5\pm2.9}$ & $\boldsymbol{86.3\pm1.8}$ & $\boldsymbol{79.1\pm4.7}$ & $91.0\pm2.6$ & $\boldsymbol{85.1\pm1.9}$ \\
    \hline
  \end{tabular}
\end{table*}

\begin{table*}
 \caption{Mean test performance of each series of 20 runs trained with the \textbf{optimized} hyperparameters}
 \label{table: perf_optimized}
%\resizebox{\textwidth}{!}{%
\centering
\scriptsize
  \begin{tabular}{|c|c||c|c|c||c|c|c|}
    \cline{3-8}
    \multicolumn{2}{c|}{} & \multicolumn{3}{c||}{ADNI} & \multicolumn{3}{c|}{AIBL} \\
    \hline
    training & \multirow{2}{*}{image type} & \multirow{2}{*}{sensitivity} & \multirow{2}{*}{specificity} & balanced & \multirow{2}{*}{sensitivity} & \multirow{2}{*}{specificity} & balanced\\
    set & & & & accuracy & & & accuracy\\
    \hline
    \hline
    \multirow{6}{*}{\textit{train-50}} & real & $75.4\pm5.0$ & $75.5\pm5.3$ & $75.5\pm2.7$ & $68.6\pm8.5$ & $82.6\pm4.2$ & $75.6\pm4.1$ \\
    & real (high-resolution) & $73.6\pm6.2$ & $70.6\pm5.9$ & $72.1\pm3.1$ & $57.8\pm12.3$ & $84.6\pm4.2$ & $71.2\pm5.1$ \\
    \cline{2-8}
    & 500 synthetic + real & $73.2\pm4.2$ & $78.0\pm3.3$ & $75.6\pm2.5$ & $69.2\pm9.4$ & $\boldsymbol{82.7\pm4.1}$ & $76.0\pm4.2$ \\
    & 2000 synthetic + real & $75.2\pm3.8$ & $\boldsymbol{78.6\pm4.4}$ & $76.9\pm2.4$ & $77.8\pm8.8$ & $82.2\pm4.5$ & $80.0\pm3.6$ \\
    & 5000 synthetic + real & $77.1\pm3.7$ & $76.7\pm4.1$ & $76.9\pm2.5$ & $80.7\pm6.1$ & $81.2\pm3.7$ & $80.9\pm2.7$ \\
    & 10000 synthetic + real & $\boldsymbol{77.8\pm4.6}$ & $78.2\pm4.9$ & $\boldsymbol{78.0\pm2.1}$ & $\boldsymbol{81.7\pm4.9}$ & $81.9\pm4.6$ & $\boldsymbol{81.9\pm2.2}$ \\
    \hline
    \hline
    \multirow{6}{*}{\textit{train-full}} & real & $82.5\pm4.2$ & $88.5\pm6.6$ & $85.5\pm2.4$ & $75.1\pm8.4$ & $88.7\pm9.0$ & $81.9\pm3.2$ \\
    & real (high-resolution) & $82.6\pm4.5$ & $88.9\pm6.3$ & $85.7\pm2.5$ & $78.9\pm5.4$ & $89.9\pm4.0$ & $84.4\pm1.7$ \\
    \cline{2-8}
    & 500 synthetic + real & $82.3\pm2.3$ & $89.8\pm2.7$ & $86.0\pm1.8$ & $74.9\pm5.0$ & $91.4\pm2.6$ & $83.2\pm2.4$ \\
    & 2000 synthetic + real & $\boldsymbol{83.1\pm4.2}$ & $\boldsymbol{91.3\pm3.2}$ & $\boldsymbol{87.2\pm1.7}$ & $76.0\pm4.7$ & $92.0\pm2.4$ & $84.0\pm2.0$ \\
    & 5000 synthetic + real & $81.9\pm3.5$ & $90.9\pm2.5$ & $86.4\pm1.3$ & $74.1\pm4.9$ & $\boldsymbol{92.9\pm1.9}$ & $83.5\pm2.2$ \\
    & 10000 synthetic + real & $82.2\pm3.4$ & $91.2\pm3.6$ & $86.7\pm1.8$ & $\boldsymbol{76.4\pm4.2}$ & $92.1\pm2.1$ & $\boldsymbol{84.3\pm1.8}$ \\
    \hline
    
  \end{tabular}%}
\end{table*}

% Higher variability on specificity and sensitivity than balanced accuracy $\rightarrow$ no advantage between AD and CN, cannot be predicted which one will be best classified.

Even though the VAE augmentation is performed on down-sampled images, the classification performance is at least as good as that of the best baseline performance, or can greatly exceed it:
\begin{itemize}
    \item \textit{train-50} and \textbf{baseline} model: balanced accuracy increases by 6.2 pts on ADNI and 8.9 pts on AIBL,
    \item \textit{train-full} and \textbf{baseline} model: balanced accuracy increases by 5.7 pts on ADNI and 4.7 pts on AIBL,
    \item \textit{train-50} and \textbf{optimized} model: balanced accuracy increases by 2.5 pts on ADNI and 6.3 pts on AIBL,
    \item \textit{train-full} and \textbf{optimized} model balanced accuracy increases by 1.5 pts on ADNI and -0.1 pts on AIBL.
\end{itemize}

Then, the performance increase thanks to DA is higher when using the \textbf{baseline} hyperparameters than the \textbf{optimized} ones. A possible explanation could be that the \textbf{optimized} network is already close to the maximum performance that can be reached with this setup and cannot be much improved with DA. Moreover, the VAE has not been subject to a similar search, which places it at a disadvantage. For both hyperparameters, the performance gain is higher on \textit{train-50} than on \textit{train-full}, which supports the results obtained in the previous section (see Fig.~\ref{Fig: Toy scalability}). The baseline balanced accuracy with the \textbf{baseline} hyperparameters on \textit{train-full}, 80.6\% on ADNI and 80.4\% on AIBL, are similar to the results of~\cite{wen_convolutional_2020}. With DA, we improve our balanced accuracy to 86.3\% on ADNI and 85.1\% on AIBL: this performance is similar to their result using autoencoder pretraining (which can be very long to compute) and longitudinal data (1830 CN and 1106 AD images) instead of baseline data (243 CN and 210 AD images) as we did.

In each table, the first two rows display the baseline performance obtained on real images only. As expected, training on high-resolution images leads to a better performance than training on down-sampled images. This is not the case for the \textbf{optimized} network on \textit{train-50}, which obtained a balanced accuracy of 72.1\% on ADNI and 71.2\% on AIBL with high-resolution images versus 75.5\% on ADNI and 75.6\% on AIBL with down-sampled images. This is explained by the fact that the hyperparameter choices are made on \textit{train-full} and so there is no guarantee that they could lead to similar results with fewer data samples.

%: by chance, it is the case for down-sampled images and not for high-resolution images.

\section{Discussion}

% Field generalization
Contrary to techniques that are specific to a field of application, our method produced relevant data for diverse data sets including 2D natural images (MNIST, EMNIST, Fashion and CIFAR) or 3D medical images (ADNI and AIBL). Moreover, we noted that the networks trained on ADNI gave similar balanced accuracies on the ADNI test subset and AIBL showing that our synthetic data learned on ADNI benefit in the same way AIBL, and that it did not overfit the characteristics of ADNI. In addition to the robustness across data sets, the relevance of synthetic data for diverse classifiers was assessed. For toy data, these classifiers were a MLP, a random forest, a k-NN algorithm and a SVM. On medical image data, two different CNNs were studied: a \textbf{baseline} one that has been only slightly optimized in a previous study and an \textbf{optimized} one found with a more extensive search (random search). All these classifiers performed best on augmented data than real data only. However, for medical image data, we noted that the data augmentation was more beneficial to the \textbf{baseline} network, than to the \textbf{optimized} one but both networks obtained a similar performance with data augmentation on the largest training set. This means that data augmentation could avoid spending time and/or resources optimizing a classifier. The ability of the model to generate relevant data and enrich the original training data was also supported by the fact that almost all classifiers could achieve a better classification performance when trained only on synthetic data than on the \emph{real} train. The method scalability to larger data sets and more complex models was also discussed. 

% Classifier independence (above)

% Synthetic vs augmented data (above)

% Result in HDLSS medical setting - reproducibility
Our generation framework appears also very well suited to perform data augmentation in a HDLSS setting (the binary classification of AD and CN subjects using T1w MRI). In all cases, the classification performance was at least as good as the maximum performance obtained with real data and could even be much better. For instance, the method allowed the balanced accuracy of the \textbf{baseline} CNN to jump from 66.3\% to 74.3\% when trained with only 50 images per class and from 77.7\% to 86.3\% when trained with 243 CN and 210 AD while still improving greatly sensitivity and specificity metrics. We witnessed a greater performance improvement than the other studies using a CNN on T1w MRI to differentiate AD and CN subjects~\cite{valliani_deep_2017, backstrom_efficient_2018, cheng_cnns_2017, aderghal_classification_2017, aderghal_classification_2018}. Indeed, these studies used simple transforms (affine and pixel-wise) that may not bring enough variability to improve the CNN performance. Though many complex methods now exist to perform data augmentation, they are still not widely adopted in the field of medical imaging. We suspect that this is mainly due to the lack of reproducibility of such frameworks. Hence we provide the source code, as well as scripts to easily reproduce the experiments of this paper from the ADNI and AIBL data set download to the final evaluation of the CNN performance. We also developed a software \footnote{\url{https://github.com/clementchadebec/pyraug}} implementing the method and making it easily accessible to the community. 

% what could be improved - limitations
However, our classification performance on synthetic data could be improved in many ways. First, we chose in this study not to spend much time optimizing the VAE's hyperparameters and so in Sec.~\ref{Sec: Medical images} we chose to work with down-sampled images to deal with memory issues. We could look for another architecture to train the VAE directly on high-resolution images leading potentially to a better performance as witnessed in experiments on real images only. Moreover, we could couple the advantages of other techniques such as autoencoder pretraining or weak supervision to our data augmentation framework. However, the advantages may not stack as observed when using DA on optimized hyperparameters. Finally, we chose to train our networks with only one image per participant, but our framework could also benefit from the use of the whole follow-up of all patients to further improve performance. However, a long follow-up is rather an exception in the context of medical imaging. This is why we assessed the relevance of our DA framework in the context of small data sets which is a main issue in this field. Nonetheless, a training set of 50 images per class can still be seen as large in the case of rare diseases and so it may be interesting to evaluate the reliability of our method on even smaller training sets.

\section{Conclusion}

In this paper, we proposed a new VAE-based data augmentation framework whose performance and robustness were validated on classification tasks on \emph{toy} and \emph{real-life} data sets. This method relies on a model combining a proper latent space modeling of the VAE seen as a Riemannian manifold and a new generation procedure exploiting such geometrical aspects. In particular, the generation method does not use the prior as is standard since we showed that, depending on its choice and the data set considered, it may lead to a very poor latent space prospecting and a degraded sampling while the proposed method does not suffer from such drawbacks. The proposed amendments were motivated, discussed and compared to other VAE models and demonstrated promising results. The model indeed appeared to be able to generate new data faithfully and demonstrated a strong generalization power which makes it very well suited to perform data augmentation even in the challenging context of HDLSS data.
For each augmentation experiment, it was able to enrich the initial data set so that a classifier performs better on augmented data than only on the \emph{real} ones. Future work would consist in building a framework able to handle longitudinal data and so able to generate not only one image but a whole patient trajectory.

\section*{Acknowledgment}

The research leading to these results has received funding from the French government under management of Agence Nationale de la Recherche as part of the ``Investissements d'avenir'' program, reference ANR-19-P3IA-0001 (PRAIRIE 3IA Institute) and reference ANR-10-IAIHU-06 (Agence Nationale de la Recherche-10-IA Institut Hospitalo-Universitaire-6). 
This work was granted access to the HPC resources of IDRIS under the allocation 101637 made by GENCI (Grand Équipement National de Calcul Intensif).

Data collection and sharing for this project was funded by the Alzheimer's Disease Neuroimaging Initiative (ADNI) (National Institutes of Health Grant U01 AG024904) and DOD ADNI (Department of Defense award number W81XWH-12-2-0012). ADNI is funded by the National Institute on Aging, the National Institute of Biomedical Imaging and Bioengineering, and through generous contributions from the following: AbbVie, Alzheimer's Association; Alzheimer's Drug Discovery Foundation; Araclon Biotech; BioClinica, Inc.; Biogen; Bristol-Myers Squibb Company; CereSpir, Inc.; Cogstate; Eisai Inc.; Elan Pharmaceuticals, Inc.; Eli Lilly and Company; EuroImmun; F. Hoffmann-La Roche Ltd and its affiliated company Genentech, Inc.; Fujirebio; GE Healthcare; IXICO Ltd.; Janssen Alzheimer Immunotherapy Research \& Development, LLC.; Johnson \& Johnson Pharmaceutical Research \& Development LLC.; Lumosity; Lundbeck; Merck \& Co., Inc.; Meso Scale Diagnostics, LLC.; NeuroRx Research; Neurotrack Technologies; Novartis Pharmaceuticals Corporation; Pfizer Inc.; Piramal Imaging; Servier; Takeda Pharmaceutical Company; and Transition Therapeutics. The Canadian Institutes of Health Research is providing funds to support ADNI clinical sites in Canada. Private sector contributions are facilitated by the Foundation for the National Institutes of Health (\url{www.fnih.org}). The grantee organization is the Northern California Institute for Research and Education, and the study is coordinated by the Alzheimer's Therapeutic Research Institute at the University of Southern California. ADNI data are disseminated by the Laboratory for Neuro Imaging at the University of Southern California.

% Can use something like this to put references on a page
% by themselves when using endfloat and the captionsoff option.
\ifCLASSOPTIONcaptionsoff
  \newpage
\fi

% trigger a \newpage just before the given reference
% number - used to balance the columns on the last page
% adjust value as needed - may need to be readjusted if
% the document is modified later
%\IEEEtriggeratref{8}
% The "triggered" command can be changed if desired:
%\IEEEtriggercmd{\enlargethispage{-5in}}

% references section

% can use a bibliography generated by BibTeX as a .bbl file
% BibTeX documentation can be easily obtained at:
% http://mirror.ctan.org/biblio/bibtex/contrib/doc/
% The IEEEtran BibTeX style support page is at:
% http://www.michaelshell.org/tex/ieeetran/bibtex/
\bibliographystyle{IEEEtran}
% argument is your BibTeX string definitions and bibliography database(s)
\bibliography{IEEEabrv,references_wo_url}

% Generated by IEEEtran.bst, version: 1.14 (2015/08/26)
\begin{thebibliography}{100}
\providecommand{\url}[1]{#1}
\csname url@samestyle\endcsname
\providecommand{\newblock}{\relax}
\providecommand{\bibinfo}[2]{#2}
\providecommand{\BIBentrySTDinterwordspacing}{\spaceskip=0pt\relax}
\providecommand{\BIBentryALTinterwordstretchfactor}{4}
\providecommand{\BIBentryALTinterwordspacing}{\spaceskip=\fontdimen2\font plus
\BIBentryALTinterwordstretchfactor\fontdimen3\font minus
  \fontdimen4\font\relax}
\providecommand{\BIBforeignlanguage}[2]{{%
\expandafter\ifx\csname l@#1\endcsname\relax
\typeout{** WARNING: IEEEtran.bst: No hyphenation pattern has been}%
\typeout{** loaded for the language `#1'. Using the pattern for}%
\typeout{** the default language instead.}%
\else
\language=\csname l@#1\endcsname
\fi
#2}}
\providecommand{\BIBdecl}{\relax}
\BIBdecl

\bibitem{button_power_2013}
K.~S. Button, J.~P. Ioannidis, C.~Mokrysz, B.~A. Nosek, J.~Flint, E.~S.
  Robinson, and M.~R. Munafò, ``Power failure: why small sample size
  undermines the reliability of neuroscience,'' \emph{Nature Reviews
  Neuroscience}, vol.~14, no.~5, pp. 365--376, 2013.

\bibitem{turner_small_2018}
B.~O. Turner, E.~J. Paul, M.~B. Miller, and A.~K. Barbey, ``Small sample sizes
  reduce the replicability of task-based {fMRI} studies,'' \emph{Communications
  Biology}, vol.~1, no.~1, pp. 1--10, 2018.

\bibitem{goodfellow_deep_2016}
I.~Goodfellow, Y.~Bengio, A.~Courville, and Y.~Bengio, \emph{Deep
  learning}.\hskip 1em plus 0.5em minus 0.4em\relax {MIT} press Cambridge,
  2016, vol.~1, issue: 2.

\bibitem{shorten_survey_2019}
C.~Shorten and T.~M. Khoshgoftaar, ``A survey on {Image} {Data} {Augmentation}
  for {Deep} {Learning},'' \emph{Journal of Big Data}, vol.~6, no.~1, p.~60,
  2019.

\bibitem{tanner_calculation_1987}
M.~A. Tanner and W.~H. Wong, ``The calculation of posterior distributions by
  data augmentation,'' \emph{Journal of the American statistical Association},
  vol.~82, no. 398, pp. 528--540, 1987.

\bibitem{chawla_smote_2002}
N.~V. Chawla, K.~W. Bowyer, L.~O. Hall, and W.~P. Kegelmeyer, ``{SMOTE}:
  synthetic minority over-sampling technique,'' \emph{Journal of artificial
  intelligence research}, vol.~16, pp. 321--357, 2002.

\bibitem{hutchison_borderline-smote_2005}
H.~Han, W.-Y. Wang, and B.-H. Mao, ``Borderline-{SMOTE}: A new over-sampling
  method in imbalanced data sets learning,'' in \emph{Advances in Intelligent
  Computing}, D.-S. Huang, X.-P. Zhang, and G.-B. Huang, Eds.\hskip 1em plus
  0.5em minus 0.4em\relax Springer Berlin Heidelberg, 2005, vol. 3644, pp.
  878--887, series Title: LNCS.

\bibitem{nguyen_borderline_2011}
H.~M. Nguyen, E.~W. Cooper, and K.~Kamei, ``Borderline over-sampling for
  imbalanced data classification,'' \emph{International Journal of Knowledge
  Engineering and Soft Data Paradigms}, vol.~3, no.~1, pp. 4--21, 2011.

\bibitem{haibo_he_adasyn_2008}
{Haibo He}, {Yang Bai}, E.~A. Garcia, and {Shutao Li}, ``{ADASYN}: Adaptive
  synthetic sampling approach for imbalanced learning,'' in \emph{2008 {IEEE}
  International Joint Conference on Neural Networks ({IEEE} World Congress on
  Computational Intelligence)}.\hskip 1em plus 0.5em minus 0.4em\relax {IEEE},
  2008, pp. 1322--1328.

\bibitem{barua_mwmote--majority_2012}
S.~Barua, M.~M. Islam, X.~Yao, and K.~Murase, ``{MWMOTE}--majority weighted
  minority oversampling technique for imbalanced data set learning,''
  \emph{{IEEE} Transactions on Knowledge and Data Engineering}, vol.~26, no.~2,
  pp. 405--425, 2012.

\bibitem{blagus_smote_2013}
R.~Blagus and L.~Lusa, ``{SMOTE} for high-dimensional class-imbalanced data,''
  \emph{{BMC} Bioinformatics}, vol.~14, no.~1, p. 106, 2013.

\bibitem{fernandez_smote_2018}
A.~Fernández, S.~Garcia, F.~Herrera, and N.~V. Chawla, ``{SMOTE} for learning
  from imbalanced data: progress and challenges, marking the 15-year
  anniversary,'' \emph{Journal of artificial intelligence research}, vol.~61,
  pp. 863--905, 2018.

\bibitem{goodfellow_generative_2014}
I.~Goodfellow, J.~Pouget-Abadie, M.~Mirza, B.~Xu, D.~Warde-Farley, S.~Ozair,
  A.~Courville, and Y.~Bengio, ``Generative adversarial nets,'' in
  \emph{Advances in Neural Information Processing Systems}, 2014, pp.
  2672--2680.

\bibitem{kingma_auto-encoding_2014}
D.~P. Kingma and M.~Welling, ``Auto-encoding variational bayes,''
  \emph{{arXiv}:1312.6114 [cs, stat]}, 2014.

\bibitem{rezende_stochastic_2014}
D.~J. Rezende, S.~Mohamed, and D.~Wierstra, ``Stochastic backpropagation and
  approximate inference in deep generative models,'' in \emph{International
  conference on machine learning}.\hskip 1em plus 0.5em minus 0.4em\relax PMLR,
  2014, pp. 1278--1286.

\bibitem{zhu_emotion_2018}
X.~Zhu, Y.~Liu, J.~Li, T.~Wan, and Z.~Qin, ``Emotion classification with data
  augmentation using generative adversarial networks,'' in \emph{Pacific-Asia
  conference on knowledge discovery and data mining}.\hskip 1em plus 0.5em
  minus 0.4em\relax Springer, 2018, pp. 349--360.

\bibitem{mariani_bagan_2018}
G.~Mariani, F.~Scheidegger, R.~Istrate, C.~Bekas, and C.~Malossi, ``{{BAGAN}}:
  {{Data Augmentation}} with {{Balancing GAN}},'' \emph{arXiv:1803.09655},
  2018.

\bibitem{antoniou_data_2018}
A.~Antoniou, A.~Storkey, and H.~Edwards, ``Data augmentation generative
  adversarial networks,'' \emph{{arXiv}:1711.04340 [cs, stat]}, 2018-03-21.

\bibitem{lim_doping_2018}
S.~K. Lim, Y.~Loo, N.-T. Tran, N.-M. Cheung, G.~Roig, and Y.~Elovici, ``Doping:
  Generative data augmentation for unsupervised anomaly detection with gan,''
  in \emph{2018 IEEE International Conference on Data Mining (ICDM)}.\hskip 1em
  plus 0.5em minus 0.4em\relax IEEE, 2018, pp. 1122--1127.

\bibitem{zhu_data_2018}
Y.~Zhu, M.~Aoun, M.~Krijn, J.~Vanschoren, and H.~T. Campus, ``Data
  {Augmentation} using {Conditional} {Generative} {Adversarial} {Networks} for
  {Leaf} {Counting} in {Arabidopsis} {Plants}.'' in \emph{{BMVC}}, 2018, p.
  324.

\bibitem{yi_generative_2019}
X.~Yi, E.~Walia, and P.~Babyn, ``Generative adversarial network in medical
  imaging: A review,'' \emph{Medical image analysis}, vol.~58, p. 101552, 2019.

\bibitem{shin_medical_2018}
H.-C. Shin, N.~A. Tenenholtz, J.~K. Rogers, C.~G. Schwarz, M.~L. Senjem, J.~L.
  Gunter, K.~P. Andriole, and M.~Michalski, ``Medical image synthesis for data
  augmentation and anonymization using generative adversarial networks,'' in
  \emph{{International Workshop on Simulation and Synthesis in Medical
  Imaging}}, ser. LNCS.\hskip 1em plus 0.5em minus 0.4em\relax Springer, 2018,
  pp. 1--11.

\bibitem{calimeri_biomedical_2017}
F.~Calimeri, A.~Marzullo, C.~Stamile, and G.~Terracina, ``Biomedical data
  augmentation using generative adversarial neural networks,'' in
  \emph{International conference on artificial neural networks}.\hskip 1em plus
  0.5em minus 0.4em\relax Springer, 2017, pp. 626--634.

\bibitem{frid-adar_gan-based_2018}
M.~Frid-Adar, I.~Diamant, E.~Klang, M.~Amitai, J.~Goldberger, and H.~Greenspan,
  ``{GAN}-based synthetic medical image augmentation for increased {CNN}
  performance in liver lesion classification,'' \emph{Neurocomputing}, vol.
  321, pp. 321--331, 2018.

\bibitem{sandfort_data_2019}
V.~Sandfort, K.~Yan, P.~J. Pickhardt, and R.~M. Summers, ``Data augmentation
  using generative adversarial networks ({CycleGAN}) to improve
  generalizability in {CT} segmentation tasks,'' \emph{Scientific reports},
  vol.~9, no.~1, p. 16884, 2019.

\bibitem{madani_chest_2018}
A.~Madani, M.~Moradi, A.~Karargyris, and T.~Syeda-Mahmood, ``Chest x-ray
  generation and data augmentation for cardiovascular abnormality
  classification,'' in \emph{Medical Imaging 2018: Image Processing}, vol.
  10574.\hskip 1em plus 0.5em minus 0.4em\relax International Society for
  Optics and Photonics, 2018, p. 105741M.

\bibitem{salehinejad_generalization_2018}
H.~Salehinejad, S.~Valaee, T.~Dowdell, E.~Colak, and J.~Barfett,
  ``Generalization of deep neural networks for chest pathology classification
  in x-rays using generative adversarial networks,'' in \emph{2018 IEEE
  International Conference on Acoustics, Speech and Signal Processing
  (ICASSP)}.\hskip 1em plus 0.5em minus 0.4em\relax IEEE, 2018, pp. 990--994.

\bibitem{waheed_covidgan_2020}
A.~Waheed, M.~Goyal, D.~Gupta, A.~Khanna, F.~Al-Turjman, and P.~R. Pinheiro,
  ``Covidgan: data augmentation using auxiliary classifier gan for improved
  covid-19 detection,'' \emph{Ieee Access}, vol.~8, pp. 91\,916--91\,923, 2020.

\bibitem{bi_synthesis_2017}
L.~Bi, J.~Kim, A.~Kumar, D.~Feng, and M.~Fulham, ``Synthesis of {{Positron
  Emission Tomography}} ({{PET}}) {{Images}} via {{Multi}}-channel {{Generative
  Adversarial Networks}} ({{GANs}}),'' in \emph{Molecular {{Imaging}},
  {{Reconstruction}} and {{Analysis}} of {{Moving Body Organs}}, and {{Stroke
  Imaging}} and {{Treatment}}}, ser. {LNCS}.\hskip 1em plus 0.5em minus
  0.4em\relax {Springer}, 2017, pp. 43--51.

\bibitem{liu_wasserstein_2019}
Y.~Liu, Y.~Zhou, X.~Liu, F.~Dong, C.~Wang, and Z.~Wang, ``Wasserstein gan-based
  small-sample augmentation for new-generation artificial intelligence: a case
  study of cancer-staging data in biology,'' \emph{Engineering}, vol.~5, no.~1,
  pp. 156--163, 2019.

\bibitem{baur_generating_2018}
C.~Baur, S.~Albarqouni, and N.~Navab, ``Generating highly realistic images of
  skin lesions with {GANs},'' in \emph{{OR} 2.0 Context-Aware Operating
  Theaters, Computer Assisted Robotic Endoscopy, Clinical Image-Based
  Procedures, and Skin Image Analysis}.\hskip 1em plus 0.5em minus 0.4em\relax
  Springer, 2018, pp. 260--267.

\bibitem{korkinof_high-resolution_2018}
D.~Korkinof, T.~Rijken, M.~O'Neill, J.~Yearsley, H.~Harvey, and B.~Glocker,
  ``High-resolution mammogram synthesis using progressive generative
  adversarial networks,'' \emph{{arXiv} preprint {arXiv}:1807.03401}, 2018.

\bibitem{wu_conditional_2018}
E.~Wu, K.~Wu, D.~Cox, and W.~Lotter, ``Conditional infilling gans for data
  augmentation in mammogram classification,'' in \emph{Image analysis for
  moving organ, breast, and thoracic images}.\hskip 1em plus 0.5em minus
  0.4em\relax Springer, 2018, pp. 98--106.

\bibitem{hsu_unsupervised_2017}
W.-N. Hsu, Y.~Zhang, and J.~Glass, ``Unsupervised domain adaptation for robust
  speech recognition via variational autoencoder-based data augmentation,'' in
  \emph{2017 IEEE Automatic Speech Recognition and Understanding Workshop
  (ASRU)}.\hskip 1em plus 0.5em minus 0.4em\relax IEEE, 2017, pp. 16--23.

\bibitem{nishizaki_data_2017}
H.~Nishizaki, ``Data augmentation and feature extraction using variational
  autoencoder for acoustic modeling,'' in \emph{2017 Asia-Pacific Signal and
  Information Processing Association Annual Summit and Conference ({APSIPA}
  {ASC})}.\hskip 1em plus 0.5em minus 0.4em\relax {IEEE}, 2017, pp. 1222--1227.

\bibitem{wu_data_2019}
Z.~Wu, S.~Wang, Y.~Qian, and K.~Yu, ``Data augmentation using variational
  autoencoder for embedding based speaker verification,'' in \emph{Interspeech
  2019}.\hskip 1em plus 0.5em minus 0.4em\relax {ISCA}, 2019, pp. 1163--1167.

\bibitem{zhuang_fmri_2019}
P.~Zhuang, A.~G. Schwing, and O.~Koyejo, ``{fMRI data augmentation via
  synthesis},'' in \emph{2019 IEEE 16th International Symposium on Biomedical
  Imaging (ISBI 2019)}.\hskip 1em plus 0.5em minus 0.4em\relax IEEE, 2019, pp.
  1783--1787.

\bibitem{liu_data_2018}
X.~Liu, Y.~Zou, L.~Kong, Z.~Diao, J.~Yan, J.~Wang, S.~Li, P.~Jia, and J.~You,
  ``Data augmentation via latent space interpolation for image
  classification,'' in \emph{2018 24th International Conference on Pattern
  Recognition (ICPR)}.\hskip 1em plus 0.5em minus 0.4em\relax IEEE, 2018, pp.
  728--733.

\bibitem{painchaud_cardiac_2019}
N.~Painchaud, Y.~Skandarani, T.~Judge, O.~Bernard, A.~Lalande, and P.-M.
  Jodoin, ``{Cardiac MRI segmentation with strong anatomical guarantees},'' in
  \emph{International Conference on Medical Image Computing and
  Computer-Assisted Intervention}.\hskip 1em plus 0.5em minus 0.4em\relax
  Springer, 2019, pp. 632--640.

\bibitem{selvan_lung_2020}
R.~Selvan, E.~B. Dam, N.~S. Detlefsen, S.~Rischel, K.~Sheng, M.~Nielsen, and
  A.~Pai, ``Lung segmentation from chest x-rays using variational data
  imputation,'' \emph{{arXiv}:2005.10052 [cs, eess, stat]}, 2020.

\bibitem{myronenko_3d_2018}
A.~Myronenko, ``{3D MRI brain tumor segmentation using autoencoder
  regularization},'' in \emph{International MICCAI Brainlesion Workshop}.\hskip
  1em plus 0.5em minus 0.4em\relax Springer, 2018, pp. 311--320.

\bibitem{jordan_introduction_1999}
M.~I. Jordan, Z.~Ghahramani, T.~S. Jaakkola, and L.~K. Saul, ``An introduction
  to variational methods for graphical models,'' \emph{Machine Learning},
  vol.~37, no.~2, pp. 183--233, 1999.

\bibitem{burda_importance_2016}
Y.~Burda, R.~Grosse, and R.~Salakhutdinov, ``Importance weighted
  autoencoders,'' \emph{{arXiv}:1509.00519 [cs, stat]}, 2016-11-07.

\bibitem{alemi_deep_2016}
A.~A. Alemi, I.~Fischer, J.~V. Dillon, and K.~Murphy, ``Deep variational
  information bottleneck,'' \emph{{arXiv} preprint {arXiv}:1612.00410}, 2016.

\bibitem{higgins_beta-vae_2017}
I.~Higgins, L.~Matthey, A.~Pal, C.~Burgess, X.~Glorot, M.~Botvinick,
  S.~Mohamed, and A.~Lerchner, ``beta-{VAE}: Learning basic visual concepts
  with a constrained variational framework.'' \emph{{ICLR}}, vol.~2, no.~5,
  p.~6, 2017.

\bibitem{cremer_inference_2018}
C.~Cremer, X.~Li, and D.~Duvenaud, ``Inference suboptimality in variational
  autoencoders,'' in \emph{International Conference on Machine Learning}.\hskip
  1em plus 0.5em minus 0.4em\relax PMLR, 2018, pp. 1078--1086.

\bibitem{zhang_advances_2018}
C.~Zhang, J.~Bütepage, H.~Kjellström, and S.~Mandt, ``Advances in variational
  inference,'' \emph{{IEEE} Transactions on Pattern Analysis and Machine
  Intelligence}, vol.~41, no.~8, pp. 2008--2026, 2018.

\bibitem{ruiz_contrastive_2019}
F.~Ruiz and M.~Titsias, ``A contrastive divergence for combining variational
  inference and mcmc,'' in \emph{International Conference on Machine
  Learning}.\hskip 1em plus 0.5em minus 0.4em\relax PMLR, 2019, pp. 5537--5545.

\bibitem{salimans_markov_2015}
T.~Salimans, D.~Kingma, and M.~Welling, ``Markov chain monte carlo and
  variational inference: Bridging the gap,'' in \emph{International Conference
  on Machine Learning}, 2015, pp. 1218--1226.

\bibitem{rezende_variational_2015}
D.~Rezende and S.~Mohamed, ``Variational inference with normalizing flows,'' in
  \emph{International Conference on Machine Learning}.\hskip 1em plus 0.5em
  minus 0.4em\relax PMLR, 2015, pp. 1530--1538.

\bibitem{neal_mcmc_2011}
R.~M. Neal and {others}, ``{MCMC} using hamiltonian dynamics,'' \emph{Handbook
  of Markov Chain Monte Carlo}, vol.~2, no.~11, p.~2, 2011.

\bibitem{caterini_hamiltonian_2018}
A.~L. Caterini, A.~Doucet, and D.~Sejdinovic, ``Hamiltonian variational
  auto-encoder,'' in \emph{Advances in Neural Information Processing Systems},
  2018, pp. 8167--8177.

\bibitem{hoffman_elbo_2016}
M.~D. Hoffman and M.~J. Johnson, ``Elbo surgery: yet another way to carve up
  the variational evidence lower bound,'' in \emph{Workshop in Advances in
  Approximate Bayesian Inference, {NIPS}}, vol.~1, 2016, p.~2.

\bibitem{nalisnick_approximate_2016}
E.~Nalisnick, L.~Hertel, and P.~Smyth, ``Approximate inference for deep latent
  gaussian mixtures,'' in \emph{NIPS Workshop on Bayesian Deep Learning},
  vol.~2, 2016, p. 131.

\bibitem{dilokthanakul_deep_2017}
N.~Dilokthanakul, P.~A.~M. Mediano, M.~Garnelo, M.~C.~H. Lee, H.~Salimbeni,
  K.~Arulkumaran, and M.~Shanahan, ``Deep unsupervised clustering with gaussian
  mixture variational autoencoders,'' \emph{{arXiv}:1611.02648 [cs, stat]},
  2017.

\bibitem{tomczak_vae_2018}
J.~Tomczak and M.~Welling, ``Vae with a vampprior,'' in \emph{International
  Conference on Artificial Intelligence and Statistics}.\hskip 1em plus 0.5em
  minus 0.4em\relax PMLR, 2018, pp. 1214--1223.

\bibitem{sonderby_ladder_2016}
C.~K. S{\o}nderby, T.~Raiko, L.~Maal{\o}e, S.~K. S{\o}nderby, and O.~Winther,
  ``Ladder variational autoencoder,'' in \emph{29th Annual Conference on Neural
  Information Processing Systems (NIPS 2016)}, 2016.

\bibitem{klushyn_learning_2019}
A.~Klushyn, N.~Chen, R.~Kurle, and B.~Cseke, ``Learning {Hierarchical} {Priors}
  in {VAEs},'' \emph{Advances in neural information processing systems}, p.~10,
  2019.

\bibitem{chen_variational_2016}
X.~Chen, D.~P. Kingma, T.~Salimans, Y.~Duan, P.~Dhariwal, J.~Schulman,
  I.~Sutskever, and P.~Abbeel, ``Variational lossy autoencoder,'' \emph{{arXiv}
  preprint {arXiv}:1611.02731}, 2016.

\bibitem{razavi_generating_2019}
A.~Razavi, A.~v.~d. Oord, and O.~Vinyals, ``Generating diverse high-fidelity
  images with vq-vae-2,'' \emph{Advances in Neural Information Processing
  Systems}, 2020.

\bibitem{pang_learning_2020}
B.~Pang, T.~Han, E.~Nijkamp, S.-C. Zhu, and Y.~N. Wu, ``Learning latent space
  energy-based prior model,'' \emph{Advances in Neural Information Processing
  Systems}, vol.~33, 2020.

\bibitem{aneja_ncp-vae_2020}
J.~Aneja, A.~Schwing, J.~Kautz, and A.~Vahdat, ``{NCP}-{VAE}: Variational
  autoencoders with noise contrastive priors,'' \emph{{arXiv}:2010.02917 [cs,
  stat]}, 2020.

\bibitem{ghosh_variational_2020}
P.~Ghosh, M.~S. Sajjadi, A.~Vergari, M.~Black, and B.~Schölkopf, ``From
  variational to deterministic autoencoders,'' in \emph{8th {International}
  {Conference} on {Learning} {Representations}, {ICLR} 2020}, 2020.

\bibitem{bauer_resampled_2019}
M.~Bauer and A.~Mnih, ``Resampled priors for variational autoencoders,'' in
  \emph{The 22nd International Conference on Artificial Intelligence and
  Statistics}.\hskip 1em plus 0.5em minus 0.4em\relax PMLR, 2019, pp. 66--75.

\bibitem{davidson_hyperspherical_2018}
T.~R. Davidson, L.~Falorsi, N.~De~Cao, T.~Kipf, and J.~M. Tomczak,
  ``Hyperspherical variational auto-encoders,'' in \emph{34th Conference on
  Uncertainty in Artificial Intelligence 2018, UAI 2018}.\hskip 1em plus 0.5em
  minus 0.4em\relax Association For Uncertainty in Artificial Intelligence
  (AUAI), 2018, pp. 856--865.

\bibitem{mathieu_continuous_2019}
E.~Mathieu, C.~Le~Lan, C.~J. Maddison, R.~Tomioka, and Y.~W. Teh, ``Continuous
  hierarchical representations with poincaré variational auto-encoders,'' in
  \emph{Advances in neural information processing systems}, 2019, pp.
  12\,565--12\,576.

\bibitem{ovinnikov_poincare_2020}
I.~Ovinnikov, ``Poincar{\'e} wasserstein autoencoder,''
  \emph{{arXiv}:1901.01427 [cs, stat]}, 2020-03-16.

\bibitem{falorsi_explorations_2018}
L.~Falorsi, P.~de~Haan, T.~R. Davidson, N.~De~Cao, M.~Weiler, P.~Forré, and
  T.~S. Cohen, ``Explorations in homeomorphic variational auto-encoding,''
  \emph{{arXiv}:1807.04689 [cs, stat]}, 2018.

\bibitem{miolane_learning_2020}
N.~Miolane and S.~Holmes, ``Learning weighted submanifolds with variational
  autoencoders and riemannian variational autoencoders,'' in \emph{Proceedings
  of the IEEE/CVF Conference on Computer Vision and Pattern Recognition}, 2020,
  pp. 14\,503--14\,511.

\bibitem{arvanitidis_locally_2016}
G.~Arvanitidis, L.~K. Hansen, and S.~Hauberg, ``A locally adaptive normal
  distribution,'' \emph{Advances in Neural Information Processing Systems}, pp.
  4258--4266, 2016.

\bibitem{chen_metrics_2018}
N.~Chen, A.~Klushyn, R.~Kurle, X.~Jiang, J.~Bayer, and P.~Smagt, ``Metrics for
  deep generative models,'' in \emph{International Conference on Artificial
  Intelligence and Statistics}.\hskip 1em plus 0.5em minus 0.4em\relax PMLR,
  2018, pp. 1540--1550.

\bibitem{shao_riemannian_2018}
H.~Shao, A.~Kumar, and P.~T. Fletcher, ``The riemannian geometry of deep
  generative models,'' in \emph{2018 {IEEE}/{CVF} Conference on Computer Vision
  and Pattern Recognition Workshops ({CVPRW})}.\hskip 1em plus 0.5em minus
  0.4em\relax {IEEE}, 2018, pp. 428--4288.

\bibitem{kalatzis_variational_2020}
D.~Kalatzis, D.~Eklund, G.~Arvanitidis, and S.~Hauberg, ``Variational
  autoencoders with riemannian brownian motion priors,'' in \emph{International
  Conference on Machine Learning}.\hskip 1em plus 0.5em minus 0.4em\relax PMLR,
  2020, pp. 5053--5066.

\bibitem{girolami_riemann_2011}
M.~Girolami and B.~Calderhead, ``Riemann manifold langevin and hamiltonian
  monte carlo methods,'' \emph{Journal of the Royal Statistical Society: Series
  B (Statistical Methodology)}, vol.~73, no.~2, pp. 123--214, 2011.

\bibitem{duane_hybrid_1987}
S.~Duane, A.~D. Kennedy, B.~J. Pendleton, and D.~Roweth, ``Hybrid monte
  carlo,'' \emph{Physics Letters B}, vol. 195, no.~2, pp. 216--222, 1987.

\bibitem{leimkuhler_simulating_2004}
B.~Leimkuhler and S.~Reich, \emph{Simulating hamiltonian dynamics}.\hskip 1em
  plus 0.5em minus 0.4em\relax Cambridge university press, 2004, vol.~14.

\bibitem{hairer_geometric_2006}
E.~Hairer, C.~Lubich, and G.~Wanner, \emph{Geometric numerical integration:
  structure-preserving algorithms for ordinary differential equations}.\hskip
  1em plus 0.5em minus 0.4em\relax Springer Science \& Business Media, 2006,
  vol.~31.

\bibitem{liu_monte_2008}
J.~S. Liu, \emph{Monte Carlo strategies in scientific computing}.\hskip 1em
  plus 0.5em minus 0.4em\relax Springer Science \& Business Media, 2008.

\bibitem{paszke_automatic_2017}
A.~Paszke, S.~Gross, S.~Chintala, G.~Chanan, E.~Yang, Z.~{DeVito}, Z.~Lin,
  A.~Desmaison, L.~Antiga, and A.~Lerer, ``Automatic differentiation in
  pytorch,'' 2017.

\bibitem{arvanitidis_latent_2018}
G.~Arvanitidis, L.~K. Hansen, and S.~Hauberg, ``Latent space oddity: On the
  curvature of deep generative models,'' in \emph{6th International Conference
  on Learning Representations, ICLR 2018}, 2018.

\bibitem{frenzel_latent_2019}
M.~F. Frenzel, B.~Teleaga, and A.~Ushio, ``Latent space cartography:
  Generalised metric-inspired measures and measure-based transformations for
  generative models,'' \emph{{arXiv} preprint {arXiv}:1902.02113}, 2019.

\bibitem{arvanitidis_geometrically_2020}
G.~Arvanitidis, S.~Hauberg, and B.~Schölkopf, ``Geometrically enriched latent
  spaces,'' \emph{{arXiv}:2008.00565 [cs, stat]}, 2020-08-02.

\bibitem{lebanon_metric_2006}
G.~Lebanon, ``Metric learning for text documents,'' \emph{{IEEE} Transactions
  on Pattern Analysis and Machine Intelligence}, vol.~28, no.~4, pp. 497--508,
  2006.

\bibitem{louis_computational_2019}
M.~Louis, ``Computational and statistical methods for trajectory analysis in a
  {Riemannian} geometry setting,'' {PhD} {Thesis}, Sorbonnes universités,
  2019.

\bibitem{dai_diagnosing_2018}
B.~Dai and D.~Wipf, ``Diagnosing and enhancing vae models,'' in
  \emph{International Conference on Learning Representations}, 2018.

\bibitem{neal_hamiltonian_2005}
R.~M. Neal, ``Hamiltonian importance sampling,'' in \emph{talk presented at the
  Banff International Research Station ({BIRS}) workshop on Mathematical Issues
  in Molecular Dynamics}, 2005.

\bibitem{xiao_fashion-mnist_2017}
H.~Xiao, K.~Rasul, and R.~Vollgraf, ``Fashion-mnist: a novel image dataset for
  benchmarking machine learning algorithms,'' \emph{arXiv preprint
  arXiv:1708.07747}, 2017.

\bibitem{cohen_emnist_2017}
G.~Cohen, S.~Afshar, J.~Tapson, and A.~Van~Schaik, ``Emnist: Extending mnist to
  handwritten letters,'' in \emph{2017 International Joint Conference on Neural
  Networks (IJCNN)}.\hskip 1em plus 0.5em minus 0.4em\relax IEEE, 2017, pp.
  2921--2926.

\bibitem{lecun_mnist_1998}
Y.~{LeCun}, ``The {MNIST} database of handwritten digits,'' 1998.

\bibitem{salimans_improved_2016}
T.~Salimans, I.~Goodfellow, W.~Zaremba, V.~Cheung, A.~Radford, and X.~Chen,
  ``Improved techniques for training gans,'' in \emph{Advances in Neural
  Information Processing Systems}, 2016.

\bibitem{heusel_gans_2017}
M.~Heusel, H.~Ramsauer, T.~Unterthiner, B.~Nessler, and S.~Hochreiter, ``Gans
  trained by a two time-scale update rule converge to a local nash
  equilibrium,'' in \emph{Advances in Neural Information Processing Systems},
  2017.

\bibitem{karras_progressive_2017}
T.~Karras, T.~Aila, S.~Laine, and J.~Lehtinen, ``Progressive growing of gans
  for improved quality, stability, and variation,'' in \emph{International
  Conference on Learning Representations ({ICLR})}, 2017.

\bibitem{lucic_are_2018}
M.~Lucic, K.~Kurach, M.~Michalski, S.~Gelly, and O.~Bousquet, ``Are {GANs}
  created equal? a large-scale study,'' in \emph{Advances in Neural Information
  Processing Systems}, 2018, p.~10.

\bibitem{shmelkov_how_2018}
K.~Shmelkov, C.~Schmid, and K.~Alahari, ``How good is my gan?'' in
  \emph{Proceedings of the European Conference on Computer Vision (ECCV)},
  2018, pp. 213--229.

\bibitem{borji_pros_2019}
A.~Borji, ``{Pros and cons of GAN evaluation measures},'' \emph{Computer Vision
  and Image Understanding}, vol. 179, pp. 41--65, 2019.

\bibitem{amos_bamosdensenetpytorch_2020}
\BIBentryALTinterwordspacing
B.~Amos, ``bamos/densenet.pytorch,'' 2020, original-date: 2017-02-09T15:33:23Z.
  [Online]. Available: \url{https://github.com/bamos/densenet.pytorch}
\BIBentrySTDinterwordspacing

\bibitem{huang_densely_2017}
G.~Huang, Z.~Liu, L.~Van Der~Maaten, and K.~Q. Weinberger, ``Densely connected
  convolutional networks,'' in \emph{2017 {IEEE} Conference on Computer Vision
  and Pattern Recognition ({CVPR})}.\hskip 1em plus 0.5em minus 0.4em\relax
  {IEEE}, 2017, pp. 2261--2269.

\bibitem{krizhevsky2009learning}
A.~Krizhevsky, G.~Hinton \emph{et~al.}, ``Learning multiple layers of features
  from tiny images,'' 2009.

\bibitem{breiman_random_2001}
L.~Breiman, ``Random forests,'' \emph{Machine Learning}, vol.~45, no.~1, pp.
  5--32, 2001.

\bibitem{kotsiantis_supervised_2007}
S.~B. Kotsiantis, I.~Zaharakis, and P.~Pintelas, ``Supervised machine learning:
  A review of classification techniques,'' \emph{Emerging artificial
  intelligence applications in computer engineering}, vol. 160, no.~1, pp.
  3--24, 2007.

\bibitem{routierClinicaOpenSourceSoftware2021}
A.~Routier, N.~Burgos, M.~D{\'i}az, M.~Bacci, S.~Bottani, O.~{El-Rifai},
  S.~Fontanella, P.~Gori, J.~Guillon, A.~Guyot, R.~Hassanaly, T.~Jacquemont,
  P.~Lu, A.~Marcoux, T.~Moreau, J.~{Samper-Gonz{\'a}lez}, M.~Teichmann,
  E.~{Thibeau-Sutre}, G.~Vaillant, J.~Wen, A.~Wild, M.-O. Habert, S.~Durrleman,
  and O.~Colliot, ``Clinica: {{An Open-Source Software Platform}} for
  {{Reproducible Clinical Neuroscience Studies}},'' \emph{Frontiers in
  Neuroinformatics}, vol.~15, p. 689675, 2021.

\bibitem{thibeau-sutreClinicaDLOpensourceDeep2021}
E.~{Thibeau-Sutre}, M.~Diaz, R.~Hassanaly, A.~M. Routier, D.~Dormont,
  O.~Colliot, and N.~Burgos, ``{{ClinicaDL}}: An open-source deep learning
  software for reproducible neuroimaging processing,'' 2021.

\bibitem{wen_convolutional_2020}
J.~Wen, E.~Thibeau-Sutre, M.~Diaz-Melo, J.~Samper-González, A.~Routier,
  S.~Bottani, D.~Dormont, S.~Durrleman, N.~Burgos, and O.~Colliot,
  ``Convolutional neural networks for classification of {Alzheimer}'s disease:
  {Overview} and reproducible evaluation,'' \emph{Medical Image Analysis},
  vol.~63, p. 101694, 2020.

\bibitem{aderghal_classification_2017}
K.~Aderghal, M.~Boissenin, J.~Benois-Pineau, G.~Catheline, and K.~Afdel,
  ``Classification of {sMRI} for {AD} diagnosis with convolutional neuronal
  networks: {A} pilot 2-{D}+$\epsilon$ study on {ADNI},'' in \emph{MultiMedia
  Modeling}, vol. 10132 LNCS, 2017, pp. 690--701.

\bibitem{aderghal_classification_2018}
K.~Aderghal, A.~Khvostikov, A.~Krylov, J.~Benois-Pineau, K.~Afdel, and
  G.~Catheline, ``Classification of {Alzheimer} {Disease} on {Imaging}
  {Modalities} with {Deep} {CNNs} {Using} {Cross}-{Modal} {Transfer}
  {Learning},'' in \emph{2018 {IEEE} 31st {International} {Symposium} on
  {Computer}-{Based} {Medical} {Systems} ({CBMS})}, 2018, pp. 345--350, iSSN:
  2372-9198.

\bibitem{islam_gan-based_2020}
J.~Islam and Y.~Zhang, ``{GAN}-based synthetic brain {PET} image generation,''
  \emph{Brain Informatics}, vol.~7, no.~1, 2020.

\bibitem{oh_classification_2019}
K.~Oh, Y.-C. Chung, K.~W. Kim, W.-S. Kim, and I.-S. Oh, ``Classification and
  {Visualization} of {Alzheimer}'s {Disease} using {Volumetric} {Convolutional}
  {Neural} {Network} and {Transfer} {Learning},'' \emph{Scientific Reports},
  vol.~9, no.~1, p. 18150, 2019.

\bibitem{liu_weakly_2020}
M.~Liu, J.~Zhang, C.~Lian, and D.~Shen, ``Weakly {Supervised} {Deep} {Learning}
  for {Brain} {Disease} {Prognosis} {Using} {MRI} and {Incomplete} {Clinical}
  {Scores},'' \emph{IEEE Transactions on Cybernetics}, vol.~50, no.~7, pp.
  3381--3392, 2020.

\bibitem{valliani_deep_2017}
A.~Valliani and A.~Soni, ``Deep {Residual} {Nets} for {Improved} {Alzheimer}'s
  {Diagnosis},'' in \emph{8th {ACM} {International} {Conference} on
  {Bioinformatics}, {Computational} {Biology},and {Health} {Informatics} -
  {ACM}-{BCB} '17}.\hskip 1em plus 0.5em minus 0.4em\relax Boston,
  Massachusetts, USA: ACM Press, 2017, pp. 615--615.

\bibitem{backstrom_efficient_2018}
K.~Bäckström, M.~Nazari, I.-H. Gu, and A.~Jakola, ``An efficient {3D} deep
  convolutional network for {Alzheimer}'s disease diagnosis using {MR}
  images,'' in \emph{2018 {IEEE} 15th {International} {Symposium} on
  {Biomedical} {Imaging} ({ISBI} 2018)}, vol. 2018-April, 2018, pp. 149--153.

\bibitem{cheng_cnns_2017}
D.~Cheng and M.~Liu, ``{CNNs} based multi-modality classification for {AD}
  diagnosis,'' in \emph{2017 10th {International} {Congress} on {Image} and
  {Signal} {Processing}, {BioMedical} {Engineering} and {Informatics}
  ({CISP}-{BMEI})}, 2017, pp. 1--5.

\bibitem{ellis_australian_2009}
K.~A. Ellis, A.~I. Bush, D.~Darby, D.~De~Fazio, J.~Foster, P.~Hudson, N.~T.
  Lautenschlager, N.~Lenzo, R.~N. Martins, P.~Maruff, C.~Masters, A.~Milner,
  K.~Pike, C.~Rowe, G.~Savage, C.~Szoeke, K.~Taddei, V.~Villemagne,
  M.~Woodward, D.~Ames, and {AIBL Research Group}, ``The {Australian}
  {Imaging}, {Biomarkers} and {Lifestyle} ({AIBL}) study of aging: methodology
  and baseline characteristics of 1112 individuals recruited for a longitudinal
  study of {Alzheimer}'s disease,'' \emph{International Psychogeriatrics},
  vol.~21, no.~4, pp. 672--687, 2009.

\bibitem{gorgolewski_brain_2016}
K.~J. Gorgolewski, T.~Auer, V.~D. Calhoun, R.~C. Craddock, S.~Das, E.~P. Duff,
  G.~Flandin, S.~S. Ghosh, T.~Glatard, Y.~O. Halchenko, D.~A. Handwerker,
  M.~Hanke, D.~Keator, X.~Li, Z.~Michael, C.~Maumet, B.~N. Nichols, T.~E.
  Nichols, J.~Pellman, J.-B. Poline, A.~Rokem, G.~Schaefer, V.~Sochat,
  W.~Triplett, J.~A. Turner, G.~Varoquaux, and R.~A. Poldrack, ``The brain
  imaging data structure, a format for organizing and describing outputs of
  neuroimaging experiments,'' \emph{Scientific Data}, vol.~3, no.~1, p. 160044,
  2016.

\bibitem{tustison_n4itk_2010}
N.~J. Tustison, B.~B. Avants, P.~A. Cook, {Yuanjie Zheng}, A.~Egan, P.~A.
  Yushkevich, and J.~C. Gee, ``{N4ITK}: {Improved} {N3} {Bias} {Correction},''
  \emph{IEEE Transactions on Medical Imaging}, vol.~29, no.~6, pp. 1310--1320,
  2010.

\bibitem{fonov_unbiased_2009}
V.~Fonov, A.~Evans, R.~McKinstry, C.~Almli, and D.~Collins, ``Unbiased
  nonlinear average age-appropriate brain templates from birth to adulthood,''
  \emph{NeuroImage}, vol.~47, p. S102, 2009.

\bibitem{fonov_unbiased_2011}
V.~Fonov, A.~C. Evans, K.~Botteron, C.~R. Almli, R.~C. McKinstry, and D.~L.
  Collins, ``Unbiased average age-appropriate atlases for pediatric studies,''
  \emph{NeuroImage}, vol.~54, no.~1, pp. 313--327, 2011.

\bibitem{avants_insight_2014}
B.~B. Avants, N.~J. Tustison, M.~Stauffer, G.~Song, B.~Wu, and J.~C. Gee, ``The
  {Insight} {ToolKit} image registration framework,'' \emph{Frontiers in
  Neuroinformatics}, vol.~8, 2014.

\bibitem{fonov_deep_2018}
V.~S. Fonov, M.~Dadar, T.~P.-A.~R. Group, and D.~L. Collins, ``Deep learning of
  quality control for stereotaxic registration of human brain {MRI},''
  \emph{bioRxiv}, p. 303487, 2018.

\bibitem{bergstra_random_2012}
J.~Bergstra and Y.~Bengio, ``Random {Search} for {Hyper}-{Parameter}
  {Optimization},'' \emph{Journal of Machine Learning Research}, vol.~13, no.
  Feb, pp. 281--305, 2012.

\bibitem{he_delving_2015}
K.~He, X.~Zhang, S.~Ren, and J.~Sun, ``Delving {Deep} into {Rectifiers}:
  {Surpassing} {Human}-{Level} {Performance} on {ImageNet} {Classification},''
  in \emph{2015 {IEEE} {International} {Conference} on {Computer} {Vision}
  ({ICCV})}.\hskip 1em plus 0.5em minus 0.4em\relax Santiago, Chile: IEEE,
  2015, pp. 1026--1034.

\bibitem{arjovsky_wasserstein_2017}
M.~Arjovsky, S.~Chintala, and L.~Bottou, ``Wasserstein {GAN},''
  \emph{{arXiv}:1701.07875 [cs, stat]}, 2017-12-06.

\bibitem{kingma_adam_2014}
D.~P. Kingma and J.~Ba, ``Adam: A method for stochastic optimization,''
  \emph{{arXiv} preprint {arXiv}:1412.6980}, 2014.

\end{thebibliography}
%
% <OR> manually copy in the resultant .bbl file
% set second argument of \begin to the number of references
% (used to reserve space for the reference number labels box)
%\begin{thebibliography}{1}

%\bibitem{IEEEhowto:kopka}
%H.~Kopka and P.~W. Daly, \emph{A Guide to {\LaTeX}}, 3rd~ed.\hskip 1em plus
%  0.5em minus 0.4em\relax Harlow, England: Addison-Wesley, 1999.

%\end{thebibliography}

% biography section
% 
% If you have an EPS/PDF photo (graphicx package needed) extra braces are
% needed around the contents of the optional argument to biography to prevent
% the LaTeX parser from getting confused when it sees the complicated
% \includegraphics command within an optional argument. (You could create
% your own custom macro containing the \includegraphics command to make things
% simpler here.)
%\begin{IEEEbiography}[{\includegraphics[width=1in,height=1.25in,clip,keepaspectratio]{mshell}}]{Michael Shell}
% or if you just want to reserve a space for a photo:

\begin{IEEEbiography}[{\includegraphics[width=1in,height=1.25in,clip,keepaspectratio]{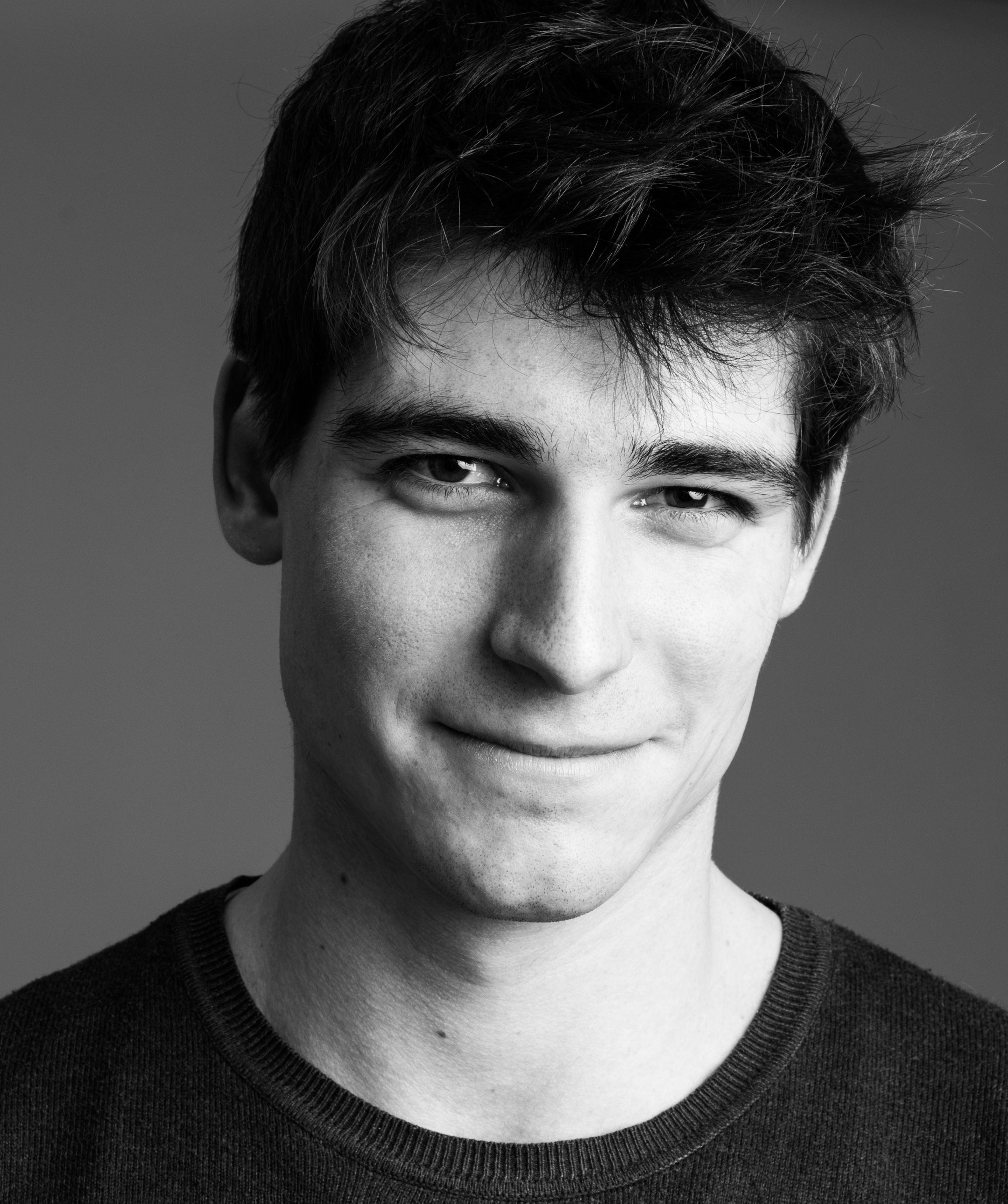}}]{Clément Chadebec} is a PhD student at Université de Paris and Inria and funded by $PR[AI]RIE$. His research interests include machine learning and in particular generative models along with Riemannian geometry and computational statistics for medicine. He received master degrees from Ecole Nationale des Mines de Paris and Ecole Normale Supérieure Paris-Saclay.
\end{IEEEbiography}
\vspace{-2em}
\begin{IEEEbiography}[{\includegraphics[width=1in,height=1.25in,clip,keepaspectratio]{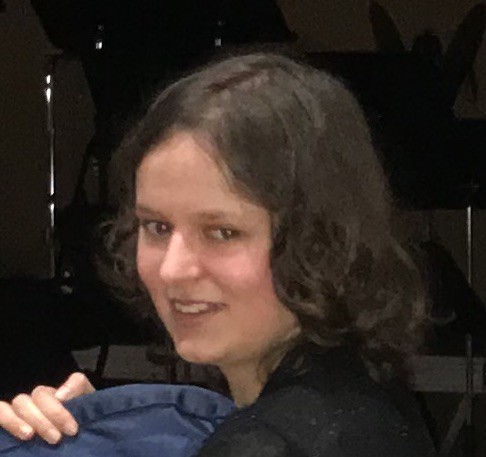}}]{Elina Thibeau-Sutre} is a PhD student at Sorbonne Université and Inria. Her research interest includes deep learning application to neuroimaging data, its interpretability and reproducibility. She received master degrees from Ecole Nationale des Mines de Paris and Ecole supérieure de physique et de chimie industrielles (Paris, France).
\end{IEEEbiography}
\vspace{-2em}
\begin{IEEEbiography}[{\includegraphics[width=1in,height=1.25in,clip,keepaspectratio]{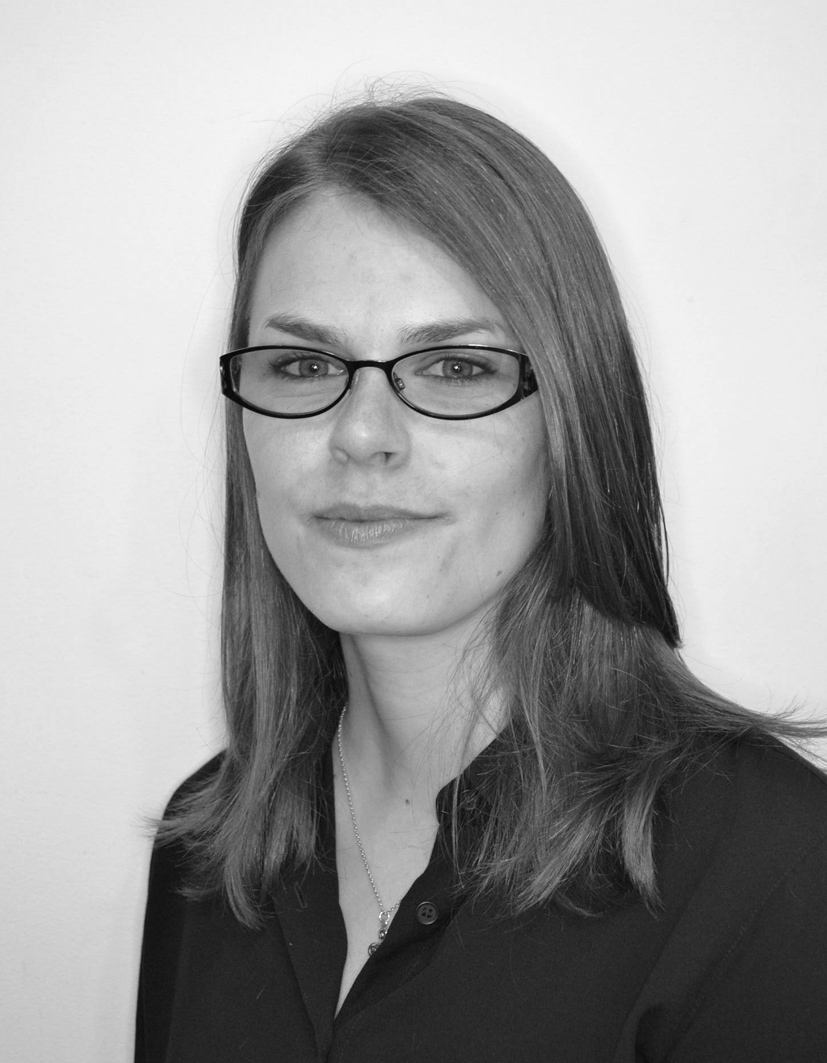}}]{Ninon Burgos}
CNRS researcher in the ARAMIS Lab, a joint laboratory between Sorbonne Université, CNRS, Inserm and Inria within the Paris Brain Institute, France. She completed her PhD at University College London, UK, in 2016. Her research focuses on the development of computational imaging tools to improve the understanding and diagnosis of dementia.
\end{IEEEbiography}
\vspace{-2em}
\begin{IEEEbiography}[{\includegraphics[width=1in,height=1.25in,clip,keepaspectratio]{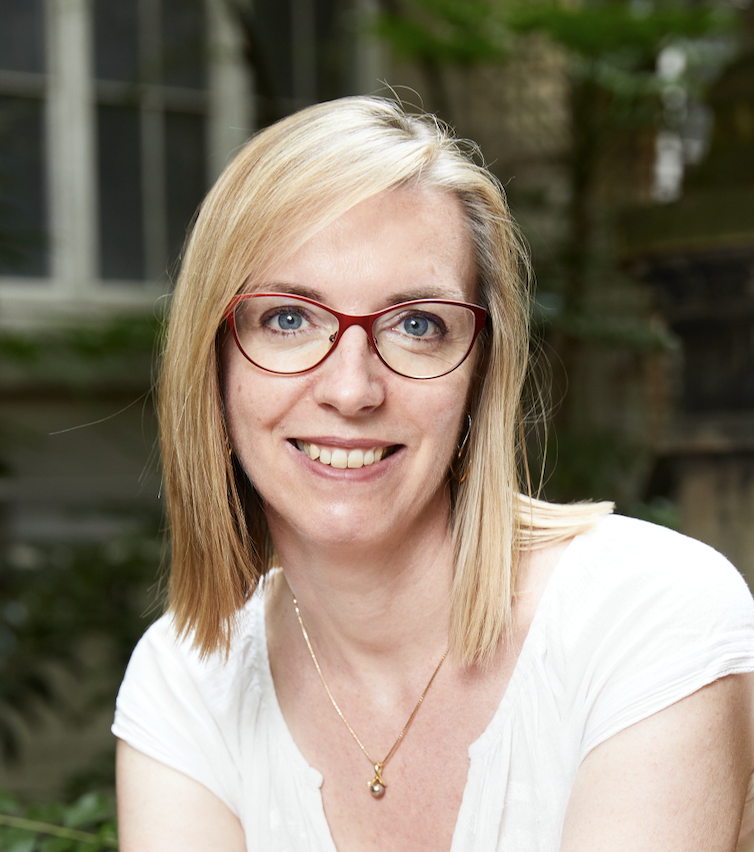}}]{Stéphanie Allassonnière}
Pr.  of  Applied  Mathematics  in  Université de Paris, $PR[AI]RIE$ fellow and deputy directeor.  She  received  her  PhD  degree  in Applied  Mathematics  (2007),  studies  one  year as postdoctoral fellow in the CIS, JHU, Baltimore. She then joined the Applied Mathematics department of Ecole Polytechnique in  2008  as  assistant  professor  and  moved  to Paris Descartes school of medicine in 2016 as Professor.  Her  researches  focus  on  statistical analysis of medical databases in order to: understanding the common features of populations, designing classification, early prediction and decision support systems.
\end{IEEEbiography}

% You can push biographies down or up by placing
% a \vfill before or after them. The appropriate
% use of \vfill depends on what kind of text is
% on the last page and whether or not the columns
% are being equalized.

%\vfill

% Can be used to pull up biographies so that the bottom of the last one
% is flush with the other column.
%\enlargethispage{-5in}
\clearpage
\appendices
\section{Riemannian Geometry}\label{appendix A}

In the framework of differential geometry, one may define a Riemannian manifold $\mathcal{M}$ as a smooth manifold endowed with a Riemannian metric $g$ that is a smooth inner product $g: p \to \langle \cdot | \cdot \rangle_p$ on the tangent space $T_p\mathcal{M}$ defined at each point of the manifold $p \in \mathcal{M}$. We call a chart (or coordinate chart) $(U, \varphi)$ a homeomorphism mapping an open set $U$ of the manifold to an open set $V$ of an Euclidean space.  The manifold is called a $d-$dimension manifold if for each chart of an atlas we further have $V \subset \mathbb{R}^d$. That is there exists a neighborhood $U$ of each point $p$ of the manifold such that $U$ is homeomorphic to $\mathbb{R}^d$. Given $p \in U$, the chart $\varphi: (x^1, \dots, x^d)$ induces a basis $\Big (\frac{\partial}{\partial x^1}, \dots, \frac{\partial}{\partial x^d} \Big)_p $ on the tangent space $T_p\mathcal{M}$. Hence, a local representation of the metric of a Riemannian manifold in the chart $(U,\varphi)$ can be written as a positive definite matrix $\mathbf{G}(p) = ( g_{i, j})_{p, 0 \leq i, j \leq d} = (\langle \frac{\partial}{\partial x^i} | \frac{\partial}{\partial x^j} \rangle_p)_{0 \leq i, j \leq d}$ at each point $p \in U$. That is for $v, w \in T_p\mathcal{M}$ and $p \in U$, we have $\langle u | w \rangle_p = u^{\top} \mathbf{G}(p) w$. Since we propose to work in the ambient-like manifold ($\mathbb{R}^d$, $g$), there exists a global chart given by $\varphi=id$. Hence, for the following, we assume that we work in this coordinate system and so $\mathbf{G}$  will refer to the metric's matrix representation in this chart.

There are two ways to apprehend manifolds. The extrinsic view assumes that the manifold is embedded within a higher dimensional Euclidean space (think of the 2-dimensional sphere $\mathcal{S}^2$ embedded within $\mathbb{R}^3$). The intrinsic view, which is adopted in this paper, does not make such an assumption since the manifold is studied using its underlying structure. For example, a curve's length cannot be interpreted using the distance defined on an Euclidean space but requires the use of the metric defined onto the manifold itself. The length of a curve $\gamma$ between two points of the manifold $z_1, z_2 \in \mathcal{M}$ and parametrized by $ t \in [0, 1]$ such that $\gamma(0) = z_1$ and $\gamma(1) = z_2$ is then given by 
\[
\]  
Curves minimizing such a length are called \textit{geodesics} and a distance $\mathrm{dist}$ between elements of a (connected) manifold can be introduced as follows:
\begin{equation}\label{Eq: app geodesic distance}
    \mathrm{dist}(z_1, z_2) = \inf_{\gamma} \mathcal{L}(\gamma) \hspace{5mm} \mathrm{s.t.} \hspace{5mm} \gamma(0) = z_1, \gamma(1) = z_2
\end{equation}
The manifold $\mathcal{M}$ is said to be \textit{geodesically complete} if all geodesic curves can be extended to $\mathbb{R}$. In other words, at each point $p$ of the manifold one may draw a \emph{straight} line (with respect to the formerly defined distance) indefinitely and in any direction.

\section{Some Further Details on Riemannian Hamiltonian Equations}\label{appendix B}

We recall that the Riemannian Hamiltonian Monte Carlo (RHMC) sampler aims at sampling from complex target probability distributions $p_{\mathrm{target}}(z)$ where $z$ is assumed to live in a Riemannian manifold $\mathcal{M}$. The main idea is to introduce a random variable $v \sim \mathcal{N}(0, \mathbf{G}(z))$ where $\mathbf{G}$ is the Riemannian metric associated to $\mathcal{M}$ and rely on Riemannian Hamiltonian dynamics. Analogous to physical systems, $z\in \mathcal{M}$ is seen as the \emph{position} and $v$ as the \emph{velocity} of a particle whose potential energy $U(z)$ and kinetic energy $K(z, v)$ are given by 
    \begin{equation*}
    \begin{aligned}
         U(z) &= - \log p_{\mathrm{target}}(z)\\
         K(v, z)&= \frac{1}{2}\Big[\log\big( (2\pi)^d|\mathbf{G}(z)|\big )+ v^{\top} \textbf{G}^{-1}(z)v \Big ]\,.
    \end{aligned}
    \end{equation*}
     These two energies give together the Hamiltonian $H(z, v)$ \cite{duane_hybrid_1987,leimkuhler_simulating_2004}. 
    \begin{equation}
        \label{eq: app riemannian hamiltonian}
        H(z, v) = U(z) + \frac{1}{2} \log((2 \pi)^D \det \mathbf{G}(z)) + \frac{1}{2} v^{\top} \mathbf{G}(z)^{-1} v\,.
    \end{equation}

     The evolution in time of such a particle is governed by Hamilton's equations which write:
     \begin{equation}\label{eq: app appednPDE Riemann}\begin{aligned}
        \frac{\partial H}{\partial v_i} = \big( \mathbf{G}^{-1}(z) v \big)_i \,, \\
        - \frac{\partial H}{\partial z_i} = \frac{\partial \log p_{\mathrm{target}}(z)}{\partial z_i} &- \frac{1}{2} \mathrm{tr} \Biggl(\mathbf{G}^{-1} \frac{\partial \mathbf{G}(z)}{\partial z_i} \Biggr)\\ 
        &+ \frac{1}{2} v^{\top} \mathbf{G}^{-1}(z) \frac{\partial \mathbf{G}(z)}{\partial z_i} \mathbf{G}^{-1}(z) v \,.
    \end{aligned} 
    \end{equation}
    These equations can be integrated using a discretization scheme known as the generalized \emph{leapfrog} integrator.
     \begin{equation}\label{Eq: app appendix Riemann Stormer Verlet}
        \begin{aligned}
            v(t + \varepsilon/2) = v(t) &- \frac{\varepsilon}{2} \nabla_z H\Bigl(z(t), v(t+\varepsilon/2)\Bigr)\,,\\
            z(t + \varepsilon)      = z(t) &+ \frac{\varepsilon}{2} \Bigl[\nabla_{v} H\Bigl(z(t), v(t + \varepsilon/2)\Bigr) \\&+ \nabla_{v} H\Bigl(z(t+\varepsilon),  v(t + \varepsilon/2)\Bigr)\Bigr]\,,\\
            v(t + \varepsilon)   = v(t + \varepsilon/2) &- \frac{\varepsilon}{2} \nabla_z H \Bigl(z(t+\varepsilon), v(t+ \varepsilon/2) \Bigr)\,,
        \end{aligned}
    \end{equation}
    where $\varepsilon$ is the integrator step size. By running $K$ times this integrator simulating the behavior of the particle, this sampler aims at creating a Markov Chain $(z^n)$ converging to the target distribution $p_{\mathrm{target}}$. In our case, the target density is set to be the joint distribution $p(z, x) = p(z) p(x|z)$ that is known thanks to the assumed generation process:
    \[\left\{\begin{aligned}
    z\sim p(z) &= \mathcal{N}(0, I_d),\\ x\sim p(x|z) &=\mathcal{N}(\mu_{\theta}(z), \sigma I_d) \Big(~\text{or}~ \emph{e.g.}~\prod_i \mathcal{B}(\pi_{\theta}(z))\Big)
     \end{aligned}\right.
    \]Hence, we can compute every terms of Eq.~\eqref{eq: app appednPDE Riemann} and so use the generalized leapfrog integrator as proposed in the manuscript.
    Finally, we also provide the full pseudo-code training algorithm of the method in Alg.~\ref{Alg: app RHVAE}. In this paper, a typical choice for $\varepsilon$ and $K$ is $\varepsilon \in [0.0001, 0.01]$ and $K\in[1, 15]$.

    \begin{algorithm*}[ht]
    \SetAlgoLined \textbf{Initialize} $\mathbf{G}$ \tcp*{We put $c_i = 0$ and
     $L_{\psi_i}=I_d$} \While{not converged}{$\mathcal{L} \leftarrow 0$ \;
     \For{$n = 1 \to N_B$}{Collect a batch of data $X_n = (x_1, \cdots,
     x_{\text{bs}})$\; $c_i \leftarrow \text{encode}(x_i)$\; $L_{\psi_i}
     \leftarrow m_{\psi}(x_i) $ \tcp*{Use the metric network to get $L_{\psi_i}$} Update the metric $\mathbf{G}$ according to
     Eq.~\eqref{eq: metric}\; $z_0 \sim \mathcal{N}(\mu(x), \Sigma(x)),$ $v_0
     \sim \mathcal{N}(0, \mathbf{G}(z_0))$\; $v_0 \leftarrow v_0 /
     \sqrt{\beta_0}$\; \For{$k = 1 \to K$}{$\bar{v} \leftarrow v_{k-1} -
     \frac{\varepsilon}{2} \nabla_{z} H(z_{k-1}, \bar{v})$ \tcp*{fixed
     point it.} $z_k \leftarrow z_{k-1} + \frac{\varepsilon}{2}
     \Big(\nabla_{v} H (z_{k-1}, \bar{v}) + \nabla_{v} H(z_k,
     \bar{v}) \Big)$ \tcp*{fixed point it.} $v' \leftarrow \bar{v}
     - \frac{\varepsilon}{2} \nabla_{z} H(z_k, \bar{v})$\; $\sqrt{\beta_k}
     \leftarrow \Big(\Big( 1 - \frac{1}{\sqrt{\beta_0}}\Big) \frac{k^2}{K^2}+
     \frac{1}{\sqrt{\beta_0}} \Big)^{-1}$ \; $v_k \leftarrow
     \frac{\sqrt{\beta_{k-1}}}{\sqrt{\beta_{k}}}v'$ \;} $p \leftarrow
     p_{\theta}(x, z_K, v_K)$ \; $q \leftarrow q_{\phi}(z_0, v_0|x) \beta_{0}^{-d/2}$\;
     $\mathcal{L}_{\mathrm{batch}} \leftarrow \log p - \log q$ \; $\mathcal{L} =
     \mathcal{L} + \mathcal{L}_{\mathrm{batch}} / N_B$ \;} Update $\theta$,
     $\phi$ and $\psi$ using gradient descent\;}
     \caption{RHVAE with metric learning}
     \label{Alg: app RHVAE}
    \end{algorithm*}

\section{On the Generation Process}\label{appendix C}
We recall that to sample from the defined target distribution given by the inverse of the volume element of the Riemannian manifold we recourse to the Hamiltonian Monte Carlo (HMC) sampler since the normalizing constant is hard to compute. Hence, we recall in this section some elements on the HMC sampler and how it applies in our specific framework.

Likewise the RHMC presented in the previous section, given a target density $p_{\mathrm{target}}$ we want to sample from, the idea behind the HMC sampler is to introduce a random variable $v \sim \mathcal{N}(0,I_d)$ independent from $z$ and rely on Hamiltonian dynamics. Analogous to physical systems, $z$ can again be seen as the \emph{position} and $v$ as the \emph{velocity} of a particle whose potential energy $U(z)$ and kinetic energy $K(v)$ are given by 
    \begin{equation*}
      U(z) = - \log p_{\mathrm{target}}(z), \hspace{5mm} K(v)= \frac{1}{2} v^{\top} v\,.
    \end{equation*}
    These two energies give together the Hamiltonian \cite{duane_hybrid_1987,leimkuhler_simulating_2004}
    \begin{equation*}
      H(z, v) = U(z) + K(v)\,.
    \end{equation*}
    The evolution in time of such a particle is governed by Hamilton's equations as follows
    \[
        \frac{\partial z_i}{\partial t} = \frac{\partial H}{\partial v_i}, \hspace{5mm}
        \frac{\partial v_i}{\partial t} = - \frac{\partial H}{\partial z_i} \,. 
    \]
    Such equations can be integrated using a discretization scheme known as the \emph{Stormer-Verlet} or \emph{leapfrog} integrator which is run $l$ times
    \begin{equation}
      \begin{aligned}\label{eq: app leapfrog}
          v(t + \gamma/2) &= v(t) - \frac{\gamma}{2} \cdot \nabla_z U(z(t))\,,\\
          z(t + \gamma) &= z(t) + \gamma \cdot v(t + \gamma/2)\,,\\
          v(t + \gamma) &= v(t + \gamma/2) - \frac{\gamma}{2} \nabla_z U(z(t + \gamma))\,,
      \end{aligned}
  \end{equation}
  where $\gamma$ is the integrator step size. The HMC sampler produces a Markov chain $(z^n)$ with the aforementioned integrator. More precisely, given $z_0^n$, the current state of the chain, an initial \emph{velocity} is sampled $v_0 \sim \mathcal{N}(0, I_d)$ and then Eq.~\eqref{eq: app leapfrog} are run $l$ times to move from ($z_0^n$, $v_0$) to ($z_l^n$, $v_l$). The proposal $z_l^n$ is then accepted with probability $\alpha = \min\Big(1, \frac{\exp (-H(z_l^n, v_l))}{\exp (-H(z_0^n, v_0))}\Big)$. It was shown that the chain $(z^n)$ is time-reversible and converges to its stationary distribution $p_{\mathrm{target}}$~\cite{duane_hybrid_1987,liu_monte_2008,neal_mcmc_2011}. 
  
  In our method $p_{\mathrm{target}}$ is given by Eq.~\eqref{eq: metric} and 
   \begin{equation}\label{Eq: app Target distribution}
    p(z) = \frac{\mathbf{1}_S(z) \sqrt{\det \mathbf{G}^{-1}(z)}}{\int \limits _{\mathbb{R}^d} \mathbf{1}_S(z) \sqrt{\det \mathbf{G}^{-1}(z) dz}}\,,
    \end{equation}
    where $S$ is a compact set\footnote{Take for instance $\{z \in \mathcal{Z}, \lVert z \rVert \leq 2 \cdot \max_i \lVert c_i \rVert \}$} so that the integral is well defined. Fortunately, since the HMC sampler allows sampling from densities known up to a normalizing constant (thanks to the acceptance ratio), the computation of the denominator of $p_{\mathrm{target}}$ is not needed and the Hamiltonian follows
  \[\begin{aligned}
    H(z, v) &= U(z) + K(v) \propto -\frac{1}{2} \log \det \mathbf{G}^{-1}(z) + \frac{1}{2} v^{\top} v \,
  \end{aligned}
  \]
  and is easy to compute. Hence, the only \emph{difficulty} left is the computation of the gradient $\nabla_z U(z)$ needed in the \emph{leapfrog} integrator which is actually pretty straightforward using the chain rule.
  %\[\begin{aligned}
  %  \frac{\partial U}{\partial z_k} &= - \frac{1}{2} %\frac{\partial \log \det \mathbf{G}^{-1}(z)}{\partial z_k}\,, = %-\frac{1}{2} \mathrm{tr}\Bigg(\mathbf{G}(z) \frac{\partial %\mathbf{G}^{-1}(z_k)}{\partial z_k}\Bigg)\,.
  %\end{aligned}
  %\]
  In this paper, a typical choice for $\gamma$ and $l$, the sampler's parameters, is $\gamma \in [0.01, 0.05]$ and $l\in[10, 15]$. We would also like to mention the recent work of~\cite{arvanitidis_geometrically_2020} where the authors used the distribution $q(z) \propto (1 + \sqrt{\det \mathbf{G}(z)})^{-1}$ to sample from a Wasserstein GAN~\cite{arjovsky_wasserstein_2017}. Nonetheless, both the framework and the metric remain quite different.

\section{Detailed Experimental Setting}\label{appendix D}

\subsection{Parameters of Sec.~3.3. Generation Comparison}\label{appendix D.1}
For this experiment and for a fair comparison, each model is trained with the same neural network architecture for the encoder and decoder presented in Table~\ref{tab: app mnist neural arch generation} along with the same latent space dimension set to 2. The main parameters for the \emph{geometry-aware} VAE are presented in Table~\ref{Table: app Hyper-parameters}. Each model is trained until the ELBO does not improve for 20 epochs with an Adam optimizer \cite{kingma_adam_2014} and a learning rate of $10^{-3}$. Since the data sets sizes are small, the training is performed in a single batch.

\begin{table}[ht]
\caption{Neural Net Architectures for MNIST, EMNIST and fashion. The same architectures are used for the VAEs, VAMP, RAE and \emph{geometry-aware} VAEs. }
    \centering
    \begin{tabular}{c|cc}
    \hline
         $\mu_{\phi} $      &  \multirow{2}{*}{($D$, 400, relu)}& (400, $d$, linear)   \\
         $\Sigma_{\phi}$    &                                 & (400, $d$, linear) \\
         \hline
         $ \pi_{\theta} $   &  ($d$, 400, relu)                 & (400, $D$, sigmoid)   \\
         \hline
        $L_{\psi}^{\text{diag.}}$  &  \multirow{2}{*}{($D$, 400, relu)}& (400, $d$, linear)   \\
        $L_{\psi}^{\text{low.}}$    &                                 & (400, $\frac{d(d-1)}{2}$, linear)\\
        \hline
        \multicolumn{3}{l}{\scriptsize{$D$: Input space dimension}}\\
        \multicolumn{3}{l}{\scriptsize{$d$: Latent space dimension}}
    \end{tabular}
    \label{tab: app mnist neural arch generation}
\end{table}

\begin{table}[!ht]
  \centering
  \caption{\emph{Geometry-aware} VAE parameters.}
  \label{Table: app Hyper-parameters}
  \begin{tabular}{c|cccccc}
  \hline
       \multirow{2}{*}{Data sets} & \multicolumn{6}{c}{Parameters} \\
       & $d^{*}$  & $K$ & $\varepsilon$ & $T$ & $\lambda$ & $\sqrt{\beta_0}$   \\
      \hline
      Synthetic shapes                  & 2 & 3 & $10^{-2}$ & 0.8 & $10^{-3}$ & 0.3 \\
      \textit{reduced} MNIST (bal.)           & 2 & 3 & $10^{-2}$ & 0.8 & $10^{-3}$ & 0.3 \\
      \textit{reduced} MNIST (unbal.)         & 2 & 3 & $10^{-2}$ & 0.8 & $10^{-3}$ & 0.3 \\
      \textit{reduced} EMNIST           & 2 & 3 & $10^{-2}$ & 0.8 & $10^{-3}$ & 0.3 \\
      \hline
      \multicolumn{7}{l}{\scriptsize{* Latent space dimension (same for the other models)}} 
  \end{tabular}
\end{table}

\subsection{Parameters of Sec.~4. Data Augmentation}\label{appendix D.2}

For this experiment, we consider a vanilla VAE, a VAE with VAMP prior, a \emph{geometry-aware} VAE using the prior to generate, a \emph{geometry-aware} VAE using the proposed method, a regularized autoencoder with a penalty on the gradient of the decoder as proposed in \cite{ghosh_variational_2020} and consider two other approaches proposed in the litterature to improve the generation from a VAE. The first one is a two stage VAE as proposed in \cite{dai_diagnosing_2018} and the second one consists in fitting a mixture of Gaussian in the latent space of the VAE post-training \cite{ghosh_variational_2020}. 

\subsubsection{MNIST, EMNIST and Fashion}
For these data sets, we use the same parameters and neural network architectures as presented in the former section and Table~\ref{tab: app mnist neural arch generation} except for \textit{reduced} Fashion where the dimension of the latent space is set to 5. As to training parameters for the VAEs, for each model we use an Adam optimizer with a learning rate set to $10^{-3}$. Since the data sets sizes are small the training is performed in a single batch. An implementation of all the models can be found at \url{https://github.com/clementchadebec/benchmark_VAE}.

\subsubsection{CIFAR}
For CIFAR, each model is trained for 500 epochs and we keep the model achieving the best ELBO. The latent space dimension is set to 5 for all models. The training is performed with an Adam optimizer \cite{kingma_adam_2014} and a learning rate of $10^{-4}$. Since the data sets sizes are small the training is performed in a single batch. All the models share again the same neural network architectures for both the encoder and decoder which is described in Table~\ref{Tab: app cifar Architectures}.

\begin{table}[ht]
\caption{Neural Net Architectures for CIFAR. The same architectures are used for the VAEs, VAMP, RAE and \emph{geometry-aware} VAEs. }
\label{Tab: app cifar Architectures}
    \centering
    \scriptsize
    \begin{sc}
    \begin{tabular}{cc}
    %\toprule
         &CIFAR10\\
    \hline
    \hline
    Encoder & (3, 32, 32) \\
    \hline
    \multirow{3}{*}{Layer 1} & Conv(128, (4, 4), stride=2)          \\
                             & Batch normalization  \\
                             & Relu   \\
    \hline
    \multirow{3}{*}{Layer 2} & Conv(256, (4, 4), stride=2)          \\
                             & Batch normalization  \\
                             & Relu \\
    \hline
    \multirow{3}{*}{Layer 3} & Conv(512, (4, 4), stride=2) \\
                             & Batch normalization         \\
                             & Relu                        \\
    \hline
    \multirow{3}{*}{Layer 4} & Conv(1024, (4, 4), stride=2) \\
                             & Batch normalization          \\
                             & Relu  \\
    \hline
    Layer 5               & Linear(4096, 10)\\
    \hline
    \hline
    Decoder                  & (10)                              \\
    \hline
    \multirow{2}{*}{Layer 1} & Linear(65536)                       \\
                             & Reshape(1024, 8, 8)                  \\
    \hline 
    \multirow{3}{*}{Layer 2} & ConvT(512, (4, 4), stride=2)         \\
                             & Batch normalization \\
                             & Relu\\
    \hline
    \multirow{3}{*}{Layer 3} & ConvT(256, (4, 4), stride=2)         \\
                             & Batch normalization \\
                             & Relu\\
    \hline
    \multirow{3}{*}{Layer 4} & ConvT(3, (4, 4), stride=1)         \\
                             & Batch normalization\\
                             & Sigmoid\\
    %\bottomrule
    \end{tabular}
\end{sc}
\end{table}

\subsubsection{Classifiers Settings}
As to the classifiers, for Sec.~4.2.2, we use a DenseNet \cite{huang_densely_2017} as benchmark for data augmentation. The implementation we use is the one proposed in \cite{amos_bamosdensenetpytorch_2020} with a \emph{growth rate} equals to 10, \emph{depth} of 20 and 0.5 \emph{reduction} and the model is trained with a learning rate of $10^{-3}$, weight decay of $10^{-4}$ and a batch size of 200. The classifier is trained until the loss does not improve on the validation set for 50 epochs and tested on the original test sets (\emph{e.g.} $\approx 1000$ samples per class for MNIST). For Sec.~4.2.3., the MLP has 400 hidden units with relu activation function. It is trained with Adam optimizer and a learning rate of  $10^{-3}$. Training is stopped if the loss does not improve on the validation set for 20 epochs. In Sec.~4.2.4, we consider a DenseNet again and increase (resp. decrease) its depth to increase (resp. decrease) the number of parameters of the classifier. Any other parameter is set to the value mentionned earlier.

%\subsubsection{Setting of Sec.~4.2.3 Robustness Across Classifiers}
%All common supervised classifiers, except the MLP, are implemented using the framework developed in \cite{pedregosa_scikit-learn_2011}. The SVM is %combined with a \emph{radial basis function} kernel. The random forest is trained with 100 trees with the \emph{gini} criteria. The $k$-Nearest Neighbour %algorithm is trained with 1 to 29 \emph{neighbours} and using the Euclidean distance. The number of nearest neighbours is selected on the model achieving %the best accuracy score on the validation set. The neural network is implemented using pytorch library \cite{paszke_automatic_2017} and is composed by %400 hidden units with relu activation. It is optimized using Adam optimizer \cite{kingma_adam_2014} with a weight decay of $10^{-4}$, learning rate of %$10^{-3}$ and batch size of 50. The model is trained until the loss does not improve on the validation set for 20 epochs. Each classifier with 5 %independent runs and tested on the original test set (\emph{i.e.} $\approx 1000$ samples per class for MNIST). An implementation is provided in the %supplementary materials.
%
%\subsubsection{Setting of Sec.~4.2.3 A Note on the Method Scalability}

\subsection{Parameters of Sec.~5 Validation on Medical Imaging}\label{appendix D.3}

To generate new data on the ADNI database we amend the neural network architectures and use the one described in Table~\ref{tab: app neural arch adni}. The parameters used in the \emph{geometry-aware} VAE are provided in Table~\ref{tab: app hyperparams adni}. An Adam optimizer with a learning rate of $10^{-5}$ and batch size of 25 are used. The VAE model is trained until the ELBO does not improve for 50 epochs. Generating 50 ADNI images takes approx. 30 s.\footnote{Depends on the length of the MCMC chain and HMC hyper-parameter, $l$. We used 300 steps with $l=15$.} with the proposed method on Intel Core i7 CPU (6x1.1GHz) and 16 GB RAM.

\begin{table}[!ht]
\caption{Neural Net Architecture}
    \centering
    \scriptsize
    \begin{tabular}{c|cccc}
    \hline
         $\mu_{\phi} $      &  \multirow{2}{*}{($D$, h1, rel)}  & (h1, h2, relu) & (h2, h3, relu) & (h3, $d$, lin)   \\
         $\Sigma_{\phi}$    &                                    & (h1, h2, relu) & (h2, h3, relu) & (h3, $d$, lin) \\
         \hline
         $ \pi_{\theta} $   &  ($d$, h3, relu)                  & (h3, h2, relu) & (h2, h1, relu) & (h1, $D$, sig)   \\
         \hline
        $L_{\psi}^{\text{diag.}}$  &  \multirow{2}{*}{($D$, h3, relu)}& (h3, $d$, lin)                 & - & - \\
        $L_{\psi}^{\text{low.}}$    &                                  & (h3, $\frac{d(d-1)}{2}$, lin) & - & - \\
        \hline
        \hline
        D &h1& h2 & h3 & d\\
        %\hline
        777504 & 500 & 500 & 400 & 10\\
        \hline
    \end{tabular}
    \label{tab: app neural arch adni}
\end{table}

\begin{table}[!ht]
  \centering
  \caption{\emph{Geometry-aware} parameters settings for ADNI database}
  \label{tab: app hyperparams adni}
  \begin{tabular}{c|cccccc}
  \hline
       \multirow{2}{*}{Data set} & \multicolumn{6}{c}{Parameters} \\
       & $d$ & $K$ & $\varepsilon$ & $T$ & $\lambda$ & $\sqrt{\beta_0}$   \\
      \hline
      ADNI       & 10 & 3 & $10^{-3}$ & 1.5 & $10^{-2}$ & 0.3 \\
      \hline
  \end{tabular}
\end{table}

\subsection{Classifiers Architectures for ADNI}\label{appendix D.4}
\begin{figure}[ht]
\centering
\includegraphics[scale=0.6]{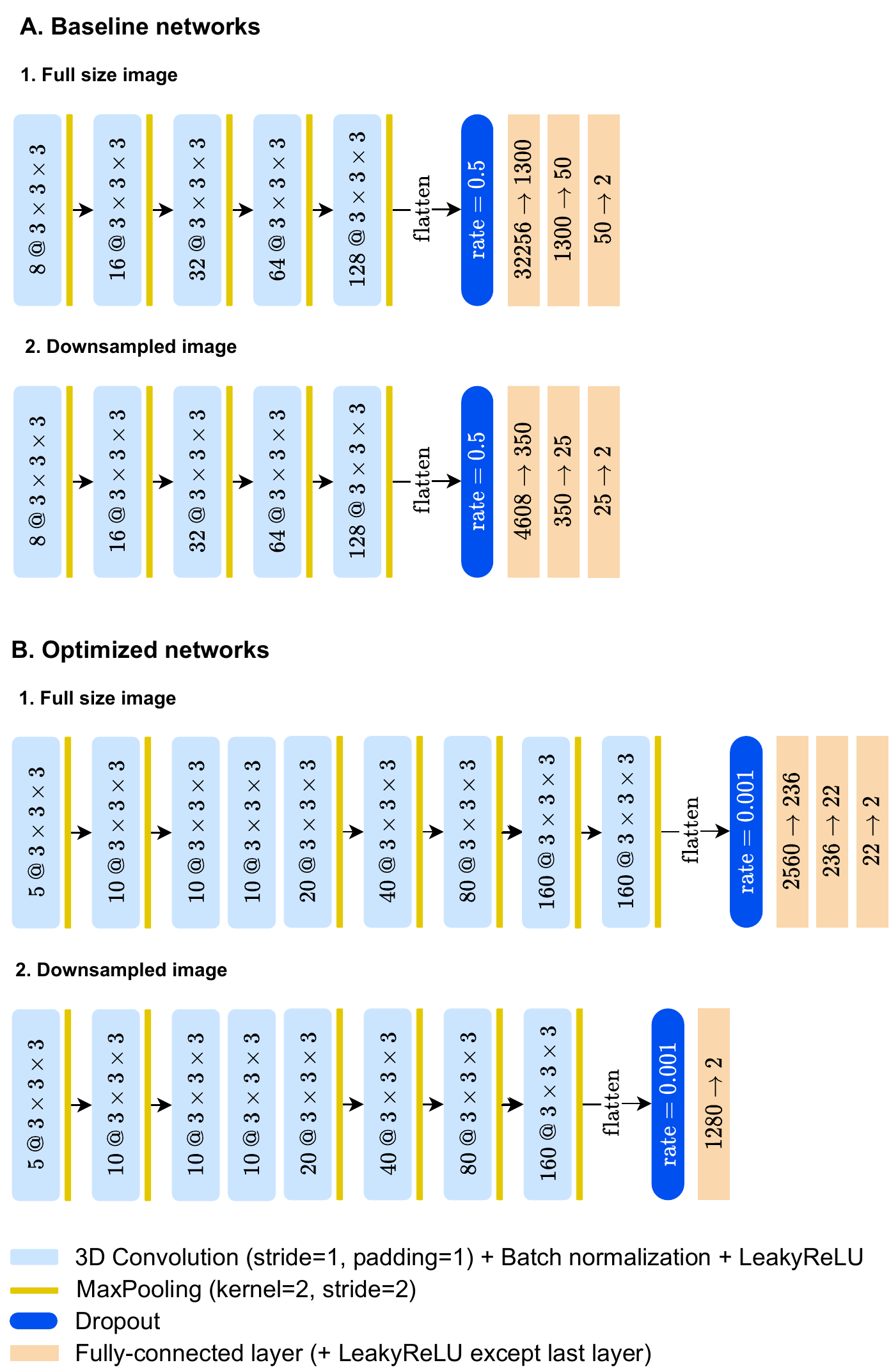}
\caption{Diagrams of the network architectures used for classification. The first \textbf{baseline} architecture (A1) is the one used in~\cite{wen_convolutional_2020}, the second one (A2) is a very similar one adapted to process smaller inputs. The \textbf{optimized} architectures (B1) and (B2) are obtained independently with two different random searches. For convolution layers we specify the number of channels @ the kernel size and for the fully-connected layers we specify the number of input nodes $\rightarrow$ the number of output nodes. Each fully-connected layer is followed by a LeakyReLU activation except for the last one. For the dropout layer, the dropout rate is specified.} 
\label{app architectures}
\end{figure}

In Fig.~\ref{app architectures} are presented the neural network architectures used for the classifier in ADNI classification tasks. As explained in the paper, we consider one architecture for each input size (\emph{i.e.} down-sampled and high-resolution images). The \textbf{baseline} architecture is taken from the study of \cite{wen_convolutional_2020} and was obtained by optimizing manually the networks on the ADNI data set for the same task (AD vs CN). The \textbf{optimized} one is obtained with a random search~\cite{bergstra_random_2012} across 100 architectures that allows exploring different hyperparameter values such as the number of convolutional blocks, the number of filters in the first layer, the number of convolutional layers in a block, the number of fully-connected layers, the dropout rate, the learning rate and the weight decay.  The architectures are trained on the 5-fold cross-validation on \textit{train-full} and for each input size we choose the architecture obtaining the best mean balanced accuracy across the validation sets of the cross-validation.
 
\section{A Few More Sampling Comparisons (Sec.~3.3)}\label{appendix E}
In addition to the comparison performed in Sec.~3.3.1, we also compare qualitatively a Vanilla VAE, a VAE with VAMP prior and a \emph{geometry-aware} VAE on four reduced data sets and in higher dimensional latent spaces of dimension 10. The first one is created with 180 binary rings and disks with different diameters and thicknesses ensuring balanced classes. The second one is composed of 120 samples of EMNIST (letter \emph{M}) and referred to as \emph{reduced} EMNIST. Another one is created with 120 samples from the classes \emph{0}, \emph{1} and \emph{2} of MNIST database ensuring balanced classes and is called \emph{reduced} MNIST. The last one, \emph{reduced} Fashion, is again composed of 120 samples from three classes (\emph{shoes}, \emph{trouser} and \emph{bag}) from FashionMNIST and ensuring balanced classes.  
The models have the same architectures as described in Table~\ref{tab: app mnist neural arch generation} and are trained with the parameters stated in Table.~\ref{Table: app Hyper-parameters more gene}. Each model is trained until the ELBO does not improve for 20 epochs with Adam optimizer, a learning rate of $10^{-3}$ and in a single batch. In Fig.~\ref{Fig: app Comparison} are presented from top to bottom: 1) an extract of the training samples for each data set; 2) samples obtained with a vanilla VAE with a Gaussian prior; 2)  data generated from a VAE with VAMP prior; 3) samples created by a \emph{geometry-aware} VAE and using the prior or 4) samples from our method. As discussed in the paper, the proposed method is again able to visually outperform peers since for all data sets it is able to create sharper and more meaningful samples even if the number of training samples is quite small.

\begin{table}[ht]
  \centering
  \caption{\emph{Geometry-aware} VAE parameters.}
  \label{Table: app Hyper-parameters more gene}
  \begin{tabular}{c|cccccc}
  \hline
       \multirow{2}{*}{Data sets} & \multicolumn{6}{c}{Parameters} \\
       & $d^{*}$  & $K$ & $\varepsilon$ & $T$ & $\lambda$ & $\sqrt{\beta_0}$   \\
      \hline
      Synthetic shapes              & 10 & 3 & $10^{-2}$ & 1.5 & $10^{-3}$ & 0.3 \\
      \textit{reduced} MNIST        & 10 & 3 & $10^{-2}$ & 1.5 & $10^{-3}$ & 0.3 \\
      \textit{reduced} EMNIST       & 10 & 3 & $10^{-2}$ & 1.5 & $10^{-3}$ & 0.3 \\
      \textit{reduced} Fashion      & 10 & 3 & $10^{-2}$ & 1.5 & $10^{-3}$ & 0.3 \\
      \hline
      \multicolumn{7}{l}{\scriptsize{* Latent space dimension (same for VAE and VAMP-VAE)}} 
  \end{tabular}
\end{table}

\section{Additional Results (Sec.4.2.3)}\label{appendix F}
\begin{figure}[ht] 
  \centering
  \subfloat[\emph{reduced} EMNIST \label{Subfig: app Classifiers rob MNIST bal}]{\includegraphics[width=3.in]{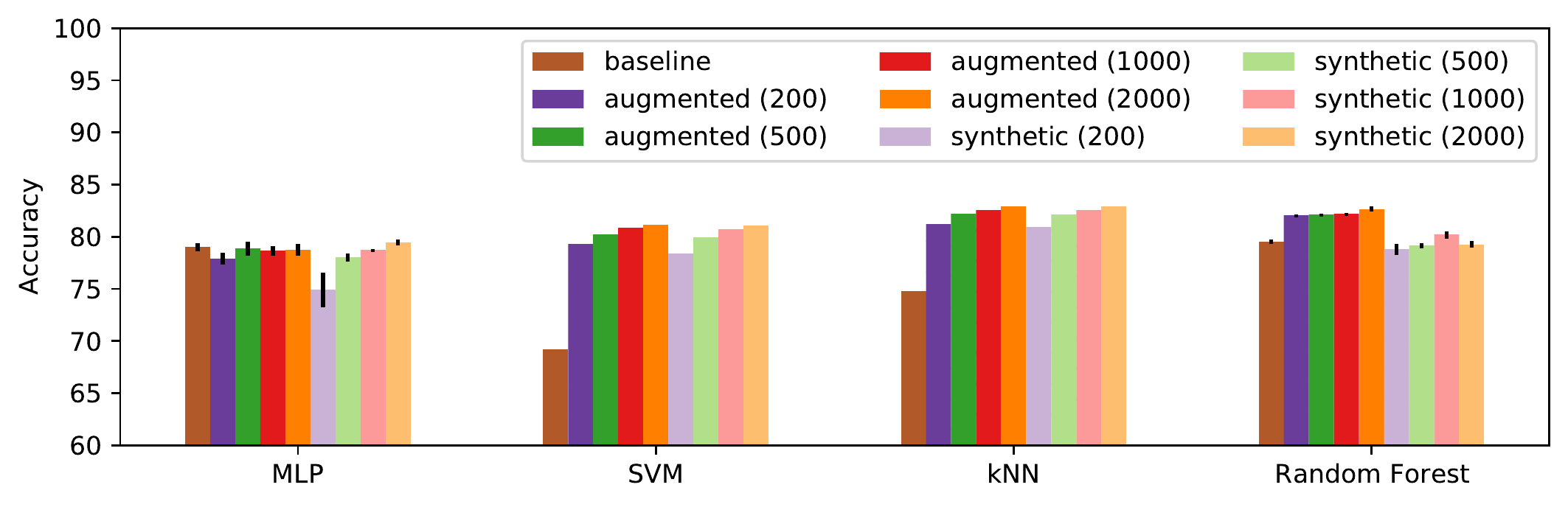}
  }
  \hfil
  \centering
  \subfloat[\emph{reduced} FashionMNIST \label{Subfig: app Classifiers rob MNIST unbal}]{\includegraphics[width=3.in]{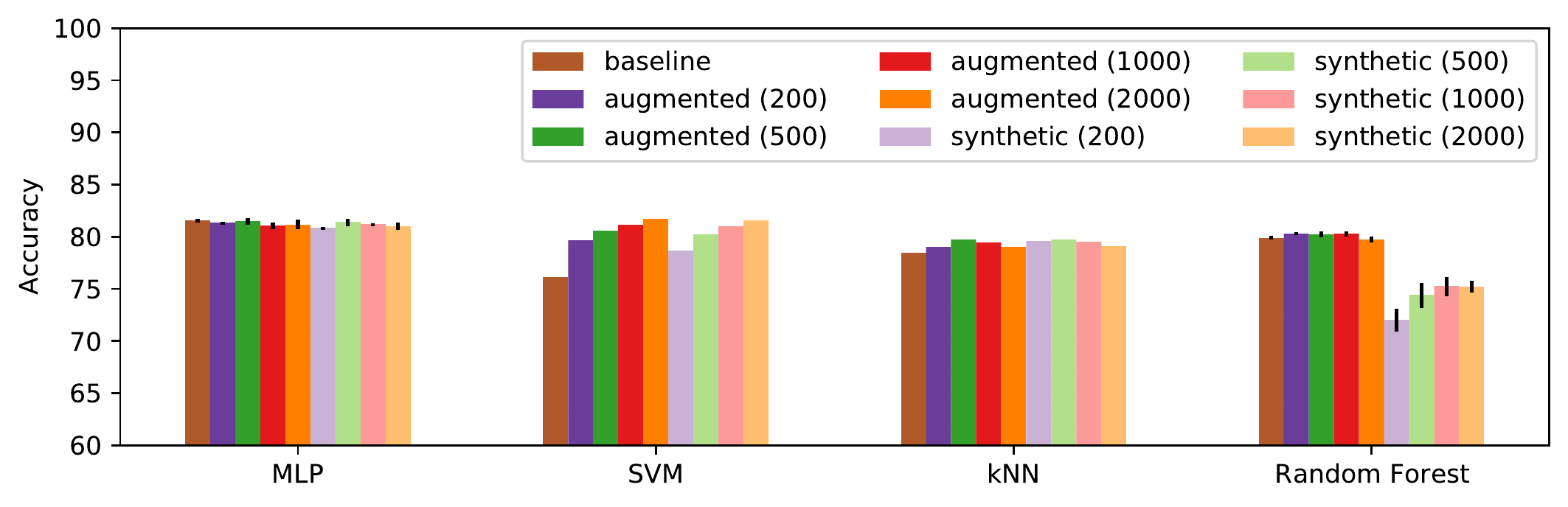}
  }
  
  \caption{Evolution of the accuracy of four benchmark classifiers on the \emph{reduced} EMNIST data set (top) and the \emph{reduced} Fashion data set (bottom). Stochastic classifiers are trained with five independent runs and we report the mean accuracy and standard deviation on the test set.}
  \label{Fig: app Classifiers rob}
  \end{figure}

\begin{figure*}[!ht]
  \centering
  \adjustbox{minipage=5.5em,raise=\dimexpr -2\height}{\small Training\\ samples}
  \captionsetup[subfigure]{position=above, labelformat = empty}
  \subfloat[\emph{reduced} EMNIST \scriptsize{(120)}]{\includegraphics[width=1.25in]{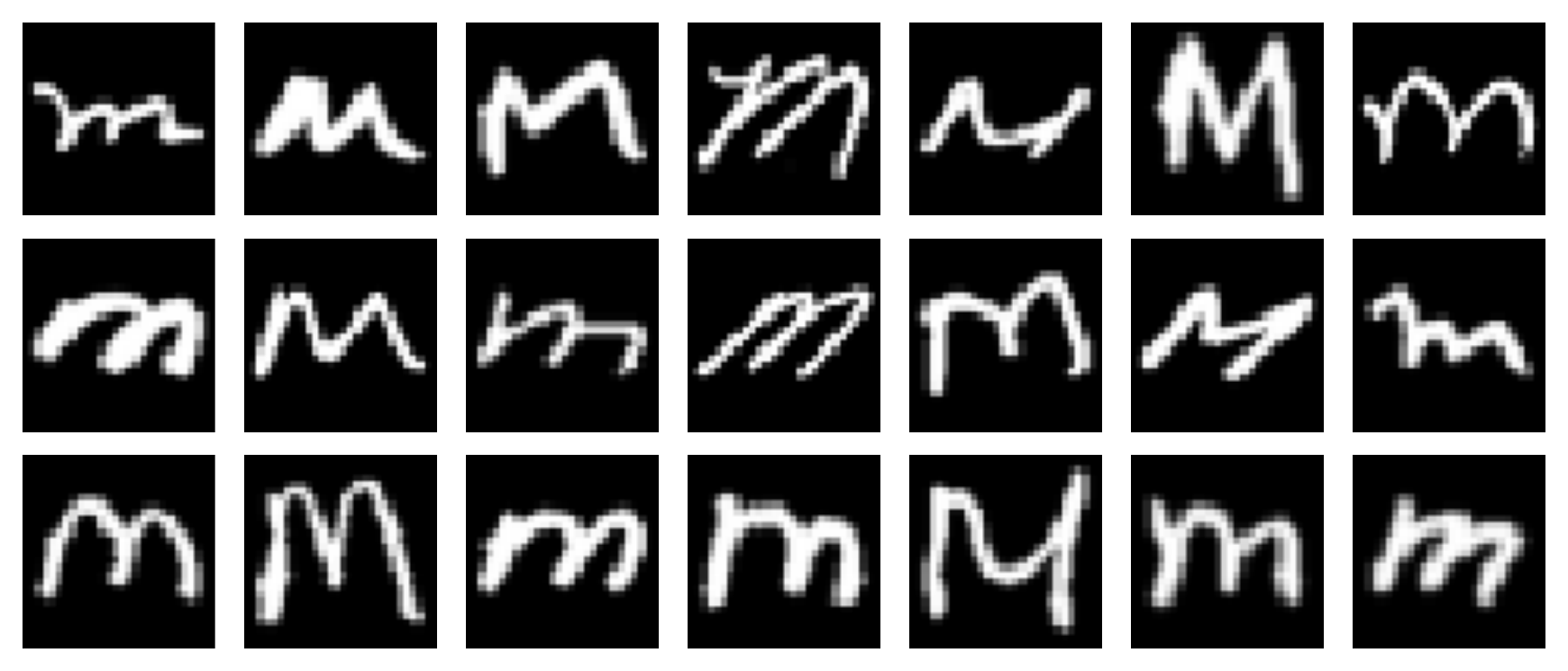}}
  \hspace{0.01in}
  \subfloat[\emph{reduced} MNIST \scriptsize{(120)}]{\includegraphics[width=1.25in]{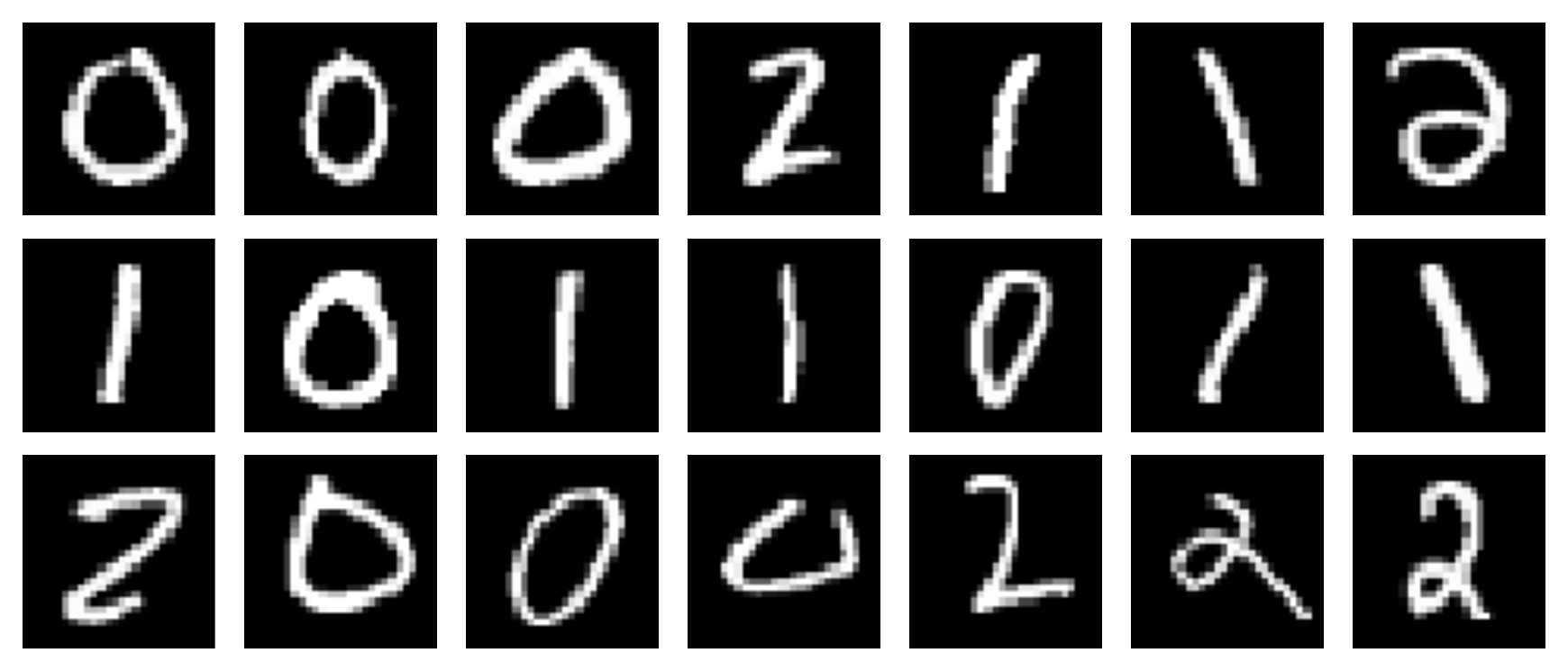}}
  \hspace{0.01in}
  \subfloat[\emph{reduced} Fashion \scriptsize{(120)}]{\includegraphics[width=1.25in]{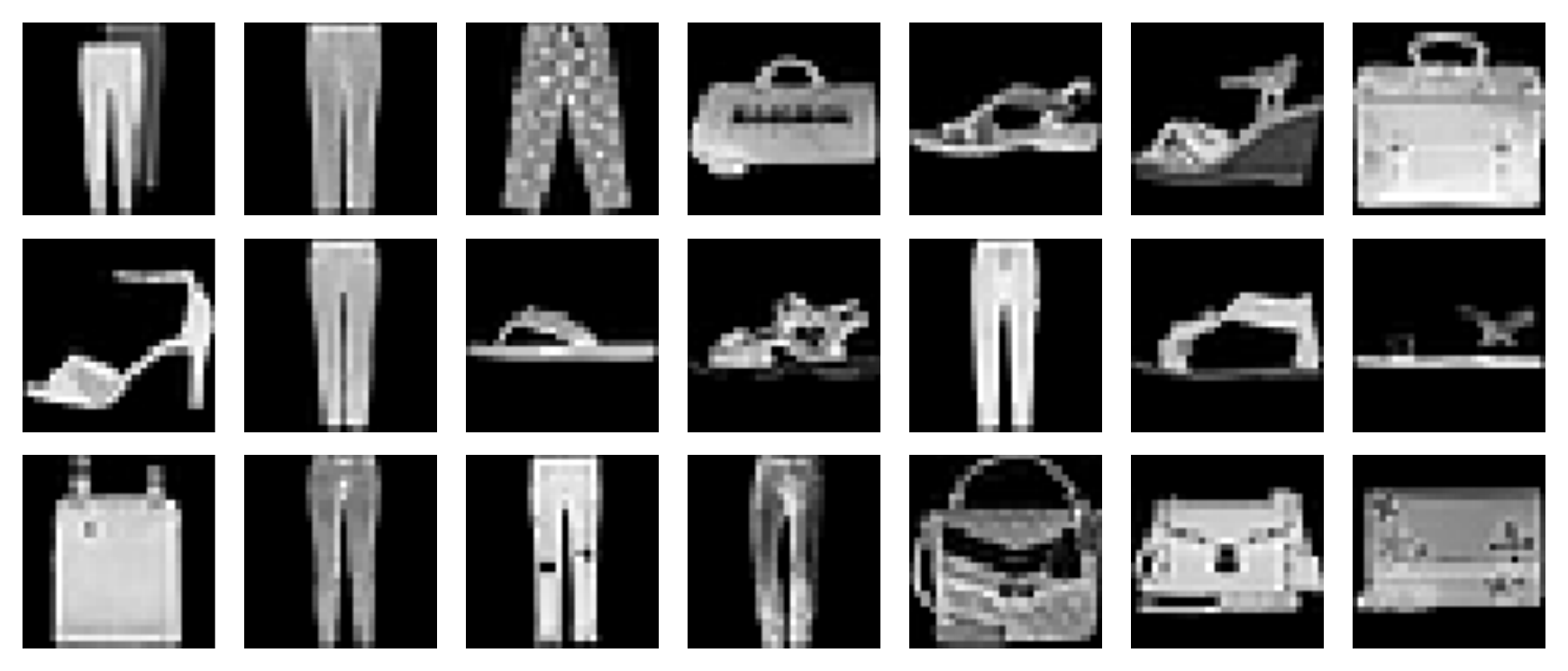}}
  \hspace{0.01in}
  \subfloat[Synthetic \scriptsize(180)]{\includegraphics[width=1.25in]{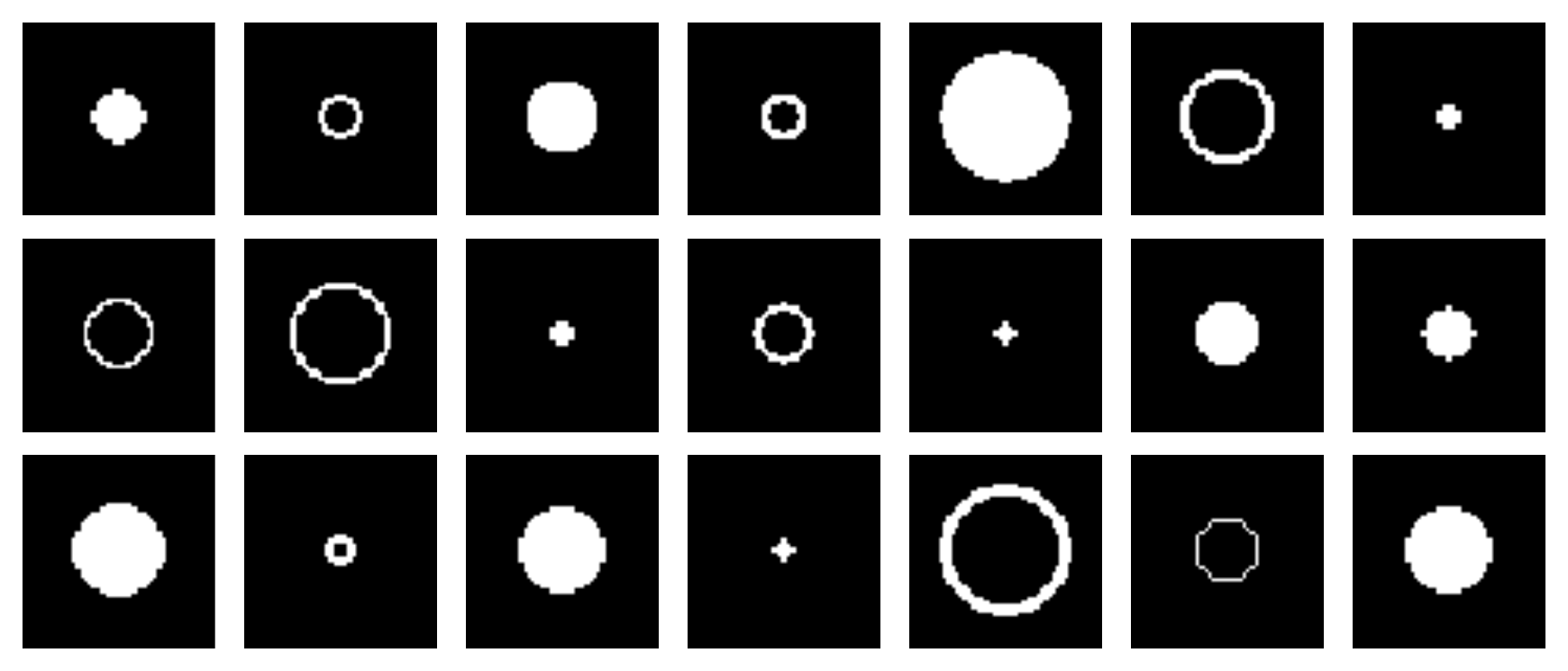}}
  \vfil
  \vspace{-0.12in}
  \adjustbox{minipage=5.5em,raise=\dimexpr -4\height}{\small VAE +\\ $\mathcal{N}(0, I_d)$}
  \subfloat{\includegraphics[width=1.25in]{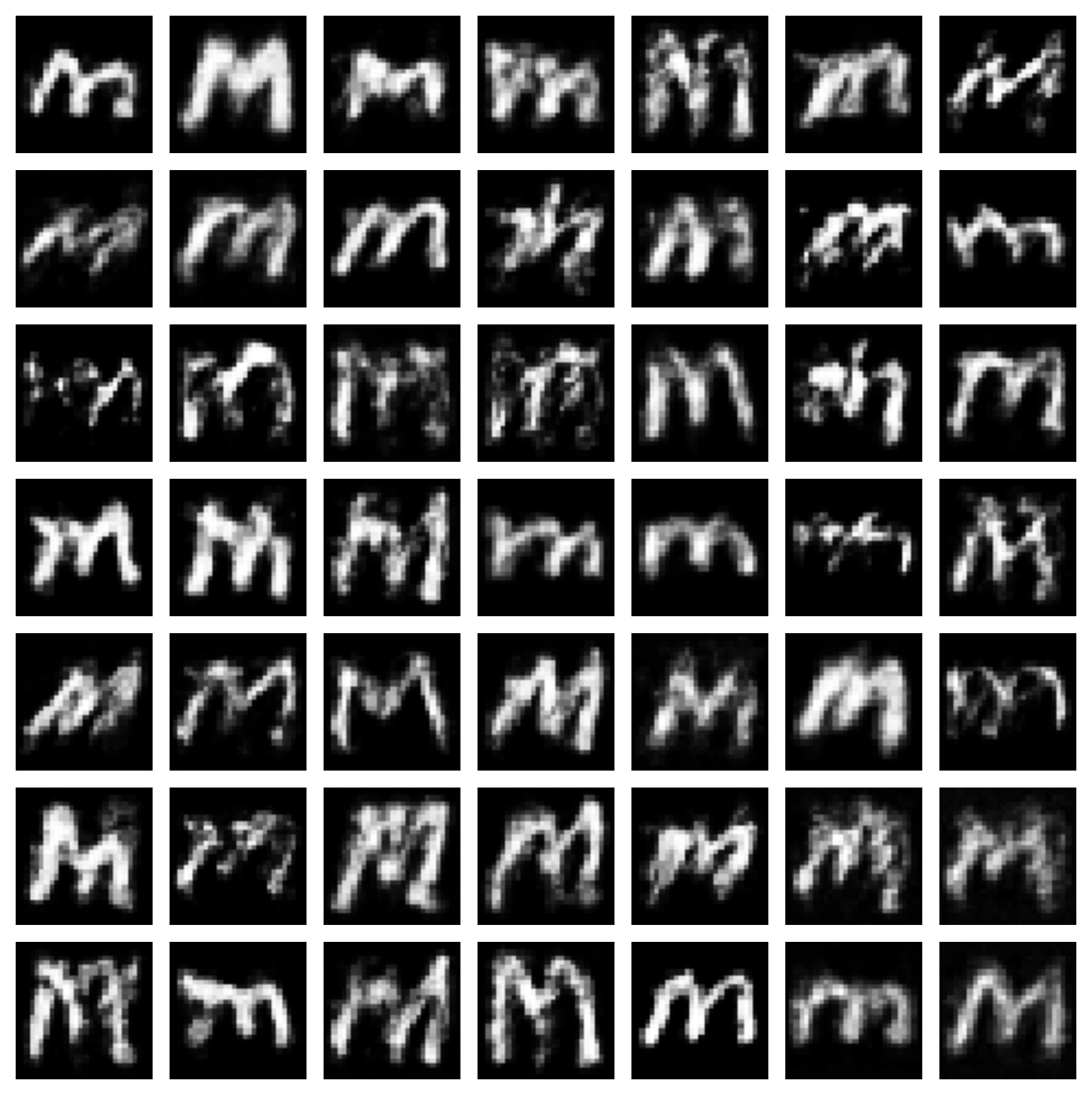}}
  \hspace{0.01in}
  \subfloat{\includegraphics[width=1.25in]{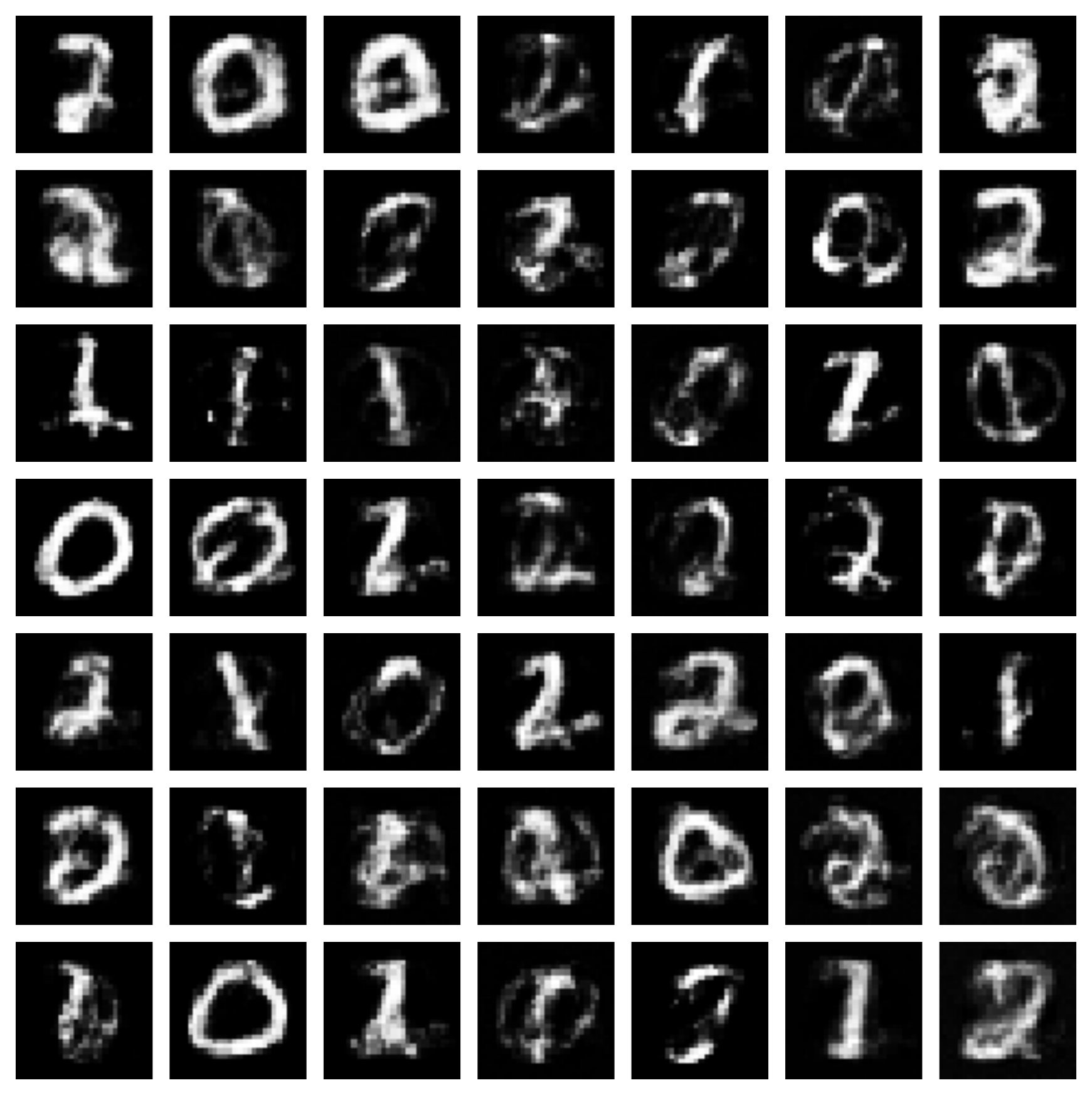}}
  \hspace{0.01in}
  \subfloat{\includegraphics[width=1.25in]{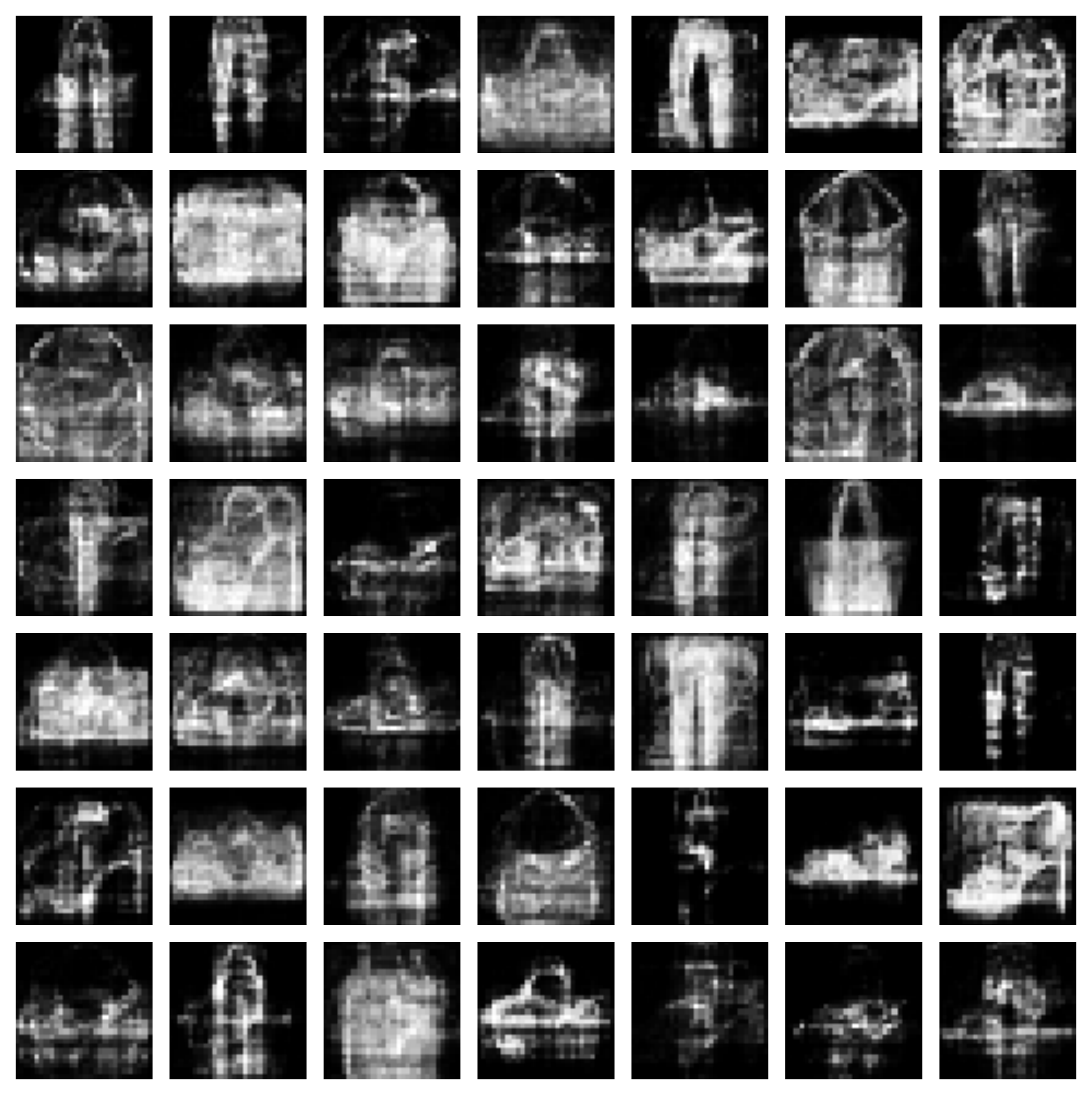}}
  \hspace{0.01in}
  \subfloat{\includegraphics[width=1.25in]{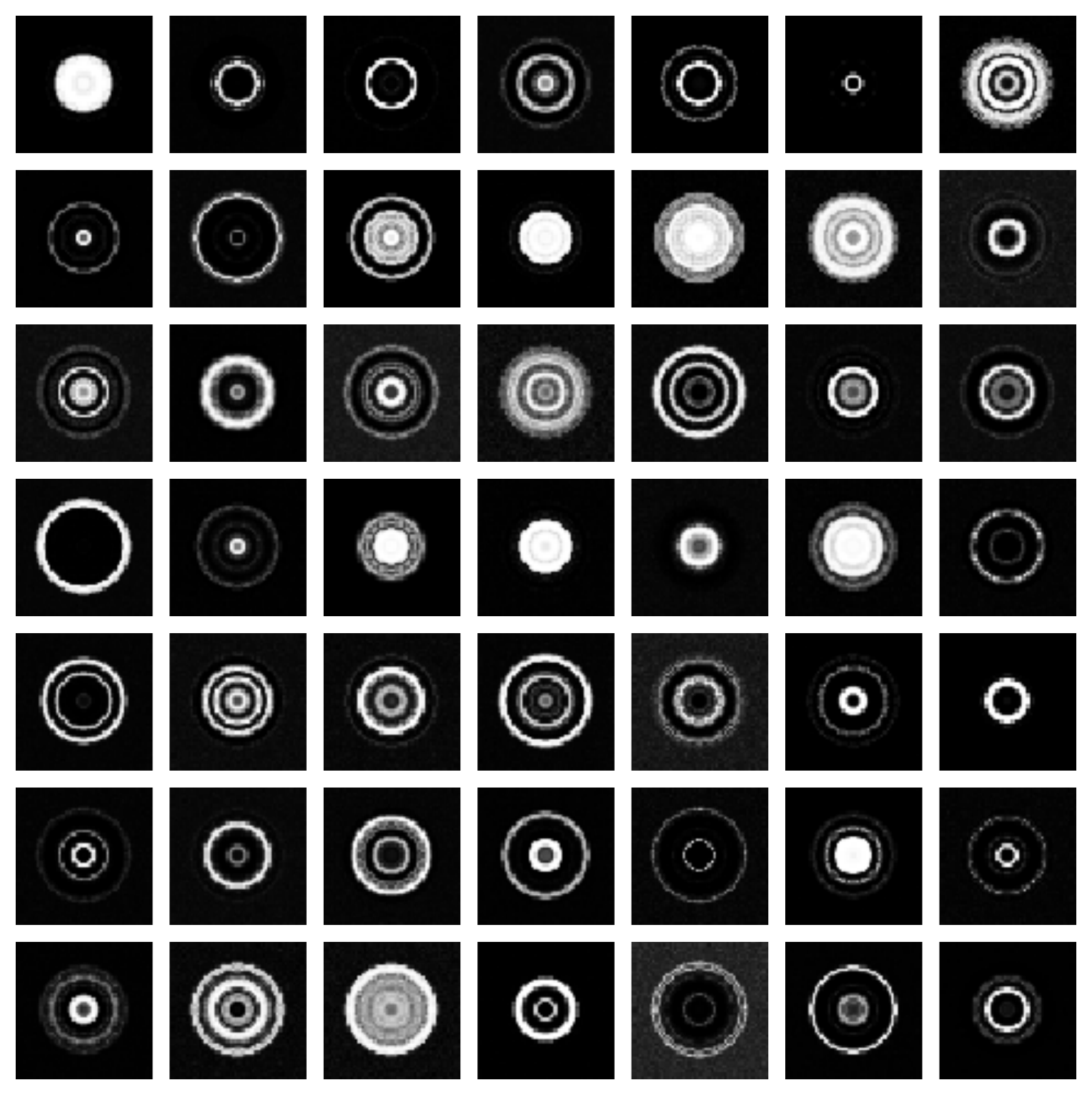}}
  \vfil
  \vspace{-0.12in}
  \adjustbox{minipage=5.5em,raise=\dimexpr -4\height}{\small VAE +\\ VAMP prior}
  \subfloat{\includegraphics[width=1.25in]{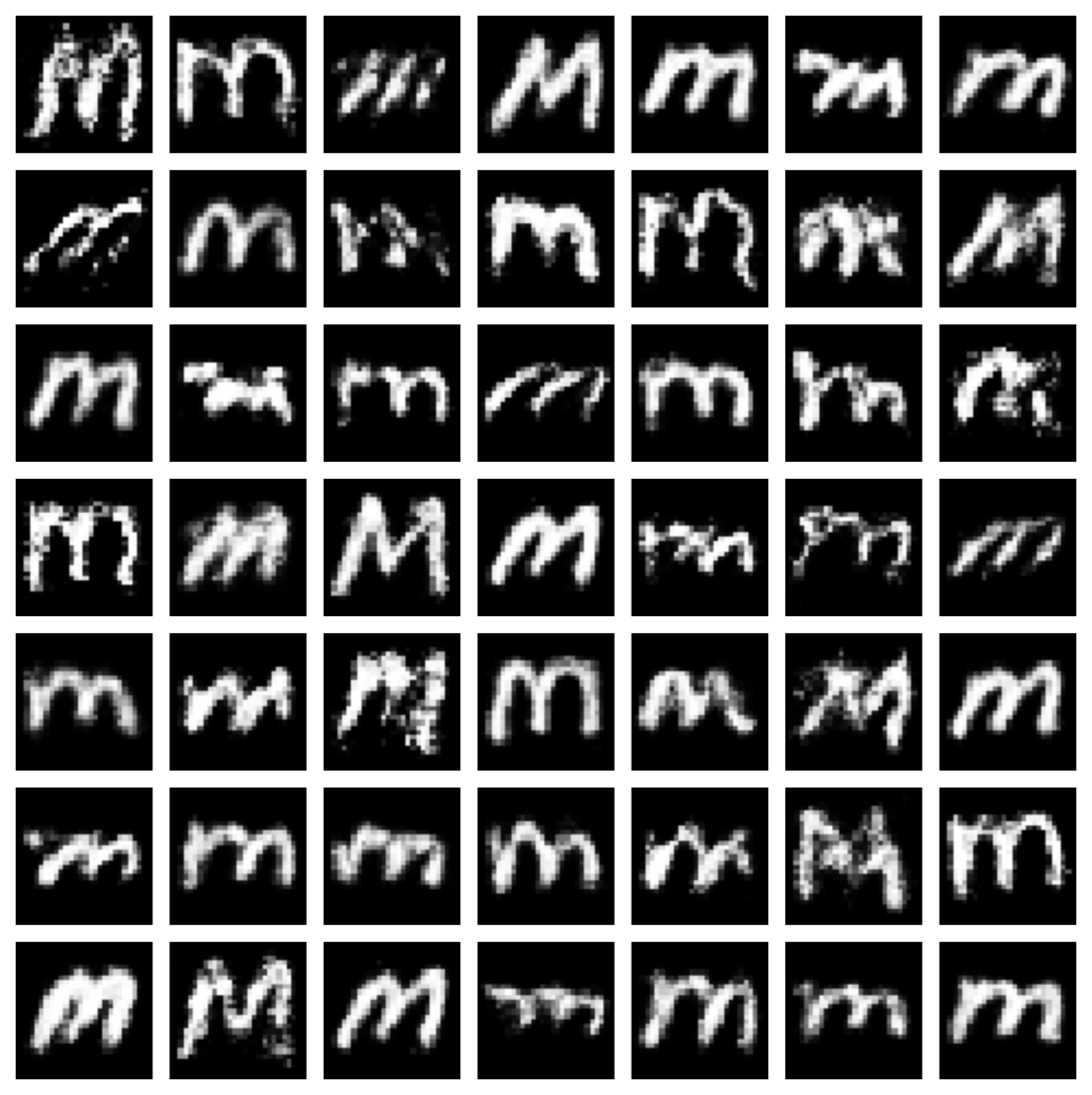}}
  \hspace{0.01in}
  \subfloat{\includegraphics[width=1.25in]{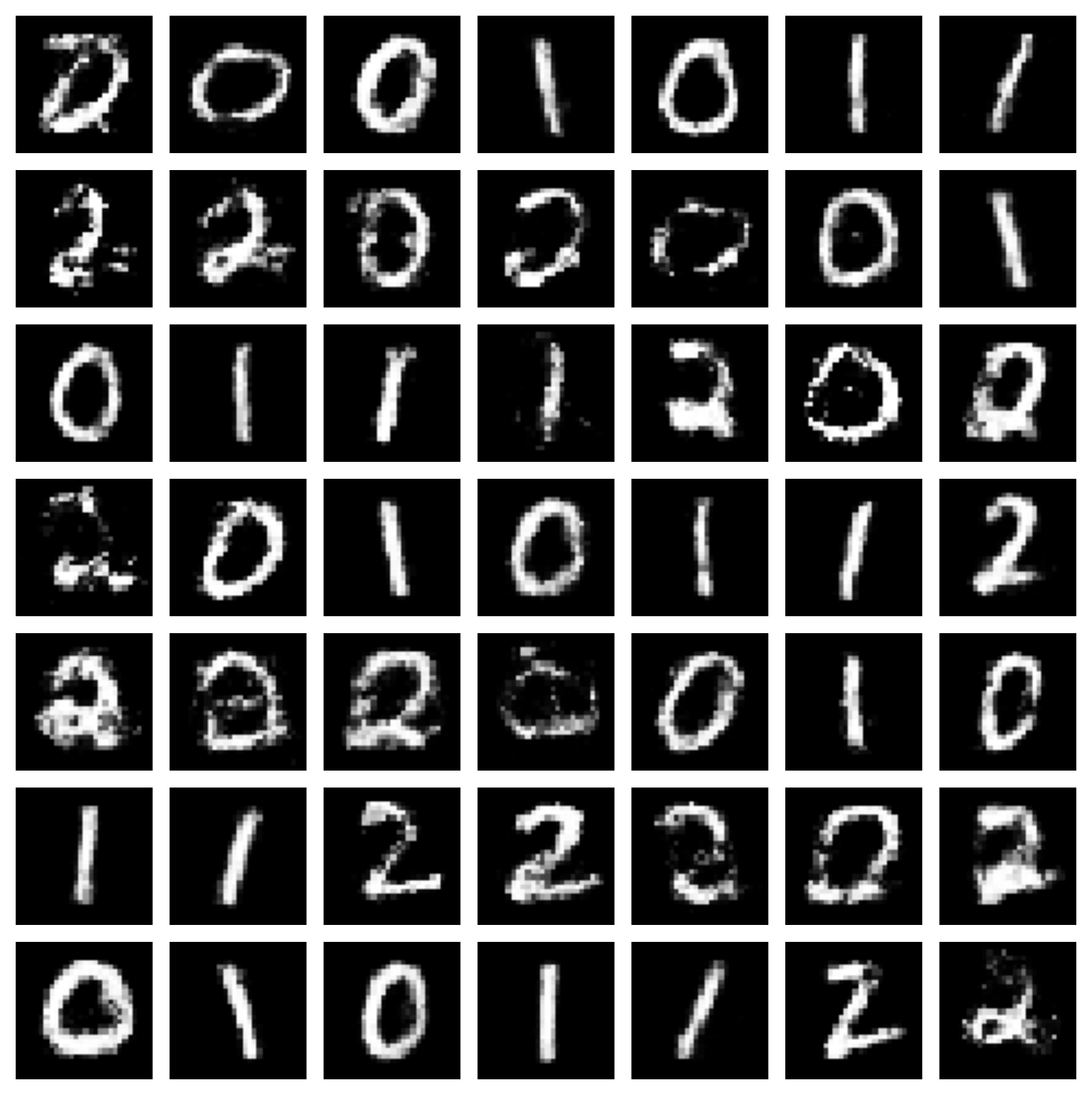}}
  \hspace{0.01in}
  \subfloat{\includegraphics[width=1.25in]{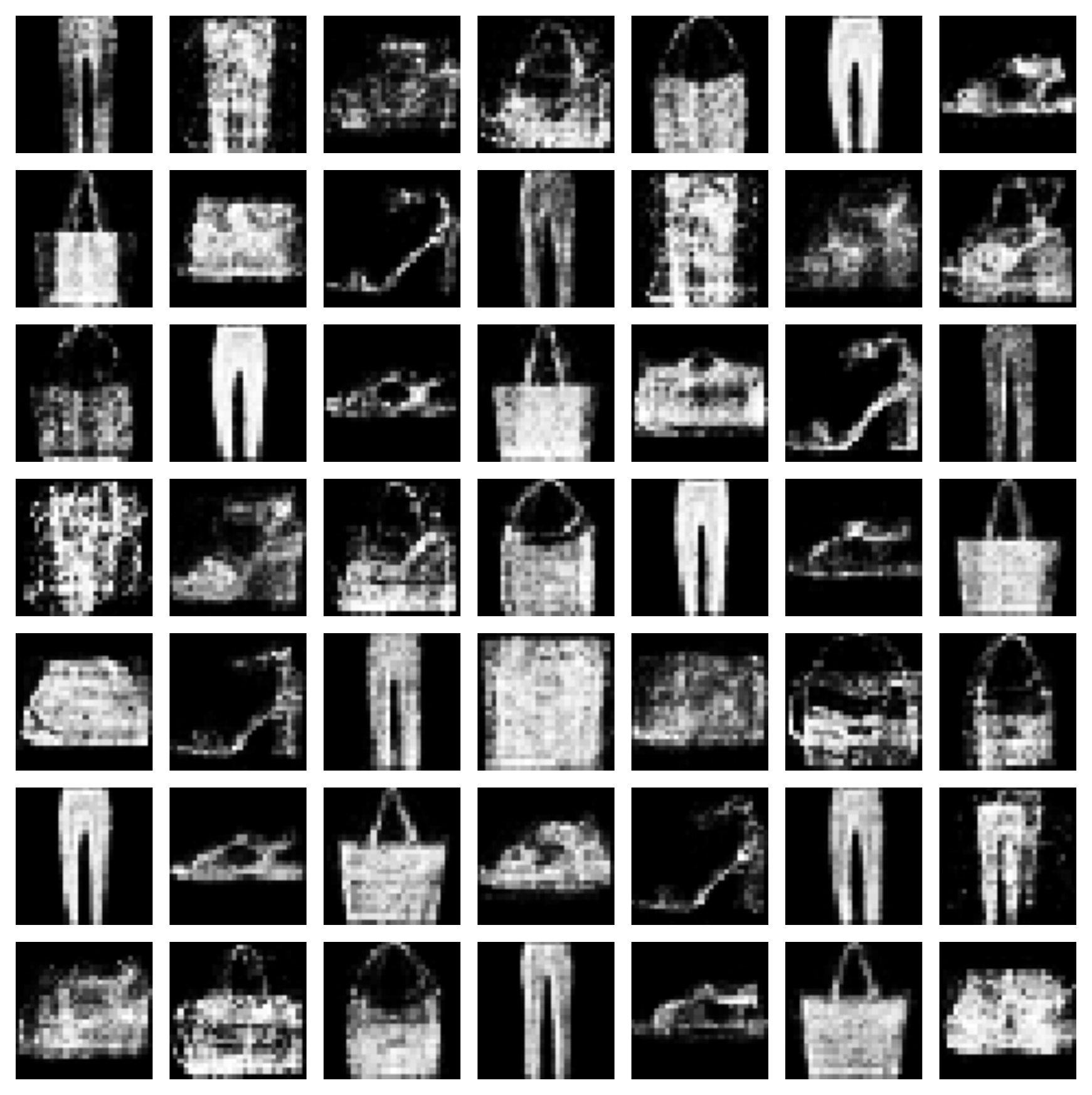}}
  \hspace{0.01in}
  \subfloat{\includegraphics[width=1.25in]{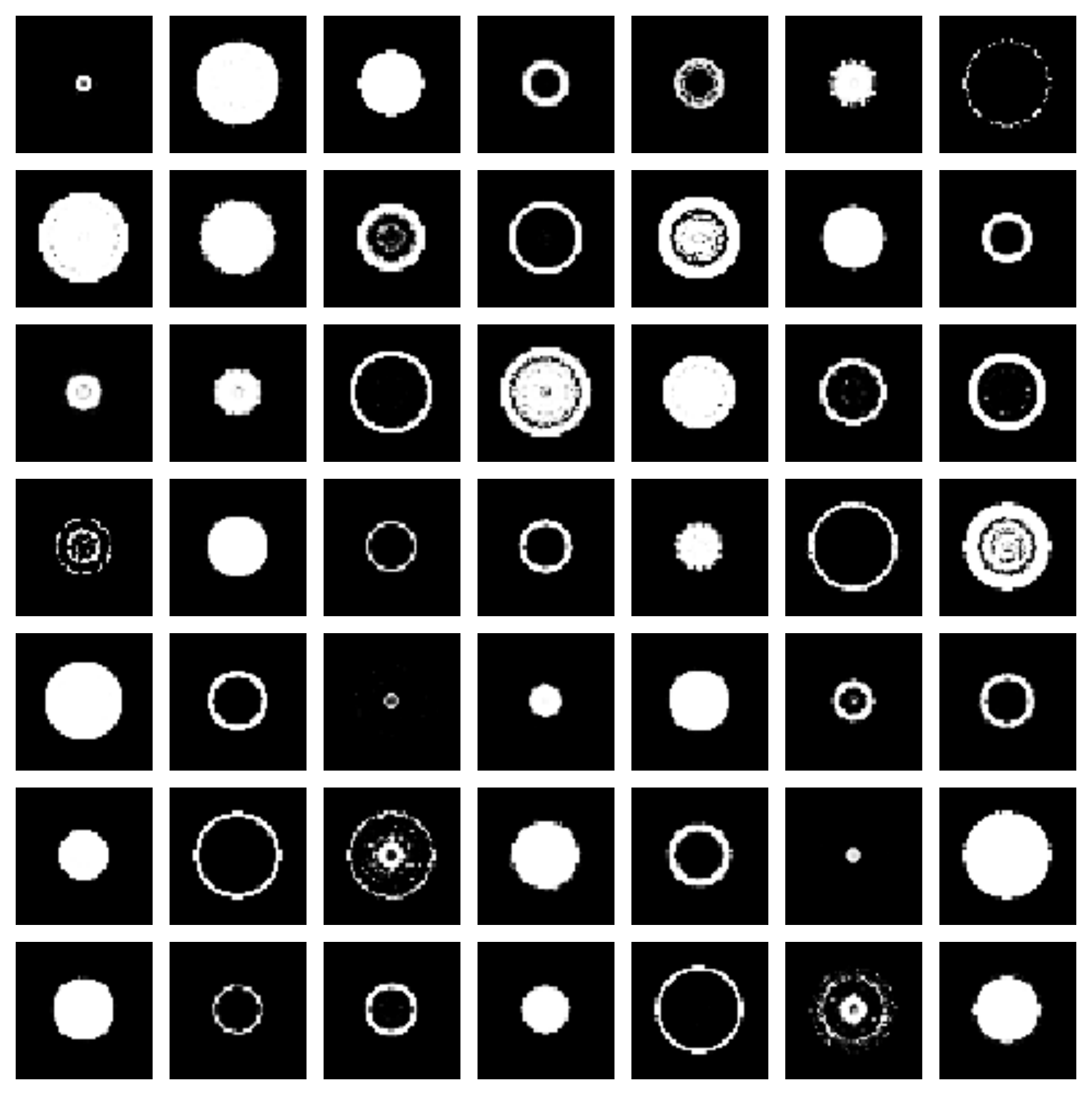}}
  \vfil
  \vspace{-0.12in}
  \adjustbox{minipage=5.5em,raise=\dimexpr -4\height}{\small RHVAE +\\ $\mathcal{N}(0, I_d)$}
  \subfloat{\includegraphics[width=1.25in]{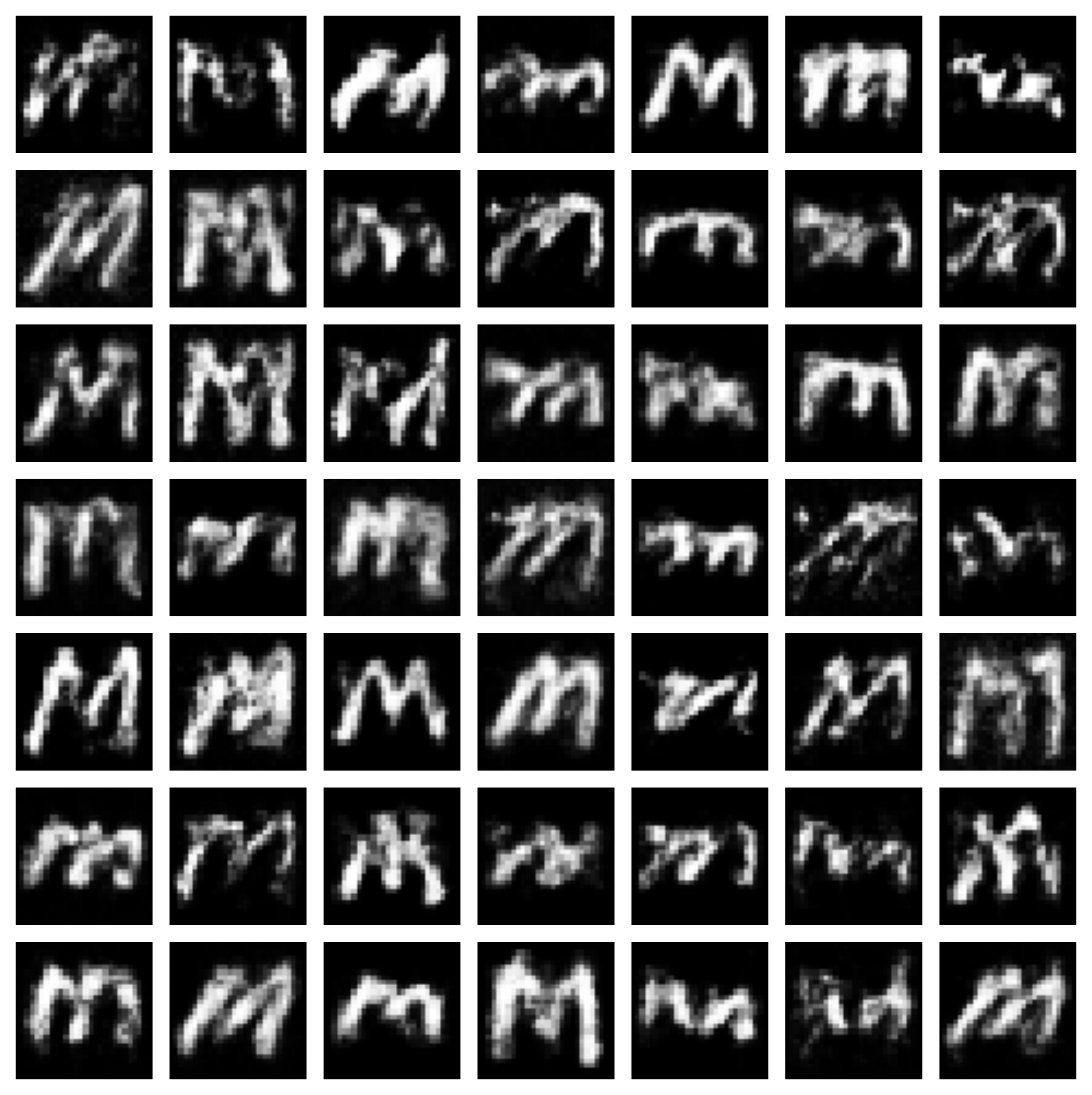}}
  \hspace{0.01in}
  \subfloat{\includegraphics[width=1.25in]{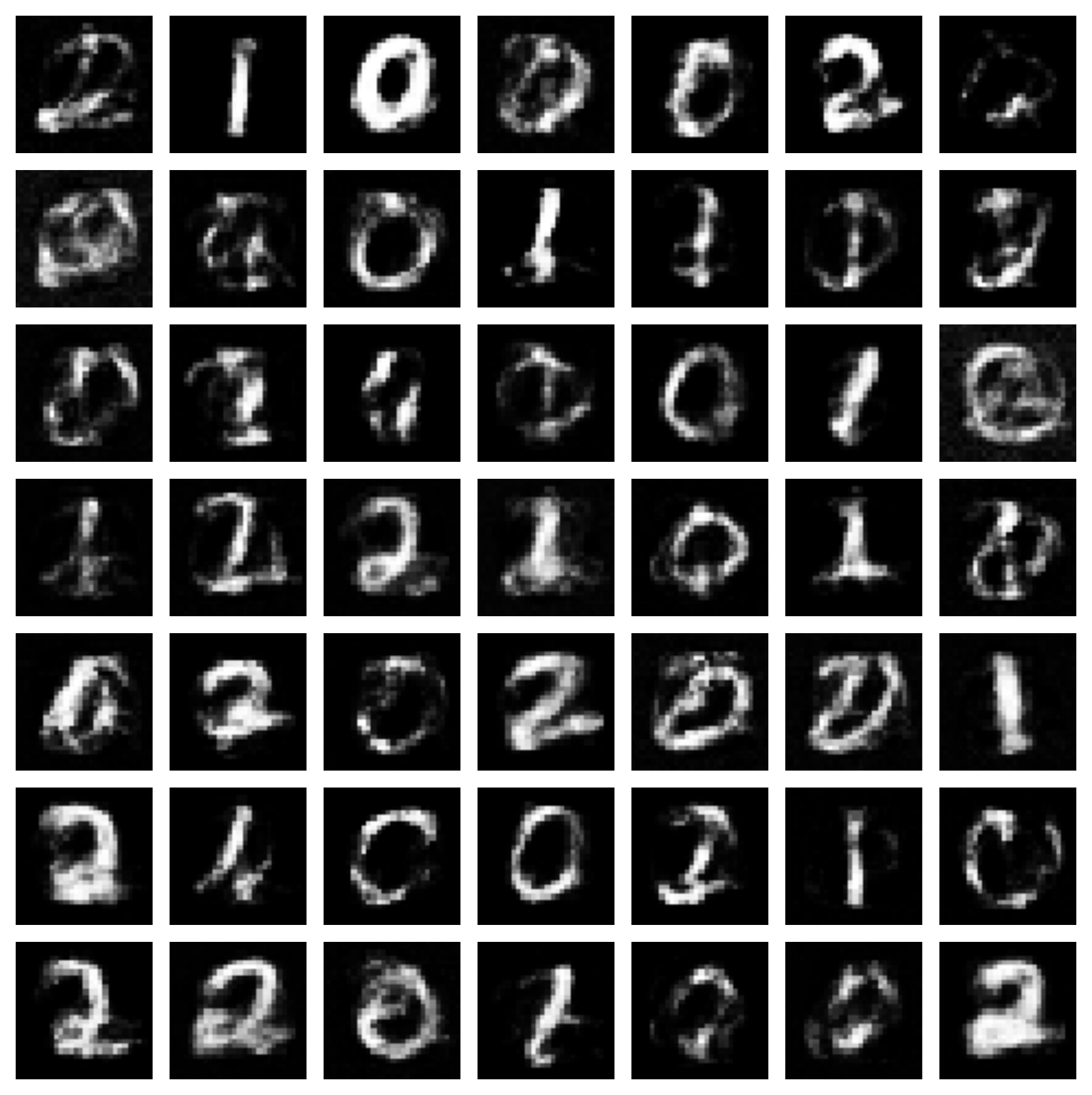}}
  \hspace{0.01in}
  \subfloat{\includegraphics[width=1.25in]{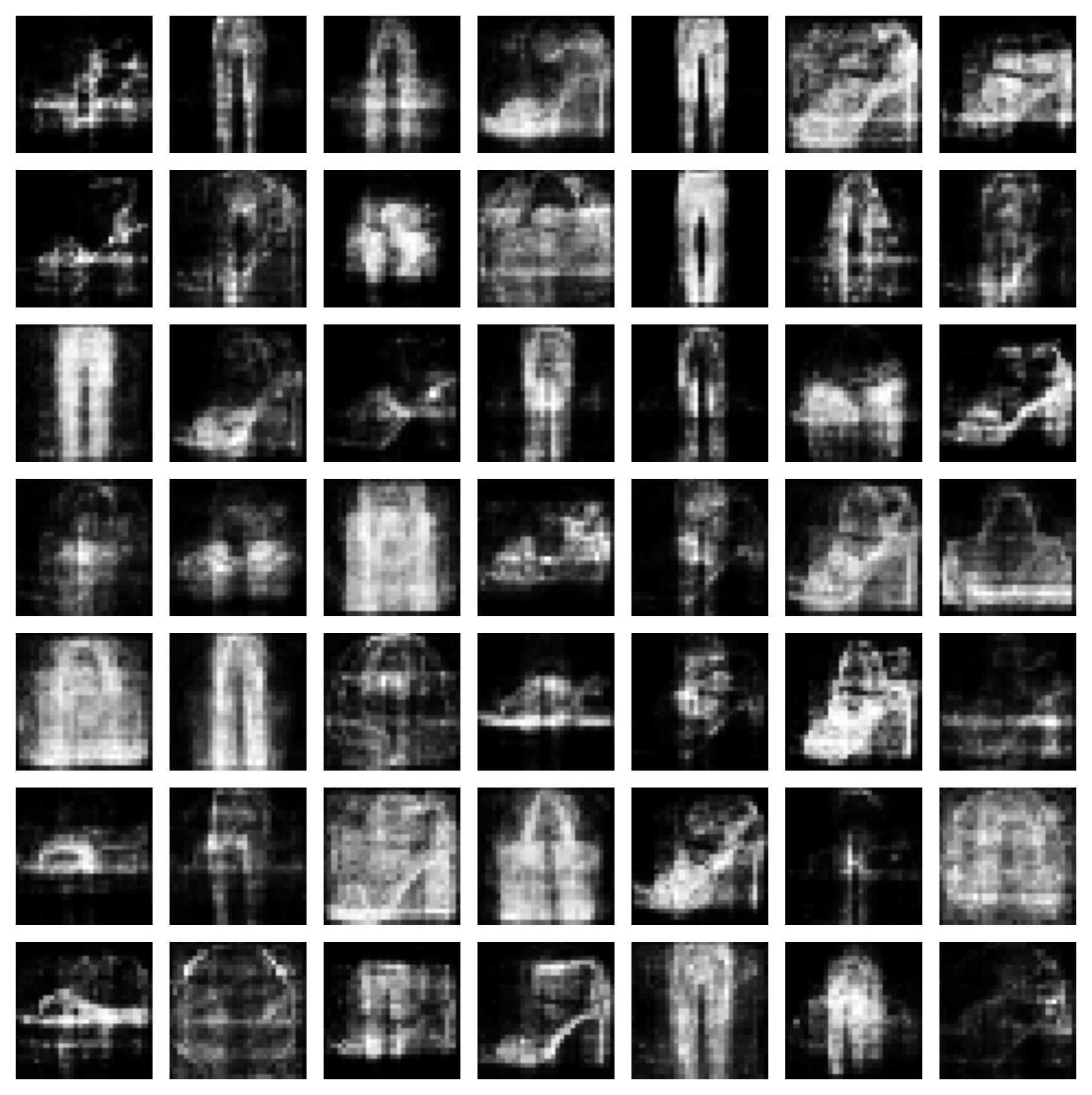}}
  \hspace{0.01in}
  \subfloat{\includegraphics[width=1.25in]{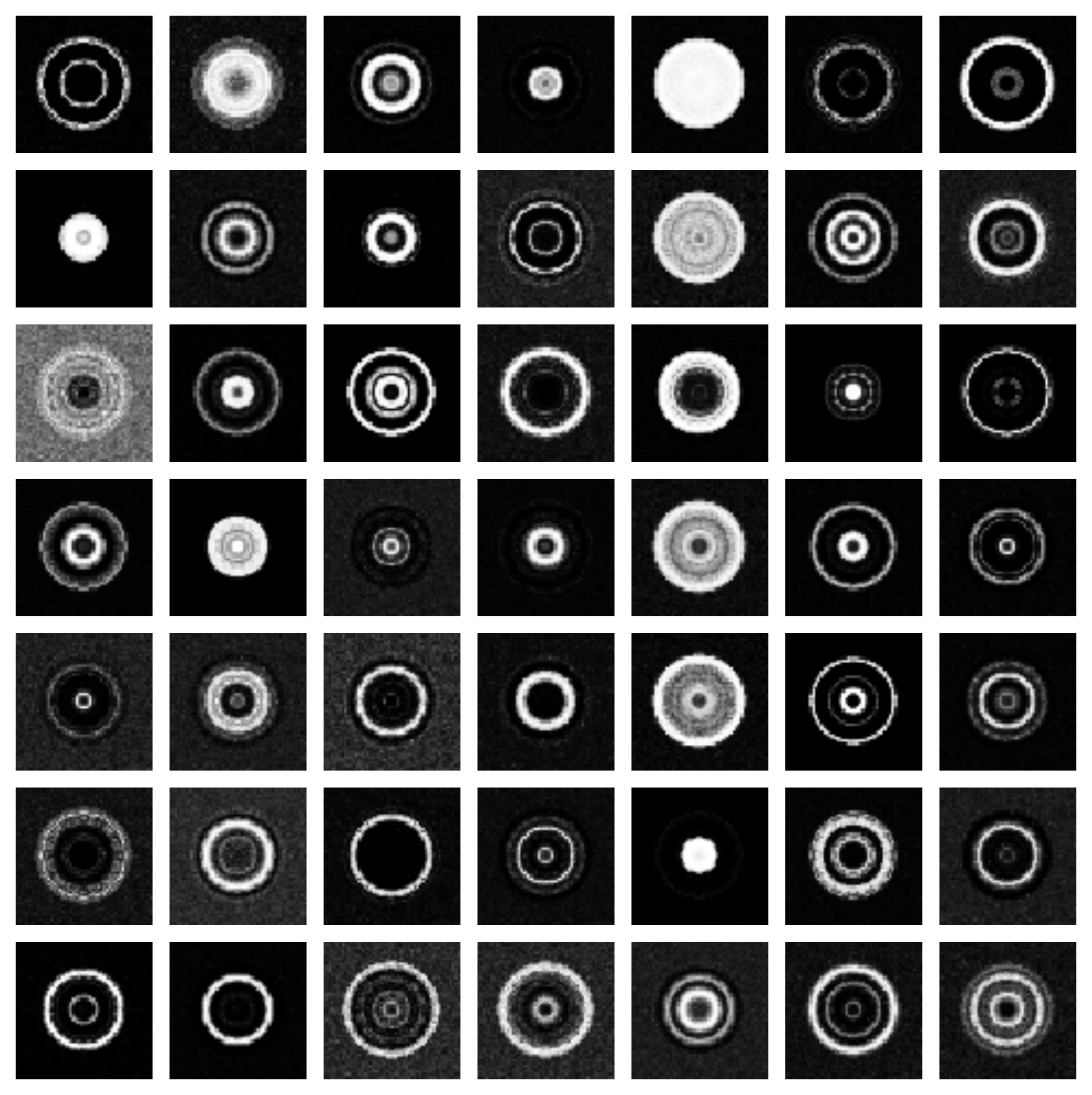}}
  \vfil
  \vspace{-0.12in}
  \adjustbox{minipage=5.5em,raise=\dimexpr -4\height}{\small RHVAE +\\ Ours}
  \subfloat{\includegraphics[width=1.25in]{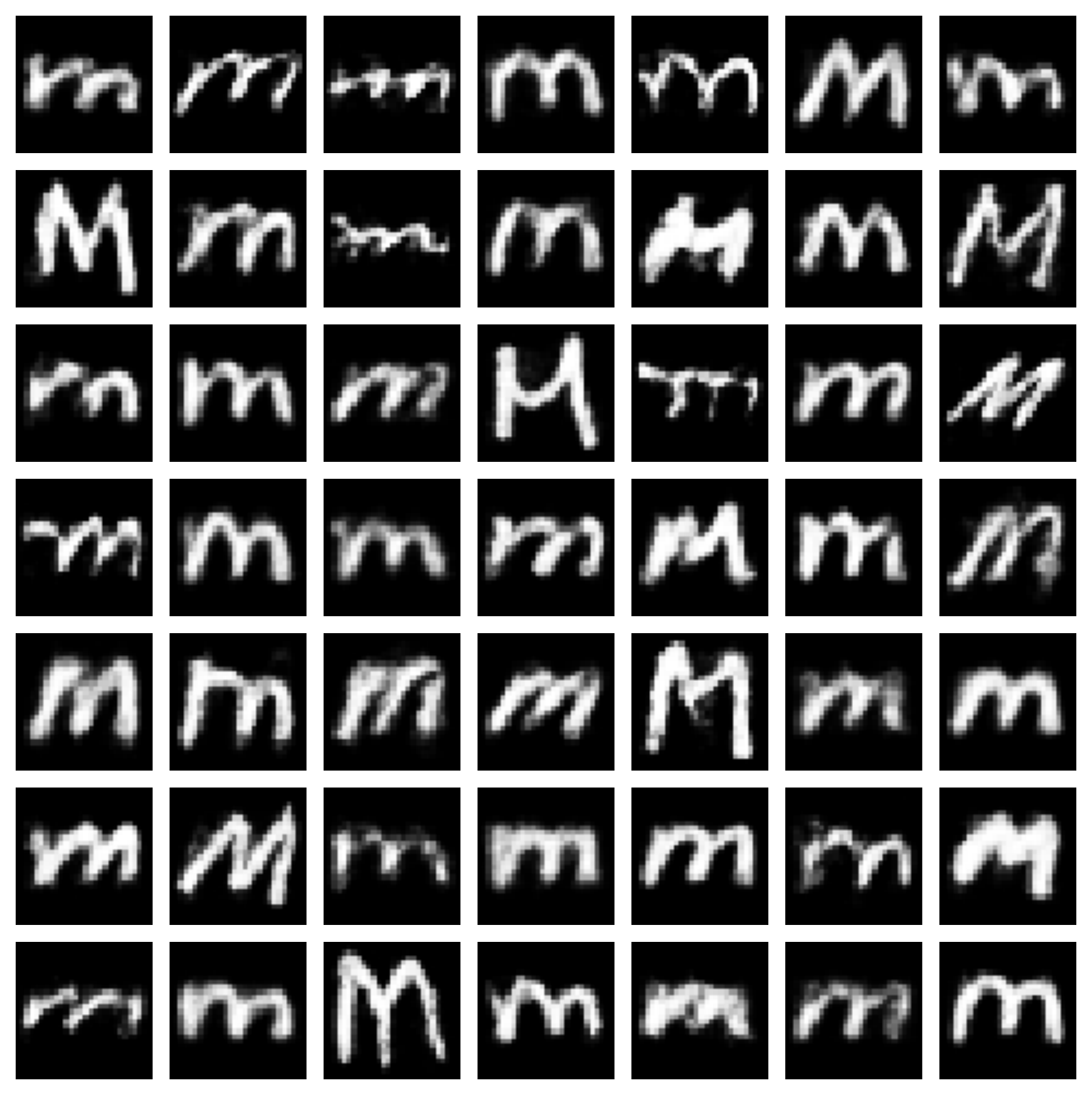}}
  \hspace{0.01in}
  \subfloat{\includegraphics[width=1.25in]{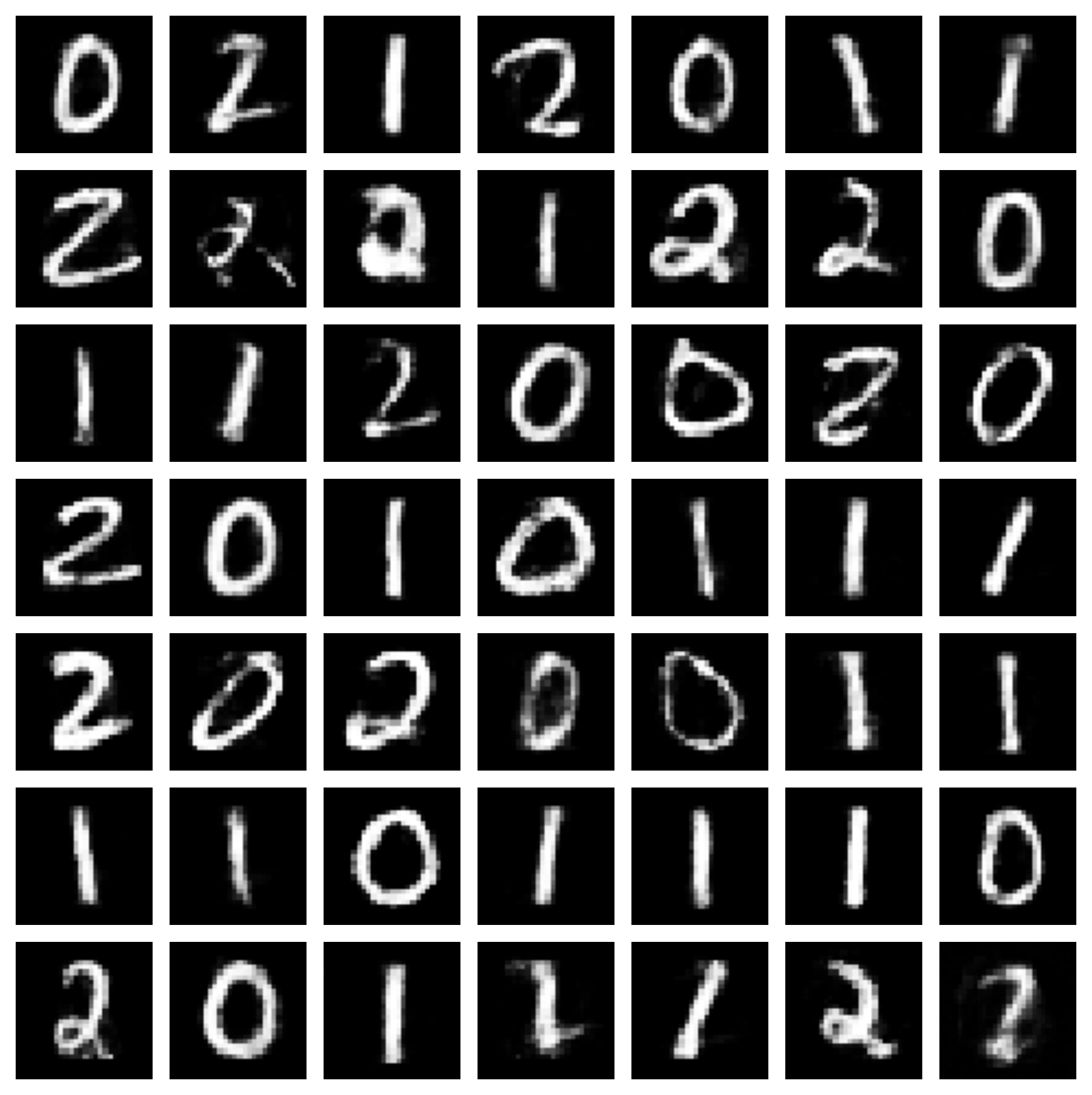}}
  \hspace{0.01in}
  \subfloat{\includegraphics[width=1.25in]{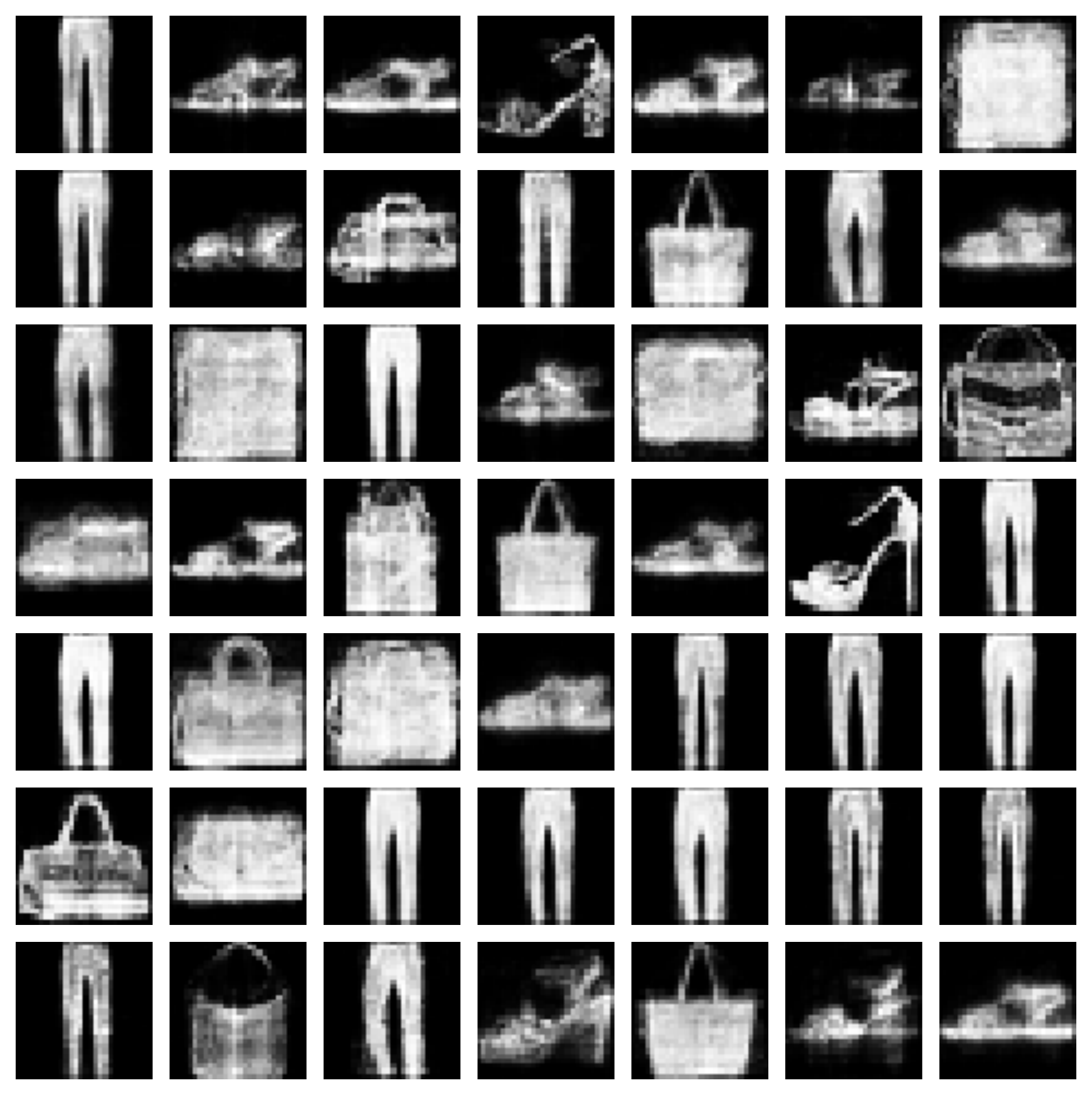}}
  \hspace{0.01in}
  \subfloat{\includegraphics[width=1.25in]{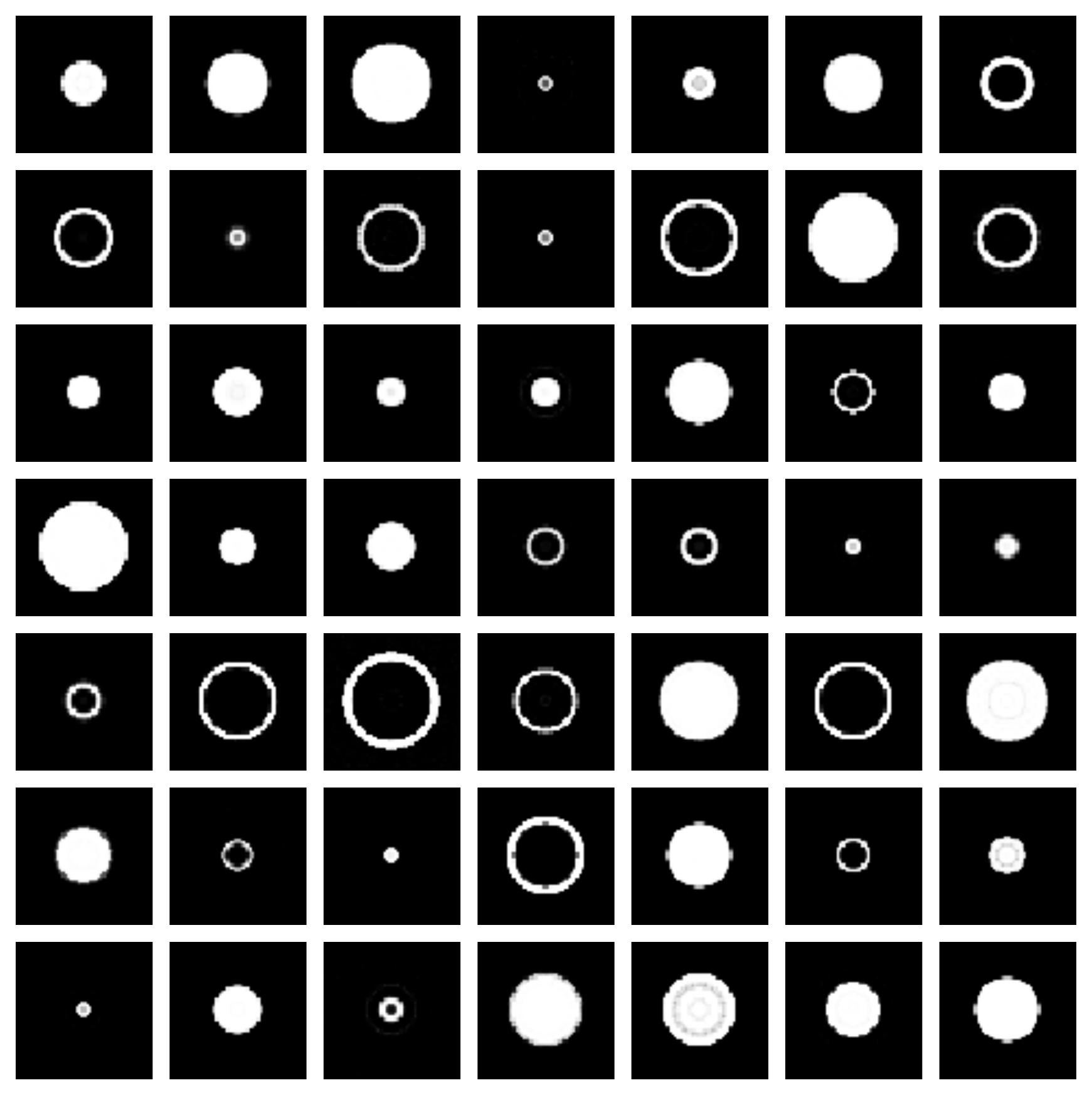}}
  \caption{Comparison of four sampling methods on \emph{reduced} EMNIST (120 letters \emph{M}), \emph{reduced} MNIST, \emph{reduced} FashionMNIST and the synthetic data sets in higher dimensional latent spaces (dimension 10). From top to bottom: 1) samples extracted from the training set; 2) samples generated with a Vanilla VAE and using the prior ($\mathcal{N}(0, I_d$)); 3) from the VAMP prior VAE ; 4) from a RHVAE and the \emph{prior-based} generation scheme and 5) from a RHVAE and using the proposed method. All the models are trained with the same encoder and decoder networks and identical latent space dimension. An early stopping strategy is adopted and consists in stopping training if the ELBO does not improve for 20 epochs. The number of training samples is noted between parenthesis.}
  \label{Fig: app Comparison}
\end{figure*}

Further to the experiments presented in Sec.~4.2.3, we also provide the results of the four classifiers on \emph{reduced} EMNIST and \emph{reduced} Fashion in Fig.~\ref{Fig: app Classifiers rob}. Again, for most classifiers the proposed method either equals or greatly outperform the \emph{baseline}. 

\section{A few More Sample Generation on ADNI}\label{appendix G}

\begin{figure*}[!ht] 
  \centering
  \subfloat{\includegraphics[width=1.02in]{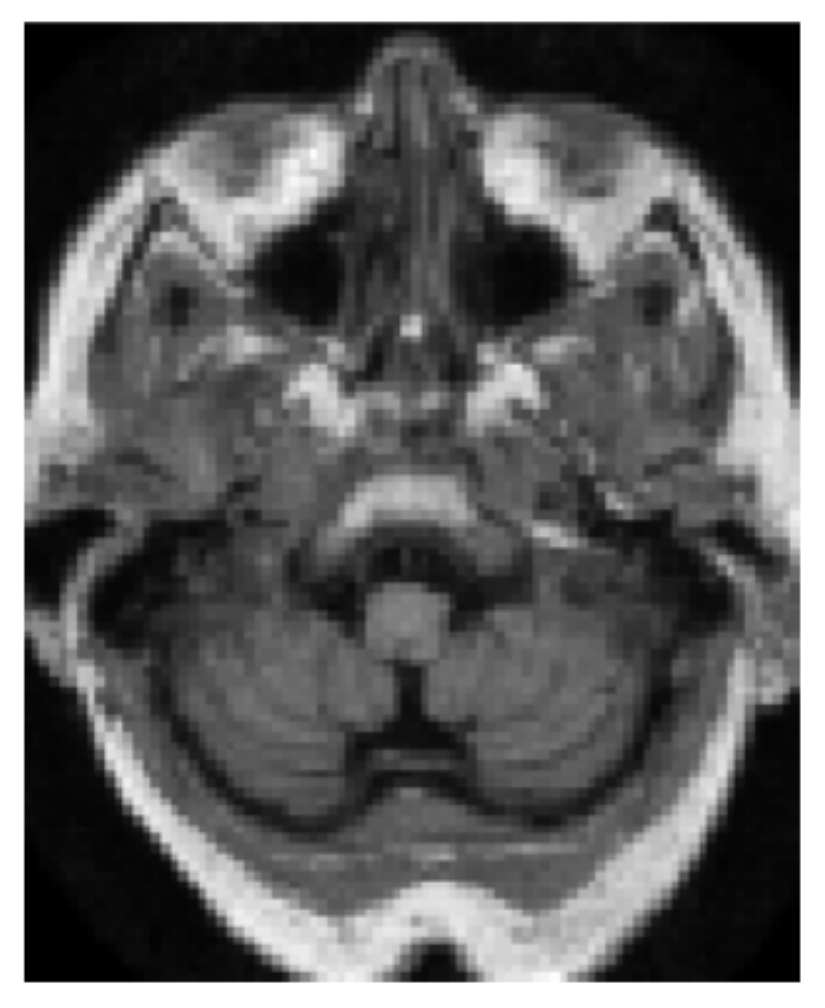}
  }
  \centering
  \subfloat{\includegraphics[width=1.455in]{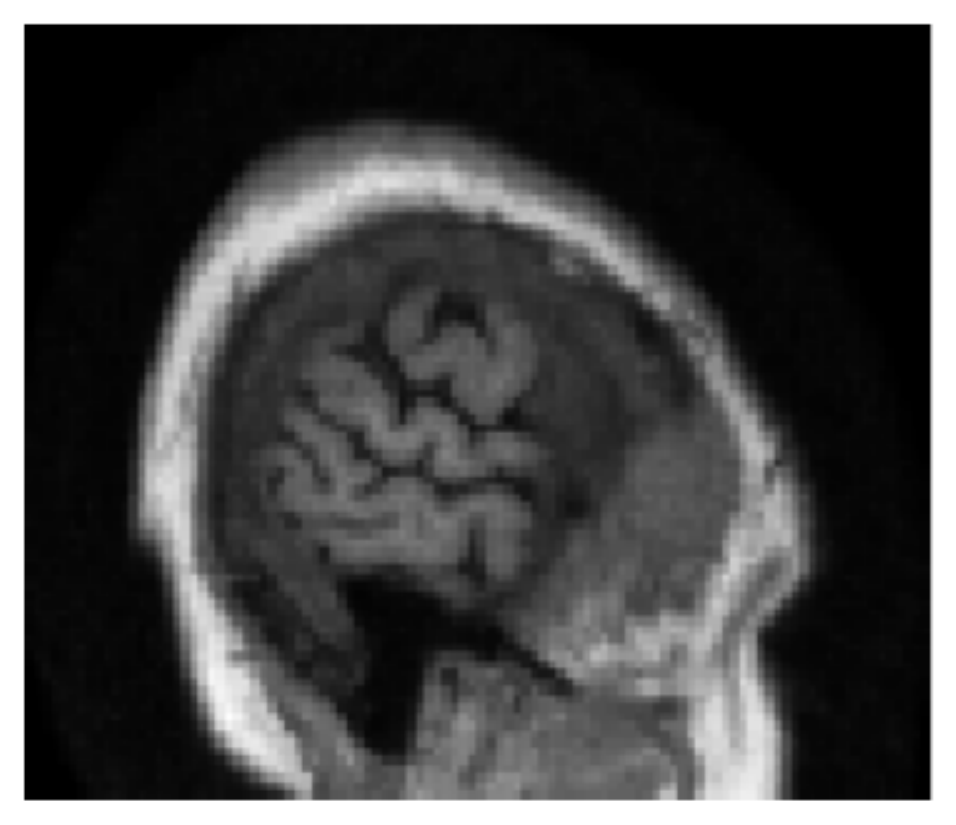}
  }
  \centering
  \subfloat{\includegraphics[width=1.19in]{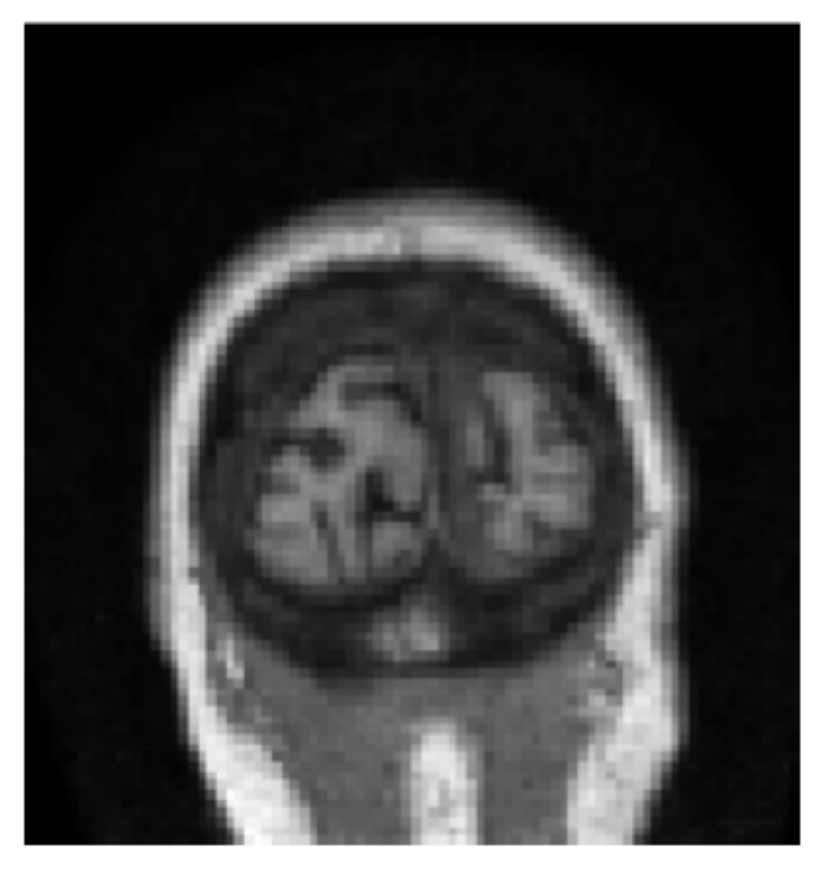}
  }
  \vfil
  \vspace{-1em}
  \centering
  \subfloat{\includegraphics[width=1.02in]{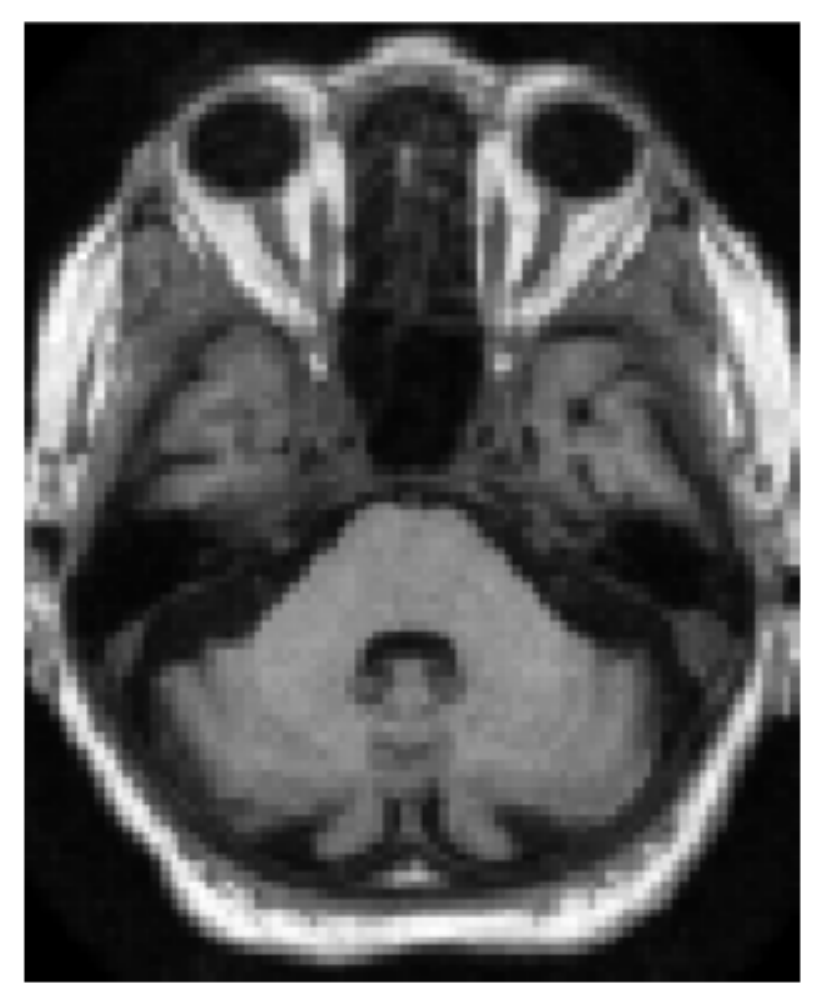}
  }
  \centering
  \subfloat{\includegraphics[width=1.455in]{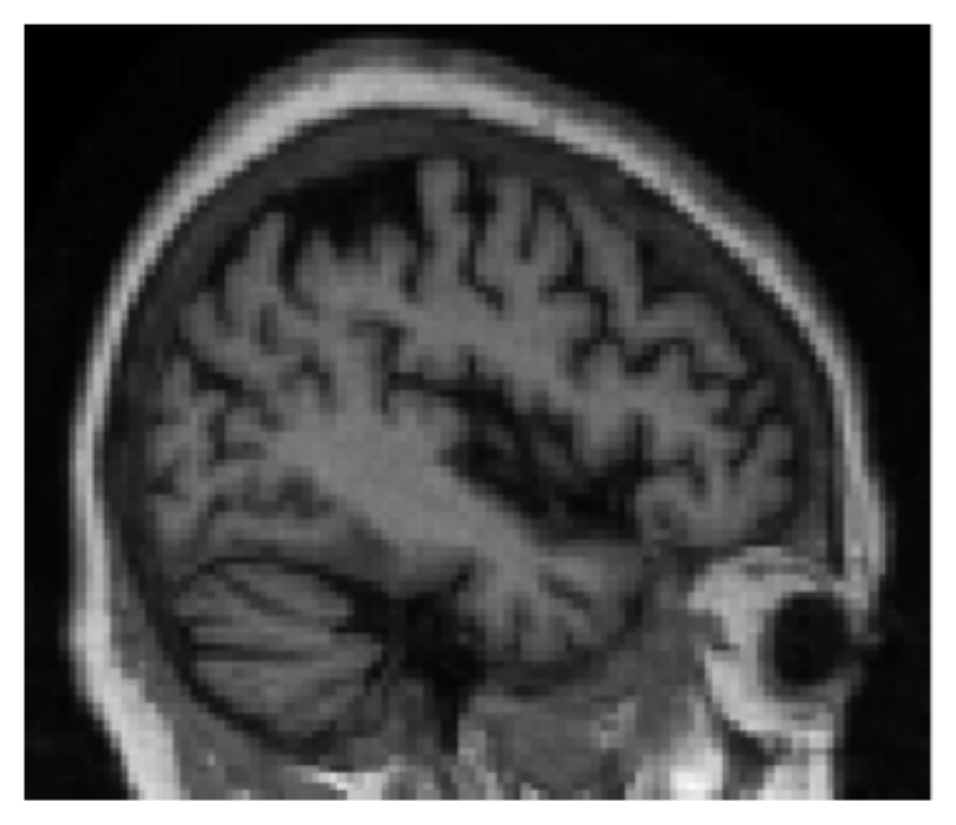}
  }
  \centering
  \subfloat{\includegraphics[width=1.19in]{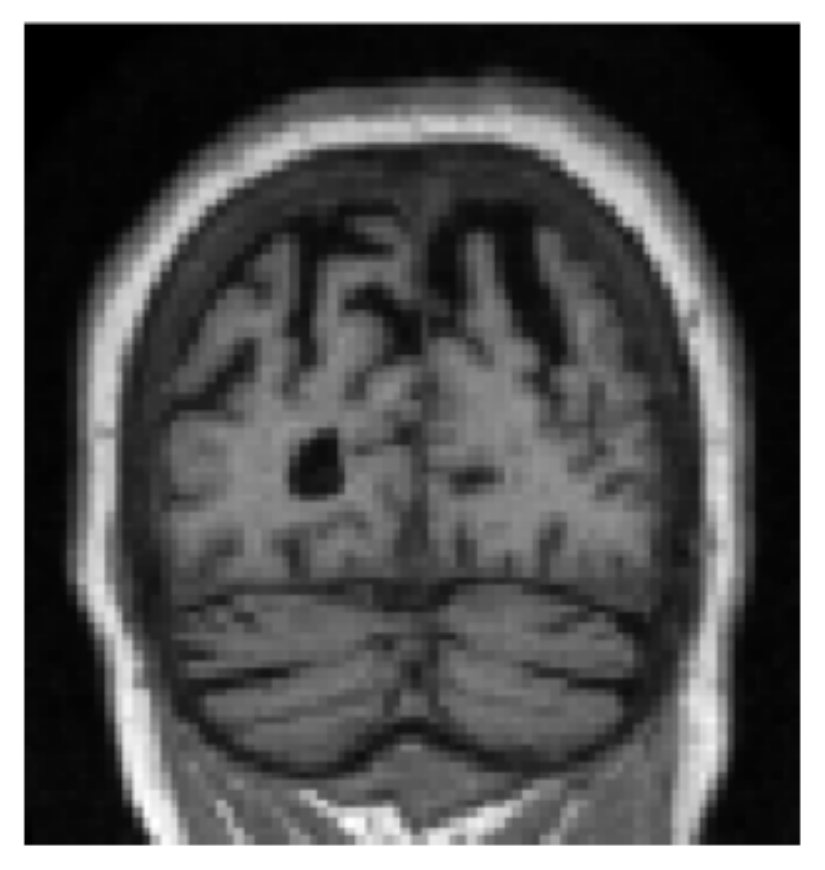}
  }
  \vfil
  \vspace{-1em}
  \centering
  \subfloat{\includegraphics[width=1.02in]{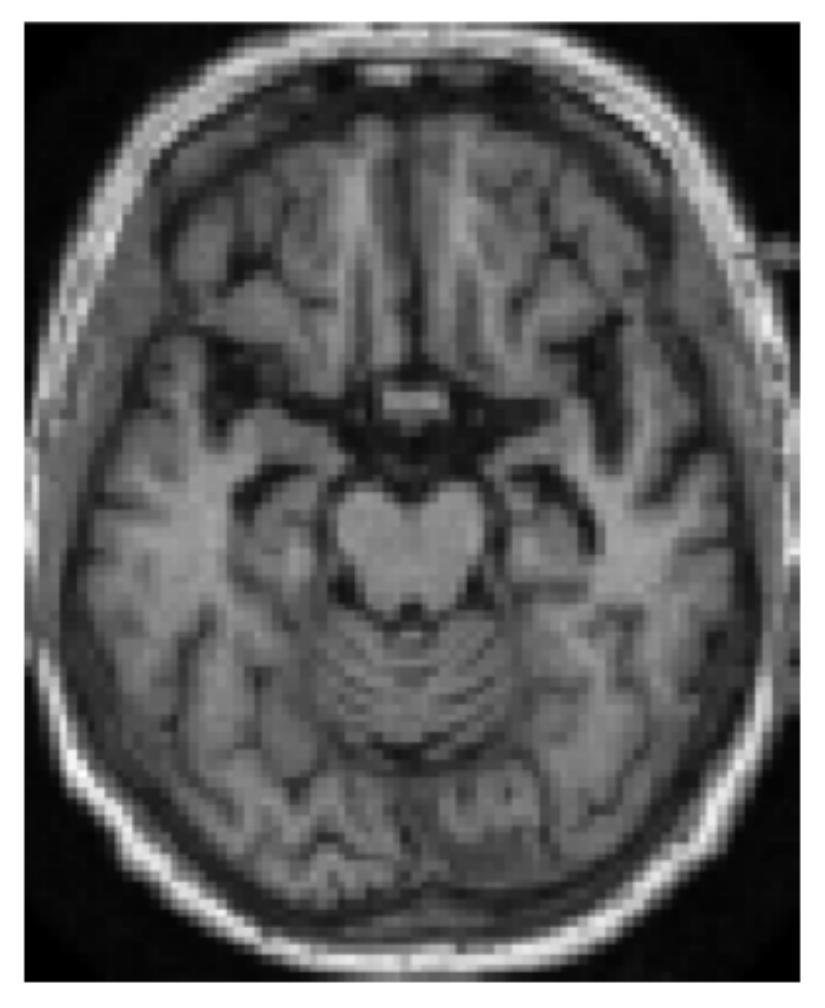}
  }
  \centering
  \subfloat{\includegraphics[width=1.455in]{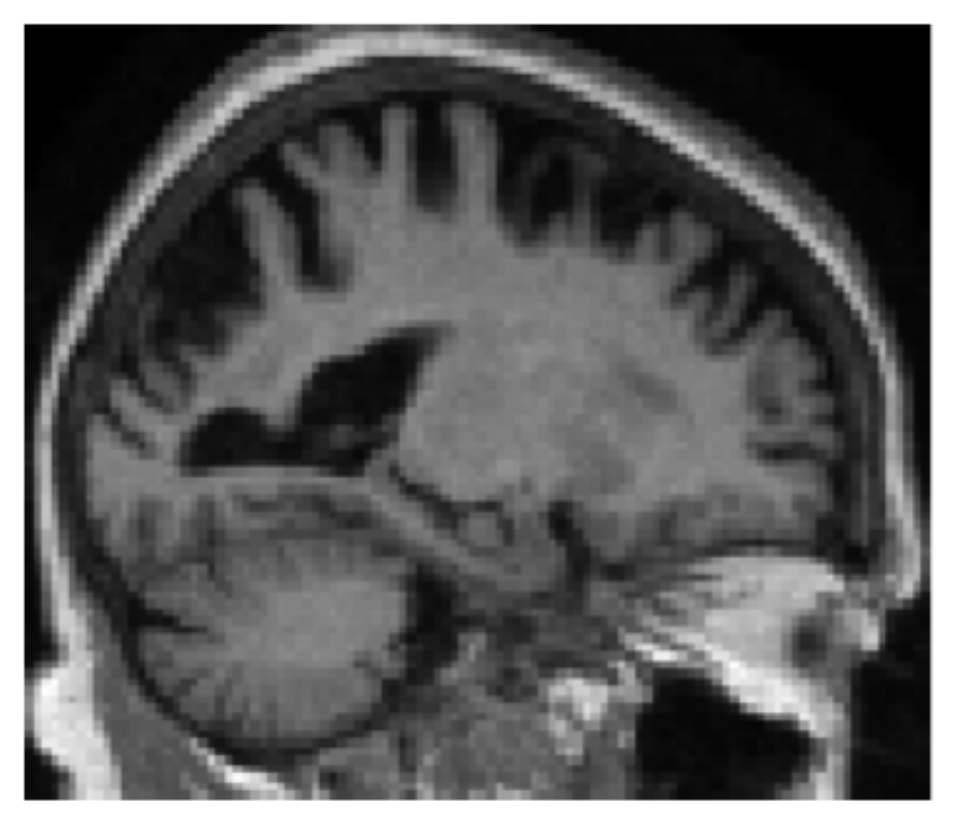}
  }
  \centering
  \subfloat{\includegraphics[width=1.19in]{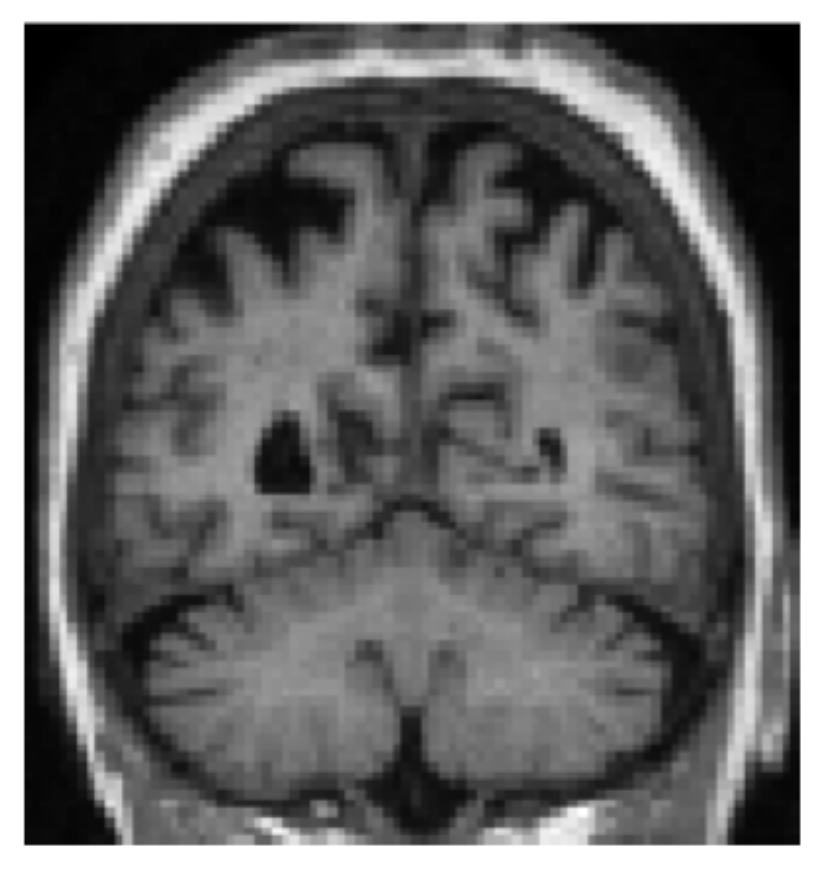}
  }     
  \vfil
  \vspace{-1em}
  \centering
  \subfloat{\includegraphics[width=1.02in]{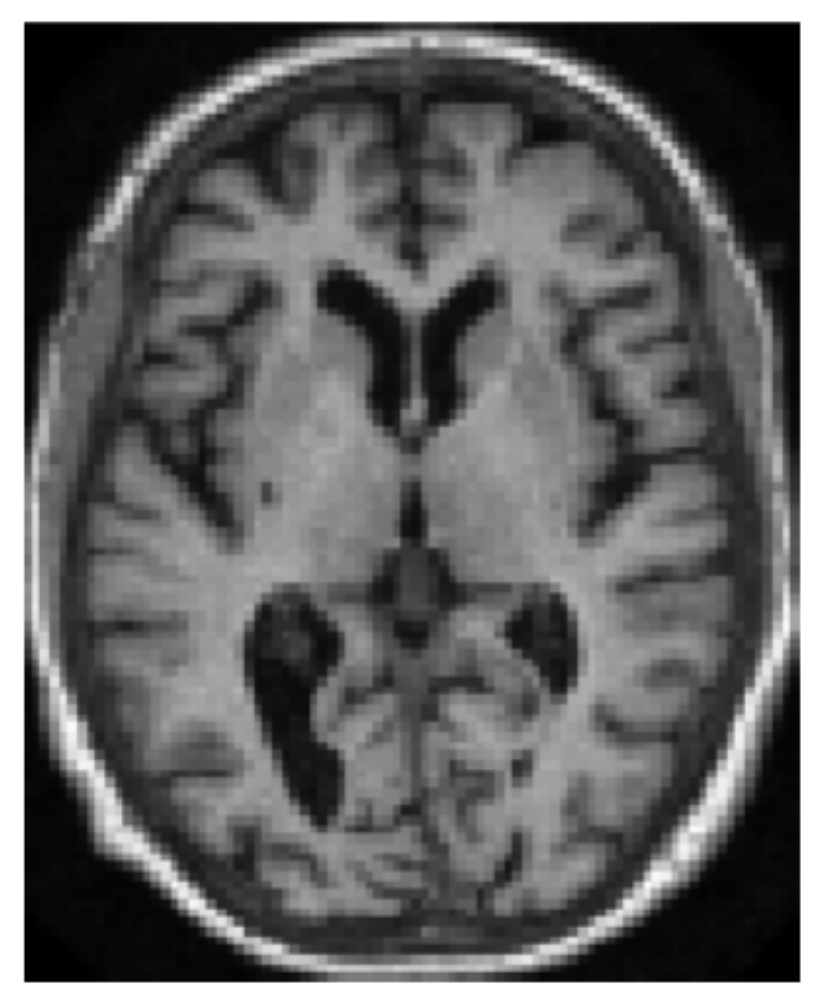}
  }
  \centering
  \subfloat{\includegraphics[width=1.455in]{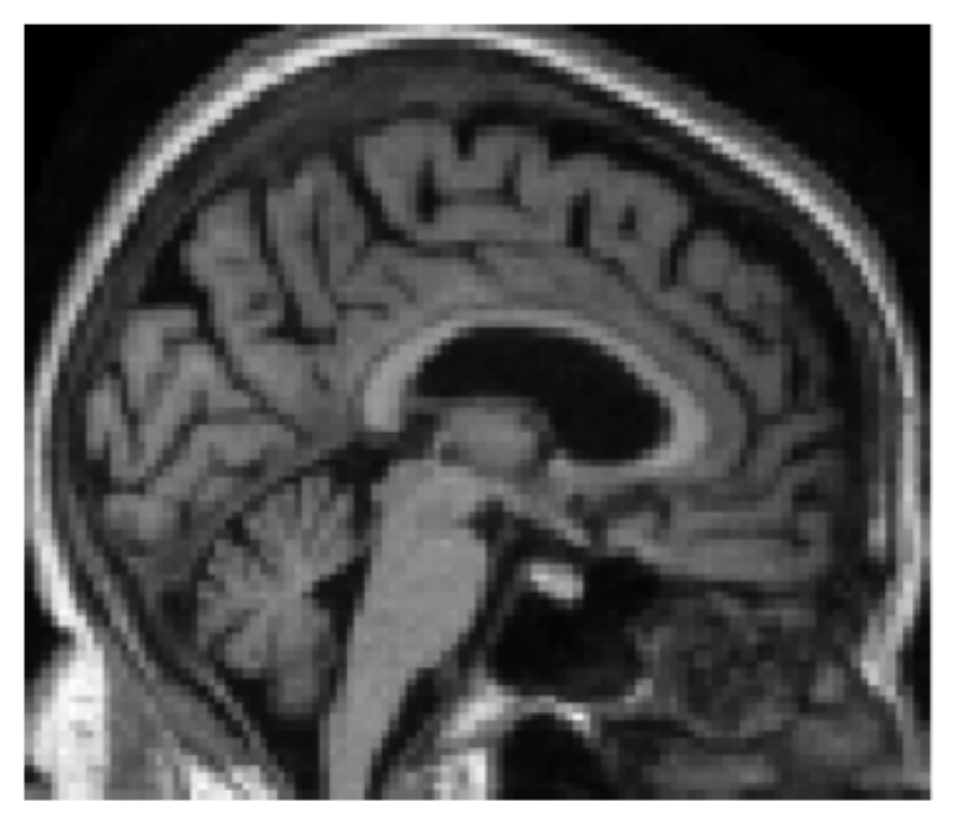}
  }
  \centering
  \subfloat{\includegraphics[width=1.19in]{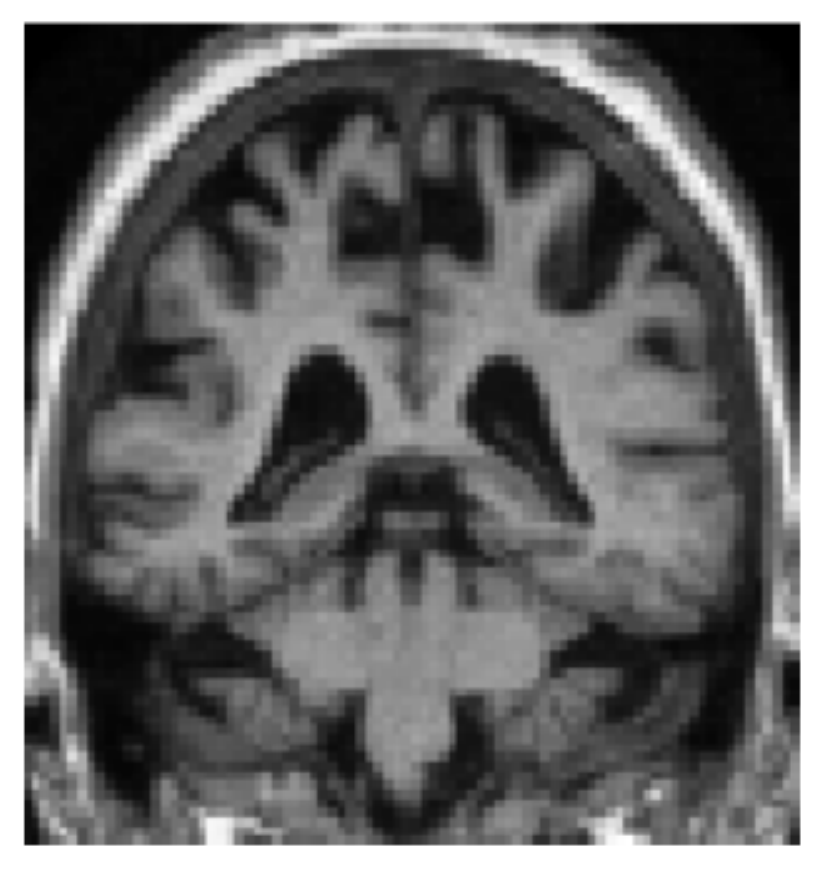}
  }     
  \vfil
  \vspace{-1em}
  \centering
  \subfloat{\includegraphics[width=1.02in]{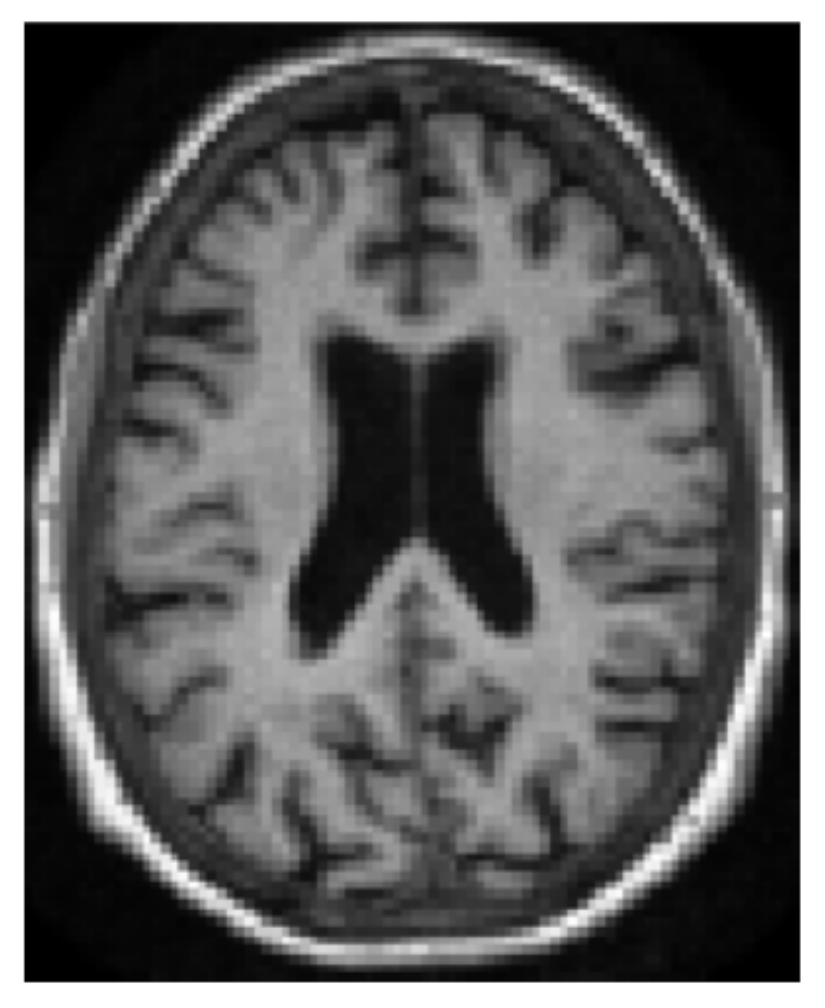}
  }
  \centering
  \subfloat{\includegraphics[width=1.455in]{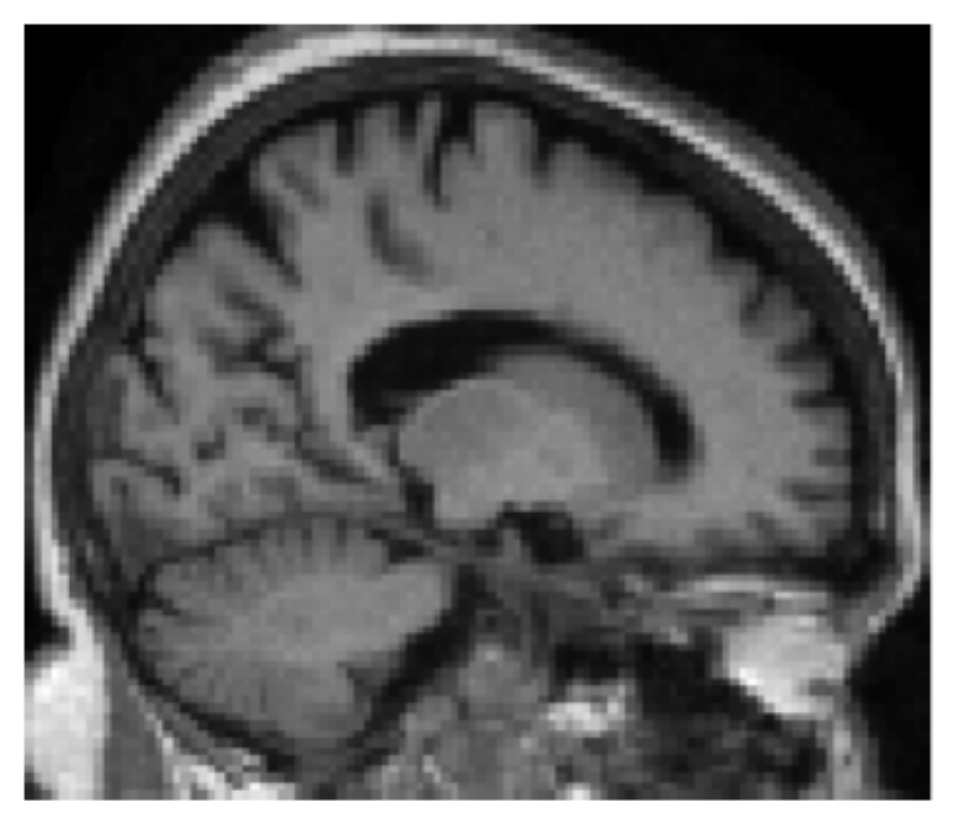}
  }
  \centering
  \subfloat{\includegraphics[width=1.19in]{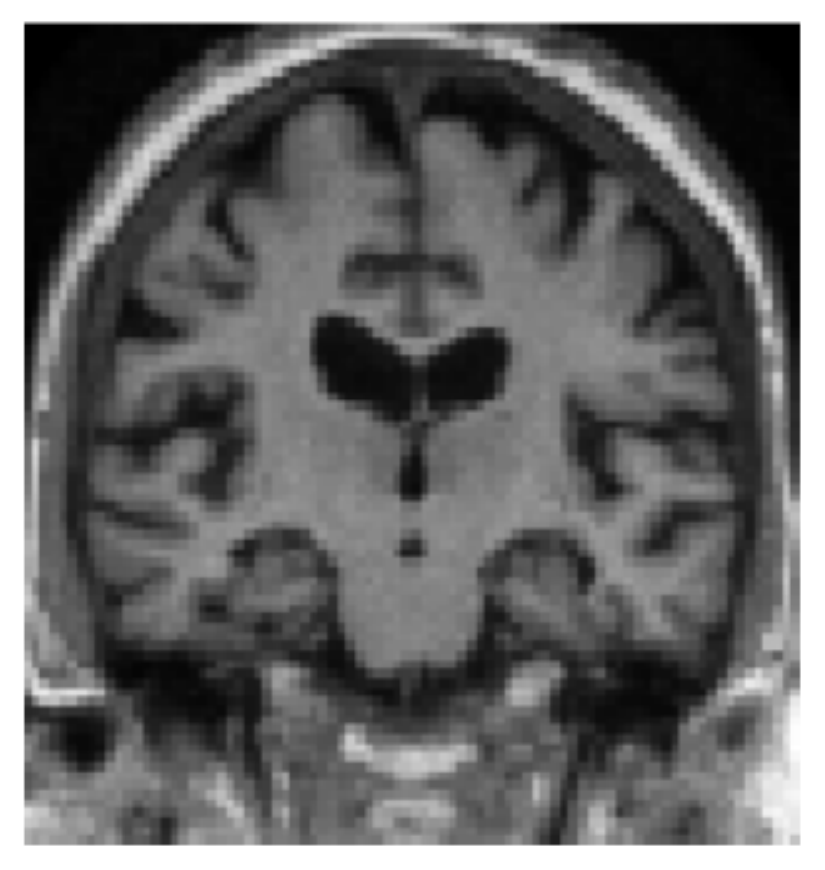}
  }     
  \vfil
  \vspace{-1em}
  \centering
  \subfloat{\includegraphics[width=1.02in]{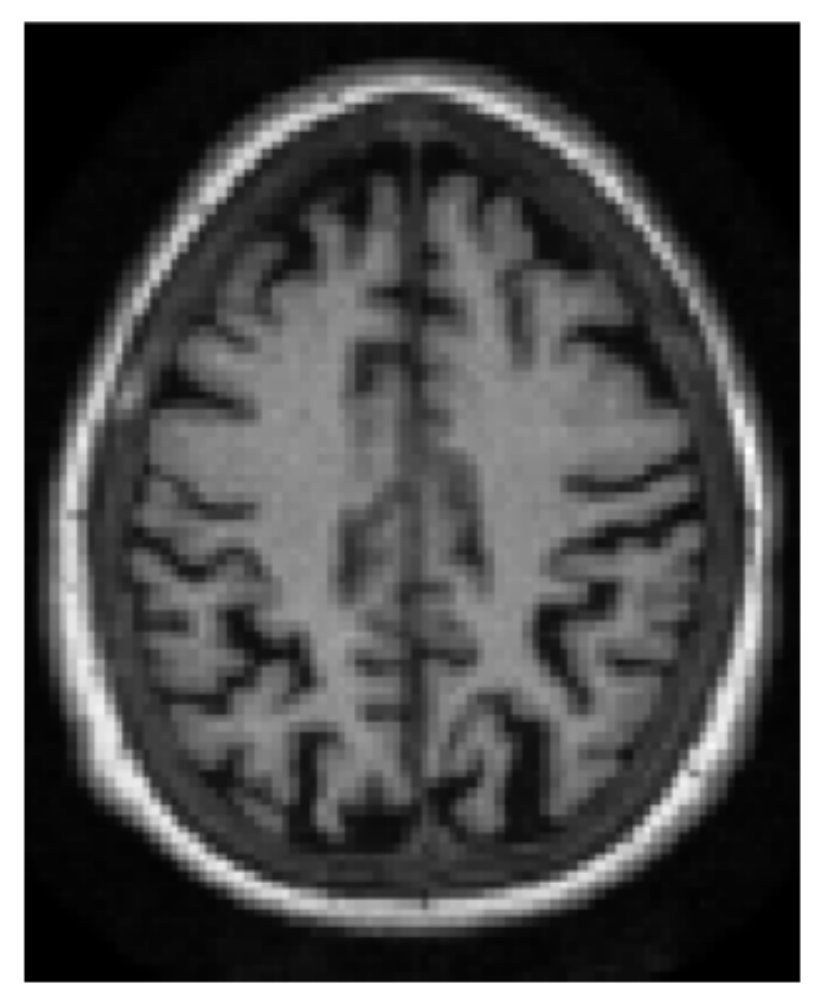}
  }
  \centering
  \subfloat{\includegraphics[width=1.455in]{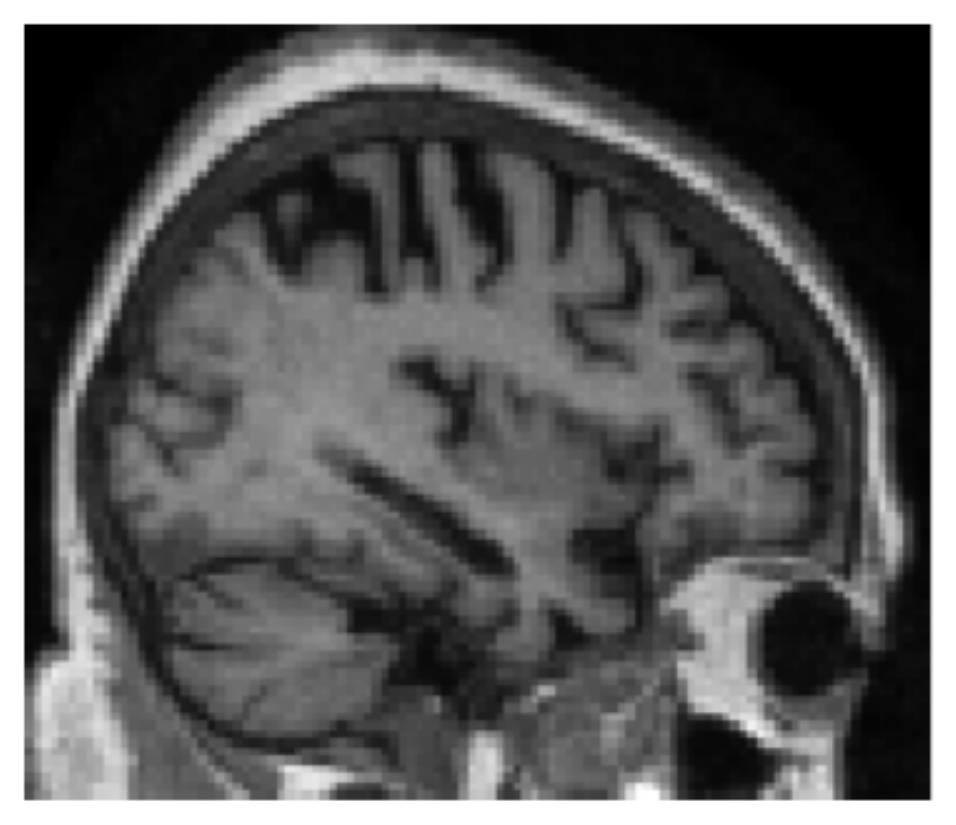}
  }
  \centering
  \subfloat{\includegraphics[width=1.19in]{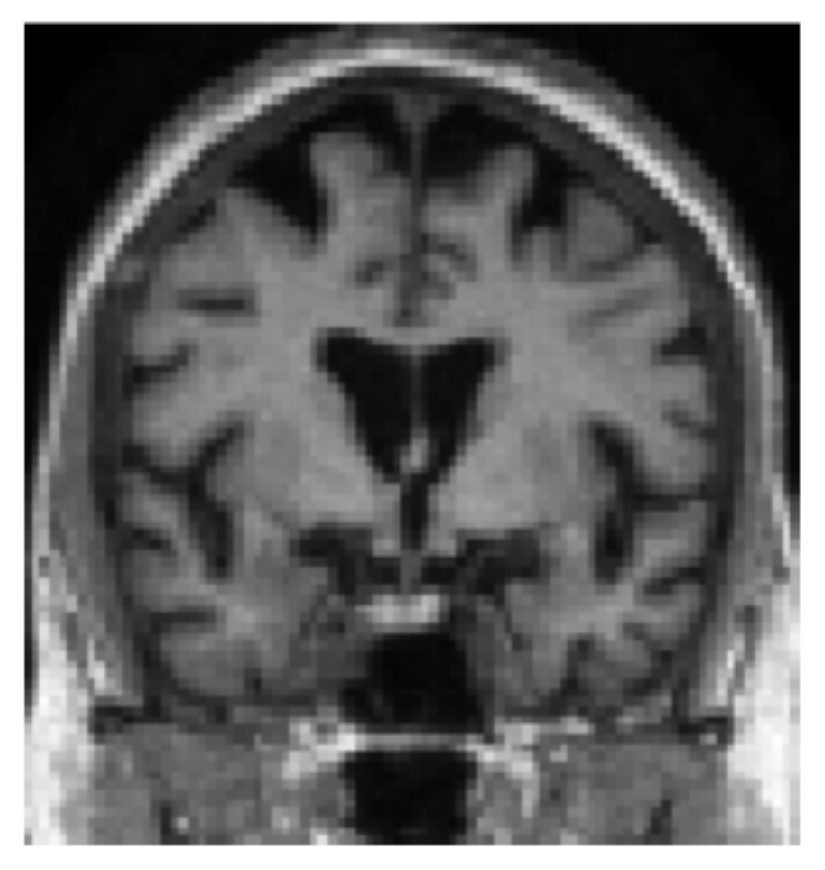}
  }    
  \vfil
  \vspace{-1em}
  \centering
  \subfloat{\includegraphics[width=1.02in]{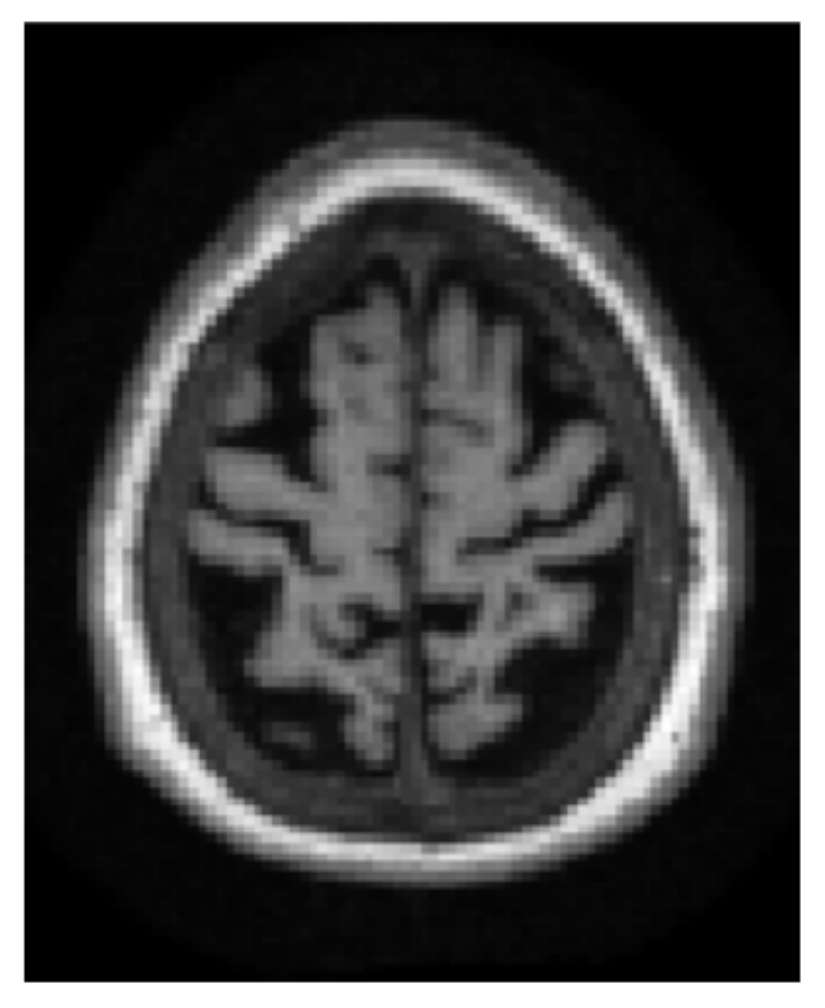}
  }
  \centering
  \subfloat{\includegraphics[width=1.455in]{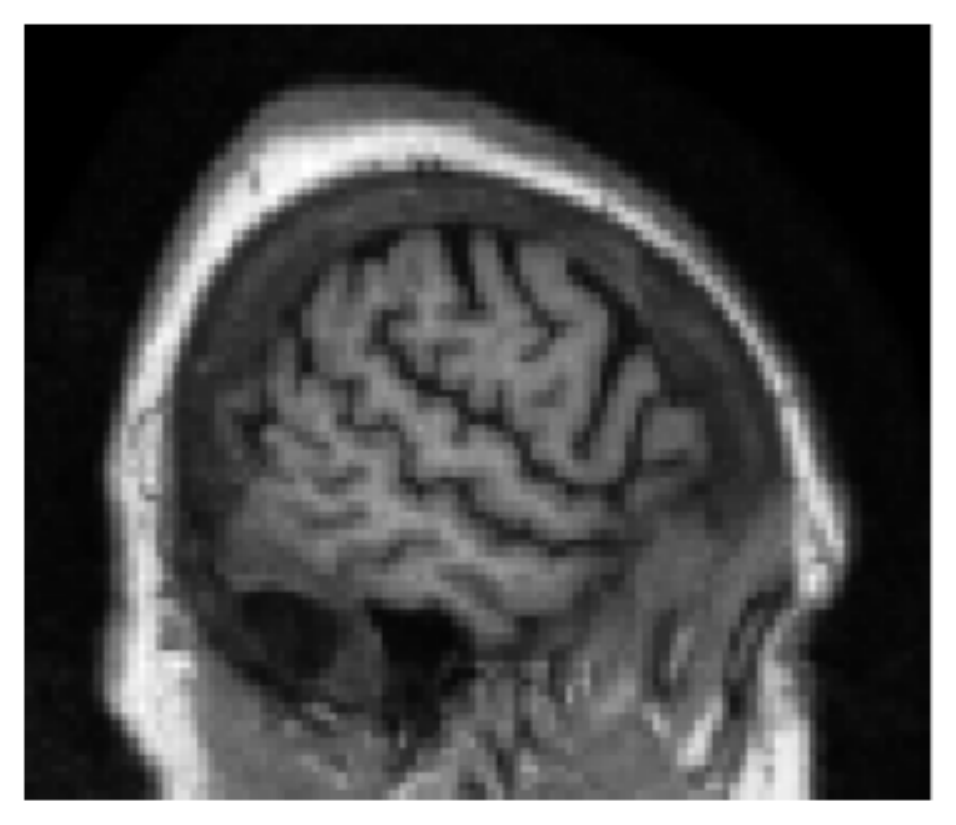}
  }
  \centering
  \subfloat{\includegraphics[width=1.19in]{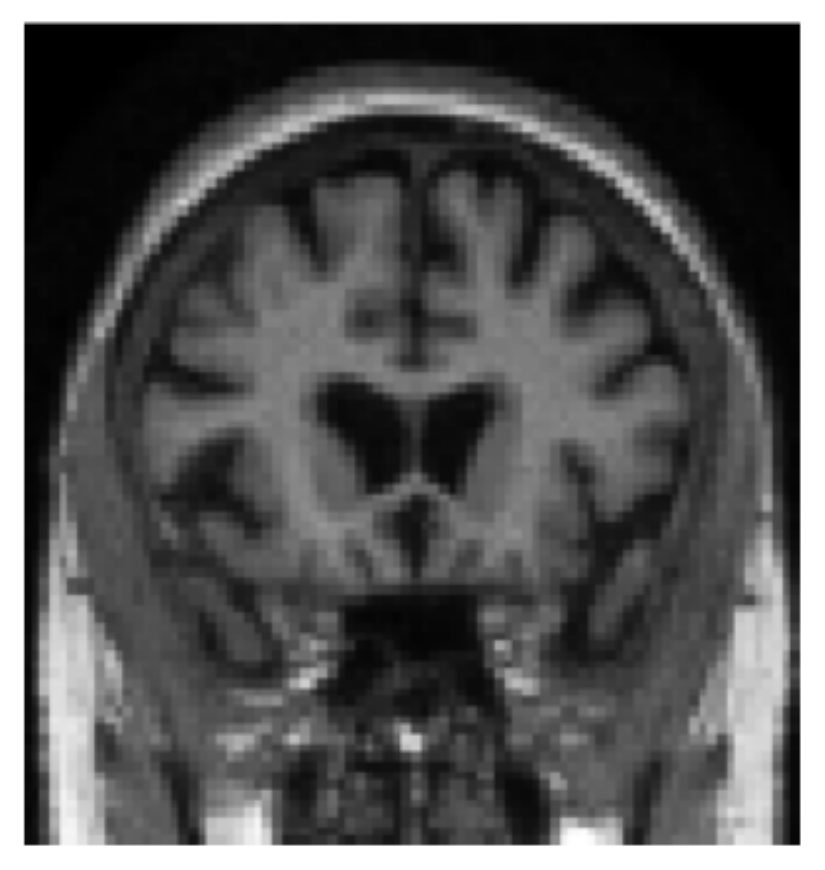}
  }    
  \caption{Several slices of a generated image. The model is trained on the AD class of \emph{train-50} (\emph{i.e.} 50 images of AD patients).}
  \label{Fig: app ADNI generation}
  \end{figure*}

\begin{figure*}[!ht] 
    \centering
    \subfloat{\includegraphics[width=1.5in]{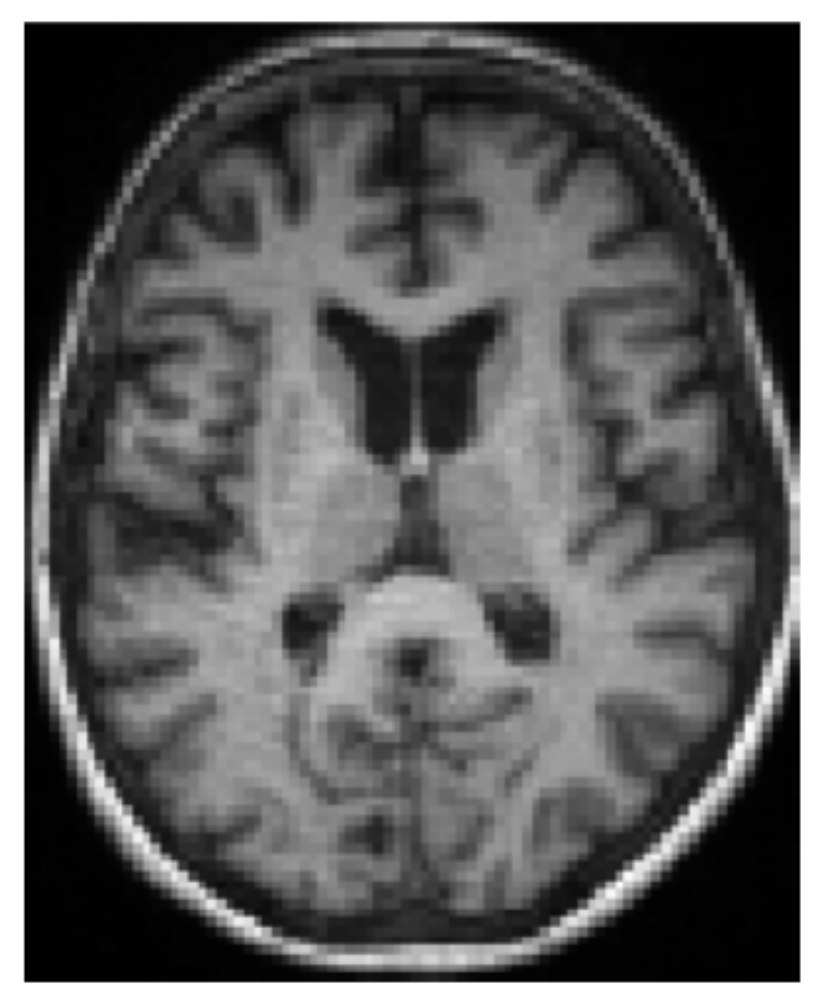}
    }
    \centering
    \subfloat{\includegraphics[width=1.5in]{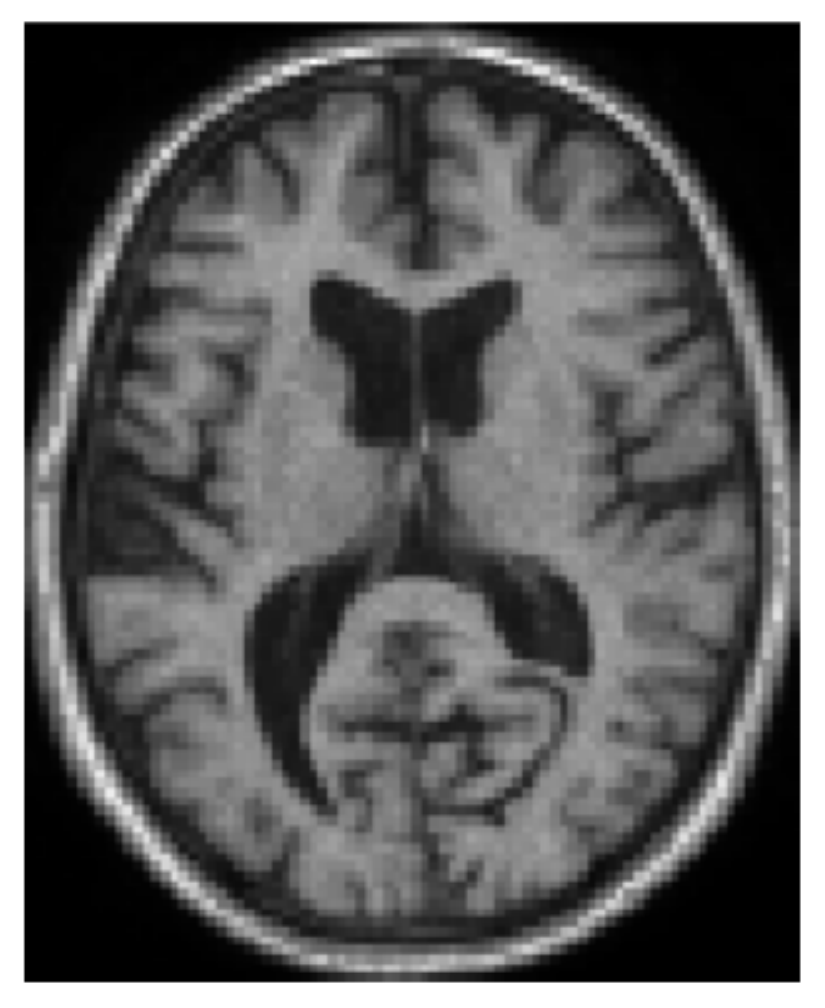}
    }
    \centering
    \hfil
    \subfloat{\includegraphics[width=1.5in]{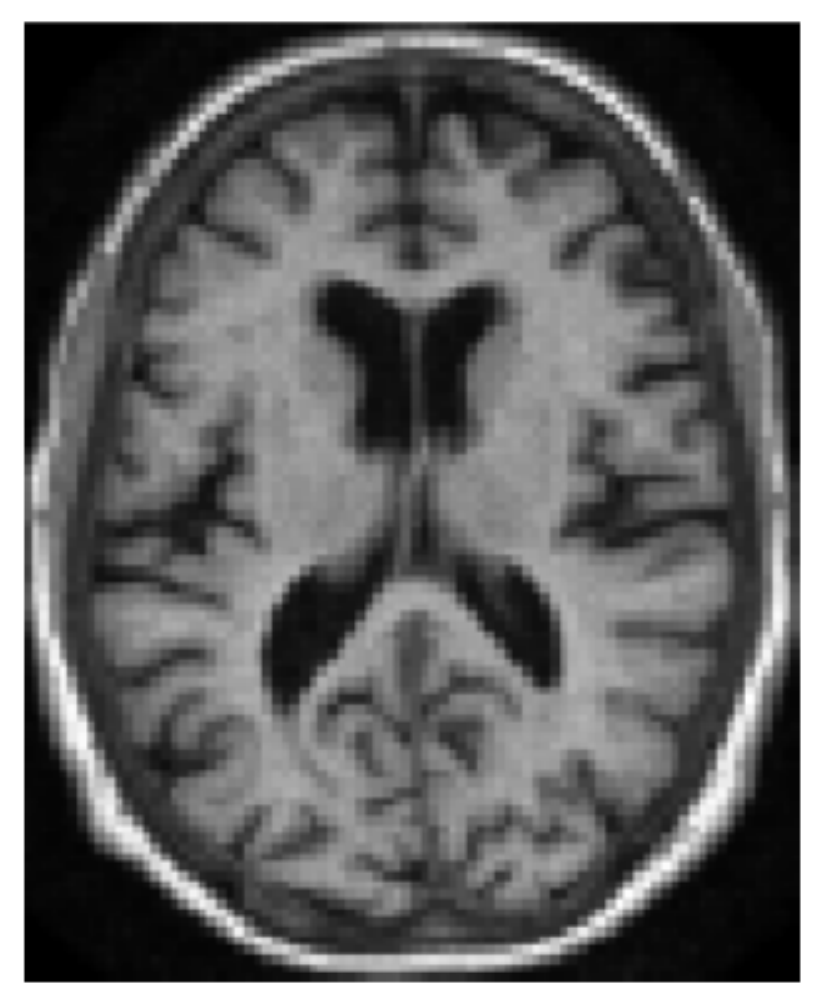}
    }
    \centering
    \subfloat{\includegraphics[width=1.5in]{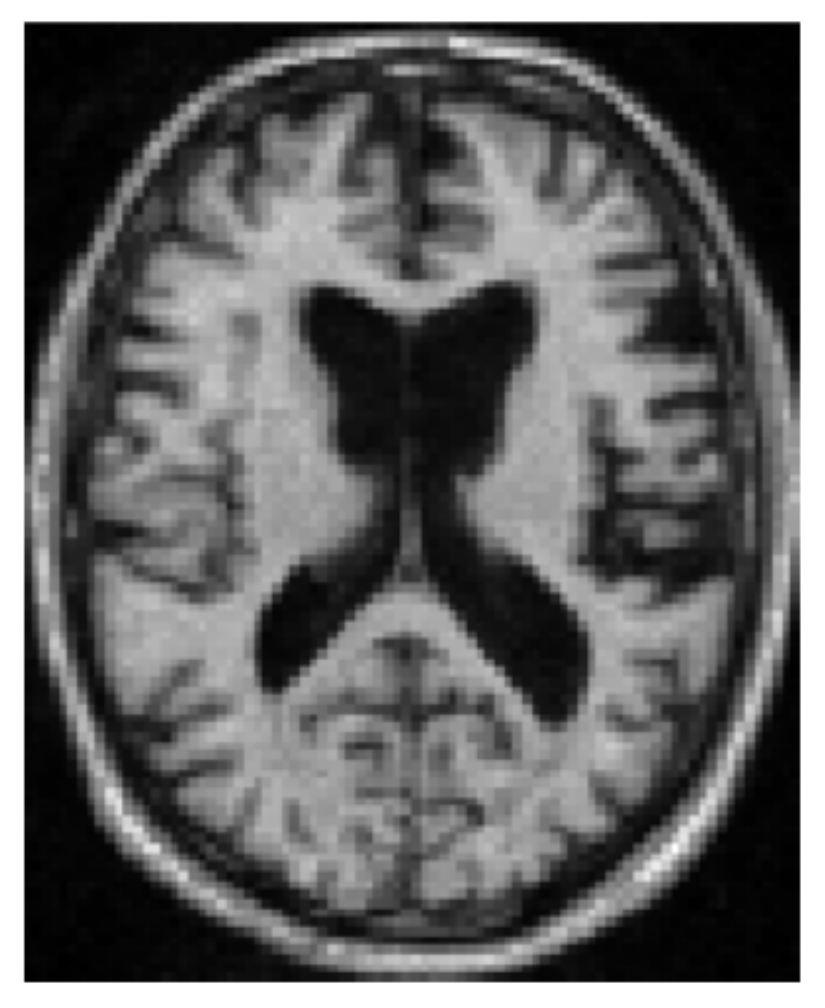}
    }
    \vfil
    \centering
    \subfloat{\includegraphics[width=1.5in]{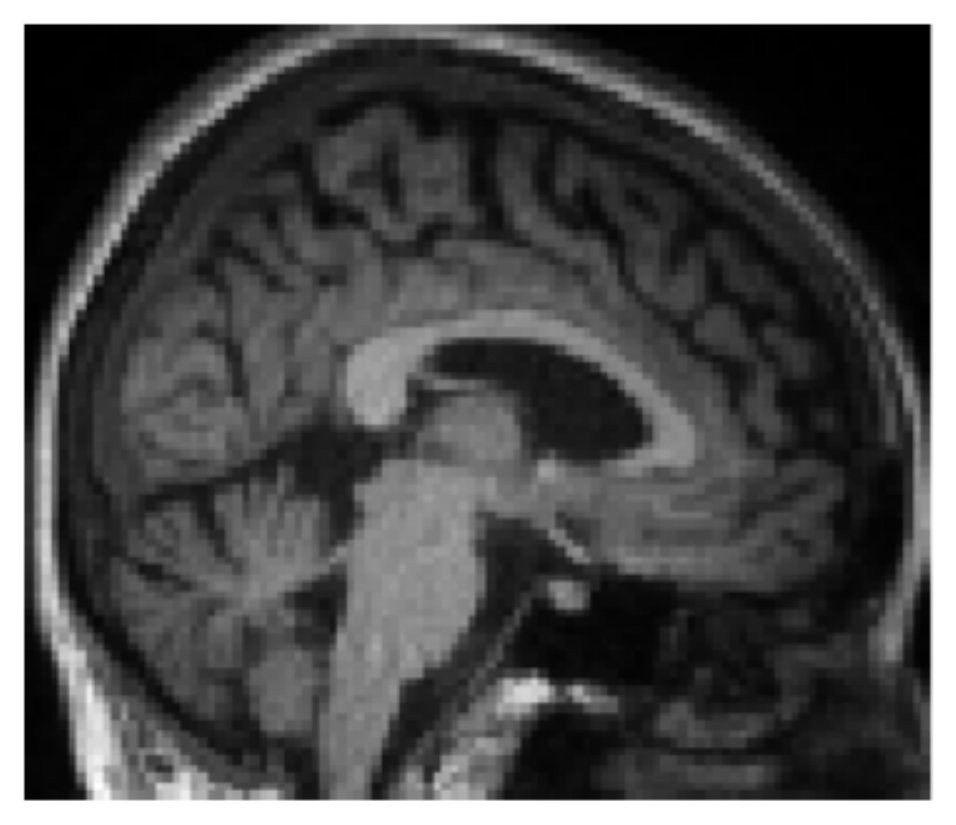}
    }
    \centering
    \subfloat{\includegraphics[width=1.5in]{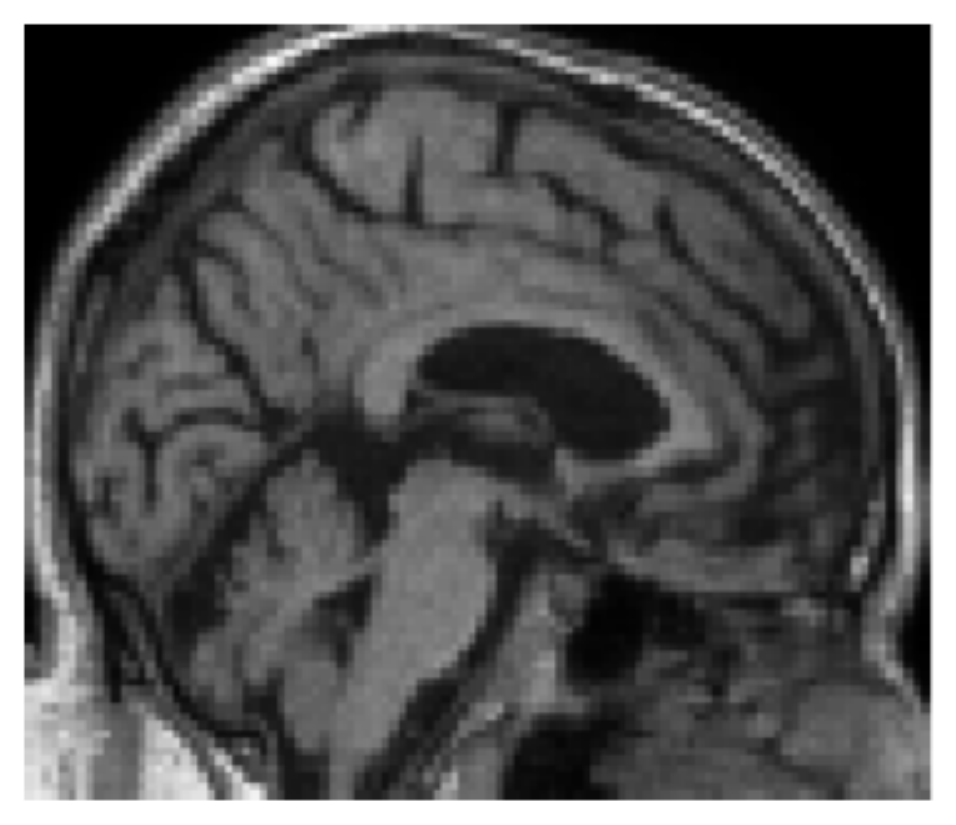}
    }
    \hfil
    \centering
    \subfloat{\includegraphics[width=1.5in]{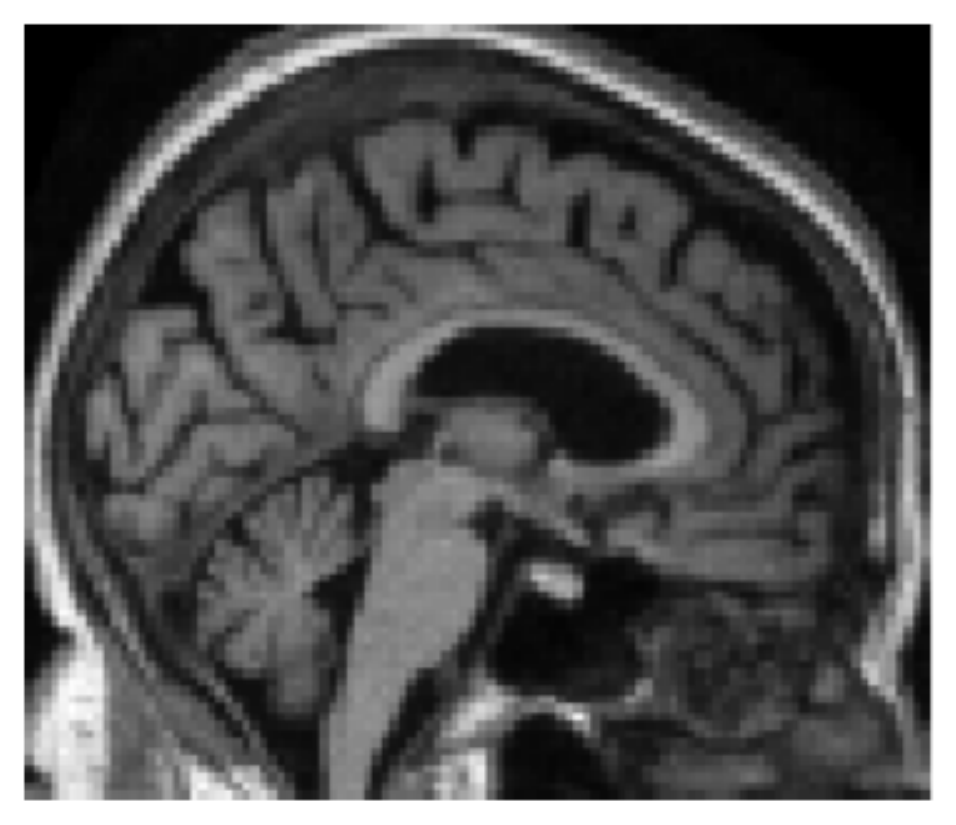}
    }
    \centering
    \subfloat{\includegraphics[width=1.5in]{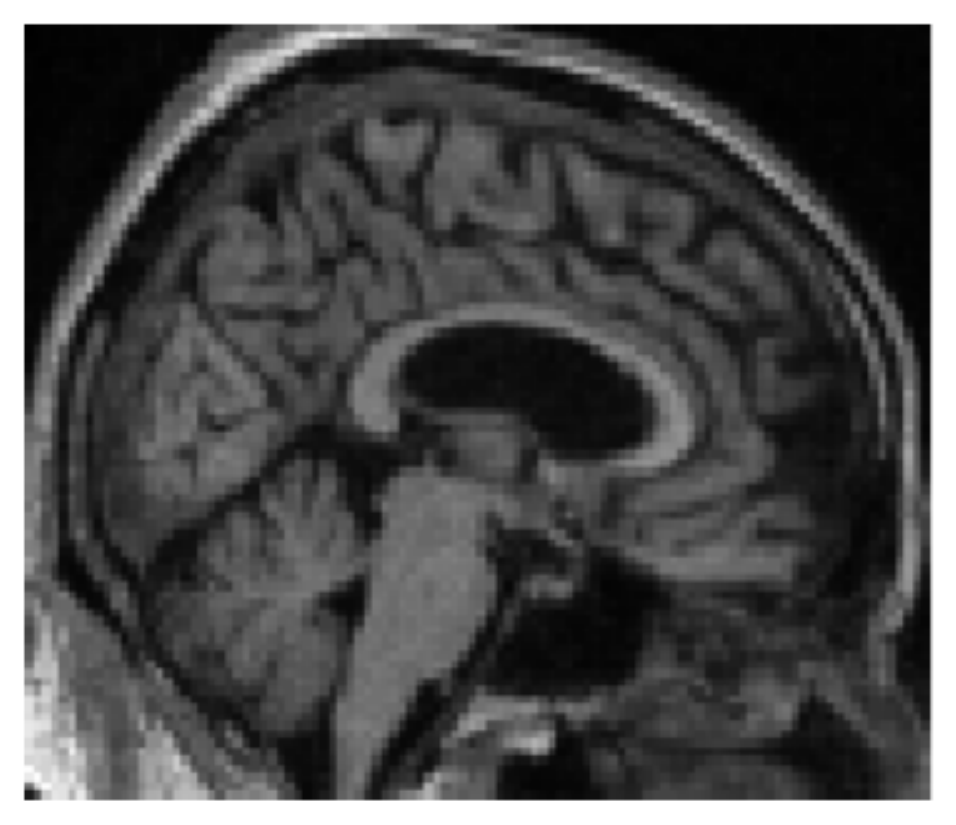}
    }   
    \vfil
    \centering
    \subfloat{\includegraphics[width=1.5in]{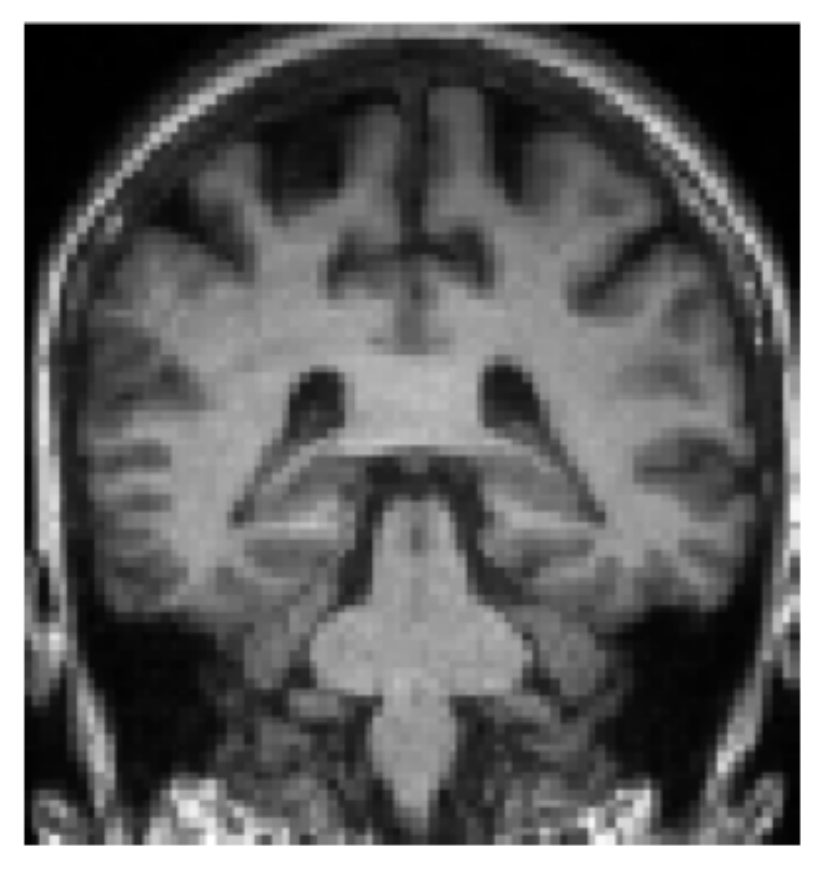}
    }
    \centering
    \subfloat{\includegraphics[width=1.5in]{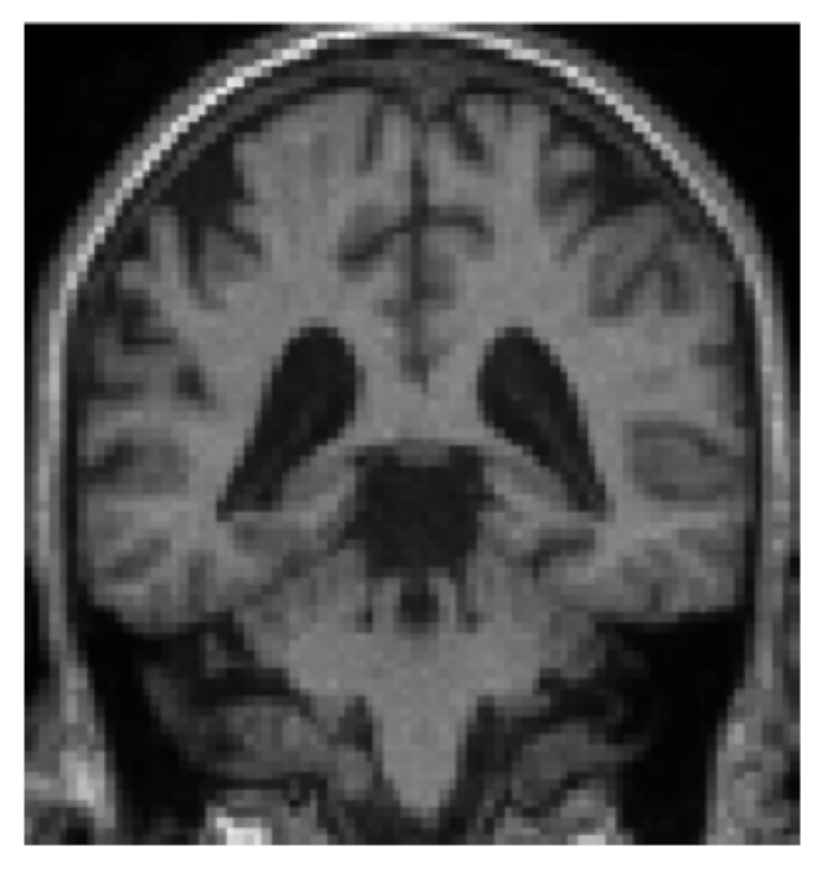}
    }
    \hfil
    \centering
    \subfloat{\includegraphics[width=1.5in]{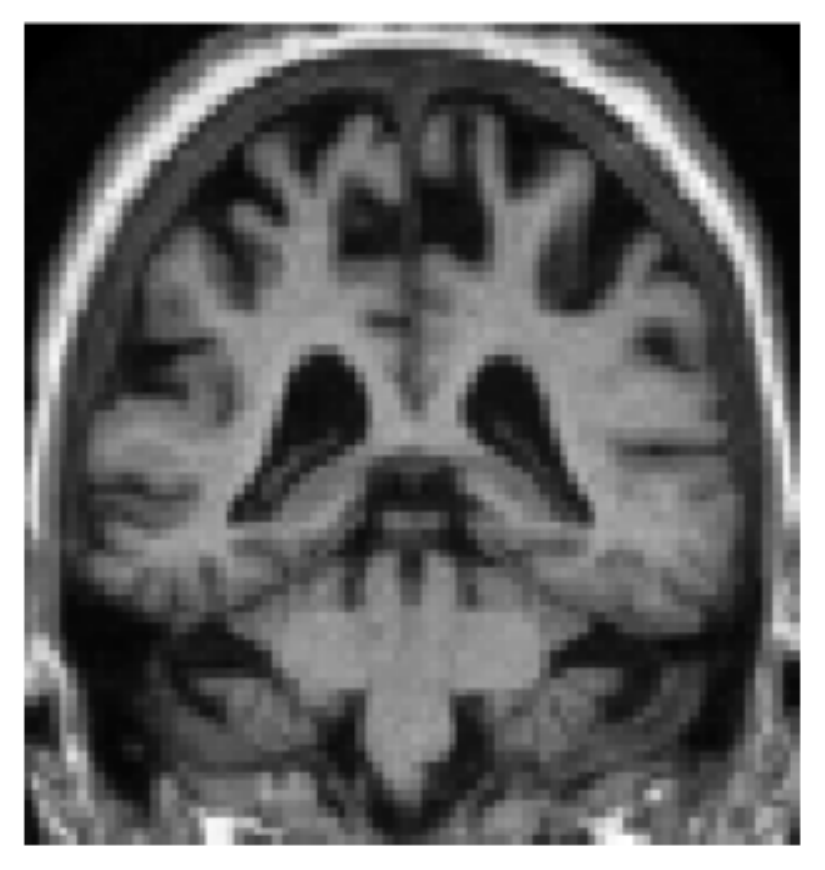}
    }
    \centering
    \subfloat{\includegraphics[width=1.5in]{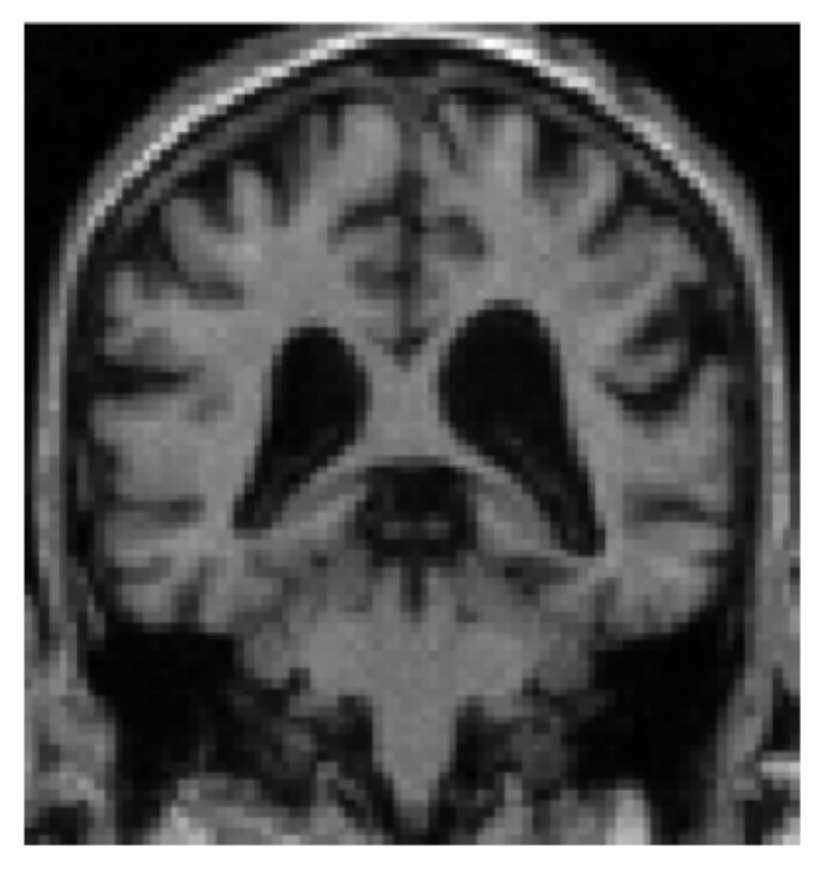}
    }        
    \caption{Images generated by our method when trained on \emph{train-50}. \emph{Left}: CN generated patients. \emph{Right}: AD generated patients.}
    \label{Fig: app ADNI train-50 generation}
    \end{figure*}

In this section, we first provide several slices of a 3D image generated by our model. The model is trained on the class AD of \emph{train-50} (\emph{i.e.} on 50 MRI of patient having been diagnosed with Alzheimer's disease). The generated image is presented in Fig.~\ref{Fig: app ADNI generation}. We also present in Fig.~\ref{Fig: app ADNI train-50 generation}, four generated patients for a model trained on \emph{train-50}. The two left images show \emph{cognitively normal} generated patients while the rightmost images represent AD generated patients.

\section{The Intruders: Answers to Fig.~7}\label{appendix H}

In Fig.~7 of the paper, the synthetic samples are the leftmost and rightmost images while the \emph{real} patients are in the middle. The model is trained on the class AD of \emph{train-full} \emph{i.e.} 210 images.

\section{Complementary Results on Medical Images}\label{appendix I}

The comprehensive results for the classification task on MRIs are added in Tables \ref{table: app train-50_Conv5F3} to \ref{table: app train-full_random}. As observed on the \emph{toy} examples, the proposed model is again able to produce meaningful synthetic samples since each CNN outperforms greatly the \emph{baseline} (\emph{i.e.} the real training data) either on \emph{train-50} or \emph{train-full}. The fact that classification performances on AIBL (which is never used for training) are better for a classifier trained on synthetic data than on the \emph{baseline} shows again that the generative model does not overfit the training data (coming from ADNI) but rather produces samples that are also relevant for another database. Moreover, we again see that the classifier is able to outperform the \emph{baseline} with only synthetic samples proof of good generalization power.

\begin{table*}[ht]
 \caption{Mean test performance of the 20 runs trained on \textit{train-50} with the baseline hyperparameters}
 \label{table: app train-50_Conv5F3}
%\resizebox{\textwidth}{!}{%
\centering
\scriptsize
  \begin{tabular}{|c|c||c|c|c||c|c|c|}
    \cline{3-8}
    \multicolumn{2}{c|}{} & \multicolumn{3}{c||}{ADNI} & \multicolumn{3}{c|}{AIBL} \\
    \hline
    \multirow{2}{*}{image type} & synthetic & \multirow{2}{*}{sensitivity} & \multirow{2}{*}{specificity} & balanced & \multirow{2}{*}{sensitivity} & \multirow{2}{*}{specificity} & balanced \\
    & images & & & accuracy & & & accuracy \\
    \hline
    \hline
    real & - & $70.3\pm12.2$ & $62.4\pm11.5$ & $66.3\pm2.4$ & $60.7\pm13.7$ & $73.8\pm7.2$ & $67.2\pm4.1$ \\
    real (high-resolution) & - & $78.5\pm9.4$ & $57.4\pm8.8$ & $67.9\pm2.3$ & $57.2\pm11.2$ & $75.8\pm7.0$ & $66.5\pm3.0$ \\
    \hline
    synthetic & 500 & $72.4\pm6.4$ & $65.6\pm8.1$ & $69.0\pm1.9$ & $56.6\pm9.9$ & $80.0\pm5.3$ & $68.3\pm3.0$ \\
    synthetic & 1000 & $75.0\pm6.2$ & $65.6\pm7.4$ & $70.3\pm2.0$ & $62.7\pm9.7$ & $78.8\pm5.3$ & $70.8\pm3.5$ \\
    synthetic & 2000 & $71.4\pm6.6$ & $70.4\pm6.6$ & $70.9\pm3.0$ & $62.1\pm8.8$ & $80.5\pm4.7$ & $71.3\pm3.6$ \\
    synthetic & 3000 & $70.6\pm5.2$ & $\boldsymbol{73.8\pm4.2}$ & $72.2\pm1.4$ & $65.7\pm6.9$ & $80.5\pm4.6$ & $73.1\pm1.8$ \\
    synthetic & 5000 & $\boldsymbol{78.1\pm6.1}$ & $69.0\pm6.9$ & $73.5\pm2.0$ & $\boldsymbol{74.5\pm7.8}$ & $77.3\pm5.4$ & $\boldsymbol{76.5\pm2.9}$ \\
    synthetic & 10000 & $75.2\pm6.8$ & $73.4\pm4.8$ & $\boldsymbol{74.3\pm1.9}$ & $73.6\pm10.8$ & $\boldsymbol{79.4\pm6.0}$ & $75.9\pm2.5$ \\
    \hline
    synthetic + real & 500 & $71.9\pm5.3$ & $67.0\pm4.5$ & $69.4\pm1.6$ & $55.9\pm6.8$ & $81.1\pm3.1$ & $68.5\pm2.5$ \\
    synthetic + real & 1000 & $69.8\pm6.6$ & $71.2\pm3.7$ & $70.5\pm2.1$ & $59.1\pm9.0$ & $82.1\pm3.7$ & $70.6\pm3.1$ \\
    synthetic + real & 2000 & $72.2\pm4.4$ & $70.3\pm4.3$ & $71.2\pm1.6$ & $66.6\pm7.1$ & $79.0\pm4.1$ & $72.8\pm2.2$ \\
    synthetic + real & 3000 & $71.8\pm4.9$ & $73.4\pm5.5$ & $72.6\pm1.6$ & $66.1\pm9.3$ & $81.1\pm5.0$ & $73.6\pm3.0$ \\
    synthetic + real & 5000 & $\boldsymbol{74.7\pm5.3}$ & $\boldsymbol{73.5\pm4.8}$ & $\boldsymbol{74.1\pm2.2}$ & $\boldsymbol{71.7\pm10.0}$ & $80.5\pm4.4$ & $\boldsymbol{76.1\pm3.6}$ \\
    synthetic + real & 10000 & $74.7\pm7.0$ & $73.4\pm6.1$ & $74.0\pm2.7$ & $69.1\pm9.9$ & $\boldsymbol{80.7\pm5.1}$ & $74.9\pm3.2$ \\
    \hline
  \end{tabular}%}
\end{table*}

\begin{table*}[ht]
 \caption{Mean test performance of the 20 runs trained on \textit{train-full} with the baseline hyperparameters}
 \label{table: app train-full_Conv5FC3}
%\resizebox{\textwidth}{!}{%
\centering
\scriptsize
  \begin{tabular}{|c|c||c|c|c||c|c|c|}
    \cline{3-8}
    \multicolumn{2}{c|}{} & \multicolumn{3}{c||}{ADNI} & \multicolumn{3}{c|}{AIBL} \\
    \hline
    \multirow{2}{*}{image type} & synthetic & \multirow{2}{*}{sensitivity} & \multirow{2}{*}{specificity} & balanced & \multirow{2}{*}{sensitivity} & \multirow{2}{*}{specificity} & balanced\\
    & images & & & accuracy & & & accuracy\\
    \hline
    \hline
    real & - & $79.1\pm6.2$ & $76.3\pm4.2$ & $77.7\pm2.5$ & $70.6\pm6.7$ & $86.3\pm3.6$ & $78.4\pm2.4$ \\
    real (high-resolution) & - & $84.5\pm3.8$ & $76.7\pm4.0$ & $80.6\pm1.1$ & $71.6\pm6.4$ & $89.2\pm2.7$ & $80.4\pm2.6$ \\
    \hline
    synthetic & 500 & $81.6\pm6.8$ & $79.5\pm5.8$ & $80.5\pm2.4$ & $74.7\pm9.3$ & $87.3\pm4.8$ & $81.0\pm3.2$ \\
    synthetic & 1000 & $82.9\pm4.5$ & $82.0\pm5.8$ & $82.4\pm1.9$ & $77.2\pm7.4$ & $88.8\pm5.2$ & $83.0\pm2.0$ \\
    synthetic & 2000 & $81.9\pm4.5$ & $87.7\pm3.4$ & $84.8\pm2.0$ & $74.7\pm6.3$ & $92.1\pm1.9$ & $83.4\pm2.7$ \\
    synthetic & 3000 & $\boldsymbol{84.9\pm3.5}$ & $87.4\pm3.5$ & $86.1\pm1.5$ & $77.4\pm5.8$ & $90.9\pm3.0$ & $84.2\pm1.8$ \\
    synthetic & 5000 & $84.0\pm3.5$ & $88.4\pm3.3$ & $86.2\pm1.7$ & $76.8\pm4.2$ & $\boldsymbol{92.2\pm1.8}$ & $\boldsymbol{84.5\pm1.8}$ \\
    synthetic & 10000 & $84.2\pm5.4$ & $\boldsymbol{88.6\pm3.9}$ & $\boldsymbol{86.4\pm1.8}$ & $\boldsymbol{77.5\pm7.4}$ & $91.0\pm3.2$ & $84.2\pm2.4$ \\
    \hline
    synthetic + real & 500 & $82.5\pm3.4$ & $81.9\pm5.4$ & $82.2\pm2.4$ & $76.0\pm6.3$ & $89.7\pm3.3$ & $82.9\pm2.5$ \\
    synthetic + real & 1000 & $84.6\pm4.4$ & $84.3\pm5.1$ & $84.4\pm1.8$ & $77.0\pm7.0$ & $90.4\pm3.4$ & $83.7\pm2.3$ \\
    synthetic + real & 2000 & $\boldsymbol{85.4\pm4.0}$ & $86.4\pm5.9$ & $85.9\pm1.6$ & $77.2\pm6.9$ & $90.4\pm3.8$ & $83.8\pm2.2$ \\
    synthetic + real & 3000 & $84.7\pm3.6$ & $86.8\pm4.5$ & $85.8\pm1.7$ & $77.2\pm4.8$ & $\boldsymbol{91.7\pm2.9}$ & $84.4\pm1.8$ \\
    synthetic + real & 5000 & $84.6\pm4.2$ & $86.9\pm3.6$ & $85.7\pm2.1$ & $76.9\pm5.2$ & $91.4\pm3.0$ & $84.2\pm2.2$ \\
    synthetic + real & 10000 & $84.2\pm2.8$ & $\boldsymbol{88.5\pm2.9}$ & $\boldsymbol{86.3\pm1.8}$ & $\boldsymbol{79.1\pm4.7}$ & $91.0\pm2.6$ & $\boldsymbol{85.1\pm1.9}$ \\
    \hline
  \end{tabular}
\end{table*}

\begin{table*}[ht]
 \caption{Mean test performance of the 20 runs trained on \textit{train-50} with the optimized hyperparameters}
 \label{table:app train-50_random}
%\resizebox{\textwidth}{!}{%
\centering
\scriptsize
  \begin{tabular}{|c|c||c|c|c||c|c|c|}
    \cline{3-8}
    \multicolumn{2}{c|}{} & \multicolumn{3}{c||}{ADNI} & \multicolumn{3}{c|}{AIBL} \\
    \hline
    \multirow{2}{*}{image type} & synthetic & \multirow{2}{*}{sensitivity} & \multirow{2}{*}{specificity} & balanced & \multirow{2}{*}{sensitivity} & \multirow{2}{*}{specificity} & balanced\\
    & images & & & accuracy & & & accuracy\\
    \hline
    \hline
    real & - & $75.4\pm5.0$ & $75.5\pm5.3$ & $75.5\pm2.7$ & $68.6\pm8.5$ & $82.6\pm4.2$ & $75.6\pm4.1$ \\
    real (high-resolution) & - & $73.6\pm6.2$ & $70.6\pm5.9$ & $72.1\pm3.1$ & $57.8\pm12.3$ & $84.6\pm4.2$ & $71.2\pm5.1$ \\
    \hline
    synthetic & 500 & $75.8\pm3.0$ & $77.6\pm5.3$ & $76.7\pm2.8$ & $73.2\pm9.0$ & $\boldsymbol{83.6\pm4.0}$ & $78.4\pm4.0$ \\
    synthetic & 1000 & $76.7\pm4.6$ & $78.5\pm4.9$ & $\boldsymbol{77.6\pm3.7}$ & $78.7\pm7.5$ & $83.2\pm4.8$ & $80.9\pm4.3$ \\
    synthetic & 2000 & $73.9\pm3.6$ & $\boldsymbol{79.8\pm4.0}$ & $76.8\pm3.0$ & $78.2\pm6.9$ & $82.4\pm3.7$ & $80.3\pm3.5$ \\
    synthetic & 3000 & $74.4\pm6.1$ & $79.8\pm4.9$ & $77.1\pm4.0$ & $76.4\pm10.1$ & $82.4\pm4.3$ & $79.4\pm4.7$ \\
    synthetic & 5000 & $77.1\pm4.5$ & $77.4\pm5.2$ & $77.2\pm2.1$ & $81.1\pm5.9$ & $82.0\pm3.9$ & $\boldsymbol{81.5\pm2.6}$ \\
    synthetic & 10000 & $\boldsymbol{77.5\pm5.3}$ & $77.3\pm4.7$ & $77.4\pm3.1$ & $\boldsymbol{81.7\pm5.4}$ & $79.7\pm4.1$ & $80.7\pm2.9$ \\
    \hline
    synthetic + real & 500 & $73.2\pm4.2$ & $78.0\pm3.3$ & $75.6\pm2.5$ & $69.2\pm9.4$ & $\boldsymbol{82.7\pm4.1}$ & $76.0\pm4.2$ \\
    synthetic + real & 1000 & $76.1\pm5.3$ & $\boldsymbol{79.5\pm2.9}$ & $77.8\pm2.3$ & $79.3\pm5.8$ & $82.5\pm4.2$ & $80.9\pm3.2$ \\
    synthetic + real & 2000 & $75.2\pm3.8$ & $78.6\pm4.4$ & $76.9\pm2.4$ & $77.8\pm8.8$ & $82.2\pm4.5$ & $80.0\pm3.6$ \\
    synthetic + real & 3000 & $76.5\pm3.8$ & $79.2\pm4.2$ & $77.8\pm1.9$ & $80.9\pm7.9$ & $81.4\pm4.2$ & $81.2\pm3.7$ \\
    synthetic + real & 5000 & $77.1\pm3.7$ & $76.7\pm4.1$ & $76.9\pm2.5$ & $80.7\pm6.1$ & $81.2\pm3.7$ & $80.9\pm2.7$ \\
    synthetic + real & 10000 & $\boldsymbol{77.8\pm4.6}$ & $78.2\pm4.9$ & $\boldsymbol{78.0\pm2.1}$ & $\boldsymbol{81.7\pm4.9}$ & $81.9\pm4.6$ & $\boldsymbol{81.9\pm2.2}$ \\
    \hline
  \end{tabular}%}
\end{table*}

\begin{table*}[ht]
 \caption{Mean test performance of the 20 runs trained on \textit{train-full} with the optimized hyperparameters}
 \label{table: app train-full_random}
%\resizebox{\textwidth}{!}{%
\centering
\scriptsize
  \begin{tabular}{|c|c||c|c|c||c|c|c|}
    \cline{3-8}
    \multicolumn{2}{c|}{} & \multicolumn{3}{c||}{ADNI} & \multicolumn{3}{c|}{AIBL} \\
    \hline
    \multirow{2}{*}{image type} & synthetic & \multirow{2}{*}{sensitivity} & \multirow{2}{*}{specificity} & balanced & \multirow{2}{*}{sensitivity} & \multirow{2}{*}{specificity} & balanced\\
    & images & & & accuracy & & & accuracy\\
    \hline
    \hline
    real & - & $82.5\pm4.2$ & $88.5\pm6.6$ & $85.5\pm2.4$ & $75.1\pm8.4$ & $88.7\pm9.0$ & $81.9\pm3.2$ \\
    real (high-resolution) & - & $82.6\pm4.5$ & $88.9\pm6.3$ & $85.7\pm2.5$ & $78.9\pm5.4$ & $89.9\pm4.0$ & $84.4\pm1.7$ \\
    \hline
    synthetic & 500 & $81.7\pm3.6$ & $90.5\pm3.9$ & $86.1\pm1.4$ & $75.5\pm7.1$ & $89.8\pm4.3$ & $82.6\pm2.9$ \\
    synthetic & 1000 & $82.8\pm3.4$ & $90.0\pm4.0$ & $86.4\pm2.1$ & $76.8\pm4.5$ & $91.5\pm2.5$ & $84.2\pm1.7$ \\
    synthetic & 2000 & $81.3\pm2.8$ & $91.2\pm2.8$ & $86.2\pm1.7$ & $76.2\pm6.7$ & $\boldsymbol{92.2\pm3.6}$ & $84.2\pm2.6$ \\
    synthetic & 3000 & $82.2\pm4.9$ & $90.6\pm4.5$ & $86.4\pm2.0$ & $77.7\pm6.3$ & $90.8\pm4.4$ & $84.3\pm2.0$ \\
    synthetic & 5000 & $80.6\pm3.4$ & $\boldsymbol{91.6\pm2.5}$ & $86.1\pm1.9$ & $75.3\pm5.4$ & $92.4\pm2.5$ & $83.8\pm2.0$ \\
    synthetic & 10000 & $\boldsymbol{84.0\pm3.8}$ & $89.1\pm3.1$ & $\boldsymbol{86.5\pm1.7}$ & $\boldsymbol{79.2\pm5.2}$ & $90.1\pm3.7$ & $\boldsymbol{84.7\pm2.3}$ \\
    \hline
    synthetic + real & 500 & $82.3\pm2.3$ & $89.8\pm2.7$ & $86.0\pm1.8$ & $74.9\pm5.0$ & $91.4\pm2.6$ & $83.2\pm2.4$ \\
    synthetic + real & 1000 & $82.5\pm3.3$ & $90.5\pm4.1$ & $86.5\pm1.9$ & $76.4\pm5.6$ & $91.0\pm3.4$ & $83.7\pm2.0$ \\
    synthetic + real & 2000 & $\boldsymbol{83.1\pm4.2}$ & $\boldsymbol{91.3\pm3.2}$ & $\boldsymbol{87.2\pm1.7}$ & $76.0\pm4.7$ & $92.0\pm2.4$ & $84.0\pm2.0$ \\
    synthetic + real & 3000 & $81.3\pm3.7$ & $90.4\pm3.4$ & $85.8\pm2.6$ & $74.9\pm7.3$ & $92.3\pm2.6$ & $83.6\pm3.2$ \\
    synthetic + real & 5000 & $81.9\pm3.5$ & $90.9\pm2.5$ & $86.4\pm1.3$ & $74.1\pm4.9$ & $\boldsymbol{92.9\pm1.9}$ & $83.5\pm2.2$ \\
    synthetic + real & 10000 & $82.2\pm3.4$ & $91.2\pm3.6$ & $86.7\pm1.8$ & $\boldsymbol{76.4\pm4.2}$ & $92.1\pm2.1$ & $\boldsymbol{84.3\pm1.8}$ \\
    \hline
  \end{tabular}%}
\end{table*}

% Can use something like this to put references on a page
% by themselves when using endfloat and the captionsoff option.
\ifCLASSOPTIONcaptionsoff
  \newpage
\fi

% that's all folks
\end{document}